%% file: MLJ_main.tex
\documentclass[pdflatex,sn-apa]{sn-jnl}
\usepackage{graphicx}
\usepackage[export]{adjustbox}
\usepackage{amsmath,amssymb,amsfonts,amsthm,bm}
\usepackage{multirow,booktabs}
\usepackage{float,wrapfig,arydshln,pifont,comment,enumitem}
\usepackage{subfig}
\usepackage{rotating}
\usepackage{xcolor}
\usepackage{url}
\makeatletter\NAT@longnamesfalse\makeatother
\newcommand{\e}{\textnormal{e}}
\definecolor{mycolor}{rgb}{0,0,0}
\allowdisplaybreaks[2]
\theoremstyle{thmstyleone}
\newtheorem{theorem}{Theorem}
\newtheorem{lemma}[theorem]{Lemma}

\theoremstyle{thmstyletwo}

\theoremstyle{thmstylethree}
\newtheorem{definition}[theorem]{Definition}

\makeatother

\begin{document}
\title[Comparing Classification and Feature Learning in Normalized Spaces]{Comparing Classification and Feature Learning by Analyzing Their Loss Functions in Normalized Spaces with Different Scale Factors}
\author[1,2,4]{\fnm{Qiufu} \sur{Li}}
\author[1,2,4]{\fnm{Huibin} \sur{Xiao}}
\author*[1,2,4]{\fnm{Linlin} \sur{Shen}}\email{llshen@szu.edu.cn}
\author[3]{\fnm{Weijia} \sur{Fan}}
\author[5]{\fnm{Xi} \sur{Jia}}
\author[6]{\fnm{Yongdong} \sur{Huang}}
\affil[1]{\orgname{School of Artificial Intelligence, Shenzhen University}, \orgaddress{\city{Shenzhen}, \postcode{518060}, \country{China}}}
\affil[2]{\orgname{National Engineering Laboratory for Big Data System Computing Technology, Shenzhen University}, \orgaddress{\city{Shenzhen}, \postcode{518060}, \country{China}}}
\affil[3]{\orgname{College of Computer Science and Software Engineering, Shenzhen University}, \orgaddress{\city{Shenzhen}, \postcode{518060}, \country{China}}}
\affil[4]{\orgname{Guangdong Provincial Key Laboratory of Intelligent Information Processing, Shenzhen University}, \orgaddress{\city{Shenzhen}, \postcode{518060}, \country{China}}}
\affil[5]{\orgname{Big Data Institute, University of Oxford}, \orgaddress{\city{Oxford}, \postcode{OX3 7LF}, \country{United Kingdom}}}
\affil[6]{\orgname{Institute of Image Understanding Research, North Minzu University}, \orgaddress{\city{Yinchuan}, \postcode{750021}, \country{China}}}

\abstract{
Classification and feature learning are two closely interrelated fundamental tasks. Feature learning in supervised settings is typically achieved by training classification models, under the common assumption that higher classification accuracy implies better feature quality. In this paper, we first point out that classification accuracy reflects the sample-wise local properties of a model on a dataset, whereas feature learning aims to learn features with good dataset-level global properties, such as intra-class compactness and inter-class separability. From the first principle of feature learning, we redesign the binary cross-entropy (BCE) loss, a counterpart of the cross-entropy (CE) loss used in classification. Then, in normalized feature space, we analyze the minima of the BCE and CE losses and show that both can lead to neural collapse, where sample features from different classes become maximally compact and maximally separated. More importantly, by investigating the convergence behaviors of the two losses as the scale factor $\gamma$ varies in the normalized space, we reveal how classification accuracy and feature quality change differently across spaces of different sizes. Under a fixed training strategy, a very large $\gamma$ achieves perfect classification accuracy but slows convergence toward the ideal feature geometry, whereas a very small $\gamma$ makes classification highly sensitive to small perturbations. A moderate $\gamma$ is therefore necessary for simultaneously achieving reliable classification and well-structured features. Extensive experiments validate these findings and demonstrate that BCE loss exhibits advantages in long-tail and open-set recognitions, as it better aligns with the global requirements of feature learning.
}


\maketitle

\section{Introduction} \label{Introduction}
In machine learning, classification and feature learning are fundamental and closely intertwined tasks in supervised scenarios, a connection that can be traced back to Fisher's linear discriminant analysis~\citep{fisher1936use}, which achieves effective classification by learning a linear projection that makes the projected feature vectors as compact as possible within classes and as separable as possible between classes.
In modern deep learning, classification tasks commonly take the cross-entropy (CE) loss to train deep feature extraction models and linear classifiers,
and they could learn desirable features, making the trained models widely applied in downstream fields that require features with well properties, such as face recognition~\citep{liu2022sphereface}, object detection and tracking~\citep{jain2024fusion}, image segmentation~\citep{Azad2024medical}, and image-text retrieval~\citep{Vasu_2024_CVPR}, etc.
For feature learning, while feature properties required in different scenarios are different, intra-class compactness and inter-class separability are most attractive in many tasks.
Intuitively, the higher classification accuracy usually implies the better feature properties,
and conversely, the better feature properties suggest the higher accuracy.

\begin{figure}[!t]
    \centering
    \includegraphics[scale=0.7, viewport=20 70 490 300, clip,max width=\linewidth,max height=0.74\textheight,keepaspectratio]{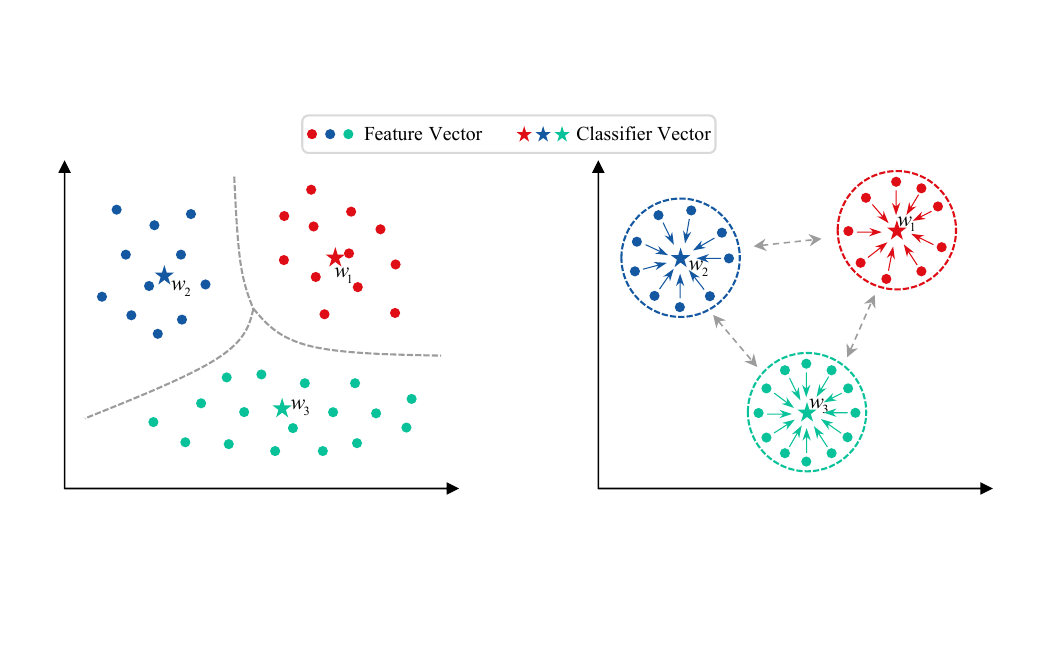}
    \caption{\color{mycolor}The intuitive difference between classification and feature learning. 
    Classification (left) requires that the feature vector of each sample be closer to its own class center/classifier vector than to those of other classes, and classification accuracy reflects the model's local properties at each sample. In contrast, intra-class compactness and inter-class separability of sample features are the model's global properties at the dataset level, and feature learning (right) drives the feature vectors of all samples to be as close as possible to their respective class centers while simultaneously being far from other class centers
    }
    \label{classification_vs_feature_learning}
\end{figure}

It should be noted that the classification accuracy of a model is calculated by checking whether each sample is correctly classified, reflecting the local properties of the model on a dataset.
In contrast, the intra-class compactness and inter-class separability are the global properties of the model on the whole dataset.
{\color{mycolor} As Fig. \ref{classification_vs_feature_learning} shows, correct classification only requires that a sample's feature vector be relatively closer to its corresponding class center/classifier vector than to other class centers; whereas feature learning demands that all samples' feature vectors be as close as possible to their own class centers while simultaneously remaining as far as possible from other class centers. 
Therefore, high classification accuracy alone is insufficient to guarantee good feature representations. That is to say, there is an essential difference between the classification and feature learning.}
However, there has not been a in-depth analysis of this difference,
and in supervised scenarios, the feature learning is usually implemented through training classification models without considering its global nature.

In unbounded Euclidean space, measuring intra-class compactness and inter-class separability of features using distance-based metrics cannot guarantee a fair comparison across different models. 
A more common approach is to use cosine similarity in a normalized space to measure the compactness and separability.
When training models in practice, normalizing features can improve the training stability, accelerate the convergence, and enhance the models' generalization, which is a frequent regularization strategy in various fields, especially in face recognition~\citep{wang2017normface}. 
Feature normalization is generally achieved by dividing by the $L_2$-norm of the feature vectors, which scales the feature space into a hypersphere of unit radius. To better adapt to downstream tasks, when training models in normalized space, the normalized features are typically multiplied by a scale factor $\gamma$ before being input into the training loss.
From a classification perspective, it is generally believed that a large $\gamma$ sharpens the probability distribution of features belonging to the true class~\citep{zheng2024knowledge}, enhancing the model's confidence in its classification decisions, while a small one smooths the probability distribution, suppressing noise effects.
While the above empirical observations exist, there is currently a lack of in-depth theoretical understanding of the scale factor or a convincing explanation of how it affects the feature learning and classification in normalized spaces.

In deep learning, although loss functions such as contrastive loss and triplet loss are commonly used for feature learning, they operate only at the local level of individual sample pairs or triplets and do not account for the global nature of intra-class compactness and inter-class separability.  Besides that, the CE loss for training classification models can also yield well features with desirable properties, and there is theoretical evidence \citep{Yaras2022_NeurIPS} that when the CE reaches its minimum, the sample feature and classifier vectors exhibit \textbf{neural collapse (NC)}, at which point the feature compactness and discriminability are maximized uniformly across different classes. 
However, the CE loss is designed for classification and does not address the global nature of feature learning; therefore, it is not always efficient in more challenging scenarios such as long-tailed recognition (LTR) and open-set recognition (OSR).
 
In this paper, in the normalized spaces, we first redesign binary cross-entropy (BCE) from the ``first principles'' of feature learning. We then compare feature learning and classification under different scale factors in normalized spaces by analyzing the minima and convergence behavior of BCE and comparing them with those of CE, which is commonly used for classification.
The main contributions are summarized as follows. 

\begin{itemize}
    {\color{mycolor} \item For the first time, we in-depth reveal the distinction between classification and feature learning. Classification is depend on sample-level relative ordering of decision scores of each sample, while feature learning expect dataset-level global properties, such as intra-class compactness and inter-class separability. With the basic requirements of feature learning, we derive a loss function which turns out to be the binary cross-entropy (BCE) loss used for multi-class tasks. Meanwhile, we prove that when BCE reaches its minima, it can uniformly maximize the compactness and separability of features across different classes. Although the CE used for classification also exhibits this property, the advantage of BCE lies in its classifier biases, which help it achieve this desirable property more rapidly.
    }
    
    \item We make an in-depth analysis of the convergence behavior of the CE and BCE as the scale factor $\gamma$ changes, to further reveal the differences between classification and feature learning in normalized spaces of different sizes. As $\gamma$ increases, the convergence rates of the two losses may decay exponentially. Consequently, in a large normalized space with a very large $\gamma$, achieving high classification performance becomes relatively easy, whereas learning features with desirable properties becomes difficult. 
    Conversely, as $\gamma$ approaches zero, the convergence rates decrease linearly. Thus, in a very small space with $\gamma$ close to $0$, the setting is unfavorable for feature learning and even more so for classification.
        
    \item We conduct extensive experiments using deep models, including ResNet and ViT. Using the fixed training strategy, when the scale factor $\gamma = 64$ is very large, all the models achieve 100\% classification accuracy on the training set, but the intra-class compactness and inter-class separability of features they extract are comparatively poor. When $\gamma = 0.1$ is very small, the models' feature properties are unsatisfactory but superior to those at $\gamma = 64$, while their classification performs very poor. When $\gamma$ is at a moderate level, the models' classification and feature properties can both reach their optimal points. 
    In addition, the superior performance of BCE in more challenging scenarios that require higher-quality features, such as LTR and OSR, further demonstrates the practical relevance of the above analysis.
\end{itemize}

In our previous work~\citep{li2025_bce} published in ICML2025, we have investigated the optimization dynamics of models in Euclidean space trained with BCE and CE losses integrated with regularization terms, demonstrating for the first time that the BCE can induce neural collapse. It further showed that the BCE loss explicitly enhances intra-class compactness and inter-class separability of features through decision score constraint and classifier bias.

Building on the ICML work~\citep{li2025_bce},  this paper addresses a different and broader question: whether classification and feature learning are equivalent in normalized spaces of different sizes. To this end, we reformulate the BCE loss based on the global requirements of feature learning, analyze the minima of the normalized BCE and CE losses, and further investigate how the scale factor $\gamma$, which controls the size of the normalized space, affects their convergence behavior. Our new theoretical analysis, study of the scale factor, comparison between local classification and global feature learning, and long-tailed/open-set recognition evaluations substantially extend the previous work.

\section{Related Works}
\textbf{Classification and feature learning}.\quad
The success of deep models such as ResNet~\citep{He2016_deep} and ViT~\citep{dosovitskiy2021an} in classification has continually driven the development of deep learning.
The deep classification models have been consistently developed, including DenseNet~\citep{Huang2017_densely}, MobileNet~\citep{howard2017mobilenets}, Swin Transformer~\citep{liu2021swin}, and ConvNeXt~\citep{liu2022convnet}, etc., all of them consist of deep feature extraction models with CNN or Transformer architectures along with linear classifiers.
In supervised scenarios, employing the CE loss and training these models on large amounts of labeled data can yield good classification alongside robust feature extraction capabilities, making them widely applicable in fields such as computer vision, natural language processing, and multi-modal large models.

For the tasks that require better feature properties, such as face recognition and long-tailed recognition, normalizing sample features and the classifier vectors can effectively enhance the performance of deep models.
NormFace~\citep{wang2017normface} and SphereFace~\citep{liu2017sphereface} are among the earliest face recognition works,
and advanced methods such as CosFace~\citep{wang2018cosface}, ArcFace~\citep{deng2019arcface}, TopoFR~\citep{dan2024topofr}, and GFace~\citep{zhao2025global} continue to be developed in the normalized space.
On imbalanced datasets, normalization can suppress the scale differences between the classifier vectors, thereby enhancing classification accuracy~\citep{liu2020deep}.
In current, various re-balancing strategies have been designed based on the normalized CE and are widely applied in long-tailed recognition (LTR) tasks~\citep{li2023fcc,han2023wrapped,liang2024nonlocal,chen2025decompose}.

\textbf{Scale factor and temperature}.\quad
In the normalized space, a scale factor $\gamma$ is typically used to adjust the size of feature space~\citep{wang2017normface}, while it is temperature coefficient $\tau$ when it is applied to the denominator in contrastive learning~\citep{he2020momentum}.
An inappropriate scale factor can significantly degrade the model performance, yet there is no clear and reasonable explanation for its effects.
A commonly circulated notion is that large scale factor sharpens the probability distribution of the models' outputs, increasing the confidence in the classification decisions, while the small one smooths the probability distribution, helping to mitigate the effects of noise and other influences.
Then, in the classification losses, the scale factor is usually greater than 1, with a typical value of $\gamma = 64$ in face recognition~\citep{deng2019arcface,zhou2023uniface}.
The temperature used in the contrastive learning plays a similar role as that of the scale factor, and it on the denominator is usually set to a value less than 1. 
In this paper, for the first time, we reveal the role of the scale factor in the deep feature learning,
which also helps people understand the temperature coefficient.

\textbf{Minima of loss function}.\quad 
When training a classification model and the loss reaches a minimum, it maximizes intra-class compactness and inter-class separability of sample features, and features from different classes form an equiangular tight frame (ETF); this phenomenon is known as {neural collapse (NC)}, first discovered in~\citet{Papyan2020_prevalence}. The NC studies on CE loss have been extended to various scenarios, including class imbalance~\citep{fang2021exploring,dang2024neural,gao2024distribution}, out-of-distribution data~\citep{chen2024neural}, and fixed classifiers~\citep{kim2024fixed}, as well as other CE variants such as focal loss, label smoothing loss.

As previously stated, we have compared BCE and CE from perspective of NC in Euclidean space~\citep{li2025_bce}. For the case in the normalized space, it has proved that the global minima of CE loss are all NC solutions~\citep{LU2022,Yaras2022_NeurIPS}.
It should be noted that existing theoretical studies on NC primarily address whether a given loss function can induce the optimal feature geometry at its global minimum. However, merely establishing the inevitability of NC does not explain why a model can achieve high classification performance before learning well-structured feature representations, why a model with optimal classification performance does not necessarily attain optimal performance on feature-contrastive tasks, or how the corresponding loss functions converge during the model training. Therefore, this paper shifts the focus from ``whether the loss function can reach NC'' to ``explaining how classification and feature learning approach their respective desired conditions across feature spaces of different sizes, by analyzing the convergence behaviors of CE and BCE.'' The normalized feature space provides a suitable framework for this analysis, while the scale factor $\gamma$ directly controls the size of the space and affects the convergence behaviors of the CE and BCE losses.

\section{Preliminaries}\label{Preliminaries}

Let $\mathcal{D}=\bigcup_{k=1}^K\big\{\boldsymbol{X}_i^{(k)}\big\}_{i=1}^{n_k}$ be a sample set from $K$ classes, where $\boldsymbol{X}_i^{(k)}$ is the $i$-th sample of the $k$-th class and $n_k$ is number of samples in the class.
In feature learning, for $\forall\boldsymbol{X}\in\mathcal D$,
a feature extractor $\mathcal M$ maps the sample into its feature vector $\boldsymbol{h}=\mathcal{M}(\boldsymbol{X})\in\mathbb R^d$, where $d$ denotes feature dimension.

For classification, a classifier $\mathcal C$ converts the feature vector into a \textbf{decision vector} $\bm z \in \mathbb R^K$, i.e.,
$\bm z = \mathcal C \big( \mathcal M(\bm X)\big) = \mathcal C(\bm h)$.
In deep learning, the classifier $\mathcal C$ is commonly represented by a fully connection layer containing a weight matrix $\bm W = [\bm w_1, \bm w_2, \cdots, \bm w_K]^\top \in \mathbb R^{K\times d}$ and a bias vector $\bm b = [b_1, b_2, \cdots, b_K]^\top \in \mathbb R^K$.
In normalized space, both sample features and classifier vectors are normalized to $\|\bm h\|_2 = 1$ and $\|\bm w_j\|_2 = 1$, for $\forall j\in[K]$.
Then, a scale factor $\gamma > 0$ is multiplied on the decision vector to adjust the size of feature space, i.e., $\bm z = \gamma \bm W \bm h - \bm b$.

In the following, we will re-derive the cross-entropy (CE) loss and binary cross-entropy (BCE) loss functions used in multiclass tasks, and show that they are fundamentally suited to classification and feature learning, respectively.

\subsection{CE for Classification}\label{sec_classification}
In a classification task, for $\forall\,\bm X \in \mathcal D$ with feature vector $\bm h = \mathcal M(\bm X)$, its predicted class label $\hat{k}$
is decided by the vector $\bm z = [\gamma\bm w_j^\top\bm h-b_j]_{j=1}^K$, and $\hat{k} = \arg\max_j \{\gamma \bm w_j^\top \bm h - b_j\}_{j=1}^K$.
For any sample $\bm X^{(k)}$ from class $k$, in its decision vector $\bm z^{(k)} = \gamma\bm W\bm h^{(k)} - \bm b$, we refer to the $k$-th component $\gamma\bm w_k^\top\bm h^{(k)} - b_k$ as its \textbf{positive decision score} and the others $\{\gamma\bm w_j^\top\bm h^{(k)}-b_j\}_{j\neq k}$ as \textbf{negative decision scores}.
Then, for correct classification, it only requires that the sample satisfy a relative ordering where its positive decision score exceeds its negative ones,
\begin{align}
    \gamma\bm w_k^\top\bm h^{(k)} - b_k > \max\{\gamma\bm w_j^\top\bm h^{(k)}-b_j\}_{j\neq k}.
    \label{eq_decision_score_for_classification}
\end{align}
Classification accuracy $\mathcal A$ is calculated by checking whether each sample is correctly classified via Eq. (\ref{eq_decision_score_for_classification}),
equal to ratio of number of correctly classified samples to sample number of dataset $\mathcal D$,
\begin{align}
    \mathcal {A}({\mathcal M, \mathcal C}) = \frac{|\mathcal D({\mathcal M, \mathcal C})|}{|\mathcal D|} \times 100\%, \label{eq_accuracy}
\end{align}
where
\begin{align}
    \mathcal D(\mathcal M, \mathcal C) = \bigcup_{k=1}^K\Big\{\boldsymbol X^{(k)}: k = \arg\max_{\ell}\{\gamma\boldsymbol w_\ell^\top \boldsymbol h^{(k)} - b_\ell\}, ~\boldsymbol X^{(k)}\in\mathcal D_k,~\boldsymbol h^{(k)} = \mathcal M(\boldsymbol X^{(k)})\Big\}
\end{align}
consisting of the samples correctly classified by $\mathcal M$ and $\mathcal C$ in $\mathcal D$.
Obviously, the classification accuracy $\mathcal A$ reflect a local property of the feature extraction model $\mathcal M$ at each sample on $\mathcal D$.

During training of a classification model, the decision vector $\bm z^{(k)}$ of sample $\bm X^{(k)}$ is converted via Softmax into predicted probabilities, $\big\{\hat{p}_j = \frac{\exp(\gamma\bm w_j\bm h^{(k)} - b_j)}{\sum_{\ell=1}^K\exp(\gamma\bm w_\ell\bm h^{(k)} - b_\ell)}\big\}_{j=1}^K$, of the sample belonging to each class.
For $\bm X^{(k)}$, its true probabilities are $p_k=1$ and $p_j=0$ for $k$ and $j$ with $j\neq k$.
Then, cross-entropy (CE) loss is calculated and minimized to drive the model training,
\begin{equation}
    \mathcal L_{\text{ce}}(\bm X^{(k)}) = -\sum_{j=1}^K p_j \log (\hat{p}_j) = - \log\bigg(\frac{\e^{\gamma \bm w_k^\top\bm h^{(k)}-b_k}}{\sum_{\ell=1}^K\e^{\gamma \bm w_\ell^\top\bm h^{(k)}-b_\ell}}\bigg).
    \label{eq_normalized_CE}
\end{equation}
The CE loss is one of the most widely used loss function in deep learning.

\subsection{BCE for Feature Learning}\label{sec_bce_deep_feature_learning}
In supervised settings, the sample features can be learned in the training of classification models using the CE loss, while this loss does not take into account the sample-independent global constrains required for inter-class compactness and inter-class separability of all sample features on dataset.

In the normalized space, to precisely measure the properties of sample features $\bigcup_{k=1}^K\big\{\bm h_i^{(k)}\big\}_{i=1}^{n_k}$, we define the metrics of intra-class compactness $\mathcal E_{\text{com}}$ and inter-class separability $\mathcal E_{\text{sep}}$ as
\begin{align}
    &\mathcal E_{\text{com}} = \frac{1}{2}\bigg[\frac{1}{K}\sum_{k=1}^K \bigg(\frac{1}{n_k^2} \sum_{i=1}^{n_k}\sum_{i'=1}^{n_k}
    \big\langle\boldsymbol h_i^{(k)} , \boldsymbol h_{i'}^{(k)} \big\rangle\bigg) +1 \bigg], \label{eq_compactness} \\
    &\mathcal E_{\text{sep}} = \frac{1}{2}\bigg[1-\frac{1}{K(K-1)} \sum_{k=1}^K\sum_{k'=1\atop k'\neq k}^K\bigg(\frac{1}{n_k}\frac{1}{n_{k'}}
    \sum_{i=1}^{n_k}\sum_{i'=1}^{n_{k'}}\big\langle\boldsymbol h_i^{(k)}, \boldsymbol h_{i'}^{(k')}\big\rangle\bigg)\bigg]. \label{eq_distinctiveness}
\end{align}
Eqs. (\ref{eq_compactness}) and (\ref{eq_distinctiveness}) compute the compactness and separability using the feature similarities of all same-class sample pairs and all different-class sample pairs, respectively, which is entirely reasonable. 
However, if all these feature similarities are taken into account in the loss function for feature learning, the computational cost increases significantly. One can observe that, if classifier vectors $\{\bm w_k\}_{k=1}^K$ are used as anchors, then the distribution of the unbiased decision scores $\bigcup_{k=1}^K\bigcup_{i=1}^{n_k}\{\bm w_j^\top \bm h_i^{(k)}\}_{j=1}^K$ for all samples could reflect the compactness and separability of features. 
Moreover, since the number of classes $K$ is usually smaller than the number of samples, $N = \sum_{k=1}^K n_k$, the number of decision scores is substantially lower than that of sample pairs, so using these scores could improve the computational efficiency.
Therefore, we re-express the two properties using the decision scores to design a loss function for feature learning.

For the sample features $\{\bm h_i^{(k)}=\mathcal M(\bm X_i^{(k)})\}_{i=1}^{n_k}$ of class $k$,
taking the $k$-th classifier vector $\bm w_k$ as an anchor, the high intra-class compactness requires that all the features should lie close to $\bm w_k$.
To explicitly measure this closeness, we choose the smallest possible unified threshold $t'_k$ such that the maximum distance from the feature vectors to the anchor is no greater than this threshold, i.e., $\max\{\|\bm w_k - \bm h_i^{(k)}\|_2^2\}_{i=1}^{n_k} < t'_k$. 
As $\|\bm w - \bm h\|_2^2 = 2-2\bm w^\top \bm h$ in the normalized space, the above compactness condition is equivalent to defining the largest possible unified threshold $t_k = 1 - {t'_k}/{2}$ to impose a uniform lower bound on cosine similarities between each sample feature and the anchor,
\begin{align}
    \min\{\bm w_k^\top \bm h_i^{(k)}\}_{i=1}^{n_k} > t_k.
    \label{eq_uniform_threshold_pos}
\end{align}
Similarly, to explicitly measure the separability between the classes $k$ and $j$ for $\forall j\neq k$, we take the $j$-th classifier vector $\bm w_j$ as an anchor, and choose the largest possible unified threshold $t'_j$ such that the minimum distance from the feature vectors of class $k$ to $\bm w_j$ is no smaller than this threshold, i.e., $\min\{\|\bm w_j - \bm h_i^{(k)}\|_2^2\} > t'_j$. This separability condition is equivalent to defining the smallest possible unified threshold $t_j = 1 - t'_j/2$ to impose a uniform upper bound on similarities between each feature vector of class $k$ and the anchor $\bm w_j$,
\begin{align}
    \max\{\bm w_j^\top \bm h_i^{(k)}\}_{i=1}^{n_k} < t_j,\quad \forall~j\neq k.
    \label{eq_uniform_threshold_neg}
\end{align}
According to the above analysis, for class $k$, the desirable feature properties require that all positive unbiased decision scores $\{\gamma\bm w_k^\top \bm h_i^{(k)}\}_{i=1}^{n_k}$ uniformly exceed as large a threshold as possible, while all negative ones $\bigcup_{j=1\atop j\neq k}^K\{\gamma\bm w_j^\top \bm h_i^{(k)}\}_{i=1}^{n_k}$ uniformly remain below as small a threshold as possible, which are \emph{sample-independent} on dataset $\mathcal D$.

In the model training, set $b_k = {\gamma t_k}$ for $\forall k\in[K]$. For $\forall\,\bm X^{(k)}$ from class $k$, applying Sigmoid on its positive decision score $\gamma\bm w_k\bm h^{(k)} - b_k$, one can compute the predicted probability that it satisfies the compactness condition in Eq. (\ref{eq_uniform_threshold_pos}),
$\hat p_{kk} = \sigma(\gamma\bm w_k^\top\bm h^{(k)} - b_k)$;
similarly, applying Sigmoid on its negative decision scores $\{\gamma\bm w_j\bm h^{(k)} - b_j\}_{j=1\atop j\neq k}^K$, one can compute the predicted probabilities that it satisfies the separability condition in Eq. (\ref{eq_uniform_threshold_neg}),
$\big\{\hat p_{kj} = 1 - \sigma(\gamma\bm w_j^\top\bm h^{(k)} - b_j): j\neq k\big\}$.
Clearly, for any reasonable threshold, the perfect feature vector always satisfies both the compactness and separability conditions; therefore, for the ground-truth labels used in feature learning, we set $p_{kj} = 1$ with $\forall j\in[K]$ for the sample $\bm X^{(k)}$ from class $k$.
Then, for feature learning, one can calculate $K$ ordinary binary cross-entropy (BCE) losses for any sample $\bm X^{(k)}$ and minimizes them together,
\begin{equation}
    \mathcal L_{\text{bce}}(\bm X^{(k)}) = -\sum_{j=1}^K p_{kj}\log\hat p_{kj} = \log\big(1+\e^{-\gamma\bm w_k^\top\bm h^{(k)} + b_k}\big) + \sum_{j=1\atop j\neq k}^K\log\big(1+\e^{\gamma\bm w_j^\top\bm h^{(k)} - b_j}\big).
    \label{eq_normalized_BCE}
\end{equation}

This BCE loss can also be derived by decomposing a multi-class classification into multiple binary classification tasks, each of which checks whether a sample $\bm X^{(k)}$ belongs to class $j$ for $\forall j\in[K]$,
which has been widely used in multi-label classification~\citep{kobayashi2023two}.
Here, we reformulate it based on the global constraints of feature learning, thereby revealing, for the first time, its connection to the global properties of sample features across the entire dataset.

\section{Main Theoretical Results}\label{Main Theoretical Results}
The above derivations of CE and BCE highlight the fundamental difference between classification and feature learning. As classification performance reflects the model's local behavior at each sample, the CE loss merely requires that, for each sample, the positive decision score be relatively higher than the negative ones. In contrast, feature learning requires the model to exhibit globally favorable properties across the entire dataset, and the classifier bias parameters in the BCE loss impose a global and absolute constraint on the feature vectors of all samples. 

In this section, we will further reveal the differences between the two tasks by analyzing the minima of CE and BCE, as well as their convergence rates to these minima in normalized spaces of different sizes.

\subsection{Minima of Normalized BCE and CE}\label{section_nc}
Following~\citet{Zhu2021_geometric} and~\citet{Yaras2022_NeurIPS}, we simplify the analysis by using unconstrained feature model (UFM) on balanced dataset. Specifically, we take the sample features $\bigcup_{k=1}^K\{\bm h_i^{(k)}\}_{i=1}^{n_k}$, classifier vectors $\{\bm w_j\}_{j=1}^K$, and classifier biases $\{b_j\}_{j=1}^K$ as free variables, without considering the parameters within the feature extraction model $\mathcal M$, and assume that $n=n_k$, for $\forall k\in[K]$.
Let
\begin{align}
    f_\mu(\bm{W},\bm{H},\bm{b})=\frac{1}{nK}\sum_{k=1}^K\sum_{i=1}^{n}\mathcal{L}_\mu(\bm X_i^{(k)}),
    \label{f_object_function}
\end{align}
where $\mu\in\{\text{ce},\text{bce}\}$ indicating the normalized CE and BCE losses in Eqs. (\ref{eq_normalized_CE}) and (\ref{eq_normalized_BCE}), and
$\bm H =\big[\bm {H}_1,\bm {H}_2,\cdots,\bm {H}_K\big]\in\mathbb{R}^{d\times{(nK)}}$ with $\bm{H}_k = \big[\bm {h}_1^{(k)},\bm {h}_2^{(k)},\cdots,\bm {h}_n^{(k)}\big]\in\mathbb{R}^{d\times{n}}$. Then, we achieve the following theorems.

\begin{theorem}\label{them:CE collapse}
    Suppose $d\geq K-1$, i.e., the feature dimension is greater than the number of classes. The loss function $f_\mathrm{ce}(\boldsymbol{W},\boldsymbol{H},\boldsymbol{b})$ defined using $\mathcal{L}_\mathrm{ce}$ in \textit{Eq.} (\ref{eq_normalized_CE}) satisfies
    \begin{align}
            f_{\mathrm{ce}}(\boldsymbol{W},\boldsymbol{H},\boldsymbol{b})
        &\geq\log\Big(1+(K-1)\e^{-\frac{\gamma K}{K-1}}\Big), \label{eq_CE_minimum_value}
    \end{align}
    and the equality is attained if and only if $\boldsymbol{W},\boldsymbol{H}$ and $\boldsymbol{b}$ satisfy
    \begin{align}
                    &\boldsymbol{w}_k = \boldsymbol{h}_i^{(k)},~~\forall k\in[K],~i\in[n],\label{eq_w=h}\\
                    &\bm w_k^\top \bm w_j = -\frac{1}{K-1},~~\forall j\neq k\in[K], \label{W_etf}\\
        \text{and}~~&b_k = {b}^{\star}\in\mathbb R,~~\forall k\in[K], \label{eq_b=b}
    \end{align}
    where $b^\star\in\mathbb R$ could be any number.
\end{theorem}
\begin{proof}
	The lower bound and its equality conditions are established in Theorem \ref{them:lemmaB.1} in Appendix.
\end{proof}

\begin{theorem}\label{them:BCE collapse}
     Suppose $d\geq K-1$. The loss function $f_\mathrm{bce}(\boldsymbol{W},\boldsymbol{H},\boldsymbol{b})$ defined using $\mathcal{L}_\mathrm{bce}$ in \textit{Eq.} (\ref{eq_normalized_BCE}) satisfies
    \begin{align}
f_{\mathrm{bce}}(\boldsymbol{W},\boldsymbol{H},\boldsymbol{b})
        \geq&~ -2\gamma- (K-2)b^{\star} +\log\big(1+\e^{\gamma-b^{\star}}\big)+(K-1)\log\Big(1+\e^{b^{\star}+\frac{\gamma}{K-1}}\Big)\nonumber\\
        =&~ \log\big(1+\e^{-\gamma+b^{\star}}\big)+(K-1)\log\Big(1+\e^{-\frac{\gamma}{K-1}-b^{\star}}\Big), \label{eq_BCE_minimum_value}
    \end{align}
    and the equality holds if and only if $\boldsymbol{W},\boldsymbol{H}$ and $\boldsymbol{b}$ satisfy Eqs. (\ref{eq_w=h}, \ref{W_etf}) and (\ref{eq_b=b}), where
    \begin{equation}
        {b^{\star}} = -\frac{\gamma}{K-1}-\log 2 + \log\Big[(K-2) + \sqrt{(K-2)^2+4(K-1)\e^{\gamma +\frac{\gamma}{K-1}}}\Big]. 
        \label{eq_transcen_eq_b_bce}
    \end{equation}
\end{theorem}
\begin{proof}
	The proof establishes a lower bound for the BCE loss and characterizes its equality conditions. The details can be found in Theorem \ref{them:BCE collapse_supp} in Appendix.
\end{proof}

Theorems \ref{them:CE collapse} and \ref{them:BCE collapse} show that CE and BCE can respectively attain their constant-valued lower bounds over the entire dataset; hence the constant values are their minima. 
Furthermore, when the CE and BCE reach their minima, the sample features and classifier vectors satisfy the conditions of NC: 
\begin{itemize}
    \item (\textbf{collapse}) The features of samples within the same class converge to the class center. From Eq. (\ref{eq_w=h}), for $\forall k\in [K]$, the features $\{\bm h_i^{(k)}\}_{i=1}^{n}$ converge to its class center, $\bar{\bm h}_k = \frac{1}{n}\sum_{i=1}^n \bm h_i^{(k)}$. 
    \item (\textbf{alignment}) The class center of each class is parallel to its corresponding classifier vector, differing only by a constant coefficient. From Eq. (\ref{eq_w=h}), $\bar{\bm h}_k = \bm w_k$, for $\forall k\in [K]$.
    \item (\textbf{equiangular}) The angle between any two class center vectors is equal and maximized, forming an equiangular tight frame (ETF). From Eqs. (\ref{eq_w=h}, \ref{W_etf}), $\bm w_k^\top \bm w_j = \bar{\bm h}_k^\top \bar{\bm h}_j = -\frac{1}{K-1},~~\forall j\neq k\in[K]$, which indicates $\{\bm w_k\}_{k=1}^K$ and $\{\bar{\bm h}_k\}_{k=1}^K$ form two perfectly overlapping ETFs.
\end{itemize}
NC indicates that, when reaching the minima of the CE or BCE, the compactness of any class is maximized with $\mathcal E_{\text{com}} = 1$, and the separability of any two classes is also maximized with $\mathcal E_{\text{sep}} = \frac{1}{2} + \frac{1}{2(K-1)}$.

{\textbf{The classifier biases.}}\quad
At their minimum points, the normalized CE and BCE share the same geometric structures of sample feature and classifier vectors. Their difference lies in the classifier biases.
In designing the loss functions, the classifier biases of the normalized CE are merely offsets of the decision scores, whereas in the normalized BCE the classifier biases are derived from the thresholds of the decision scores.

When the losses reach their minimum points, the classifier bias vector $\bm b\in\mathbb R^K$ become multiples of the all-one vector $\bm 1_K\in\mathbb R^K$.
For the normalized CE loss, the multiple factor $b^\star\in\mathbb R$ can be any number.
As long as $b_k = b_j,\forall k\neq j$, they do not affect the classification results, and they can be canceled out in the Softmax of the CE loss. Therefore, in the CE used for classification tasks, the $K$ classifier biases can take any identical value when the loss achieving minimum value.
This analysis also implies that, when training models with CE, the classifier biases cannot meaningfully contribute to the unbiased decision scores that reflect the feature properties.
In previous works~\citep{Yaras2022_NeurIPS,LU2022}, the authors ignore the classifier biases in the CE when analyzing its minimum value.

In contrast, for the normalized BCE, the final value of classifier biases, $b^\star$, is figured out via Eq. (\ref{eq_transcen_eq_b_bce}), indicating the BCE loss has only one minimum point in terms of the classifier biases.
On the contrary, the unique converged value of $b^\star$ indicates that, during BCE minimization, the classifier biases play a substantive role by effectively contributing to the update of decision scores that reflect the feature properties, thereby accelerating loss convergence and improving training efficiency.
Furthermore, with a fixed class number $K$, the curve of $b^\star$ in terms of $\gamma$ defined in Eq. (\ref{eq_transcen_eq_b_bce}) has an asymptote, i.e.,
\begin{align}
    b^\star \rightarrow\frac{1}{2}\log(K-1) + \frac{K-2}{K-1}\frac{\gamma}{2}, \label{eq_asymptote_bias_theory}
\end{align}
indicating that the final bias $b^\star$ increases almost linearly as $\gamma$ approaches positive infinity.

\subsection{Convergence of Normalized CE and BCE} \label{section_normalization_practical}
When the classifier biases are equal, i.e., $b_k = b_j,\forall k,j\in[K]$,
the unbiased decision scores $\{\gamma \bm w_j^\top\bm h_i^{(k)}\}_{j,k,i}$ not only determine each sample's classification but also reflect the feature properties.
According to Theorems \ref{them:CE collapse} and \ref{them:BCE collapse}, at the minimum points of normalized CE or BCE, the positive and negative unbiased decision scores will converge two fixed values related to the radius $\gamma$ of normalized space,
\begin{align}
    &\gamma\bm w_k^\top\bm h_i^{(k)} \rightarrow \gamma, ~~\forall k\in[K],~\forall i\in[n],\label{eq_pos_decision_score_theory}\\
    &\gamma\bm w_j^\top\bm h_i^{(k)} \rightarrow -\frac{\gamma}{K-1}, ~~\forall k\neq j,~\forall i\in[n].\label{eq_neg_decision_score_theory}
\end{align}
Clearly, as $\gamma$ increases, both the positive and negative decision scores at the minima increase linearly.
Then, we can analyze the convergence of the losses by analyzing the convergence behavior of the unbiased decision scores, which will help to compare classification and feature learning in the normalized spaces with different scale factors.
Before these analysis, we first define two critical conditions (CC) for the unbiased decision scores.

\begin{definition}\label{def_critical_con_A}
    \textbf{CC I} states that for every sample in dataset $\mathcal D$, its positive unbiased decision score exceeds the negative ones, i.e., for $\forall k\in[K],~i\in[n]$,
    \begin{align}
        \gamma\bm w_k^\top \bm h_i^{(k)} > \max\big\{\gamma \bm w_j^\top \bm h_i^{(k)}:j\in[K],~j\neq k\big\}.
    \end{align}
\end{definition}
\begin{definition}\label{def_critical_con_B}
    \textbf{CC II} states that for all the samples of $\mathcal D$, the positive unbiased decision scores are greater than zero and the negative ones are less than zero, i.e., for $\forall k\in[K],~i\in[n]$,
    \begin{align}
        \gamma\bm w_k^\top\bm h_i^{(k)} > 0 > \max\big\{\gamma\bm w_j^\top\bm h_i^{(k)}: j\in[K],~j\neq k\big\}.
    \end{align}
\end{definition}
When the classifier biases are equal, CC I is equivalent to all the samples being correctly classified,
while CC II not only implies that the classification accuracy is 100\%, but also that their positive and negative unbiased decision scores are uniformly bounded by the threshold $t = 0$, indicating better intra-class compactness and inter-class separability of sample features.

\begin{theorem} \label{them:CE_decay_of_decision_score}
(1) When training models using the normalized CE loss defined in Eq. (\ref{eq_normalized_CE}), as $\gamma$ approaches zero,
its linear decrease leads to a linear decay in the convergence rate of unbiased decision scores.

(2) Once the {CC I} is satisfied, as $\gamma$ linearly approaches positive infinity, the convergence rate of unbiased decision scores decay exponentially.
\end{theorem}
\begin{proof}
	The result follows from the gradient analysis and asymptotic estimates given in Theorem \ref{them:CE_decay_of_decision_score_supp} in Appendix.
\end{proof}

\begin{theorem}\label{them:BCE_decay_of_decision_score}
(1) When training models using the BCE loss defined in Eq. (\ref{eq_normalized_BCE}),
as $\gamma$ approaches zero, its linear decrease leads to a linear decay in the convergence rate of unbiased decision scores.

(2) In contrary, once the {CC II} is satisfied, as $\gamma$ linearly approaches positive infinity, the convergence rate of unbiased decision scores decay exponentially.
\end{theorem}
\begin{proof}
	The result follows from the gradient analysis and asymptotic estimates given in Theorem \ref{them:BCE_decay_of_decision_score_supp} in Appendix.
\end{proof}

{\color{mycolor}
Using the unbiased decision scores, Theorems \ref{them:CE_decay_of_decision_score} and \ref{them:BCE_decay_of_decision_score} reveal the convergence rates of the CE and BCE in normalized spaces of varying sizes: as the space radius $\gamma$ decreases linearly toward zero, the convergence rates of CE and BCE decay linearly; whereas once CCs I and II are satisfied, respectively, their convergence rates decay exponentially. 
Meanwhile, based on these two theorems, one can also analyze the differences between classification and feature learning across normalized spaces of varying sizes.

\begin{itemize}
    \item[(1)] \textbf{When the radius of normalized space is very small, for example $0< \gamma < 1$, it is not suitable for feature learning, and even less suitable for classification.} 
     As $\gamma$ decreases linearly, the theoretical margin between the positive and negative decision scores of each sample also decreases linearly. In this case, any small perturbation during training or even testing can easily change their relative magnitudes, thereby significantly affecting the classification results; however, this does not necessarily mean that the learned feature properties are also extremely poor. Nevertheless, since the convergence rate of the loss function decreases linearly as $\gamma$ decreases, the number of training epochs required for the feature properties to reach their theoretical maximum increases linearly. 
     Therefore, under the same training strategy, the smaller the value of $\gamma$, the poorer the learned feature properties will be.
    
    \item[(2)] \textbf{When the radius of normalized space is very large, it is beneficial for classification but detrimental to feature learning.} 
     When $\gamma$ is very large, the theoretical margin between the positive and negative decision scores of each sample also becomes very large. With training, it is then easy to satisfy the relative ordering required for correct classification, and the larger $\gamma$ is, the easier this becomes. However, as $\gamma$ increases linearly, the convergence rate of the loss function decays exponentially, which means that the number of training epochs required for the feature properties to reach their theoretical optimum grows exponentially. Therefore, under the same training strategy, increasing the spatial size does not affect classification accuracy, but it significantly degrades the quality of the learned features.

    \item[(3)] \textbf{The model's classification performance is misaligned with the properties of the sample features.} 
    With a fixed training strategy, the classification accuracy $\mathcal A$ of the trained model increases monotonically with the radius $\gamma$ of the normalized space, while the final feature qualities including $\mathcal E_{\text{com}}$ and $\mathcal E_{\text{sep}}$ first increase and then decrease as $\gamma$ grows. Therefore, a model's classification accuracy cannot be used to reliably indicate its performance on downstream tasks that require feature comparison, such as face recognition, image retrieval, person re-identification, and object tracking, etc.

    \item[(4)] \textbf{BCE outperforms CE in larger normalized spaces.}
    Theorems \ref{them:CE_decay_of_decision_score} and \ref{them:BCE_decay_of_decision_score} stem from the fact that, once the decision scores of the samples satisfy CCs I and II, respectively, their contributions to the convergence rates of CE and BCE decay exponentially.
    Therefore, in a larger space, once a sample is correctly classified and thus satisfies CC I, its contribution to the convergence of CE becomes very small. At this point, however, its contribution to the convergence of BCE may still remain relatively large, and only when its decision score further satisfies CC II will its contribution to the convergence of BCE also become very small. This analysis implies that, in larger spaces, BCE can converge faster than CE. Consequently, using the same training strategy, BCE can train the model into a more favorable state than CE, thereby achieving better classification accuracy and feature properties.
\end{itemize}
Through the above analysis, we highlight the differences between classification and feature learning and depict the variation in classification performance and feature properties with space size; meanwhile, we demonstrate that BCE performs better than CE in larger spaces. In the following, we will empirically validate these conclusions.
}

\section{Experiments}\label{Experiments}
To validate the conclusions about classification and feature learning as well as the differences between the CE and BCE,
we trained CNN~\citep{He2016_deep} and Transformer~\citep{dosovitskiy2020_vit} on MNIST~\citep{Lecun1998_gradient}, CIFAR10, and CIFAR100~\citep{krizhevsky2009_learning}.

\begin{figure}[!t]
	\centering
	\subfloat[$\mathcal{NC}_1$, $\mathcal{NC}_2$, and $\mathcal{NC}_3$ of ResNet18 on CIFAR10]{
		\includegraphics[width=0.32\linewidth,max width=\linewidth,max height=0.74\textheight,keepaspectratio]{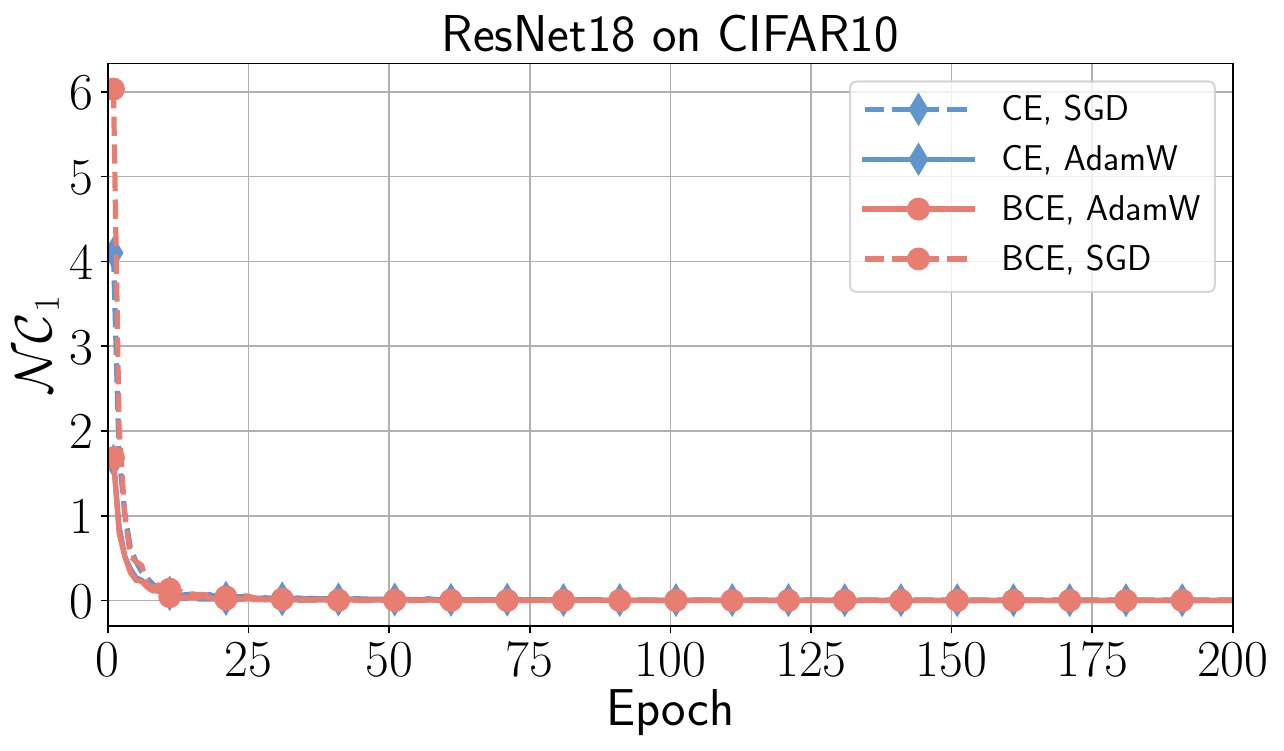}
		\includegraphics[width=0.32\linewidth,max width=\linewidth,max height=0.74\textheight,keepaspectratio]{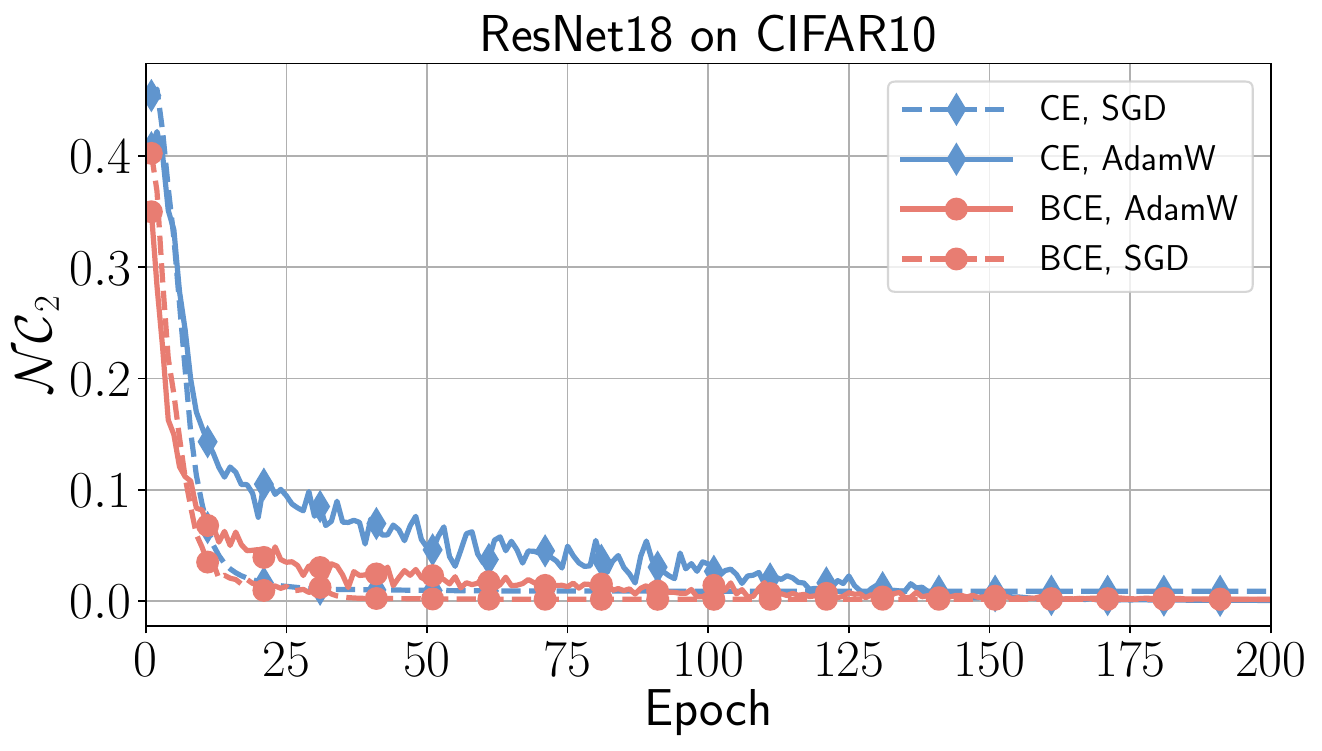}
		\includegraphics[width=0.32\linewidth,max width=\linewidth,max height=0.74\textheight,keepaspectratio]{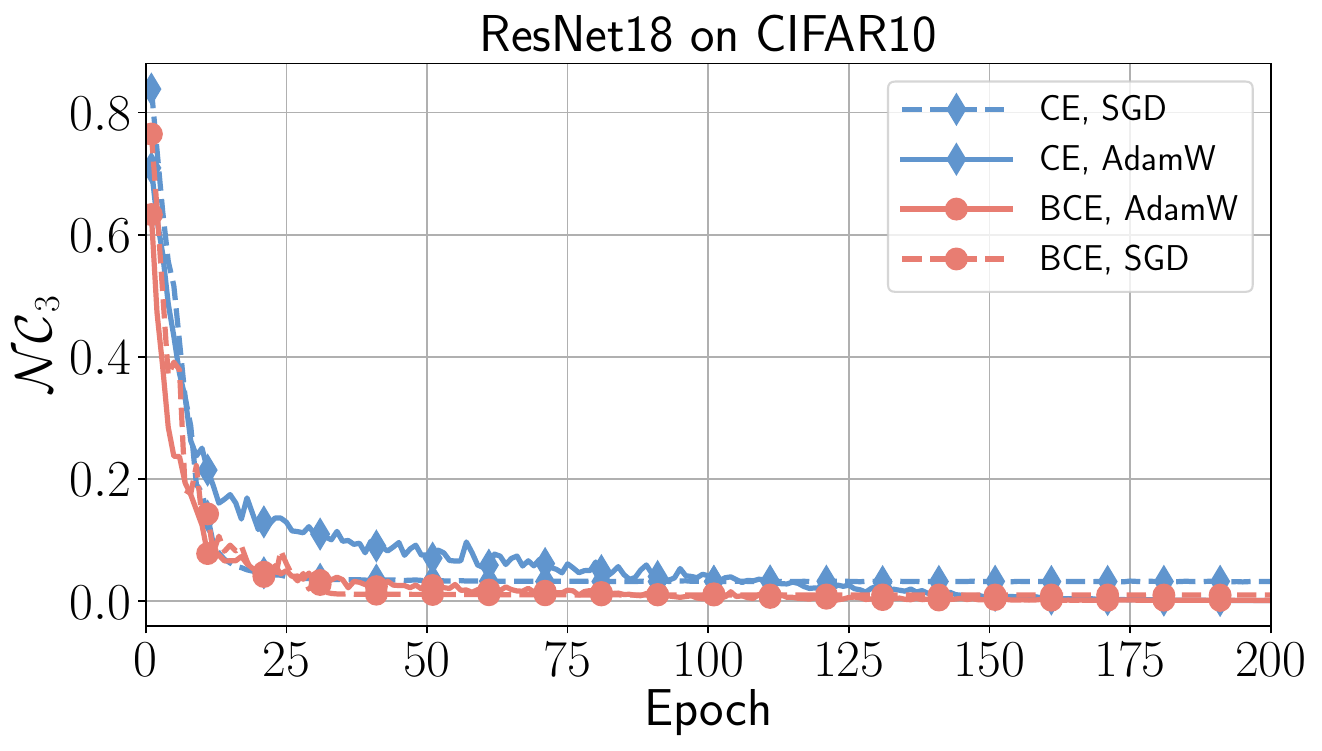}
		\label{fig_resnet18_cifar10_NC}}
	
	\subfloat[$\mathcal{NC}_1$, $\mathcal{NC}_2$, and $\mathcal{NC}_3$ of ResNet50 on CIFAR10]{
		\includegraphics[width=0.32\linewidth,max width=\linewidth,max height=0.74\textheight,keepaspectratio]{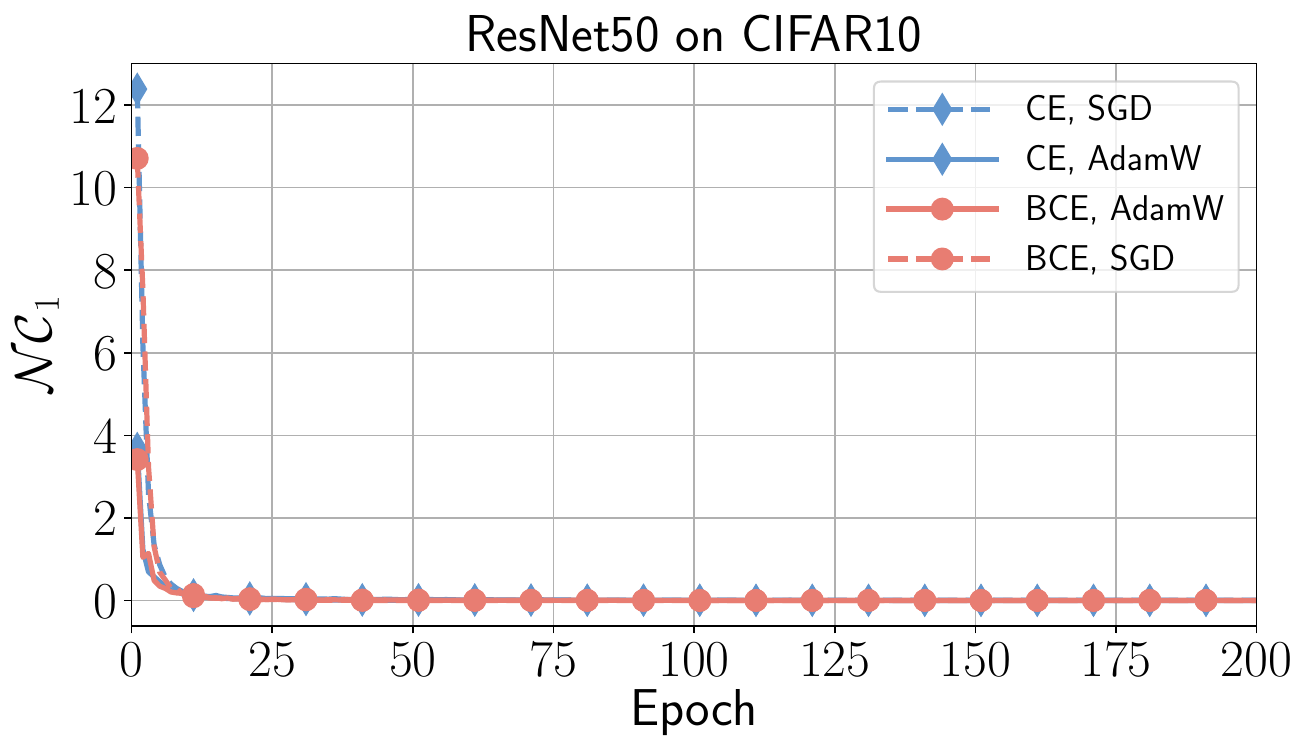}
		\includegraphics[width=0.32\linewidth,max width=\linewidth,max height=0.74\textheight,keepaspectratio]{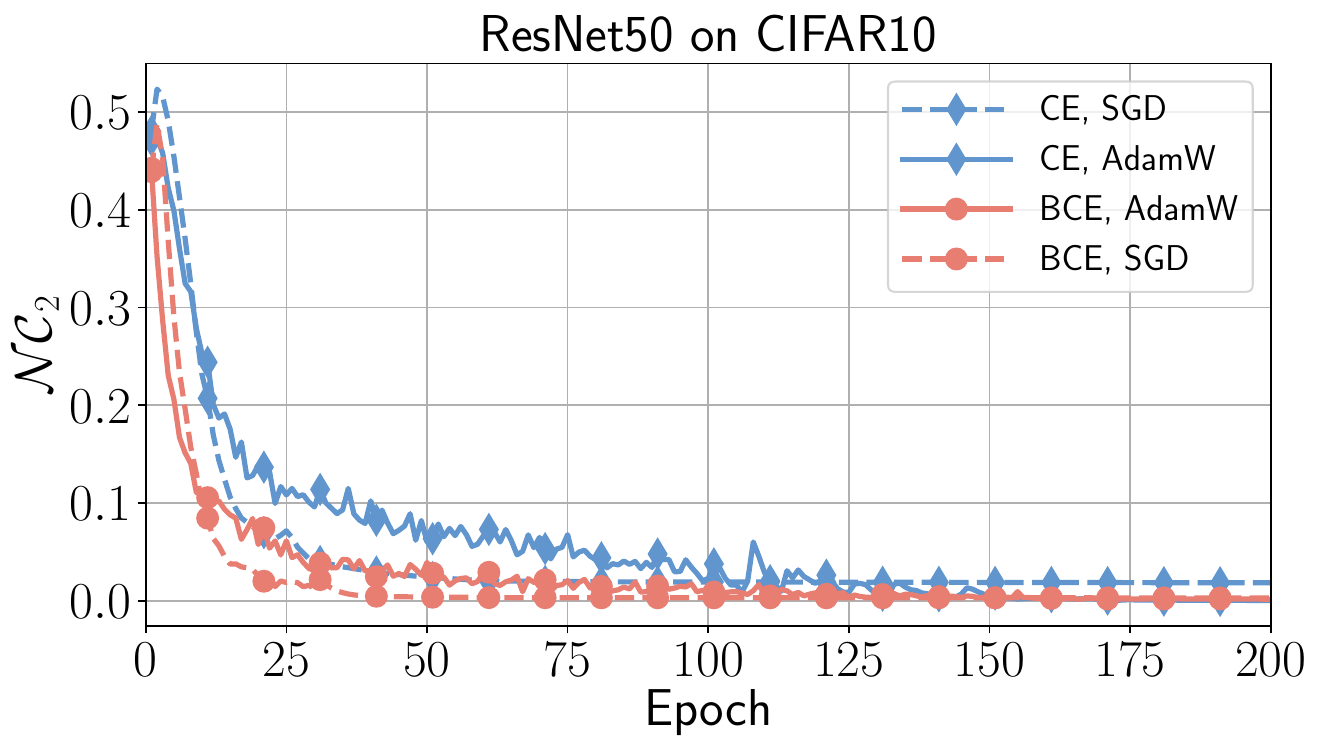}
		\includegraphics[width=0.32\linewidth,max width=\linewidth,max height=0.74\textheight,keepaspectratio]{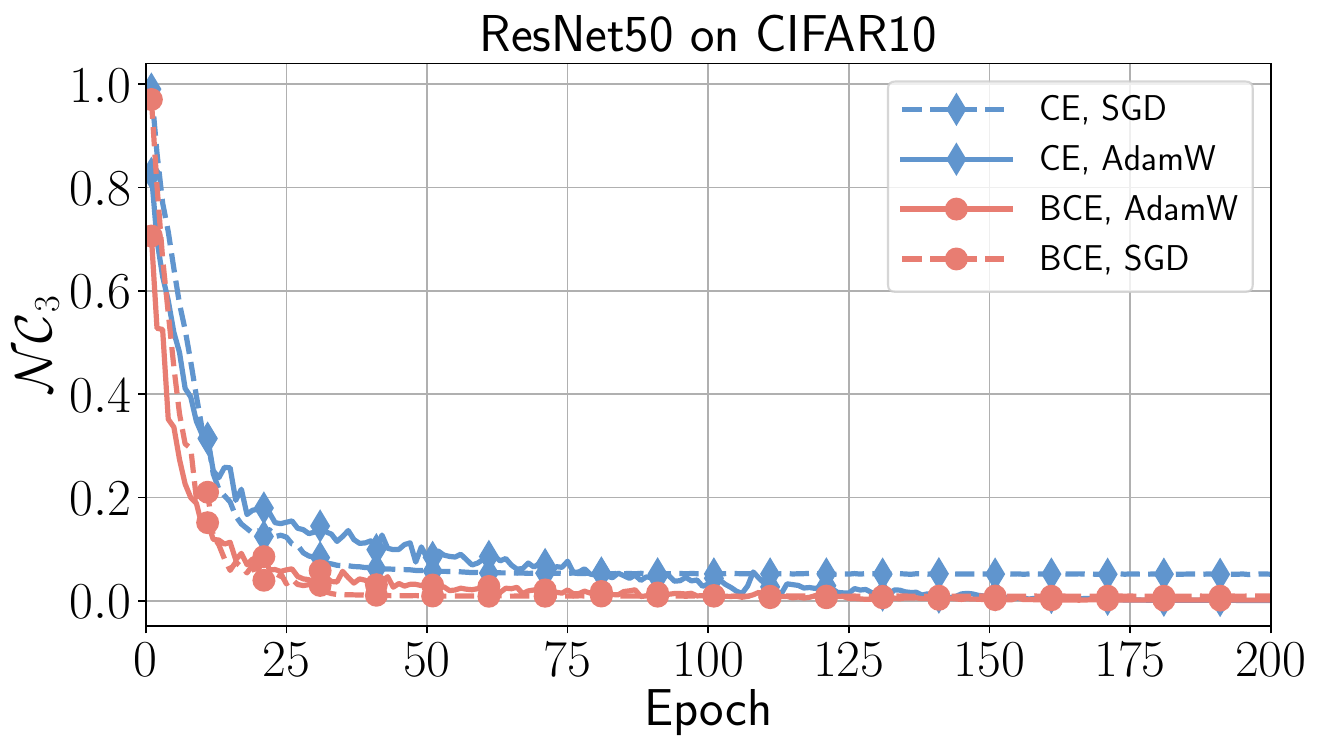}
		\label{fig_resnet50_cifar10_NC}}
	
	\subfloat[$\mathcal{NC}_1$, $\mathcal{NC}_2$, and $\mathcal{NC}_3$ of DenseNet121 on CIFAR10]{
		\includegraphics[width=0.32\linewidth,max width=\linewidth,max height=0.74\textheight,keepaspectratio]{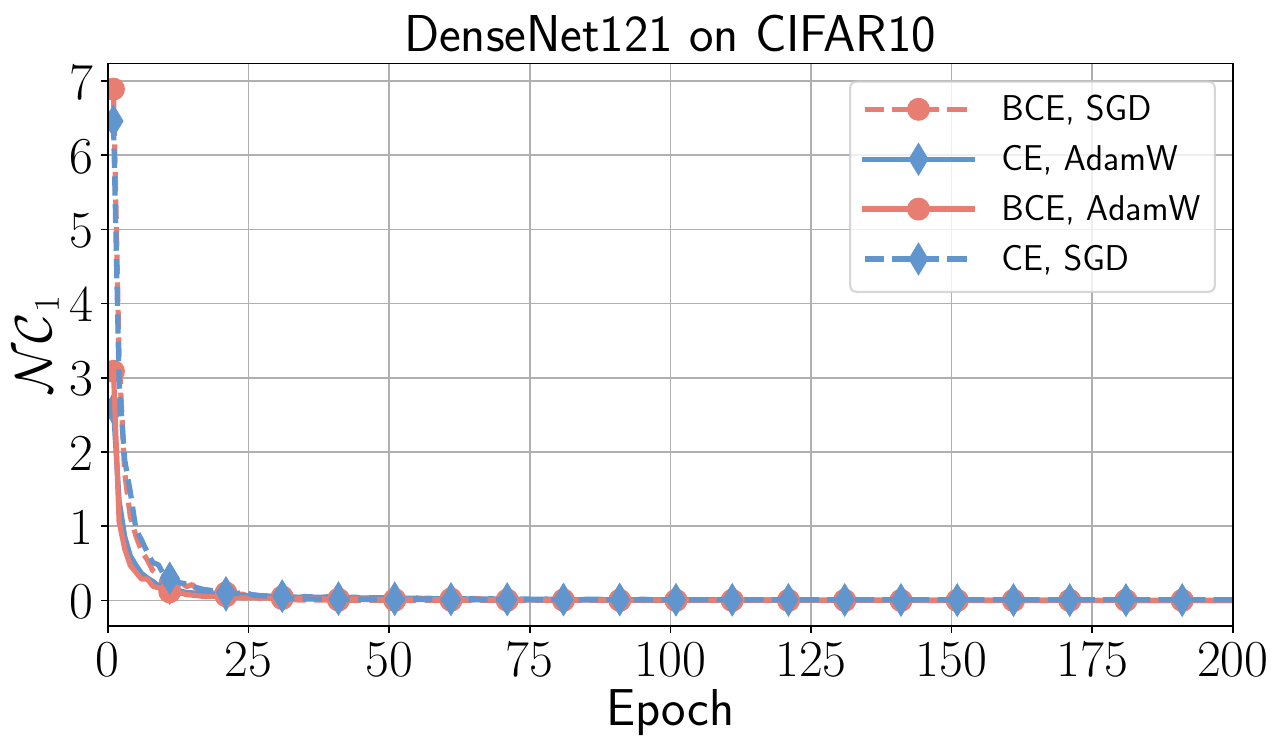}
		\includegraphics[width=0.32\linewidth,max width=\linewidth,max height=0.74\textheight,keepaspectratio]{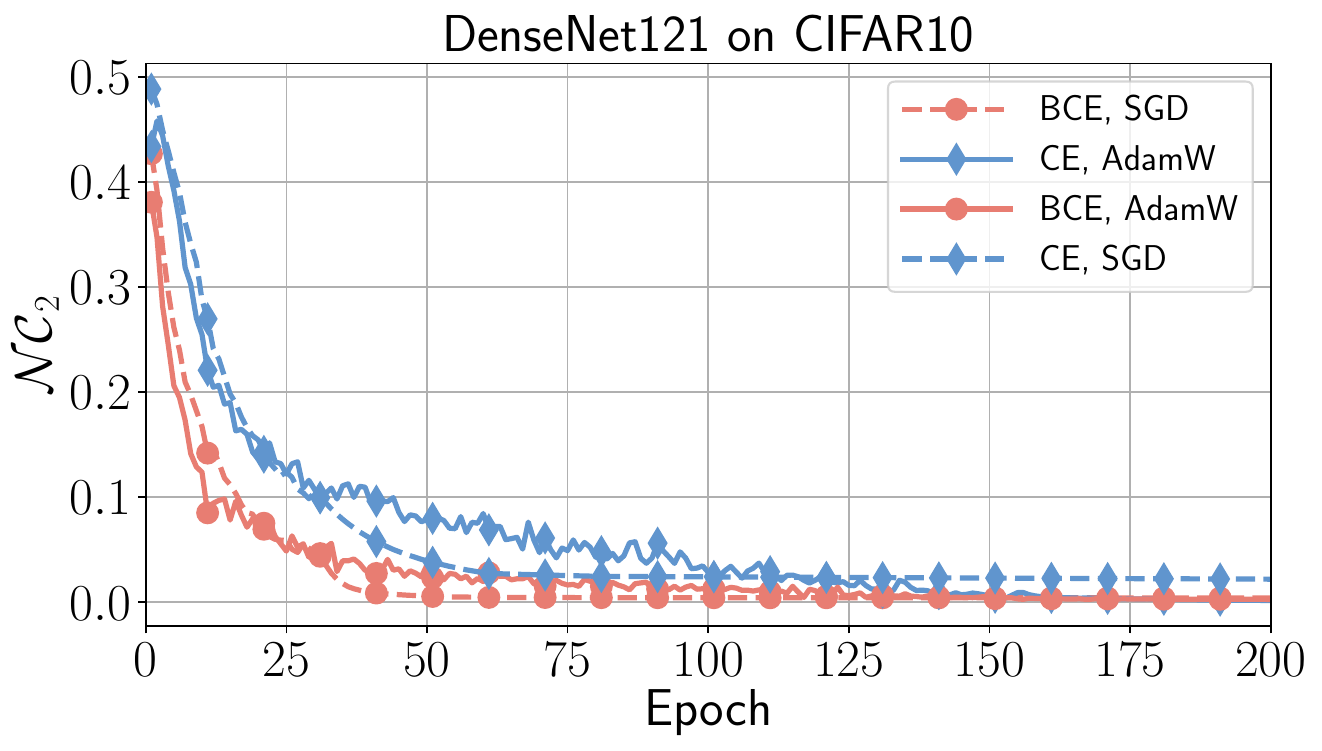}
		\includegraphics[width=0.32\linewidth,max width=\linewidth,max height=0.74\textheight,keepaspectratio]{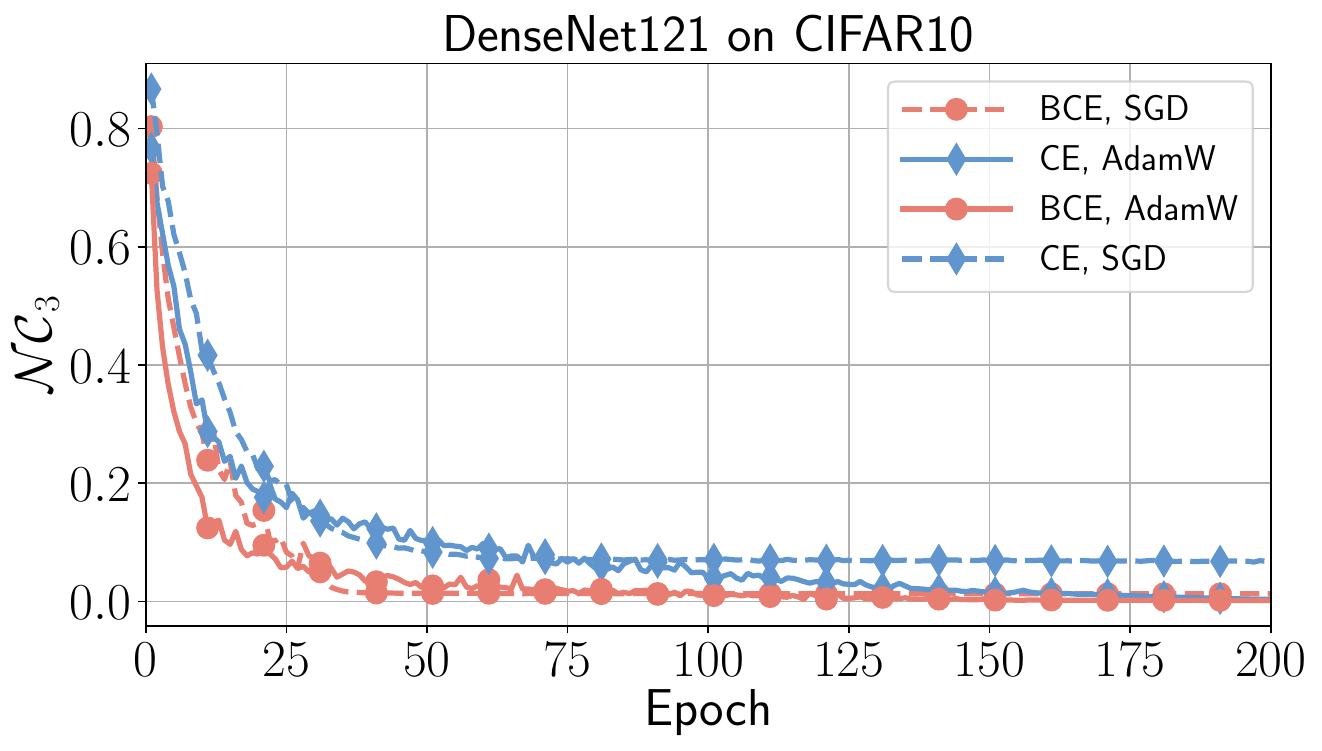}
		\label{fig_densenet121_cifar10_NC}}
	
	\subfloat[$\mathcal{NC}_1$, $\mathcal{NC}_2$, and $\mathcal{NC}_3$ of ViT on CIFAR10]{
		\includegraphics[width=0.32\linewidth,max width=\linewidth,max height=0.74\textheight,keepaspectratio]{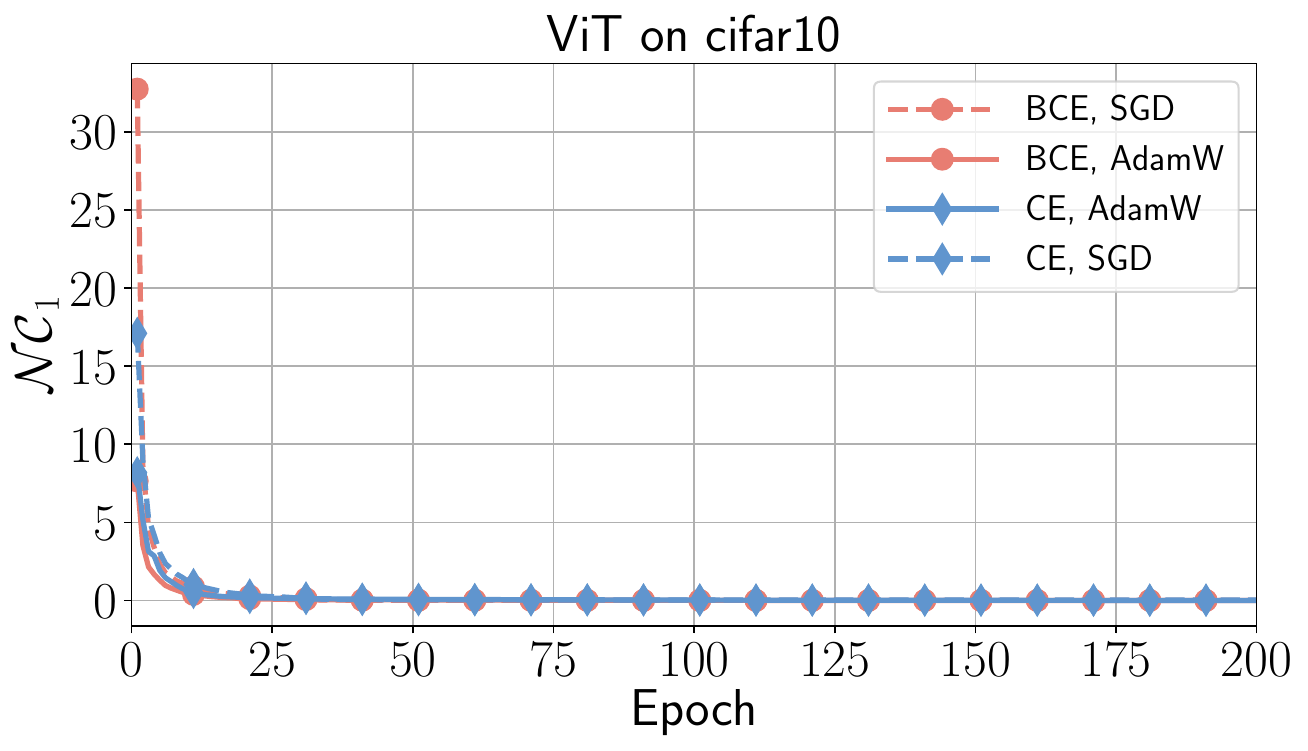}
		\includegraphics[width=0.32\linewidth,max width=\linewidth,max height=0.74\textheight,keepaspectratio]{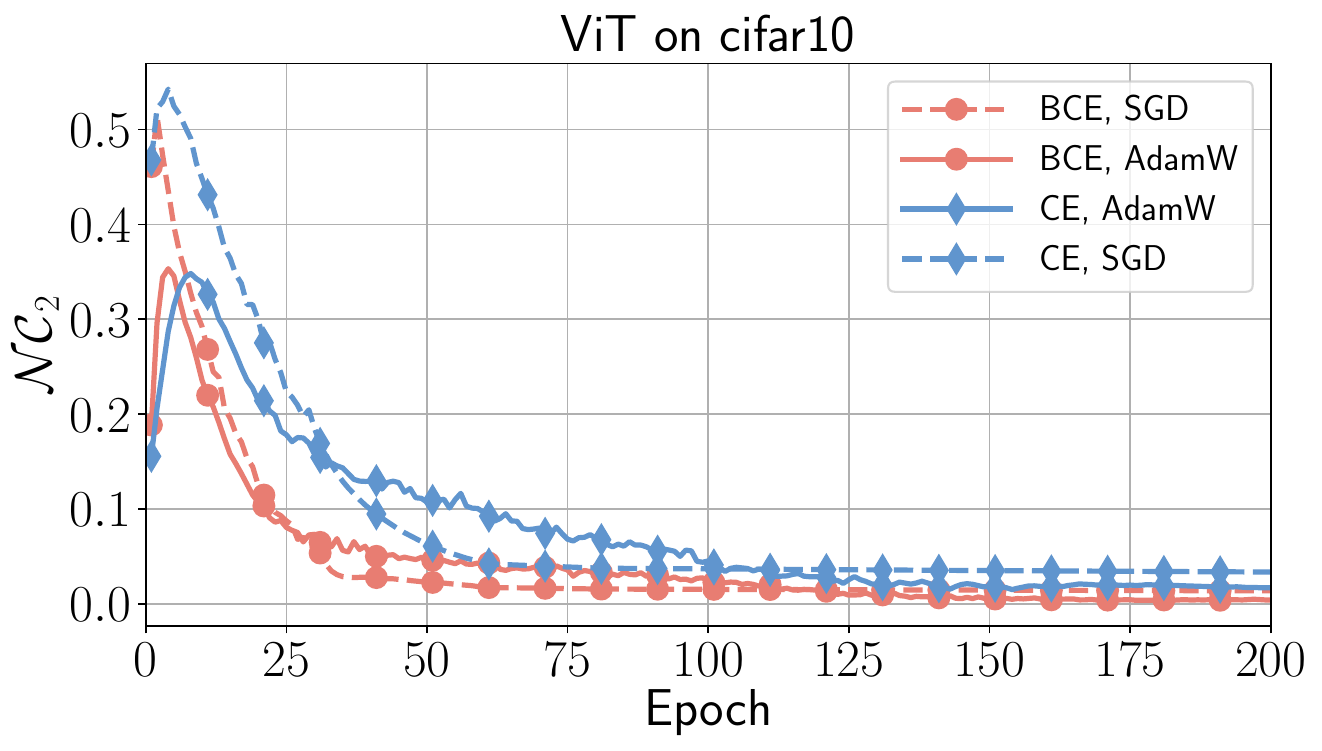}
		\includegraphics[width=0.32\linewidth,max width=\linewidth,max height=0.74\textheight,keepaspectratio]{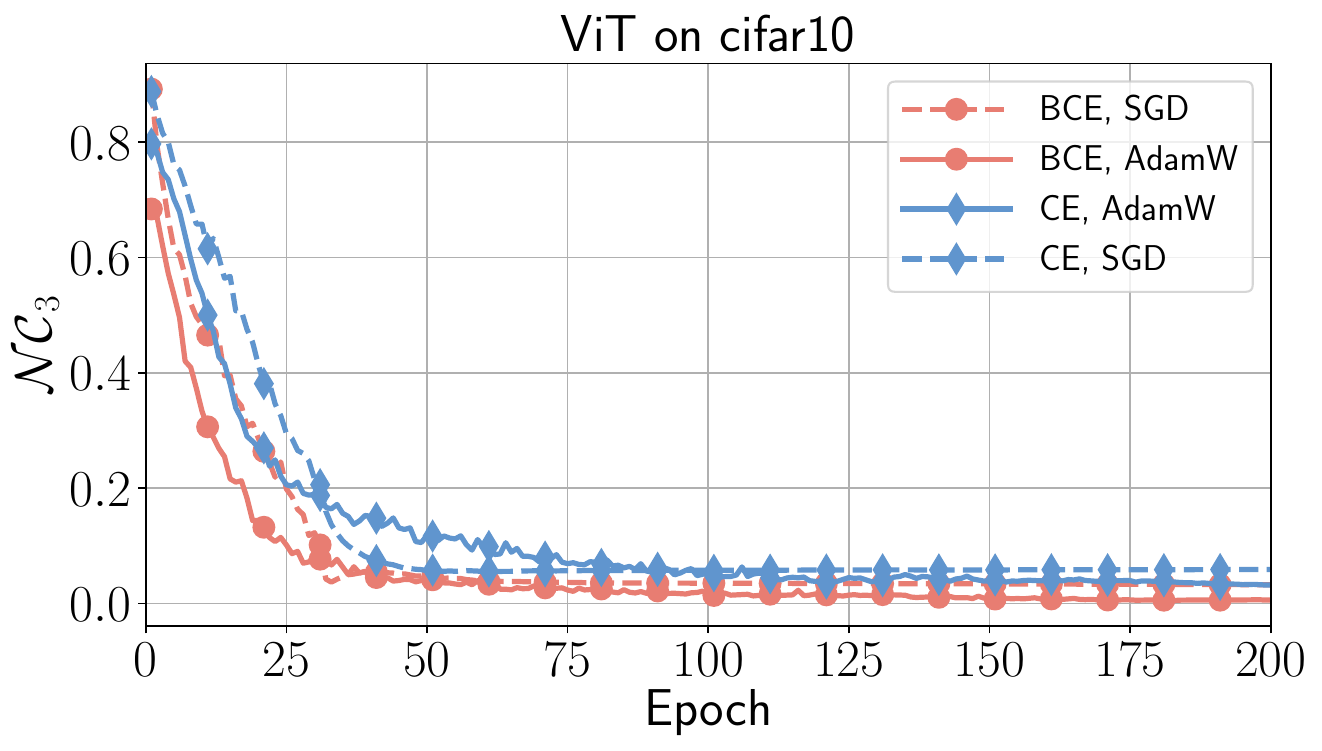}
		\label{fig_vit_cifar10_NC}}
	\caption{The evolution of the three NC metrics in the training of ResNet18, ResNet50, DenseNet121, and ViT on the training data of CIFAR10 driven by the normalized CE and BCE losses with $\gamma = 8$ using SGD and AdamW, respectively}
	\label{fig_NC_result_of_resnet_and_densenet_cifar10}
\end{figure}
\subsection{Minima of the normalized CE and BCE}

{\color{mycolor} We first verify that the normalized CE and BCE losses can lead to the NC geometry predicted by Theorems \ref{them:CE collapse} and \ref{them:BCE collapse}. We trained ResNet18, ResNet50, DenseNet121, and ViT on MNIST, CIFAR10, and CIFAR100 using SGD and AdamW optimizers when the scale factor $\gamma=8$. 
We applied the metrics of $\mathcal {NC}_1, \mathcal {NC}_2$, and $\mathcal {NC}_3$ defined in~\citet{Zhu2021_geometric} and~\citet{liu2023generalizing} to measure the evolution of NC during the training. 
The smaller values indicate the stronger NC properties.

Fig. \ref{fig_NC_result_of_resnet_and_densenet_cifar10} shows the evolution of the three NC metrics on the training data of CIFAR10. For different models and optimizers, both the CE and BCE steadily reduce the NC metrics during the training, which verifies that the two normalized losses can lead to NC when approaching their global minima. 
The results are consistent with our theorems, where samples from the same class collapse to their class center, class centers align with their corresponding classifier vectors, and the different class centers form an ETF structure.

Although the sample features deduced by the CE and BCE share the same geometric structure at their respective minima, the BCE typically reaches this structure faster during the early training. 
As Figure \ref{fig_NC_result_of_resnet_and_densenet_cifar10} shows, within the first $50$ epochs, the NC metrics for the BCE are generally lower than those for the CE, suggesting that the BCE more efficiently pushes the feature distribution toward the global NC structure, rather than only improving sample-level classification. 
This advantage of BCE comes from its explicit constraints on the positive and negative decision scores of samples. Compared with CE, which mainly relies on the relative ordering of the decision scores for each sample, BCE uses the classifier biases as the thresholds to uniformly distinguish the positive and negative decision scores across the entire dataset.

Similar results can be observed on MNIST and CIFAR100, as reported in Sec. \ref{NC_visualization} in Appendix. Overall, these experiments support the minima analysis in the normalized spaces. Both the CE and BCE can induce NC, while the BCE tends to reach the desired compact and separable feature geometry more rapidly.}

\subsection{Convergence of the normalized CE and BCE}\label{gamma_decay_experiment_icml}
Theorems \ref{them:CE_decay_of_decision_score} and \ref{them:BCE_decay_of_decision_score} describe the convergence behaviors of the CE and BCE.
To directly verify these two theorems, we set the NC structure as the stopping criterion, and train ResNet18 on MNIST with various scale factor $\gamma$. In these experiments, the stopping criterion is set as
\begin{align}
    &\big|\boldsymbol{w}_k^{\top}\boldsymbol{h}_i^{(k)}-1\big| \leq t,~~ \forall k\in[K],~\forall i\in[n], \label{gamma_stop_condition_1}\\
  \text{and} \quad &\Big|\boldsymbol{w}_j^{\top}\boldsymbol{h}_i^{(k)}+\frac{1}{K-1}\Big| \leq t,~~ \forall j\neq k\in[K], \label{gamma_stop_condition_2}
\end{align}
where for $\gamma < 5$, $t = 0.001$, and for $\gamma\geq 5$, $t=0.0035$.
Moreover, only SGD optimizer was applied and the learning rate was fixed at $0.01$.
Table \ref{gamma_theorem_table_icml} 
shows the epoch numbers required by the models trained with the CE and BCE losses when they satisfying the stopping criterion for $\gamma > 1$.

For the cases with $\gamma < 0.5$ and $\gamma > 5$, we took six and five data points to respectively fit the dependence of epoch number on $\gamma$ using linear, quadratic, and cubic polynomial functions, as well as exponential and logarithmic functions, while the remaining data points were used to evaluate the prediction quality of the fitted functions. 
For $\gamma > 5$, Table \ref{big_gamma_fit_results} shows the fitting results, together with the metrics of $R^2$ and RMSE on the fitting and prediction data. 
It can be observed that for the CE, although the exponential function exhibits weaker performance metrics than the quadratic and cubic polynomial functions on the fitting data, it better reflects the varying trend of epoch number on the prediction data. For the BCE, the exponential function achieves the best fitting results among the five types of functions, both on the fitting and prediction data. These results indicate that when $\gamma$ is relatively large, as it increases, the number of epochs required for the CE and BCE convergence grows exponentially, meaning their convergence rates decay exponentially and aligning with Theorems \ref{them:CE_decay_of_decision_score} and \ref{them:BCE_decay_of_decision_score}.

Moreover, in the two fitted exponential functions, the coefficient ($0.52$) of BCE is less than that ($0.87$) of CE, indicating that the BCE converges faster than the CE in the large feature spaces with larger scale factor $\gamma$.
The results for the case with $\gamma < 0.5$ are presented in Sec. \ref{gamma_decay_experiment} of Appendix, which also align with the theorems. 

\begin{table}[tbp]
    \centering
    \color{black}
    \caption{The number of training epochs required by ResNet18 to reach the stopping criterion of \textit{Eqs.} (\ref{gamma_stop_condition_1}) and (\ref{gamma_stop_condition_2}) on MNIST trained by the normalized CE and BCE with different $\gamma > 1$. ``$*$'' indicating that the data is used for fitting the functions of epoch number on $\gamma$. {\color{mycolor} When $\gamma \geq 10$, training ResNet18 with the CE loss requires more than $4,000$ epochs to converge, so the experiment was not performed}}\label{gamma_theorem_table_icml}
    \tabcolsep=0.1cm
    \begin{tabular}{c ccccccccccccccccccc}
        \toprule
        $\gamma$ & $2$ & $3$ & $4$ & $5$ & $6^*$ & $6.5$ & $7^*$ & $7.5$ & $8^*$ & $8.5$ & $9^*$ & $9.5^*$ & $10^*$ & $11$ & $12^*$ & $13$ & $14^*$ & $15$ & $16^*$ \\
        {CE} & $28$ & $31$ & $64$ & $31$ & $77$ & $145$ & $180$ & $308$ & $435$ & $741$ & $1448$ & $2083$ & - & - & - & - & - & - & -  \\
        {BCE} & $24$ & $22$ & $22$ & $14$ & $15$ & - & $21$ & - & $37$ & - & $53$ & - & $83$ & $137$ & $243$ & $411$ & $737$ & $1165$ & $2076$ \\
        \bottomrule
    \end{tabular}
\end{table}

\begin{table}[!htp]
    \centering
    \caption{Fitted functions and their evaluation results for the epoch number used by the CE and BCE with $\gamma > 5$}
    \label{big_gamma_fit_results}
    \begin{tabular}{c|c|cccc}
        \toprule
        \multirow{2}{*}{Loss} & \multirow{2}{*}{Fitted function} & \multicolumn{2}{c}{Fitting data} & \multicolumn{2}{c}{Predicted data} \\
        \cmidrule(lr){3-4} \cmidrule(lr){5-6}
        &  & $R^2 \uparrow$ & RMSE $\downarrow$& $R^2 \uparrow$ & RMSE $\downarrow$ \\
        \hline
        \multirow{5}{*}{CE} & $568.39\gamma-3645.68$ & $0.8556$ & $299.08$ & $-0.5945$ & $317.58$ \\
        & $236.01\gamma^2-3098.39\gamma+10205.49$ & $0.9939$ & $61.59$ & $0.7463$ & $126.69$ \\
        & $40.65\gamma^3-706.28\gamma^2+4076.80\gamma-7729.46$ & $\boldsymbol{0.9969}$ & $\boldsymbol{43.85}$ & $0.9038$ & $77.99$ \\
        & $0.54\exp(0.87\gamma)-61.83$ & $0.9932$ & $64.98$ & $\boldsymbol{0.9253}$ & $\boldsymbol{68.72}$ \\
        & $4.22\log(0.32\gamma)-3.01$ & $0.8094$ & $343.40$ & $-0.9326$ & $349.63$ \\
        \hline
        \multirow{5}{*}{BCE} & $236.60\gamma-2204.00$ & $0.7714$ & $364.33$ & $0.5566$ & $849.72$ \\
        & $52.14\gamma^2-1014.83\gamma+4887.43$ & $0.9812$ & $104.54$ & $0.8140$ & $550.31$ \\
        & $7.61\gamma^3-221.98\gamma^2+2171.11\gamma-7027.87$ & $0.9996$ & $15.55$ & $0.8837$ & $435.17$ \\
        & $0.49\exp(0.52\gamma)-3.79$ & $\boldsymbol{0.9999}$ & $\boldsymbol{7.56}$ & $\boldsymbol{0.9128}$ & $\boldsymbol{376.77}$ \\
        & $2.58\log(0.25\gamma)-2.11$ & $0.6871$ & $426.22$ & $0.4633$ & $934.77$ \\
        \bottomrule
    \end{tabular}
\end{table}

\begin{figure}[!tbp]
	\centering
	\subfloat[ResNet18 trained on CIFAR10]{
		\includegraphics[width=0.98\linewidth,max height=0.70\textheight,keepaspectratio]{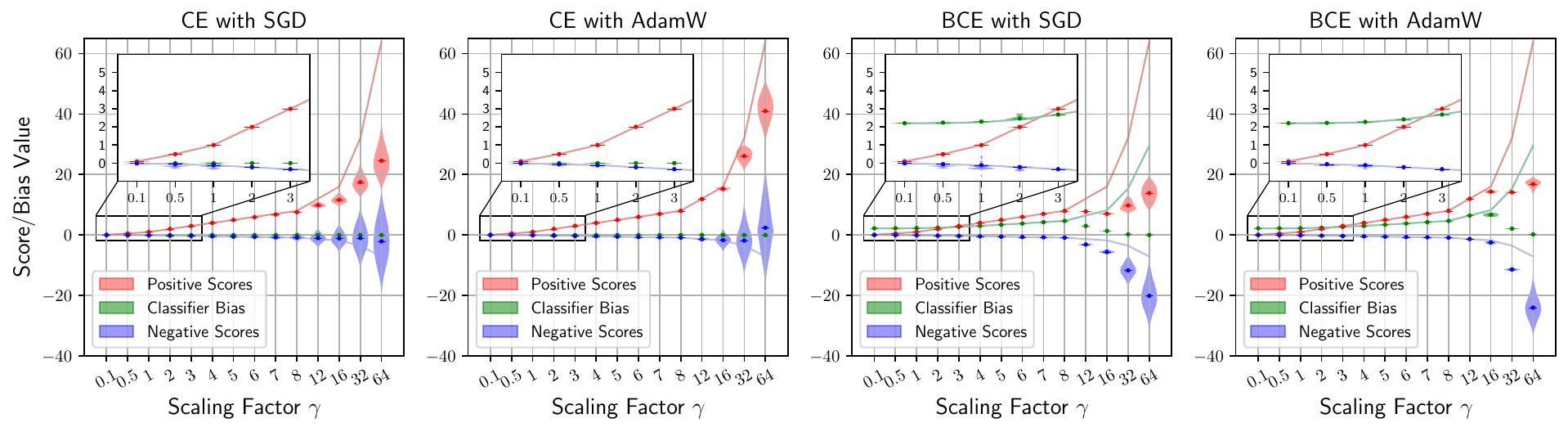}
	}
	
	\subfloat[ViT trained on CIFAR10]{
		\includegraphics[width=0.98\linewidth,max height=0.70\textheight,keepaspectratio]{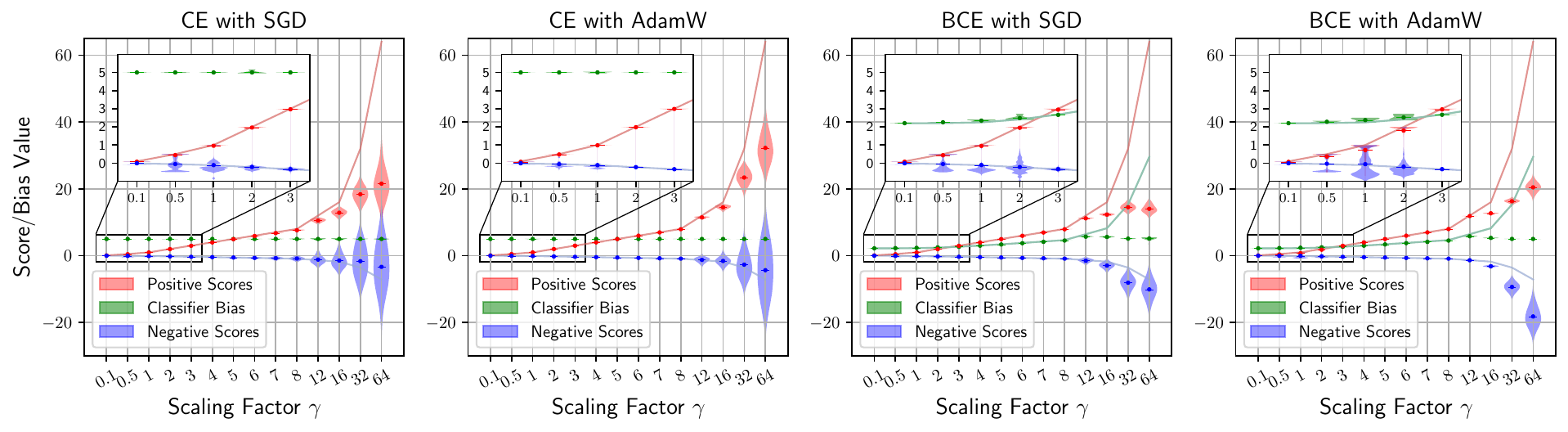}
	}
	
	\caption{The distribution of the final classifier biases and the positive/negative unbiased decision scores for 56 ResNet18s (top, without the final ReLU activation) and 56 ViTs (bottom) trained on CIFAR10, with various scale factors $\gamma$. 
		The maximum number of training epochs is fixed at 200.
		The initial mean ($\bar{b} = \frac{1}{K}\sum_{j=1}^K b_j$) of the biases is $0$ for ResNet18 and $5$ for ViT, and the initial variance is both $0$. The solid line represents the theoretical extremum when achieving NC}
	\label{resnet18_vit_cifar10_gamma}
\end{figure}

\subsection{Classification accuracy vs. feature property}
For a model, its classification and feature learning perform differently in the normalized space with different scale factor $\gamma$, as we analyzed in Sec. \ref{section_normalization_practical}.
To verify this difference, using the CE and BCE with SGD and AdamW, we trained 56 ResNet18s and 56 ViTs on CIFAR10 with 200 epochs by setting $\gamma$ to 0.1, 0.5, 1, 2, 3, 4, 5, 6, 7, 8, 12, 16, 32, and 64, respectively.
The detailed experimental settings are shown in Table \ref{tab_experimental_setting} in Appendix. 
Fig. \ref{resnet18_vit_cifar10_gamma} shows the distributions of the final unbiased decision scores and classifier biases on the training data, while the red, blue, and green solid lines represent the theoretical values of the final positive/negative unbiased decision scores and classifier biases for different $\gamma$.

As Fig. \ref{resnet18_vit_cifar10_gamma} shows, when $\gamma \leq 2$, although the mean of all positive/negative unbiased decision scores eventually aligns with their theoretical final values, their variance is relative large, particularly for the negative unbiased scores of ViT, which indicates the losses has not fully converged to their minima with 200 epochs.
When $3 \leq \gamma \leq 8$, all the final unbiased decision scores fall along their theoretical extreme values on the corresponding real lines, as well as the final classifier biases for the BCE, which indicates that the losses has basically converged at this stage.
As $\gamma$ continues to increase to $12$ or more, the unbiased decision scores of the trained models gradually deviate from their theoretical extremum, and they are distributed over an increasingly larger areas, particularly for models trained with the CE, with an even broader distribution, which indicates that the losses have totally not converged.

Since the substantial role of classifier biases in the BCE during the model training, when $\gamma \leq 8$, they eventually converges to their theoretical extremum. In contrast, the final biases of the CE are almost entirely dependent on their initial value, i.e. $0$ for ResNet18 and $5$ for ViT in the experiments.

\begin{tableorg}[!t]
\tablebodyfont
\caption{The classification accuracy (\%) and feature properties of ResNet18 (without the final ReLU) with different $\gamma$ on CIFAR10. The accuracies of $\mathcal A$ and $\mathcal A^*$ are calculated using the decision scores $\{\gamma\bm w_j^\top\bm h^{(k)}-b_j\}_{j=1}^K$ and the unbiased ones $\{\gamma\bm w_j^\top\bm h^{(k)}\}_{j=1}^K$, respectively. The higher $\mathcal E_{\text{com}}$ and $\mathcal E_{\text{sep}}$ respectively indicate the better compactness and separability of features}
\label{tab_cifar10_resnet18_reslts}
\begin{minipage}{\linewidth}\centering\textit{(a) Results on the training data of CIFAR10 --- CE with SGD.}\par\smallskip
	\centering
\begin{tabular}{c|ccccccc}
\hline
$\gamma$ & $0.1$ & $0.5$ & $1$ & $2$ & $16$ & $32$ & $64$ \\
$\mathcal A$ & $26.24$ & $54.28$ & $79.50$ & $99.97$ & $100.0$ & $100.0$ & $100.0$ \\
$\mathcal A^*$ & $27.27$ & $57.00$ & $80.01$ & $99.97$ & $100.0$ & $100.0$ & $100.0$ \\
$\mathcal E_{\text{com}}$ & $0.965$ & $0.995$ & $0.998$ & $0.999$ & $0.862$ & $0.789$ & $0.766$ \\
$\mathcal E_{\text{sep}}$ & $0.526$ & $0.554$ & $0.555$ & $0.555$ & $0.394$ & $0.288$ & $0.221$ \\
\hline
\end{tabular}\par\medskip\par
\end{minipage}

\begin{minipage}{\linewidth}\centering\textit{(a) Results on the training data of CIFAR10 --- BCE with SGD.}\par\smallskip
	\centering
\begin{tabular}{c|ccccccc}
\hline
$\gamma$ & $0.1$ & $0.5$ & $1$ & $2$ & $16$ & $32$ & $64$ \\
$\mathcal A$ & $25.47$ & $52.07$ & $76.73$ & $93.79$ & $100.0$ & $100.0$ & $100.0$ \\
$\mathcal A^*$ & $27.77$ & $54.75$ & $77.76$ & $93.78$ & $100.0$ & $100.0$ & $100.0$ \\
$\mathcal E_{\text{com}}$ & $0.965$ & $0.995$ & $0.997$ & $0.999$ & $0.973$ & $0.899$ & $0.842$ \\
$\mathcal E_{\text{sep}}$ & $0.526$ & $0.554$ & $0.555$ & $0.553$ & $0.410$ & $0.354$ & $0.313$ \\
\hline
\end{tabular}\par\medskip
\par
\end{minipage}

\begin{minipage}{\linewidth}\centering\textit{(b) Results on the test data of CIFAR10 --- CE with SGD.}\par\smallskip
	\centering
\begin{tabular}{c|ccccccc}
\hline
$\gamma$ & $0.1$ & $0.5$ & $1$ & $2$ & $16$ & $32$ & $64$ \\
$\mathcal A$ & $25.81$ & $44.18$ & $61.54$ & $78.95$ & $78.91$ & $77.84$ & $77.46$ \\
$\mathcal A^*$ & $25.62$ & $46.94$ & $61.60$ & $78.94$ & $78.91$ & $77.79$ & $77.44$ \\
$\mathcal E_{\text{com}}$ & $0.889$ & $0.829$ & $0.824$ & $0.821$ & $0.725$ & $0.691$ & $0.685$ \\
$\mathcal E_{\text{sep}}$ & $0.515$ & $0.535$ & $0.535$ & $0.535$ & $0.361$ & $0.266$ & $0.204$ \\
\hline
\end{tabular}\par\medskip\par
\end{minipage}

\begin{minipage}{\linewidth}\centering\textit{(b) Results on the test data of CIFAR10 --- BCE with SGD.}\par\smallskip
	\centering
\begin{tabular}{c|ccccccc}
\hline
$\gamma$ & $0.1$ & $0.5$ & $1$ & $2$ & $16$ & $32$ & $64$ \\
$\mathcal A$ & $24.98$ & $44.12$ & $58.41$ & $76.76$ & $82.14$ & $81.69$ & $82.54$ \\
$\mathcal A^*$ & $26.55$ & $45.67$ & $59.71$ & $76.75$ & $82.12$ & $81.72$ & $82.56$ \\
$\mathcal E_{\text{com}}$ & $0.889$ & $0.824$ & $0.825$ & $0.837$ & $0.807$ & $0.749$ & $0.726$ \\
$\mathcal E_{\text{sep}}$ & $0.514$ & $0.535$ & $0.536$ & $0.534$ & $0.368$ & $0.318$ & $0.293$ \\
\hline
\end{tabular}\par\medskip\par
\end{minipage}
\end{tableorg}

Table \ref{tab_cifar10_resnet18_reslts} presents part of results for ResNet18 trained by CE and BCE with small and large $\gamma$ on the training and test data of CIFAR10, and more results can be found in Sec. \ref{sec_resnet_vit_on_cifar10-app} in Appendix.
The expressions of accuracy without the bias, $\mathcal A^*$, are presented in Sec. \ref{sec_classification_fea_supp} in Appendix.
When $\gamma = 0.1$ or $0.5$, both the accuracy $\mathcal A$ and the unbiased one $\mathcal A^*$ are very low, but they differ clearly, indicating the slight variance in the classifier biases significantly affect the classification. However, with $\gamma=0.1$ or $0.5$, the final feature properties are not so bad, as the compactness $\mathcal E_{\text{com}}$ and separability $\mathcal E_{\text{sep}}$ are high, although they have not yet reached their maximum. In contrast, when $\gamma = 32$ or $64$, the models' accuracies are very high reaching $100\%$ on the training data, while the intra-class compactness and inter-class separability of their features are comparatively poor and worse than that learned with small $\gamma$. 
As Table \ref{tab_cifar10_resnet18_reslts} shows, a similar phenomenon is also presented in the evaluation results on the test dataset. 

The above results are entirely consistent with the analytical conclusions in this paper for classification and feature learning: with a fixed training strategy in practice, classification performs well in a relatively large feature space, whereas in a very small feature space it is more susceptible to interference from random factors, leading to a marked reduction in the final accuracy. In contrast, feature learning that pursues intra-class compactness and inter-class separability requires an appropriately smaller feature space. In summary, a model's classification accuracy cannot reflect its performance in downstream tasks that require feature comparison.

\subsection{Performance of BCE and CE on LTR and OSR}\label{comparison_performance_BCE_CE}
The applications of the BCE in the face recognition have been explored by~\citet{wensphereface2} and~\citet{zhou2023uniface},
and we here explore its advantages over the CE on other more challenging tasks such as \textbf{long-tailed recognition (LTR)} and \textbf{open-set recognition (OSR)}. Based on practical experience, we set $\gamma=32$ in these two tasks.

\begin{table}[!t]
	\centering
	\makebox[\linewidth][c]{%
		\begin{minipage}[t]{0.48\linewidth}
			\centering
			\caption{LTR results (\%) on CIFAR10-LT and CIFAR100-LT}
			\label{tab4_long_tailed}
			\fontsize{8}{10}\selectfont
			\tabcolsep=0.8mm
			\begin{tabular}{c ccc ccc}
				\toprule
				\multirow{2}{*}{IF} & \multicolumn{3}{c}{CIFAR10} & \multicolumn{3}{c}{CIFAR100} \\
				\cmidrule(lr){2-4} \cmidrule(lr){5-7}
				& $10$ & $50$ & $100$ & $10$ & $50$ & $100$ \\
				\midrule
				CE & $93.68$ & $87.80$ & $83.37$ & $69.30$ & $55.15$ & $49.47$ \\
				BCE & $\bf{93.96}$ & $\bf{88.75}$ & $\bf{84.47}$ & $\bf{69.49}$ & $\bf{58.53}$ & $\bf{52.15}$ \\
				\bottomrule
			\end{tabular}
		\end{minipage}
		\hspace{0.01\linewidth}
		\begin{minipage}[t]{0.48\linewidth}
			\centering
			\caption{The OSR results under different setups} 
			\label{tab_osr_reslts}
			\fontsize{8}{10}\selectfont
			\tabcolsep=0.7mm
			\begin{tabular}{ccccccccc}
				\toprule
				& \multicolumn{2}{c}{MNIST} & \multicolumn{2}{c}{SVHN} & \multicolumn{2}{c}{CIFAR10} & \multicolumn{2}{c}{CIFAR+50} \\
				\cmidrule(lr){2-3} \cmidrule(lr){4-5} \cmidrule(lr){6-7} \cmidrule(lr){8-9}
				& CE & BCE & CE & BCE & CE & BCE & CE & BCE \\
				\midrule
				AUROC & $98.6$ & $\boldsymbol{99.2}$ & $94.3$ & $\boldsymbol{95.1}$ & $84.2$ & $\boldsymbol{85.8}$ & $87.8$ & $\boldsymbol{90.2}$ \\
				OSCR & $98.5$ & $\boldsymbol{99.0}$ & $92.4$ & $\boldsymbol{93.4}$& $81.9$ & $\boldsymbol{83.7}$ & $85.7$ & $\boldsymbol{88.5}$ \\
				\bottomrule
			\end{tabular}
		\end{minipage}%
	}
\end{table}

\textbf{LTR.}\quad
On the imbalanced datasets, CIFAT10-LT and CIFAR100-LT, when the imbalance factors (IF) of the training sets are $10$, $50$, and $100$, we trained ResNet32 using the normalized CE and BCE with SGD, respectively.
Table \ref{tab4_long_tailed} shows the classification results on their balanced test sets.
The BCE consistently achieves better LTR results on the six pairs of models.
We believe that, compared to the CE, which couples the $K$ decision scores of each sample into one Softmax, the BCE decouples them using $K$ Sigmoids, mitigating the imbalance effects caused by the imbalanced datasets and improving the LTR performance.

\textbf{OSR.}\quad
In the open-set experiments, we evaluate the performances of the CE and BCE on the MNIST, SVHN, CIFAR10 and CIFAR+50 datasets, and the models and training details follow APRL~\citep{chen2021_arpl}.
Table \ref{tab_osr_reslts} presents the OSR results in terms of the area under the receiver operating characteristic (AUROC) and the open set classification rate (OSCR). On these OSR experiments, the BCE consistently achieves higher performance across the two metrics. We believe that in the BCE, the uniform and explicit constraints of the classifier biases on the positive/negative decision scores are beneficial for learning a clear decision boundary for sample features, which in turn aids the models in discriminating unknown classes and enhances the performance in the OSR.

\section{Discussion, Limitation, and Future Work}\label{discussion}

\textbf{Weight decay factor}.\quad
In Euclidean space without normalization on sample features or classifier vectors, $L_2$ regularization with weight decay factor $\lambda$ are typically added on the CE and BCE losses, which can also constrain the features into a bounded space.
Intuitively, a small $\lambda$ results in a large feature space, while a large $\lambda$ leads to a small one.
In \citet{li2025_bce}, we have compared the CE and BCE in Euclidean space but did not reveal the difference of classification and feature learning.
We conjecture that, similar to the scale factor $\gamma$ in normalized space, when $\lambda\geq0$ is very small,
it is difficult for the losses to reach their minima, resulting in poor feature properties;
when $\lambda$ is very large, the small theoretical gap between the positive and negative decision scores will harm the classification.
However, in Euclidean space, it is not easy to rigorously analyze the losses' convergence behavior.

\textbf{Temperature coefficient}.\quad
In contrastive learning (CL), a temperature $\tau$ is applied on the denominator of the feature cosine similarity.
In CL, the Softmax-based losses are typically applied, similar to the CE loss in classification.
When the CL losses reach their minima,
the scaled feature similarity $\frac{1}{\tau}\langle\bm h, \bm h_*\rangle$ of any positive or negative sample pair converge to fixed values, according to NC studies in~\citet{graf2021dissecting} and~\citet{koromilas2024bridging}.
We here conjecture that the impact of the temperature $\tau$ on the convergence rate of Softmax-based losses parallels that of the scale factor $\gamma$ on the convergence rate of CE loss in classification,
i.e., too large or too small $\tau$ would result in slow convergence of the CL losses and thereby poor feature properties.

\textbf{Minima of losses within more challenging scenarios}.\quad
In more challenging scenarios, such as low-dimensional spaces where the feature dimension is smaller than the number of classes (i.e.,
$d<K$) and in class-imbalanced settings, theoretically analyzing the minima and convergence of the losses becomes more difficult.
Currently, we are investigating the minima of the BCE loss on imbalanced datasets.

\textbf{Feature learning in other scenarios.}\quad
In this paper, within the normalized, supervised, closed-set, and high-dimensional feature spaces, we analyze the convergence behavior of the CE and BCE, to compare classification and feature learning.
When the setting changes, the convergence behavior of the losses may also change, thereby altering the conclusions regarding the two tasks.
However, we conjecture that in both supervised and self-supervised contrastive learning (CL), as scale factor $\gamma$ linearly increases, the convergence rate of the existing Softmax-based losses decays exponentially, then, too large feature spaces are less favorable for the CL.

\section{Conclusions}\label{Conclusion}
This paper conducts an in-depth comparison for classification and feature learning in normalized space, by theoretically analyzing the minima and convergence of CE and BCE losses.
While the CE has been widely used for classification, the BCE is specially designed from the first principle of feature learning for the first time.
Although both the CE and BCE can lead to neural collapse at their global minima, the former constrains the relative values of the positive and negative decision scores for samples through probabilistic constraints, whereas the latter imposes uniform thresholds on the absolute values of the positive and negative decision scores for all samples. By analyzing the convergence behavior of the CE and BCE, we find that the scale factor $\gamma$, i.e., the size of the normalized space, can significantly affect classification and feature learning of a model trained by fixed strategy. A larger $\gamma$ allows the trained model to achieve higher classification accuracy even without well-structured feature geometry, while a small $\gamma$ makes classification unstable, yet may yield samples with higher intra-class compactness and inter-class separability in feature space. 
Experiments on CNNs and ViTs support our conclusions. Furthermore, as the BCE could obtain better features for samples, it outperforms the CE in more challenging tasks such as long-tailed and open-set recognitions.

\backmatter

\begin{appendices}
	\numberwithin{figure}{section}
	\numberwithin{table}{section}
	\numberwithin{equation}{section}
	\numberwithin{theorem}{section}
	\newpage
	\input{appendix}
\end{appendices}

\bmhead{Acknowledgements}
The authors acknowledge the support provided by the Intelligent Computing Center of Shenzhen University.

\bmhead{Funding}
This work was supported by the National Natural Science Foundation of China
(Grant Nos. 62576217 and 62576216),
Guangdong Provincial Key Laboratory
(Grant No. 2023B1212060076),
and the Natural Science Foundation of Ningxia Province
(Grant No. 2025AAC020002).

\bmhead{Data availability}
The datasets used in this study are publicly available.
CIFAR-10 and CIFAR-100 can be obtained from
\url{https://cave.cs.toronto.edu/kriz/cifar-10-python.tar.gz}
and \url{https://cave.cs.toronto.edu/kriz/cifar-100-python.tar.gz}, respectively,
MNIST from \url{https://storage.googleapis.com/cvdf-datasets/mnist/},
SVHN from \url{http://ufldl.stanford.edu/housenumbers/},
and Tiny ImageNet from
\url{http://cs231n.stanford.edu/tiny-imagenet-200.zip}.
The corresponding dataset references are
\citet{krizhevsky2009_learning},
\citet{Lecun1998_gradient},
\citet{netzer2011reading},
and \citet{wu2017tiny}, respectively.

\bibliography{MLJ_references}

\end{document}

%% file: appendix.tex
\section{Neural Collapse and Feature Properties}
\subsection{Neural Collapse}

{Neural collapse} (NC) was first found in~\citet{Papyan2020_prevalence} that occurs in the terminal training phase of the classification model, manifesting as elegant geometric structures in the sample features and classifier vectors.
With NC, the geometric structure of sample features and classifier vectors manifests as follows:
\begin{itemize}
  \item For any class $k$, all the sample features $\big\{\bm h_i^{(k)}\big\}_{i=1}^{n_k}$ of this class converge to the feature center $\bar{\bm h}_k = \frac{1}{n_k}\sum_{i=1}^{n_k} \bm h_i^{(k)}$, i.e., $\bm h_i^{(k)} = \bm  h_j^{(k)}$, for $\forall i,j\in[n_k] = \{1,2,\cdots, n_k\}$;
  \item The feature center vectors of any two classes have the same angular distance and equal magnitudes, i.e., $\|\bar{\bm h}_k\| = \|\bar{\bm h}_\ell\|$ and $\frac{\langle\bar{\bm h}_k,\bar{\bm h}_\ell\rangle}{\|\bar{\bm h}_k\|\|\bar{\bm h}_\ell\|} = \frac{\langle\bar{\bm h}_{k'},\bar{\bm h}_{\ell'}\rangle}{\|\bar{\bm h}_{k'}\|\|\bar{\bm h}_{\ell'}\|}$, for $\forall k\neq\ell, k'\neq\ell'\in[K]$, forming an equiangular tight frame (ETF);
  \item The feature center vector of any class is parallel to its corresponding classifier vector with a fixed constant between them, i.e., $\exists\alpha>0$, such that $\bar{\bm h}_k = \alpha\bm w_k$, for $\forall k\in [K]$.
\end{itemize}
In~\citet{Zhu2021_geometric} and~\citet{Zhou2022_areall}, the authors defined metrics of $\mathcal {NC}_1$, $\mathcal {NC}_2$, $\mathcal {NC}_3$ to evaluate the NC properties.
\begin{align}
    \mathcal{NC}_1 &:= \frac{1}{K}\operatorname{trace}\left(\boldsymbol{\Sigma}_W \boldsymbol{\Sigma}_B^{\dagger}\right), \\
    \mathcal{NC}_2 &:= \bigg\|\frac{\tilde{\boldsymbol{W}}\tilde{\boldsymbol{W}}^{\top}}{\|\tilde{\boldsymbol{W}}\tilde{\boldsymbol{W}}^{\top}\|_{F}}-\frac{1}{\sqrt{K-1}}(\boldsymbol{I}_K-\frac{1}{K}\mathbf{1}_K\mathbf{1}_K^{\top})\bigg\|_{F}, \\
    \mathcal{NC}_3 &:=\bigg\|\frac{\boldsymbol{W}\tilde{\boldsymbol{H}}}{\|\boldsymbol{W}\tilde{\boldsymbol{H}}\|_{F}}-\frac{1}{\sqrt{K-1}}(\boldsymbol{I}_K-\frac{1}{K}\mathbf{1}_K\mathbf{1}_K^{\top})\bigg\|_{F},
\end{align}
where
\begin{align}
\bm{\Sigma}_{W} &= \frac{1}{\sum_k n_k}\sum_{k=1}^{K}\sum_{i=1}^{n_k}\left(\boldsymbol{h}_i^{(k)}-\bar{\boldsymbol{h}}_k\right)\left(\boldsymbol{h}_i^{(k)}-\bar{\boldsymbol{h}}_k\right)^{\top}, \nonumber \\
\bm{\Sigma}_{B} &= \frac{1}{K}\sum_{k=1}^{K}\left(\bar{\boldsymbol{h}}_k-\bar{\boldsymbol{h}}\right)\left(\bar{\boldsymbol{h}}_k-\bar{\boldsymbol{h}}\right)^{\top}, \nonumber \\
\tilde{\boldsymbol{W}} &= [\boldsymbol{w}_1 - \bar{\boldsymbol{w}},~ \boldsymbol{w}_2 - \bar{\boldsymbol{w}},~\cdots,~\boldsymbol{w}_K - \bar{\boldsymbol{w}}]^T 
\in \mathbb{R}^{K \times d}, \nonumber \\
\tilde{\boldsymbol{H}} &= [\bar{\boldsymbol{h}}_1 - \bar{\boldsymbol{h}},~\bar{\boldsymbol{h}}_2 - \bar{\boldsymbol{h}},~\cdots,~\bar{\boldsymbol{h}}_K - \bar{\boldsymbol{h}}] \in \mathbb{R}^{d \times K}, \nonumber 
\end{align}
and $\bar{\boldsymbol{w}} = \frac{1}{K} \sum_{k=1}^{K} \boldsymbol{w}_k$, $\bar{\bm h}_k = \frac{1}{n_k}\sum_{i=1}^{n_k} h_i^{(k)}$, $\bar{\bm h} = \frac{1}{\sum_k n_k}\sum_{k=1}^K\sum_{i=1}^{n_k} \bm{h}_i^{(k)}$.
In this paper, we take $\mathcal{NC}_1, \mathcal{NC}_2$, and $\mathcal{NC}_3$ to compare the evolution of NC of the CE and BCE in the normalized feature space.

\subsection{Classification}\label{sec_classification_fea_supp}

In our experiments, we apply various metrics to comprehensively compare the performance of the BCE and CE losses in the normalized feature space, i.e., classification accuracy $\mathcal A$, unbiased classification accuracy $\mathcal A^{*}$, features compactness $\mathcal E_{\text{com}}$, and features separability $\mathcal E_{\text{sep}}$. These metrics will be maximized when reaching NC.

For a classification task, suppose a dataset
\begin{align}
    \mathcal{D} = \bigcup_{k=1}^K\mathcal D_k = \bigcup_{k=1}^K\big\{\boldsymbol X_i^{(k)}\big\}_{i=1}^{n_k}
\end{align}
from $K$ categories, where $\boldsymbol X_i^{(k)}$ denotes the $i$th sample from the class $k$.
For any sample $\boldsymbol X_i^{(k)}$, a model $\mathcal {M}$ converts it into its feature $\boldsymbol h_i^{(k)} = \mathcal {M}(\boldsymbol X_i^{(k)}) \in \mathbb{R}^d$, where $d$ is dimension of the feature space.
To classify the sample, a linear, full connection classifier $\mathcal C=\big\{(\boldsymbol w_k, b_k)\big\}_{k=1}^K$ transform the feature into $K$ decision scores $\big\{\gamma\boldsymbol w_j^\top \boldsymbol h_i^{(k)} - b_j\big\}_{j=1}^K$, where $\{\bm w_k\}_{k=1}^K$ are classifier vectors and $\{b_k\}_{k=1}^K$ are classifier biases.
Then, the sample is classified into class $\hat{k}$,
\begin{align}
    \hat{k} = \arg\max_j \big\{\gamma\boldsymbol w_j^\top \boldsymbol h_i^{(k)} - b_j\big\}_{j=1}^K,
\end{align}
and correct classification for the sample is achieved when $k=\hat{k}$, which is equivalent to
\begin{align}
    \gamma\boldsymbol w_k^\top \boldsymbol h_i^{(k)} - b_k = \max\big\{\gamma\boldsymbol w_j^\top \boldsymbol h_i^{(k)} - b_j\big\}_{j=1}^K. \label{eq:classification_condition_2-supp}
\end{align}

\textbf{Classification accuracy.}\quad
The commonly used classification accuracy $\mathcal A$ is defined as
\begin{align}
    \mathcal {A}({\mathcal M, \mathcal C}) = \frac{|\mathcal D({\mathcal M, \mathcal C})|}{|\mathcal D|} \times 100\%, \label{eq_accuracy_appendix}
\end{align}
where
\begin{align}
    \mathcal D(\mathcal M, \mathcal C) = \bigcup_{k=1}^K\Big\{\boldsymbol X^{(k)}: k = \arg\max_{\ell}\{\gamma\boldsymbol w_\ell^\top \boldsymbol h^{(k)} - b_\ell\},
                    \boldsymbol X^{(k)}\in\mathcal D_k, \boldsymbol h^{(k)} = \mathcal M(\boldsymbol X^{(k)})\Big\},
\end{align}
consisting of all the samples correctly classified by $\mathcal M$ and $\mathcal C$ in $\mathcal D$.

\textbf{Unbiased classification accuracy.}\quad
Similarly, the unbiased classification accuracy $\mathcal{A}^{*}$ is computed without using the classifier biases $\{b_k\}_{k=1}^K$, i.e.,
\begin{equation}
    \mathcal {A}^{*}({\mathcal M, \mathcal C}) = \frac{|\mathcal D^{*}({\mathcal M, \mathcal C})|}{|\mathcal D|} \times 100\%, \label{eq_accuracy_unbiased}
\end{equation}
where
\begin{align}
    \mathcal D^{*}(\mathcal M, \mathcal C) = \bigcup_{k=1}^K\Big\{\boldsymbol X^{(k)}: k = \arg\max_{\ell}\{\gamma\boldsymbol w_\ell^\top \boldsymbol h^{(k)}\}, \boldsymbol X^{(k)}\in\mathcal D_k, \boldsymbol h^{(k)} = \mathcal M(\boldsymbol X^{(k)})\Big\}.
\end{align}
The difference between the classification accuracy $\mathcal A$ and unbiased one $\mathcal A^{*}$ reflects the affect of the variance of the biases $\{b_j\}_{j=1}^K$ to the classification.

\subsection{Feature Properties}
There is a close and intricate relationship between the sample classification and their feature properties.
To precisely measure the properties of sample features, we define their intra-class compactness $\mathcal E_{\text{com}}$ and inter-class separability $\mathcal E_{\text{sep}}$,
\begin{align}
    &\mathcal E_{\text{com}} = \frac{1}{2}\bigg[\frac{1}{K}\sum_{k=1}^K \bigg(\frac{1}{n_k^2} \sum_{i=1}^{n_k}\sum_{i'=1}^{n_k}
    \big\langle\boldsymbol h_i^{(k)} , \boldsymbol h_{i'}^{(k)} \big\rangle\bigg) +1 \bigg], \label{eq_compactness_supp} \\
    &\mathcal E_{\text{sep}} = \frac{1}{2}\bigg[1-\frac{1}{K(K-1)} \sum_{k=1}^K\sum_{k'=1\atop k'\neq k}^K\bigg(\frac{1}{n_k}\frac{1}{n_{k'}}
    \sum_{i=1}^{n_k}\sum_{i'=1}^{n_{k'}}\big\langle\boldsymbol h_i^{(k)}, \boldsymbol h_{i'}^{(k')}\big\rangle\bigg)\bigg]. \label{eq_distinctiveness_supp}
\end{align}
where $\bar{\boldsymbol h} = \frac{1}{|\mathcal D|}\sum_{k=1}^{K}\sum_{i=1}^{n_k}\boldsymbol h_i^{(k)}$ is the global feature center.

Due to the properties of neural collapse, when approaching the minima of the CE or BCE, the intra-class compactness $\mathcal E_{\text{com}}$ might be higher than $\frac{1}{2} - \frac{1}{2(K-1)}$, and the inter-class separability $\mathcal E_{\text{sep}}$ might be lower than $\frac{1}{2} + \frac{1}{2(K-1)}$, for the model $\mathcal M$ and classifier $\mathcal C$ trained on the dataset $\mathcal D$.


\section{Experiments} \label{additional experiments}
\subsection{Experimental Setting Details}
In Sec. \ref{Experiments}, we conducted extensive experiments with multiple models across various datasets to validate our theoretical findings and analyses. We present additional experimental details in Table \ref{tab_experimental_setting}.

\textbf{Neural collapse on balanced classification}. By default, to validate the phenomenon of neural collapse and the convergence of models, we train ResNet18, ResNet50~\citep{He2016_deep}, and DenseNet121 on the MNIST~\citep{Lecun1998_gradient}, CIFAR10, and CIFAR100~\citep{krizhevsky2009_learning} by using settings-1 and 2, and we train ViT~\citep{dosovitskiy2020_vit} from scratch on the CIFAR10 dataset using settings-3 and 4, without applying additional regularization or data augmentation techniques during the training.

To assess classification performance on balanced datasets, we train ResNet18 and ResNet50 on the CIFAR10, CIFAR100, and Tiny-ImageNet~\citep{wu2017tiny} datasets using settings-5 and 6, employing the commonly used data augmentation techniques for a fair comparison. When training ViT models on the CIFAR10 and CIFAR100 datasets with the AdamW optimizer, we follow the weight decay setting of~\citet{s-chh_ViT_python}, and conduct a learning rate search based on setting-5 and 6. For training ViT on the Tiny-ImageNet dataset, we follow the network architecture and hyperparameter settings of~\citet{lee2021vision}. The classification results of the models trained using CE and BCE losses are shown in Table \ref{balanced_classification}.

\textbf{Long-tailed recognition}. The long-tailed datasets, CIFAR10-LT and CIFAR100-LT , are produced by sampling the training samples of the original datasets, using an exponential decay imbalance mode across classes, following the work of~\citet{cao2019learning}. For each of them, we produced three variants of long-tailed datasets by using three different imbalance factors (IF), $10, 50, ~\text{and} ~100$. The IF is defined as the number of training samples in the largest class divided by the smallest. The experimental setting details for long-tailed recognition can be found in the work of~\citet{alshammari2022long}.

\textbf{Open-set recognition}. We sample from original datasets to generate the corresponding known and unknown classes, a simple summary is provided: (1) For MNIST, SVHN~\citep{netzer2011reading} and CIFAR10, six known classes and four unknown classes are randomly sampled. (2) For the CIFAR+$50$ experiments, four classes are sampled from CIFAR10 for training and $50$ nonoverlapping classes are used as unknown classes, which are sampled from the CIFAR100 dataset. We use the area under the receiver operating characteristic (AUROC) curve~\citep{neal2018open} and open set classification Rate (OSCR)~\citep{chen2021_arpl} as evaluation metrics in the experiments. The experimental setting details for open-set recognition can be found in the work of~\citet{chen2021_arpl}.

\begin{table}[!t]
    \caption{Detailed experimental settings}
    \label{tab_experimental_setting}
    \centering
        \begin{tabular}{c|l||cccc||cc}
            \hline
            \multicolumn{2}{c||}{} & \multicolumn{4}{c||}{Neural Collapse} & \multicolumn{2}{c}{Classification} \\ \cmidrule(lr){3-8}
            \multicolumn{2}{c||}{} & setting-1 & setting-2 & setting-3 & setting-4 & setting-5 & setting-6 \\ 
            \hline
            \multirow{8}{*}{\rotatebox{90}{Hyper-parameter}} & epochs & 200 & 200 & 200 & 200 & 200 & 200 \\
            & optimizer & SGD & AdamW & SGD & AdamW & SGD & AdamW \\
            & batch size & 128 & 128 & 128 & 128 & 128 & 128 \\
            & learning rate & $0.01$ & $0.001$ & $0.03$ & $0.0003$ & $0.1$ & $0.005$ \\
            & learning rate decay & step & cosine & step & cosine & step & cosine \\
            & weight decay $\lambda$ & \ding{55} & \ding{55} & \ding{55} & \ding{55} & $0.0001$ & $0.05$ \\
            & warmup epochs & 0 & 0 & 0 & 0 & 0 & 0 \\ \hline
            \multirow{6}{*}{\rotatebox{90}{Data Aug.}}
            & random cropping & \ding{55} & \ding{55} & \ding{55} & \ding{55} & \ding{51} & \ding{51} \\
            & horizontal flipping & \ding{55} & \ding{55} & \ding{55} & \ding{55} & 0.5 & 0.5 \\
            & random rotation & \ding{55} & \ding{55} & \ding{55} & \ding{55} & 15 & 15 \\ \cmidrule(lr){2-8}
            &label smoothing & \ding{55} & \ding{55} & \ding{55} & \ding{55} & 0.1 & 0.1 \\
            & mixup alpha & \ding{55} & \ding{55} & \ding{55} & \ding{55} & 0.8 & 0.8 \\
            & cutmix alpha & \ding{55} & \ding{55} & \ding{55} & \ding{55} & 1.0 & 1.0 \\
            & mixup prob. & \ding{55} & \ding{55} & \ding{55} & \ding{55} & 0.8 & 0.8 \\ \cmidrule(lr){2-8}
            & normalization & \multicolumn{6}{c}{mean $= [0.4914, 0.4822, 0.4465]$, std $= [0.2023, 0.1994, 0.2010]$} \\ \hline
        \end{tabular}
\end{table}

\subsection{Neural Collapse of BCE and CE Losses} \label{NC_visualization}

This section presents more results for the neural collapse (NC) that support the conclusions drawn in the paper. These results are computed on the training data from the the datasets. As we have mentioned before, we take $\mathcal {NC}_1$, $\mathcal {NC}_2$, and $\mathcal {NC}_3$ presented in~\citet{Zhu2021_geometric} and~\citet{Zhou2022_areall} to evaluate the evolution of NC for the CE and BCE.

\begin{figure}[!t]
	\centering
        \subfloat[$\mathcal{NC}_1$, $\mathcal{NC}_2$, and $\mathcal{NC}_3$ of ResNet18 on MNIST]{
            \includegraphics[width=0.3\textwidth,max width=\linewidth,max height=0.74\textheight,keepaspectratio]{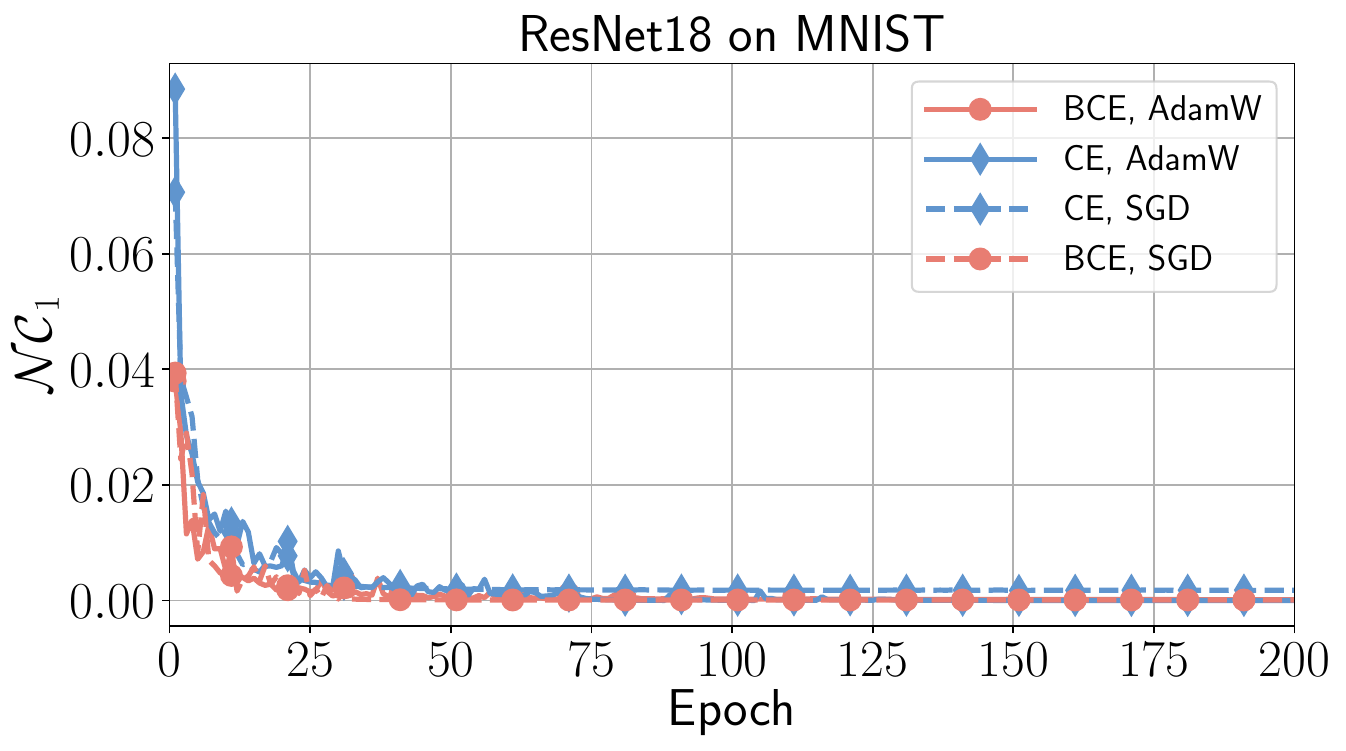}
            \includegraphics[width=0.3\textwidth,max width=\linewidth,max height=0.74\textheight,keepaspectratio]{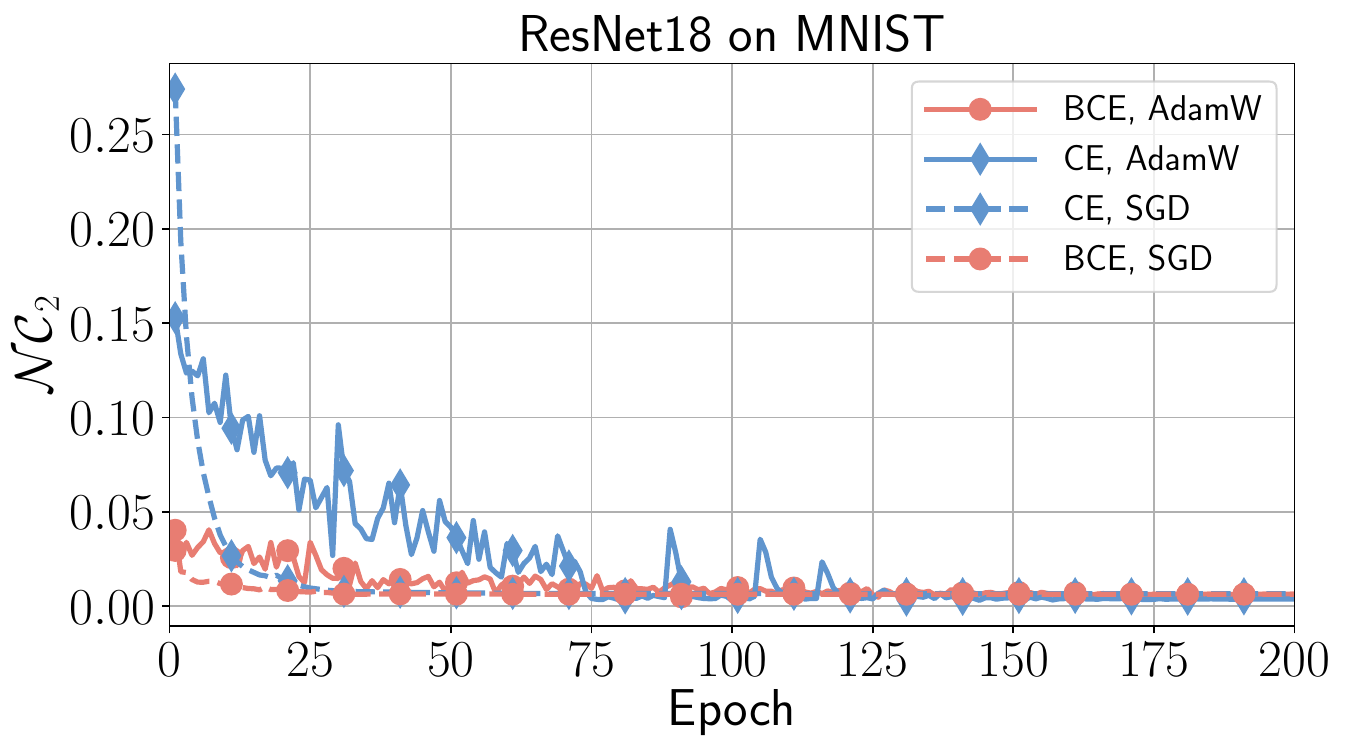}
            \includegraphics[width=0.3\textwidth,max width=\linewidth,max height=0.74\textheight,keepaspectratio]{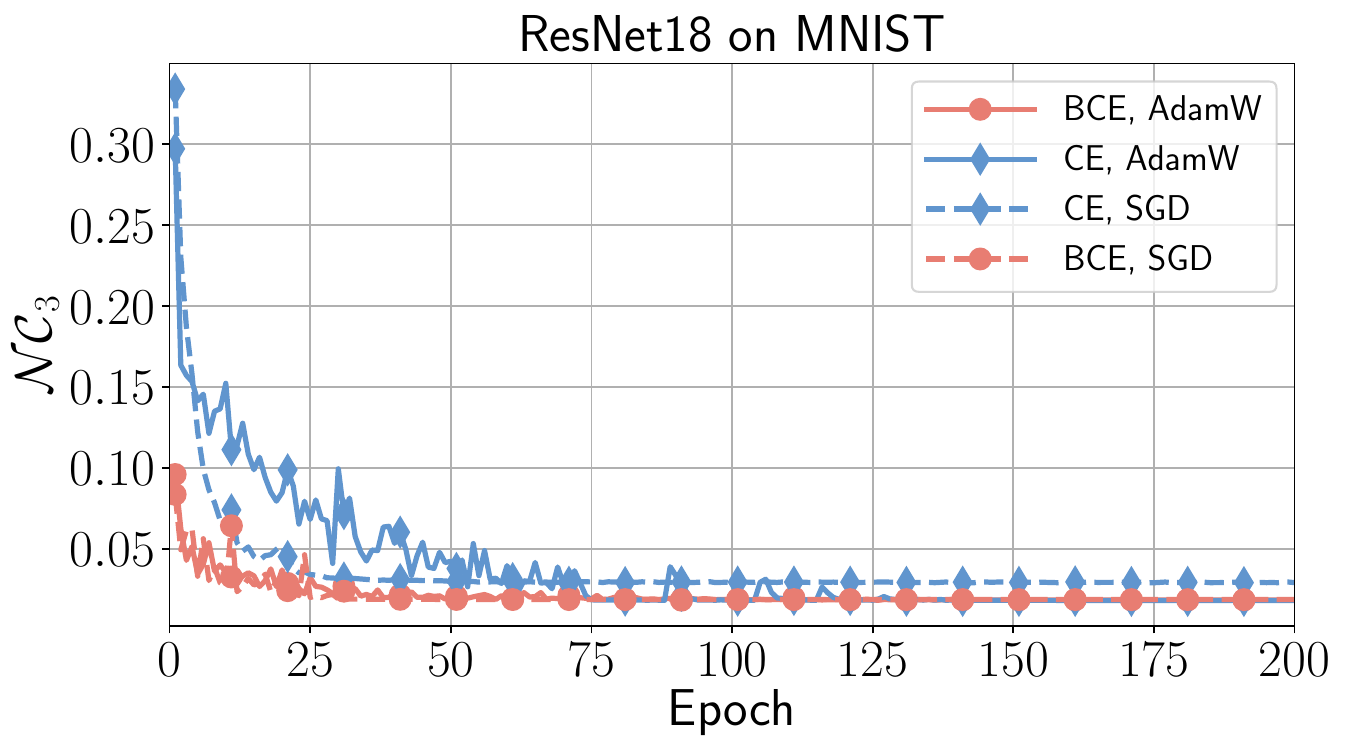}
            \label{fig_resnet18_mnist_NC-supp}}

        \subfloat[$\mathcal{NC}_1$, $\mathcal{NC}_2$, and $\mathcal{NC}_3$ of ResNet50 on MNIST]{
            \includegraphics[width=0.3\textwidth,max width=\linewidth,max height=0.74\textheight,keepaspectratio]{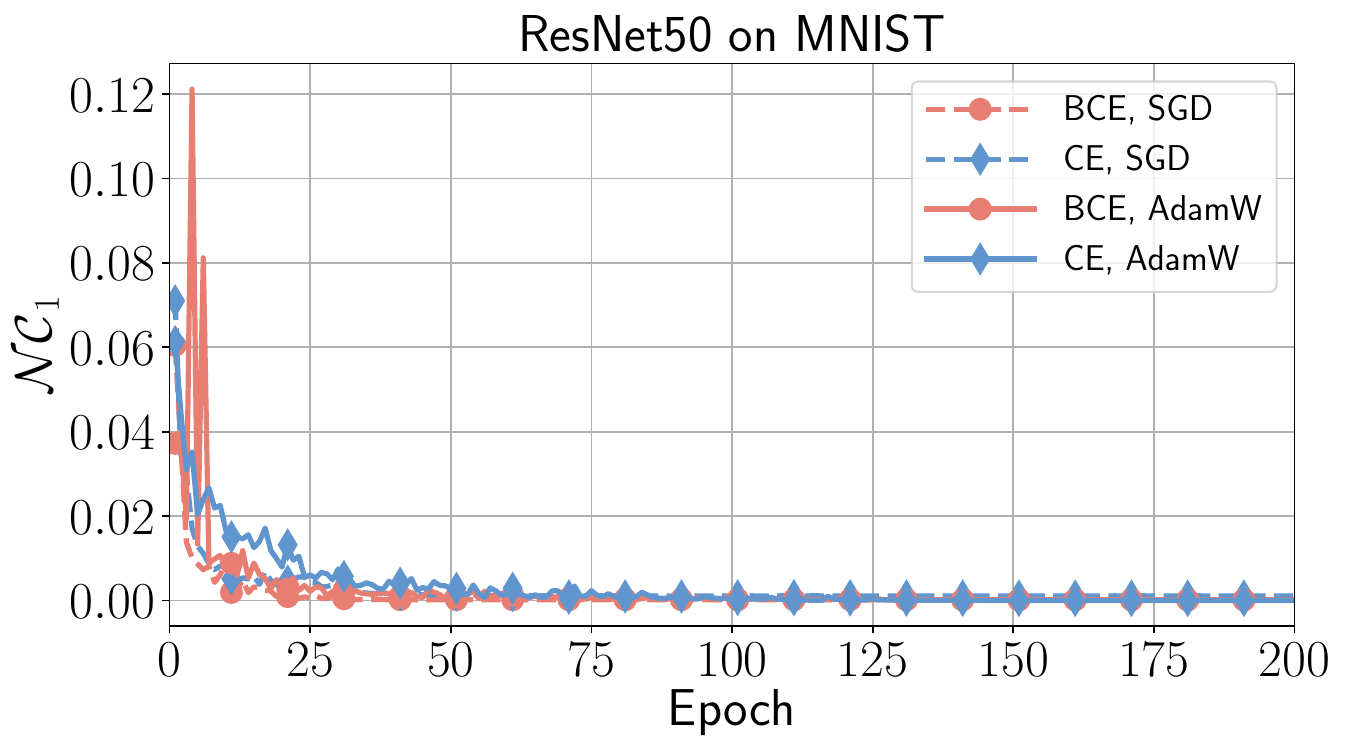}
            \includegraphics[width=0.3\textwidth,max width=\linewidth,max height=0.74\textheight,keepaspectratio]{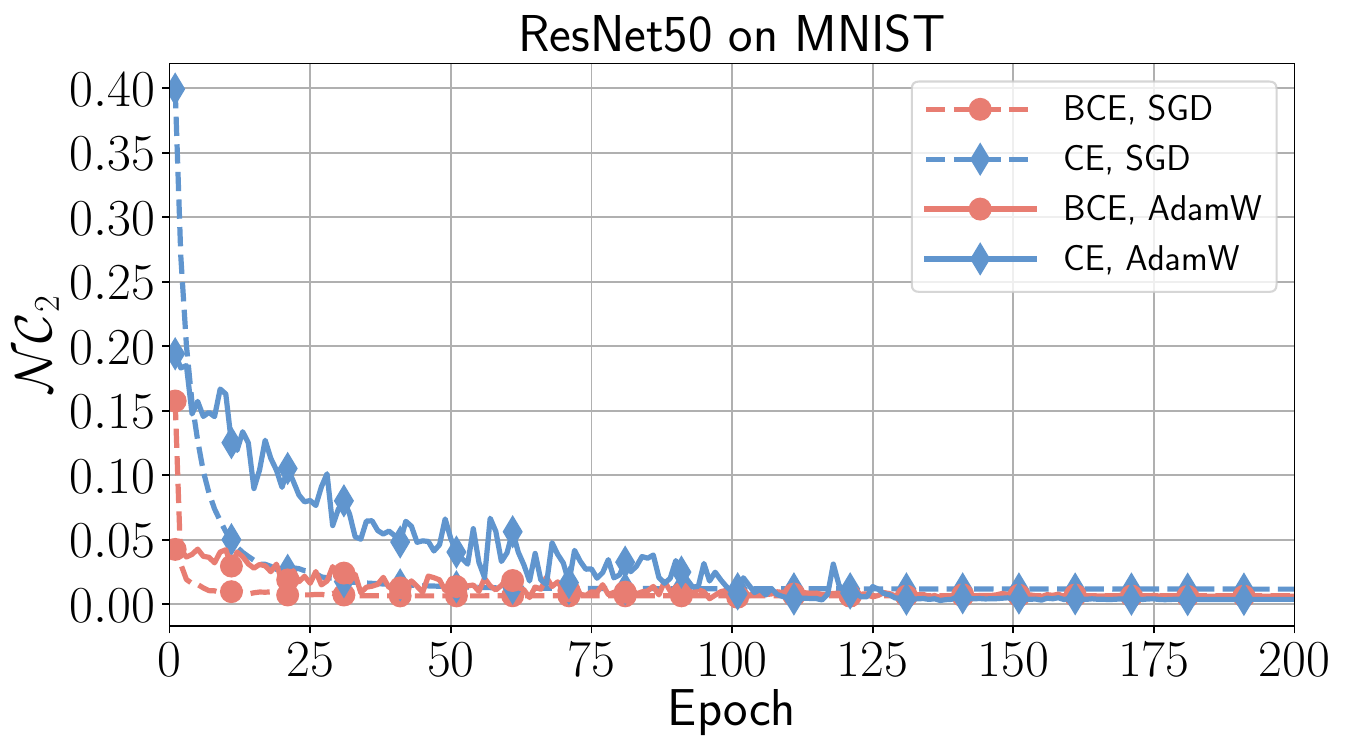}
            \includegraphics[width=0.3\textwidth,max width=\linewidth,max height=0.74\textheight,keepaspectratio]{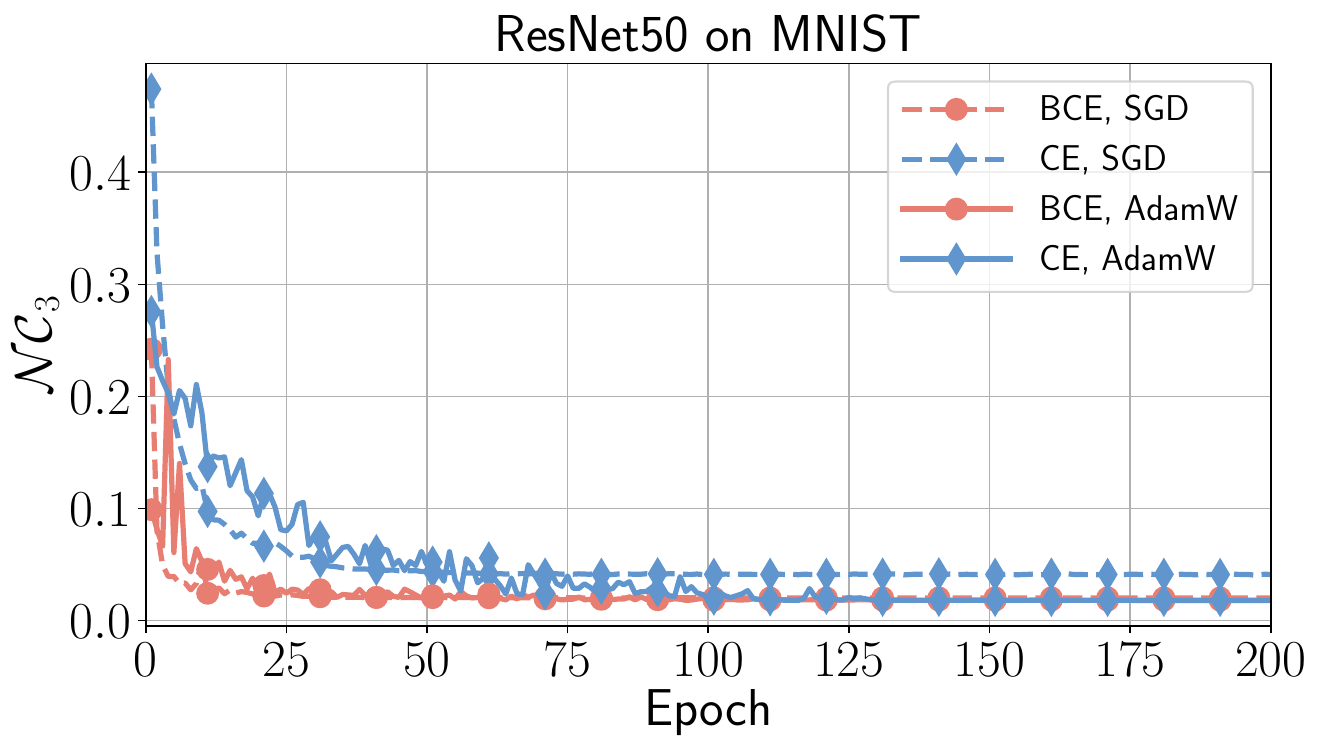}
            \label{fig_resnet50_mnist_NC-supp}}

        \subfloat[$\mathcal{NC}_1$, $\mathcal{NC}_2$, and $\mathcal{NC}_3$ of DenseNet121 on MNIST]{
            \includegraphics[width=0.3\textwidth,max width=\linewidth,max height=0.74\textheight,keepaspectratio]{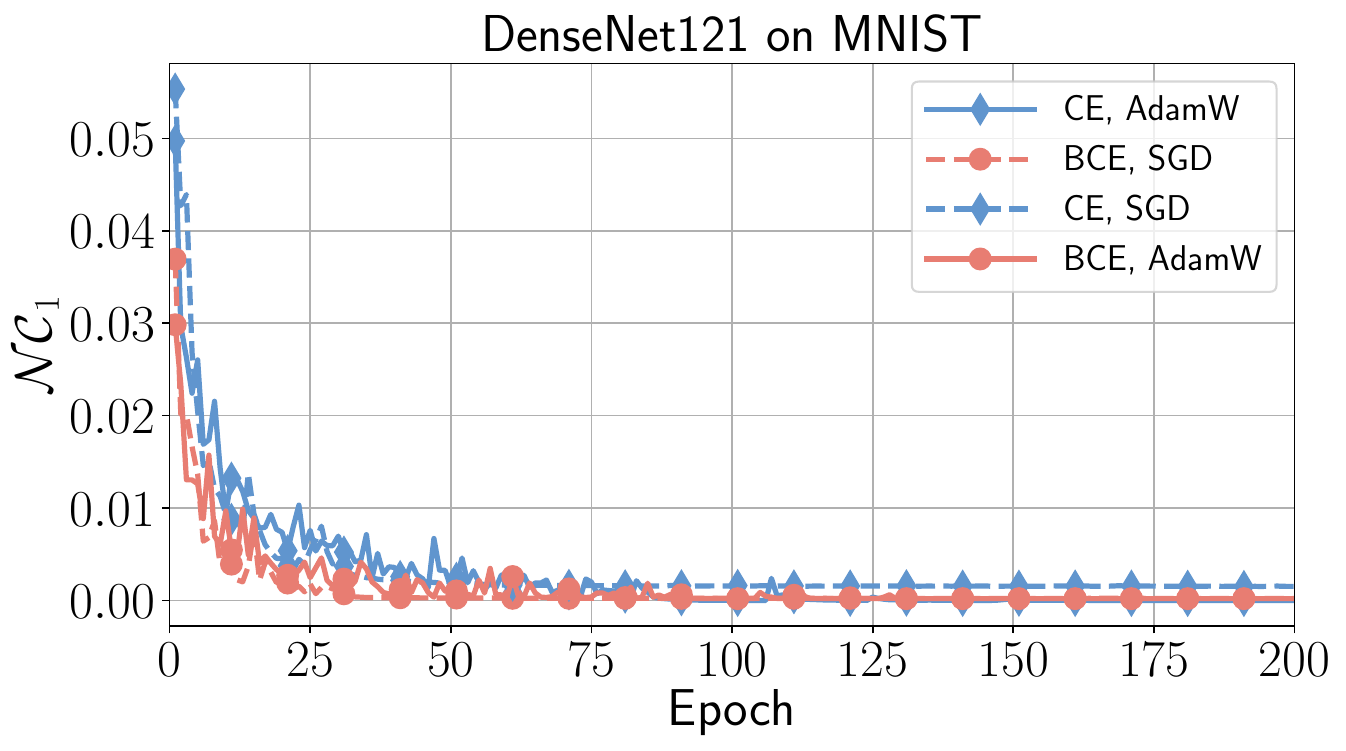}
            \includegraphics[width=0.3\textwidth,max width=\linewidth,max height=0.74\textheight,keepaspectratio]{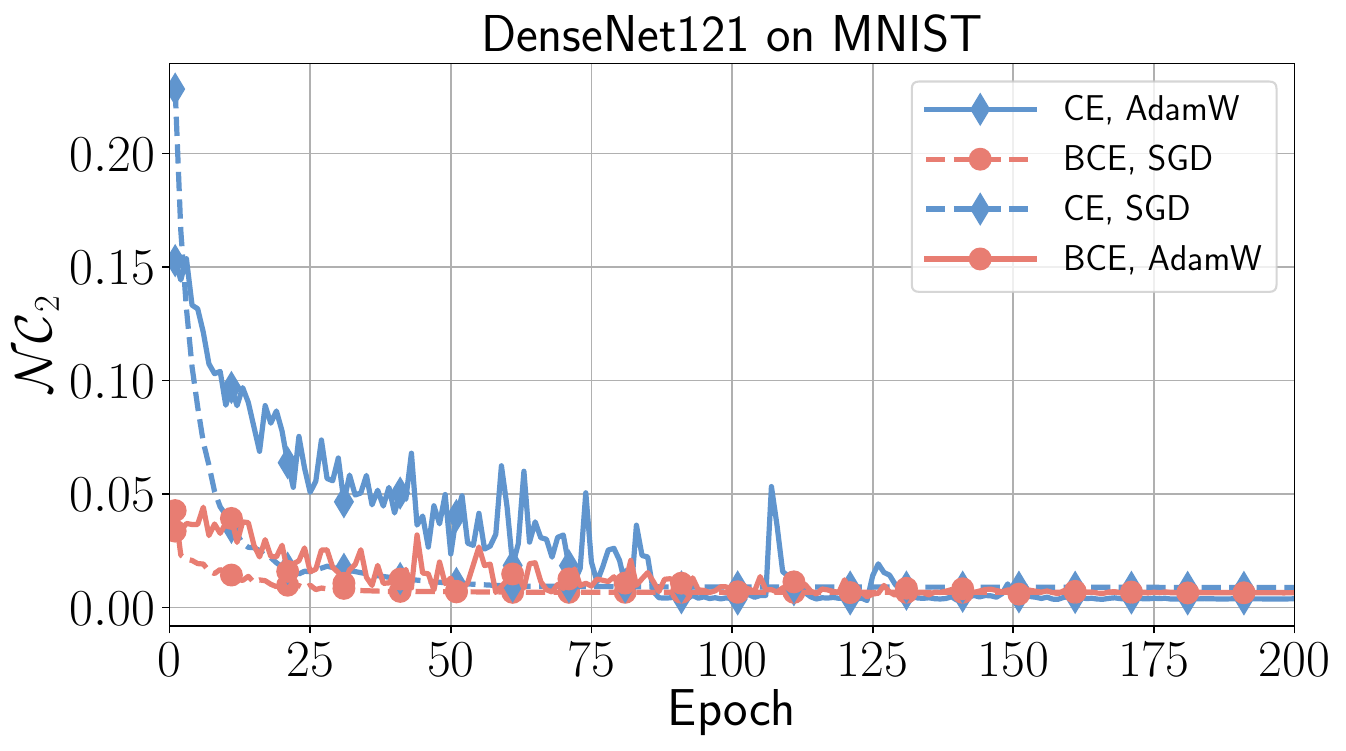}
            \includegraphics[width=0.3\textwidth,max width=\linewidth,max height=0.74\textheight,keepaspectratio]{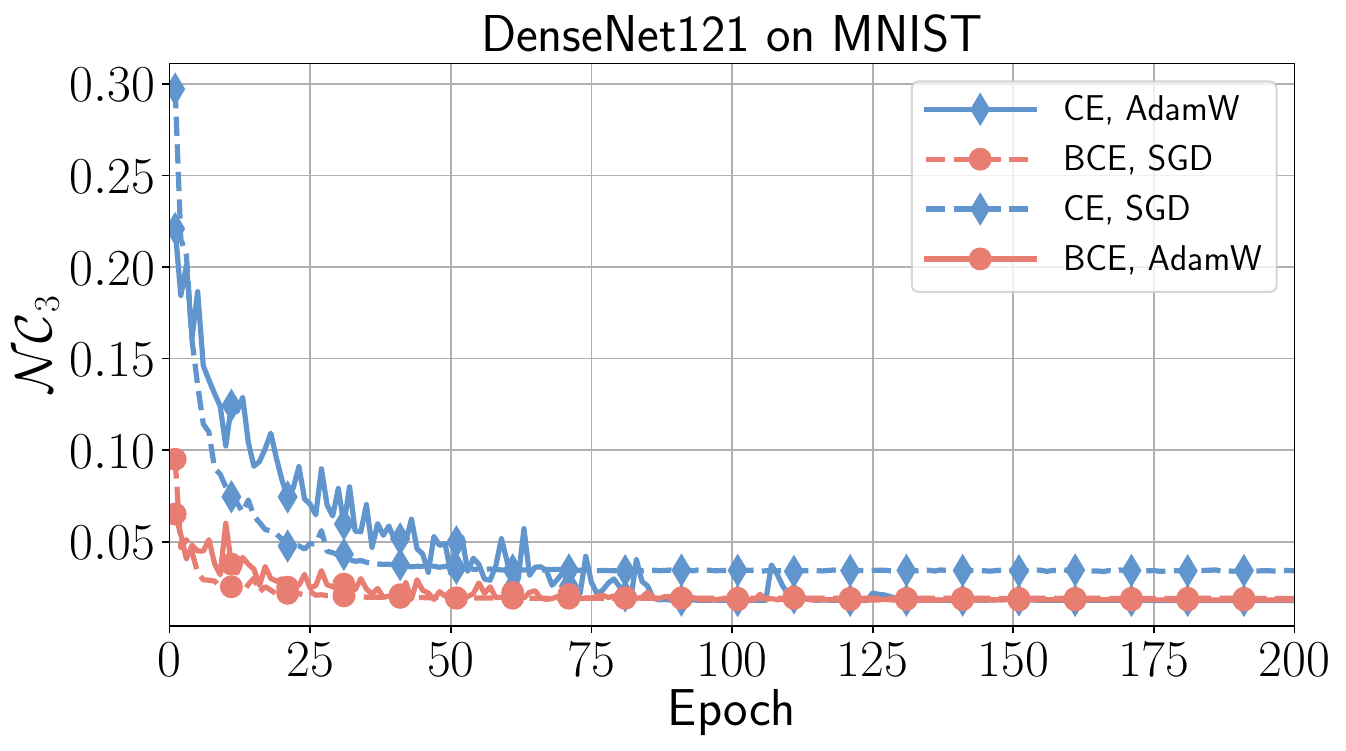}
            \label{fig_densenet121_mnist_NC-supp}}
	\caption{The evolution of the three NC metrics in the training of ResNet18 (top), ResNet50 (middle) and DenseNet121 (bottom) on MNIST with CE and BCE losses using SGD and AdamW, respectively}
	\label{fig_NC_result_of_resnet_and_densenet_mnist-supp}
\end{figure}

\begin{figure}[!t]
	\centering
        \subfloat[$\mathcal{NC}_1$, $\mathcal{NC}_2$, and $\mathcal{NC}_3$ of ResNet18 on CIFAR10]{
            \includegraphics[width=0.3\textwidth,max width=\linewidth,max height=0.74\textheight,keepaspectratio]{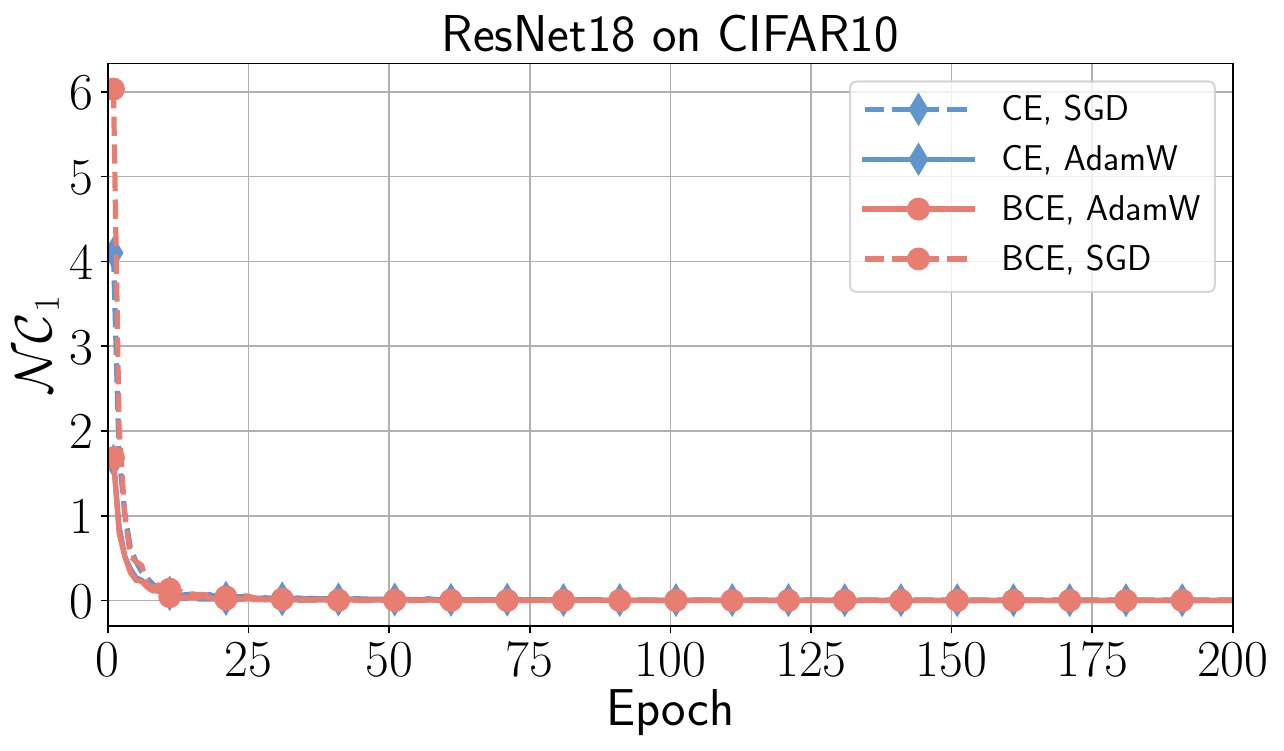}
            \includegraphics[width=0.3\textwidth,max width=\linewidth,max height=0.74\textheight,keepaspectratio]{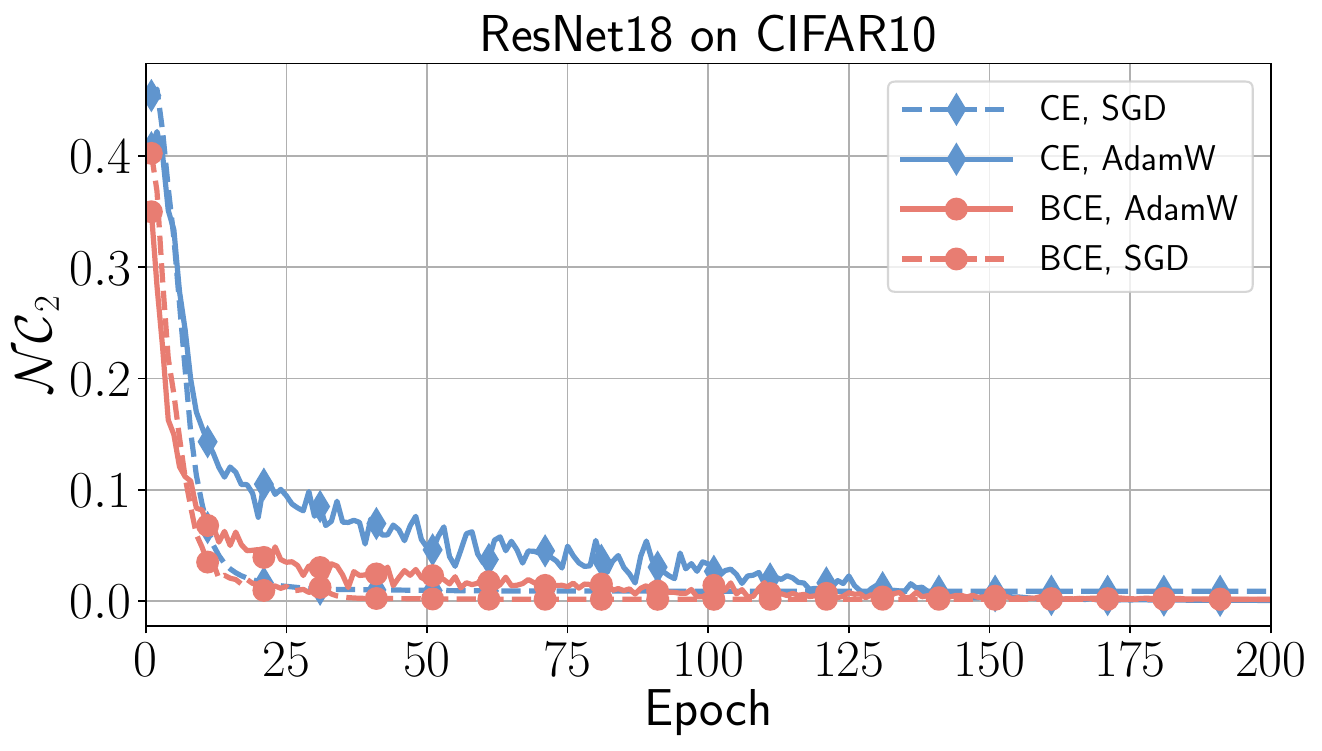}
            \includegraphics[width=0.3\textwidth,max width=\linewidth,max height=0.74\textheight,keepaspectratio]{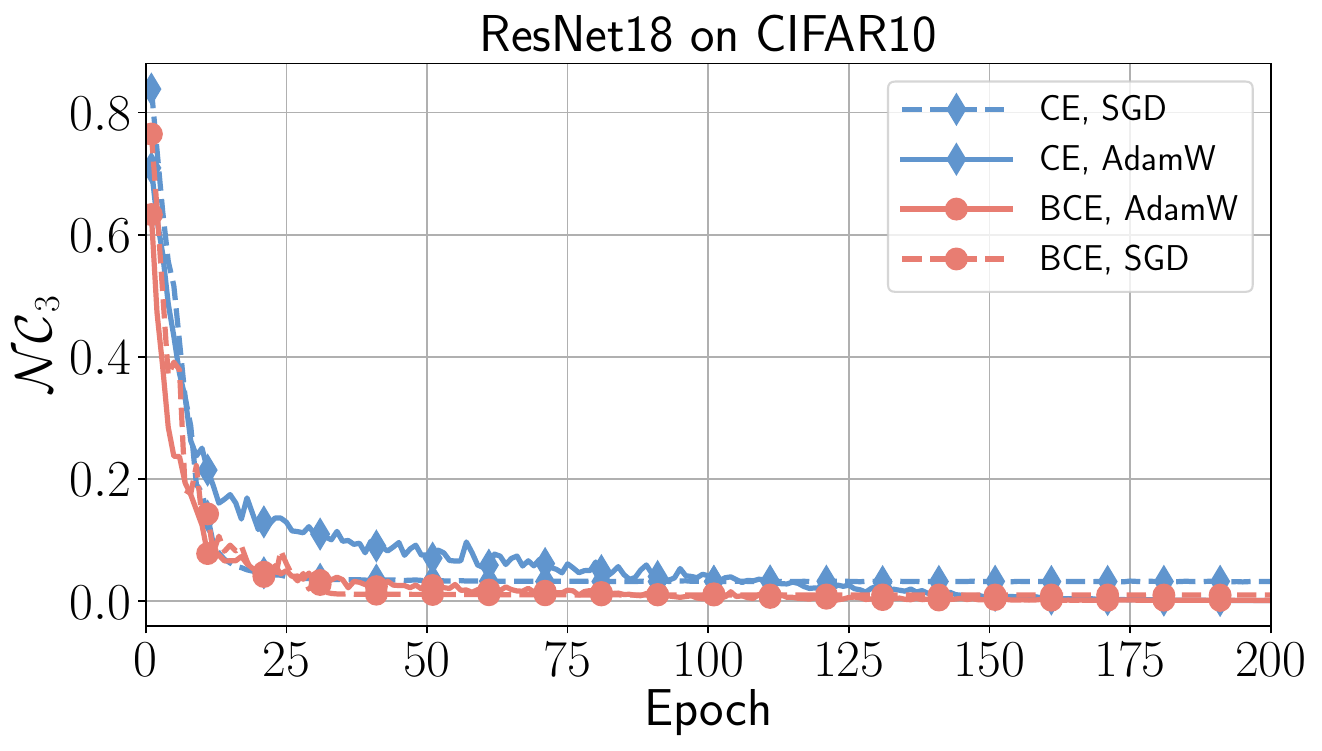}
            \label{fig_resnet18_cifar10_NC-supp}}

        \subfloat[$\mathcal{NC}_1$, $\mathcal{NC}_2$, and $\mathcal{NC}_3$ of ResNet50 on CIFAR10]{
            \includegraphics[width=0.3\textwidth,max width=\linewidth,max height=0.74\textheight,keepaspectratio]{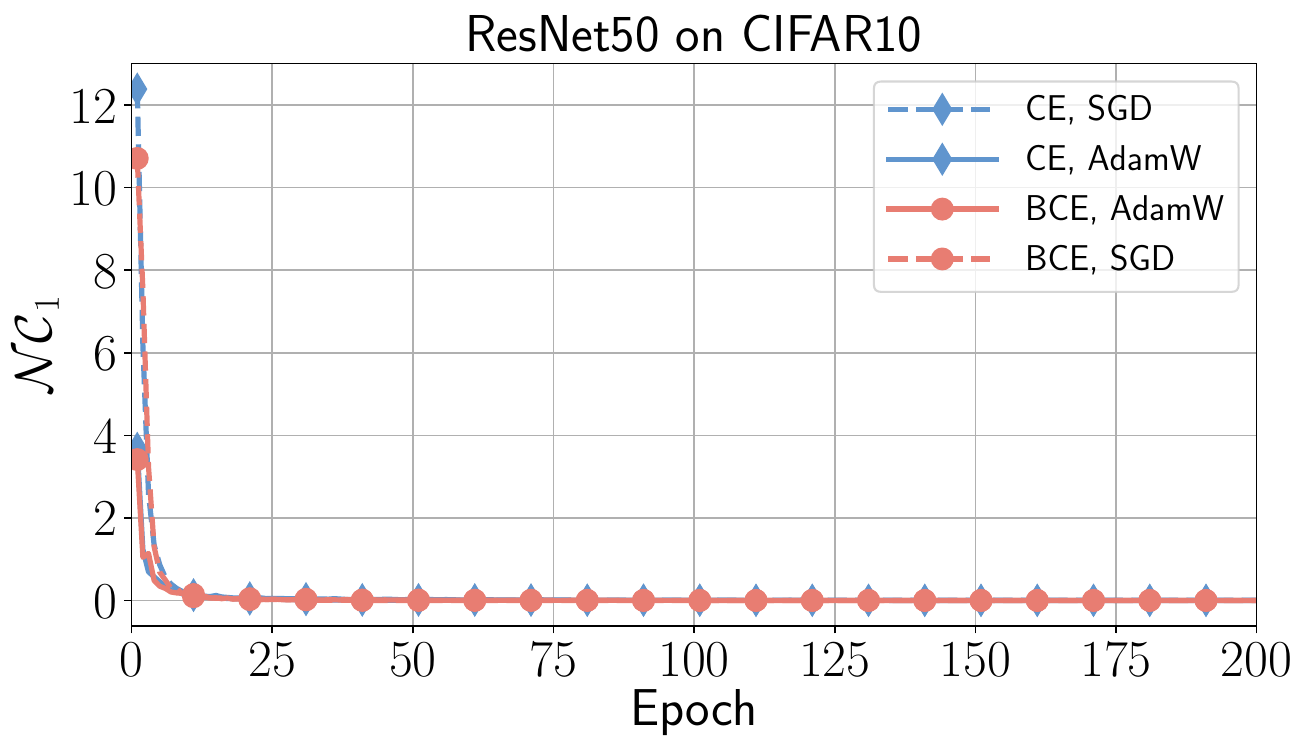}
            \includegraphics[width=0.3\textwidth,max width=\linewidth,max height=0.74\textheight,keepaspectratio]{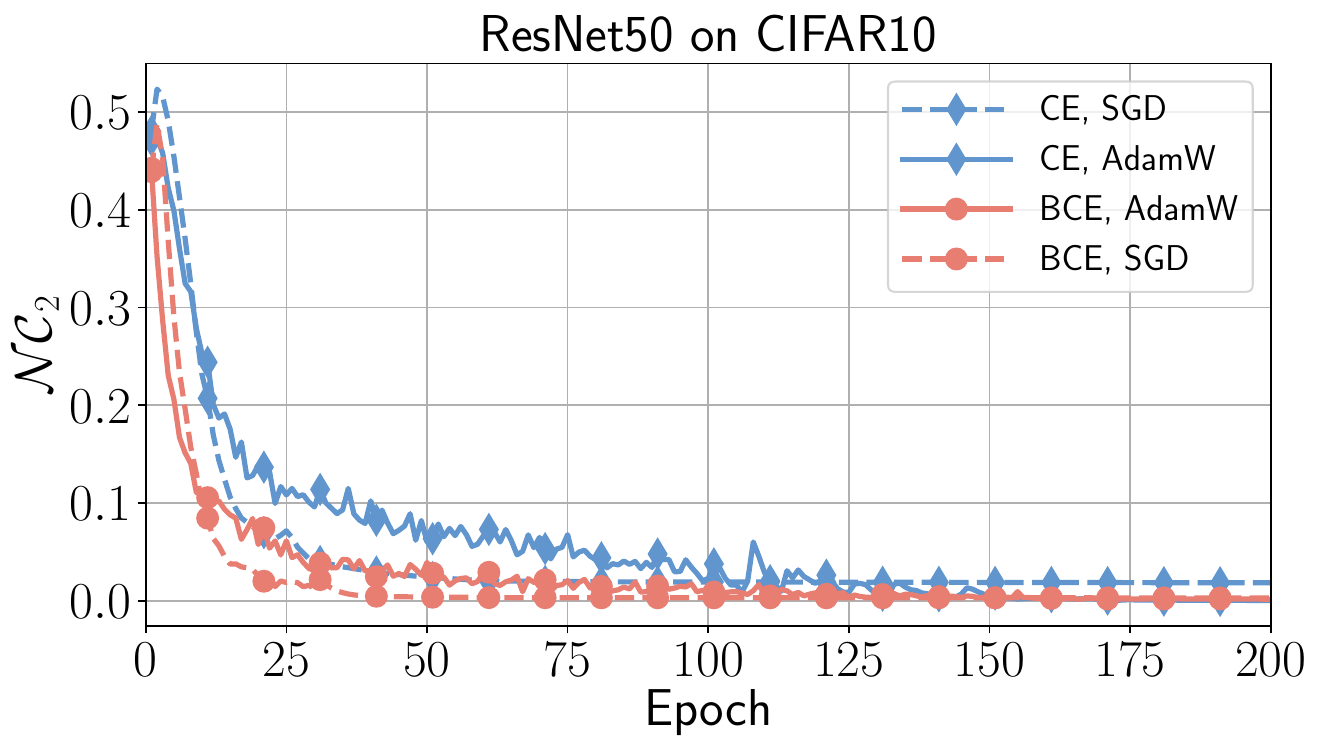}
            \includegraphics[width=0.3\textwidth,max width=\linewidth,max height=0.74\textheight,keepaspectratio]{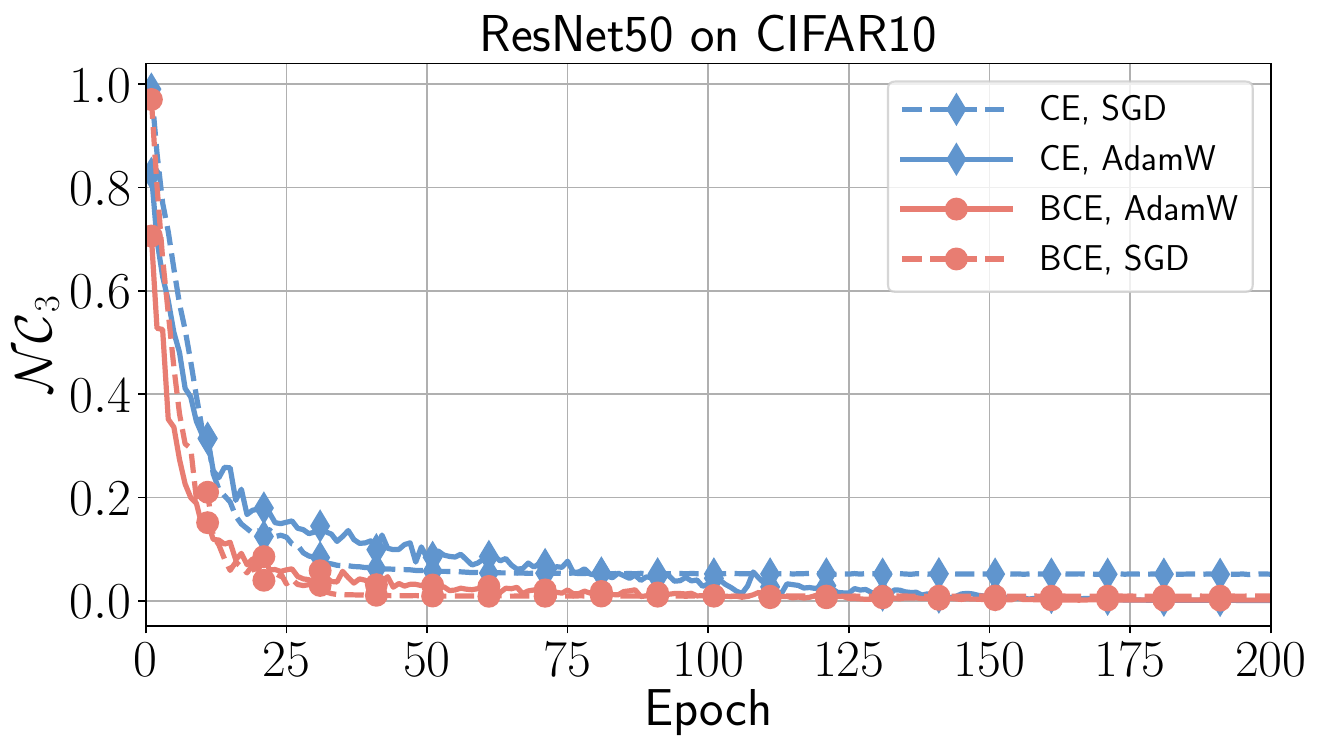}
            \label{fig_resnet50_cifar10_NC-supp}}

        \subfloat[$\mathcal{NC}_1$, $\mathcal{NC}_2$, and $\mathcal{NC}_3$ of DenseNet121 on CIFAR10]{
            \includegraphics[width=0.3\textwidth,max width=\linewidth,max height=0.74\textheight,keepaspectratio]{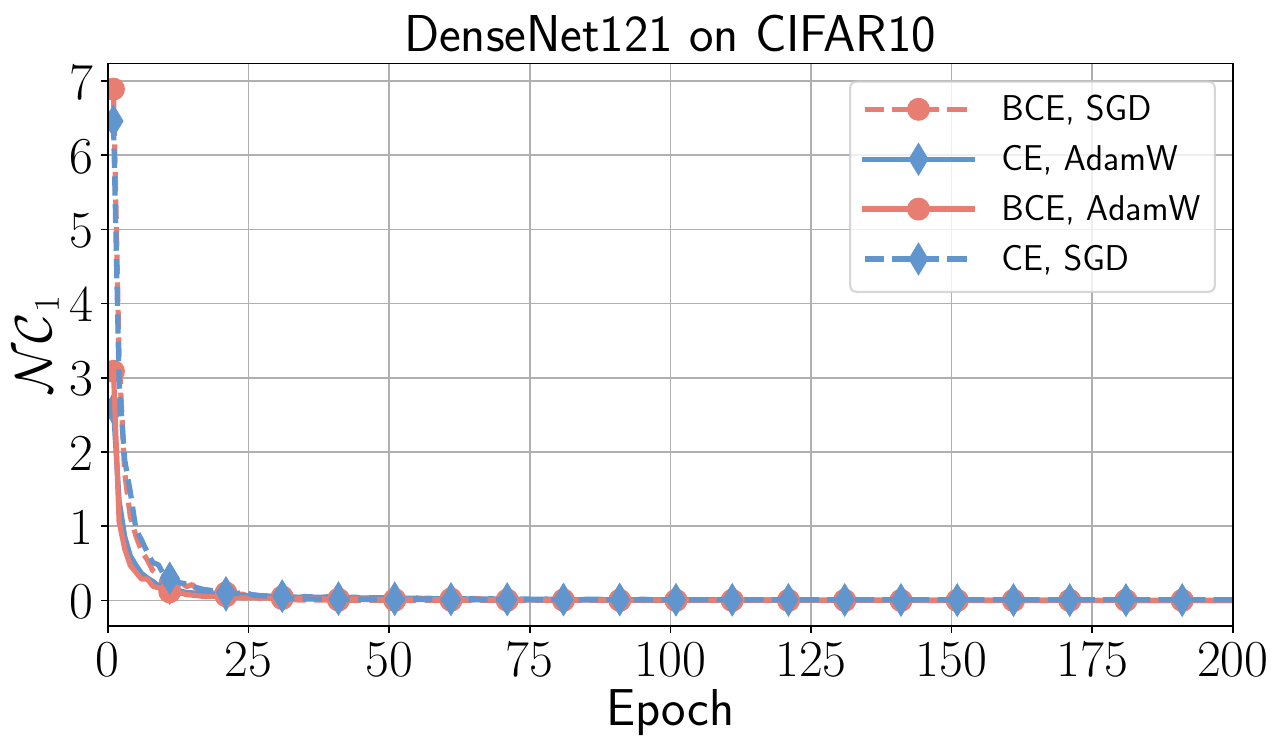}
            \includegraphics[width=0.3\textwidth,max width=\linewidth,max height=0.74\textheight,keepaspectratio]{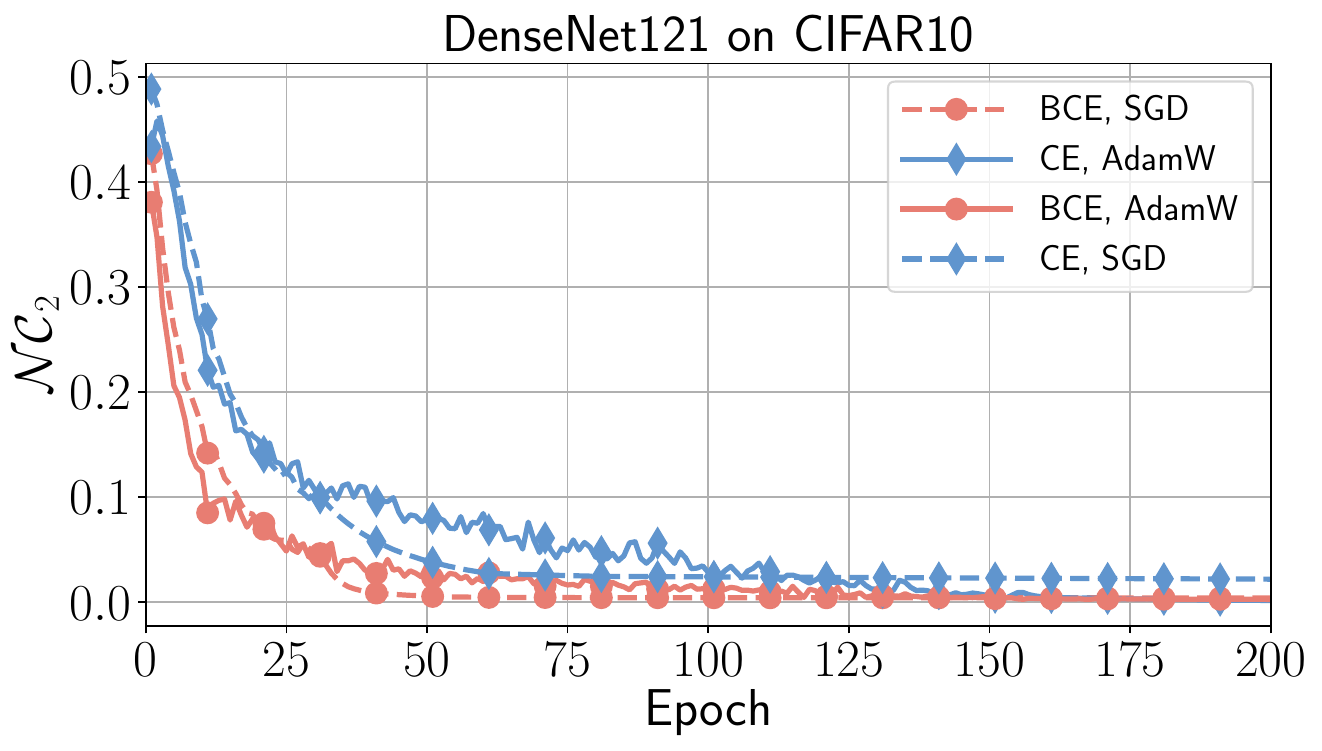}
            \includegraphics[width=0.3\textwidth,max width=\linewidth,max height=0.74\textheight,keepaspectratio]{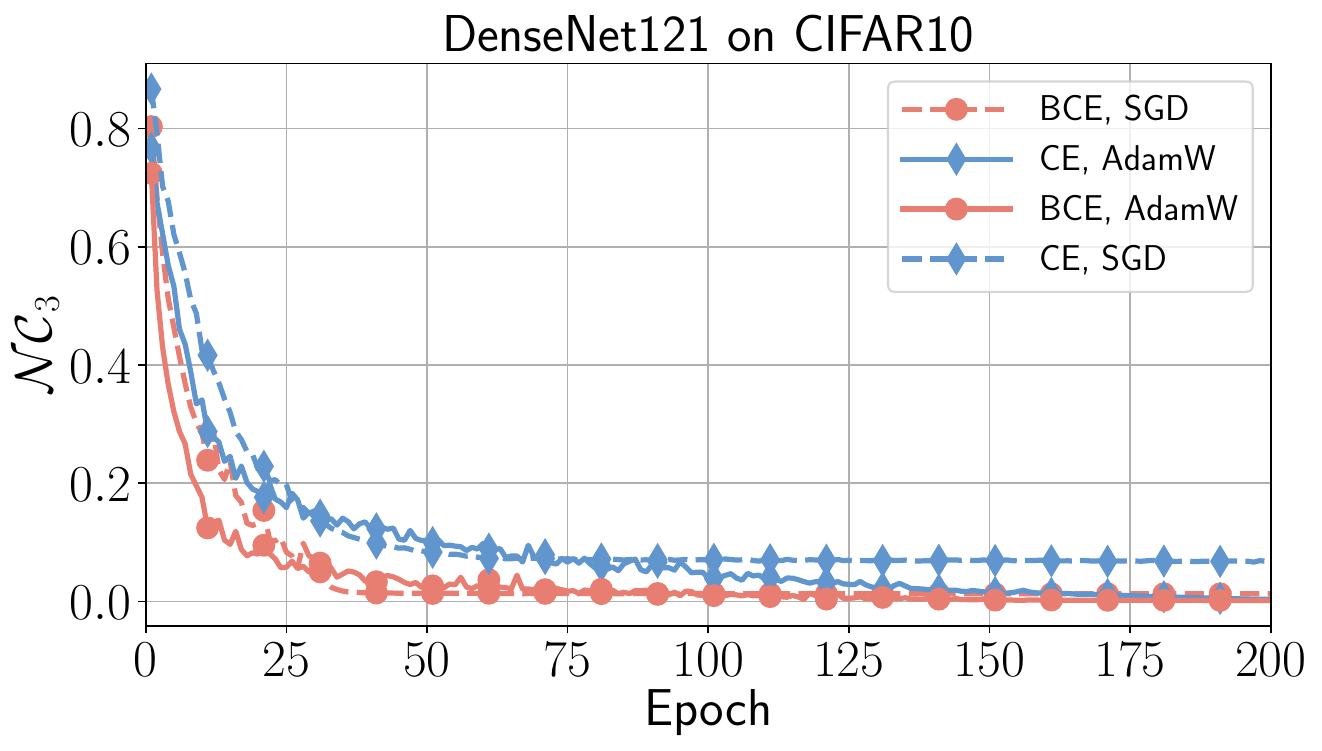}
            \label{fig_densenet121_cifar10_NC-supp}}
    \caption{The evolution of the three NC metrics in the training of ResNet18 (top), ResNet50 (middle) and DenseNet121 (bottom) on CIFAR10 with CE and BCE losses using SGD and AdamW, respectively}
	\label{fig_NC_result_of_resnet_and_densenet_cifar10-supp}
\end{figure}

\begin{figure}[!t]
	\centering
        \subfloat[$\mathcal{NC}_1$, $\mathcal{NC}_2$, and $\mathcal{NC}_3$ of ResNet18 on CIFAR100]{
            \includegraphics[width=0.3\textwidth,max width=\linewidth,max height=0.74\textheight,keepaspectratio]{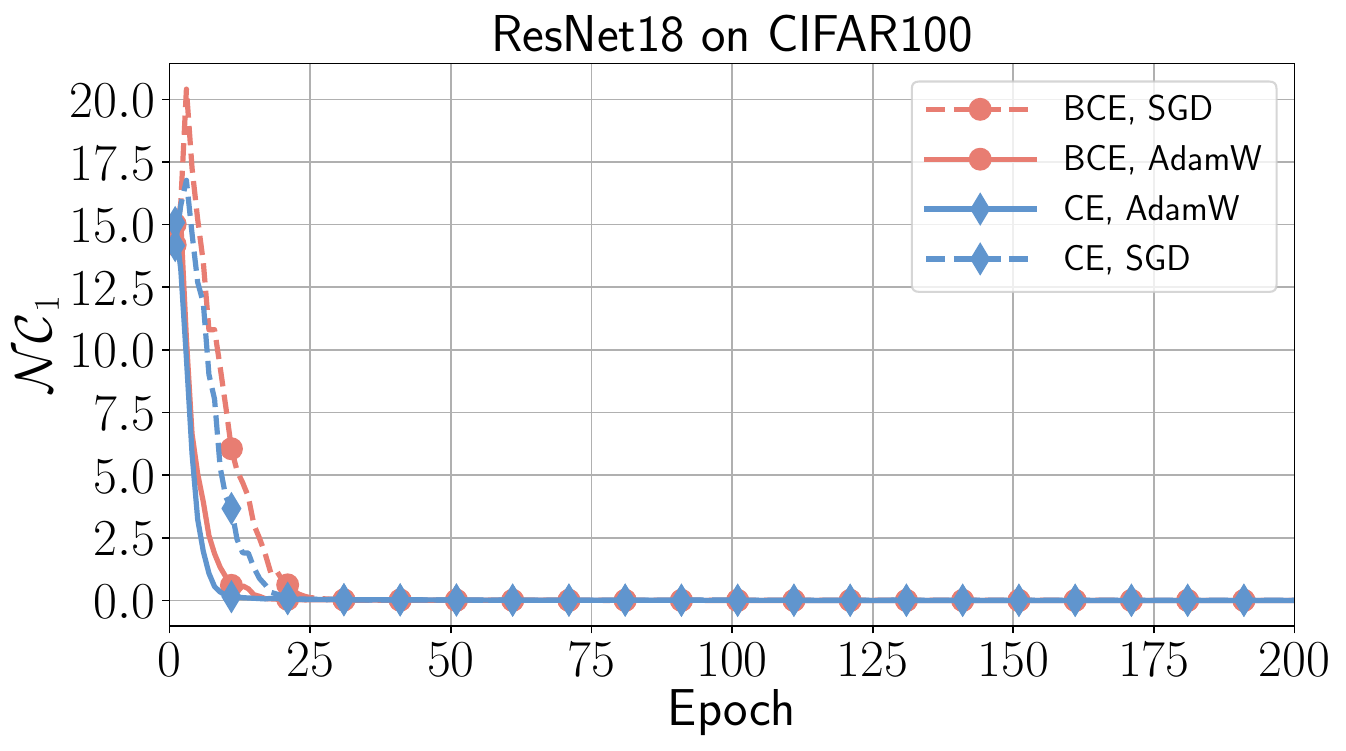}
            \includegraphics[width=0.3\textwidth,max width=\linewidth,max height=0.74\textheight,keepaspectratio]{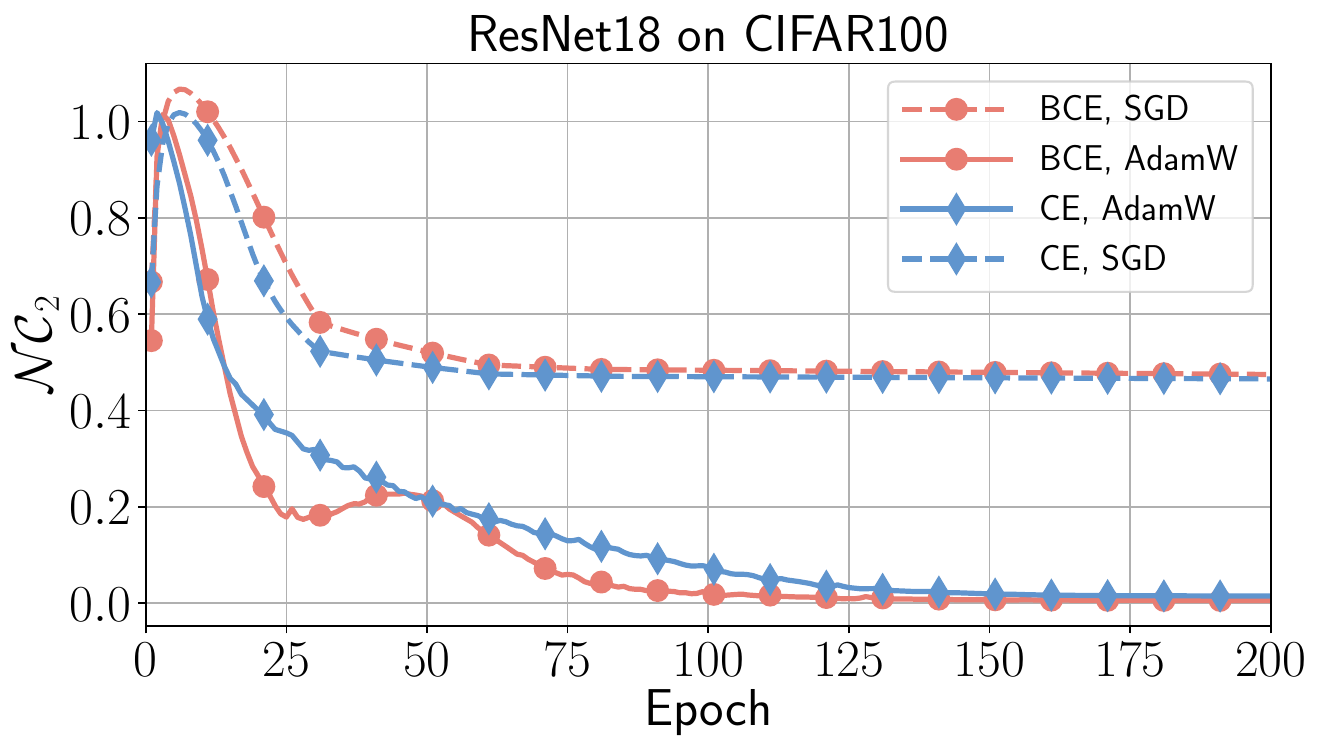}
            \includegraphics[width=0.3\textwidth,max width=\linewidth,max height=0.74\textheight,keepaspectratio]{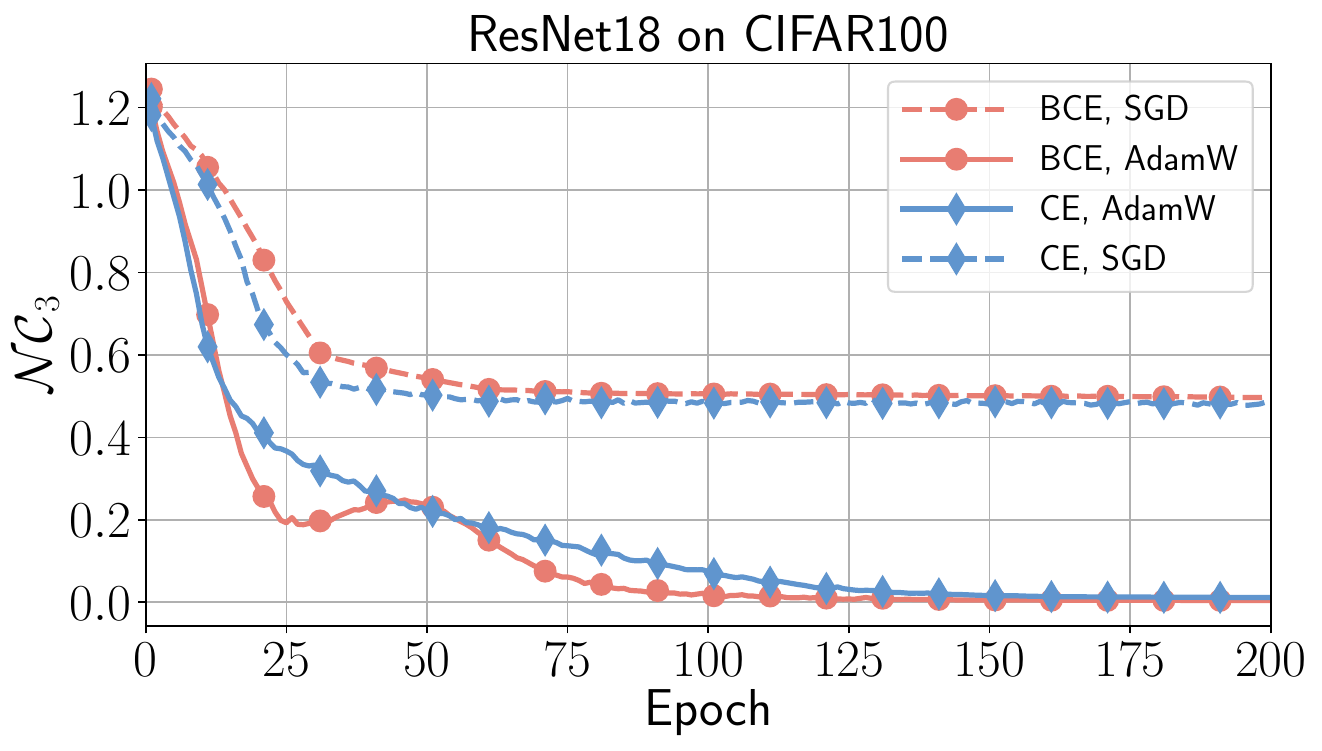}
            \label{fig_resnet18_cifar100_NC-supp}}

        \subfloat[$\mathcal{NC}_1$, $\mathcal{NC}_2$, and $\mathcal{NC}_3$ of ResNet50 on CIFAR100]{
            \includegraphics[width=0.3\textwidth,max width=\linewidth,max height=0.74\textheight,keepaspectratio]{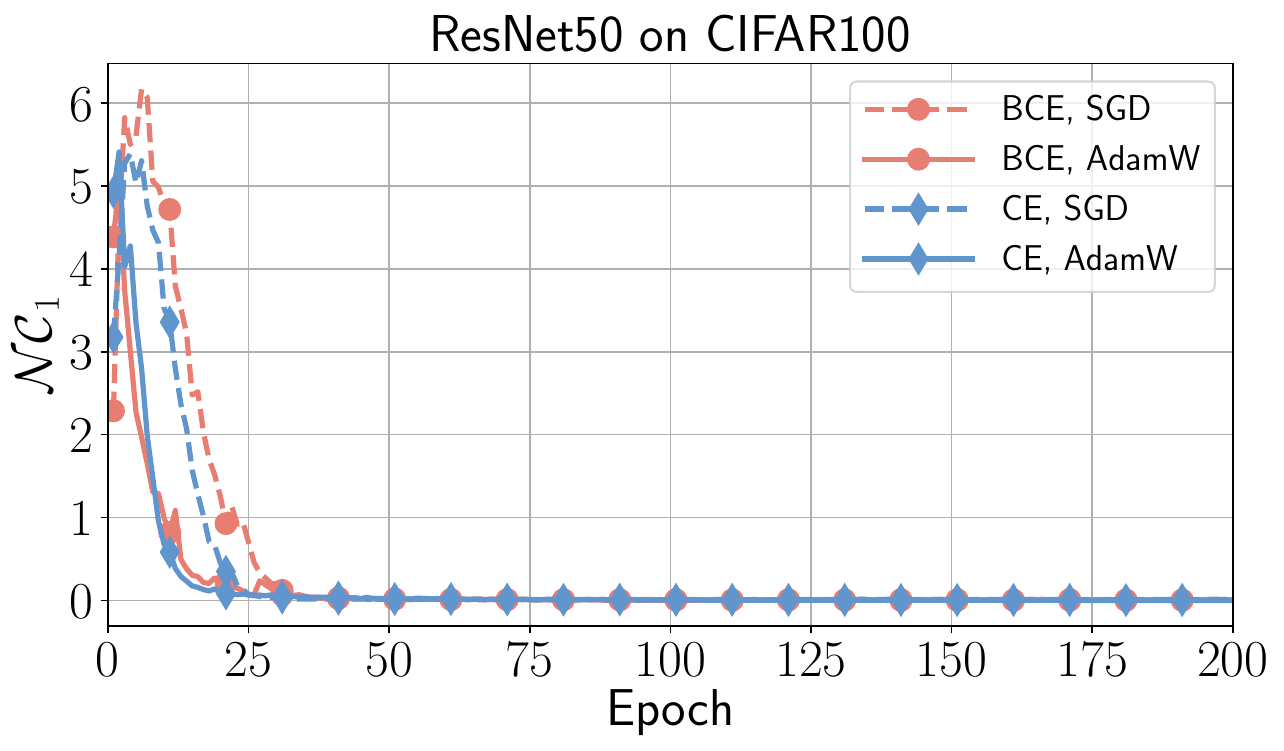}
            \includegraphics[width=0.3\textwidth,max width=\linewidth,max height=0.74\textheight,keepaspectratio]{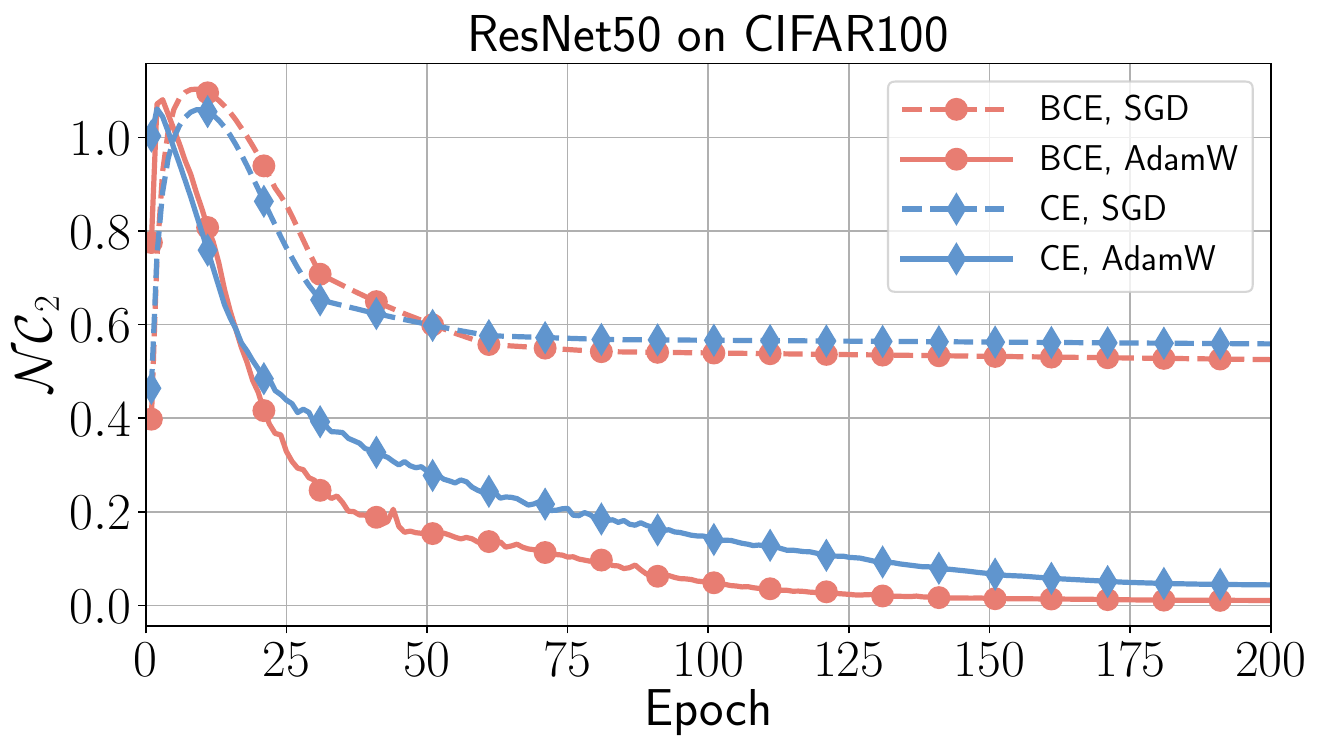}
            \includegraphics[width=0.3\textwidth,max width=\linewidth,max height=0.74\textheight,keepaspectratio]{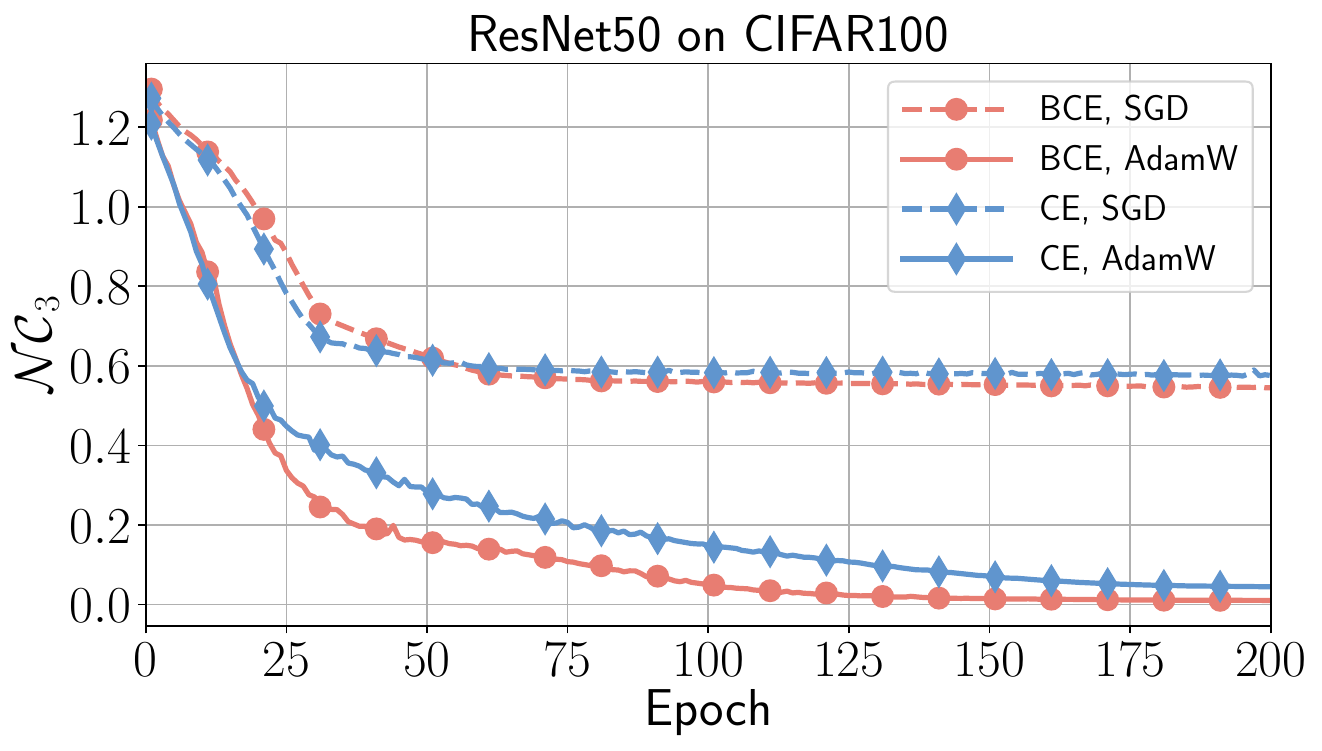}
            \label{fig_resnet50_cifar100_NC-supp}}

        \subfloat[$\mathcal{NC}_1$, $\mathcal{NC}_2$, and $\mathcal{NC}_3$ of DenseNet121 on CIFAR100]{
            \includegraphics[width=0.3\textwidth,max width=\linewidth,max height=0.74\textheight,keepaspectratio]{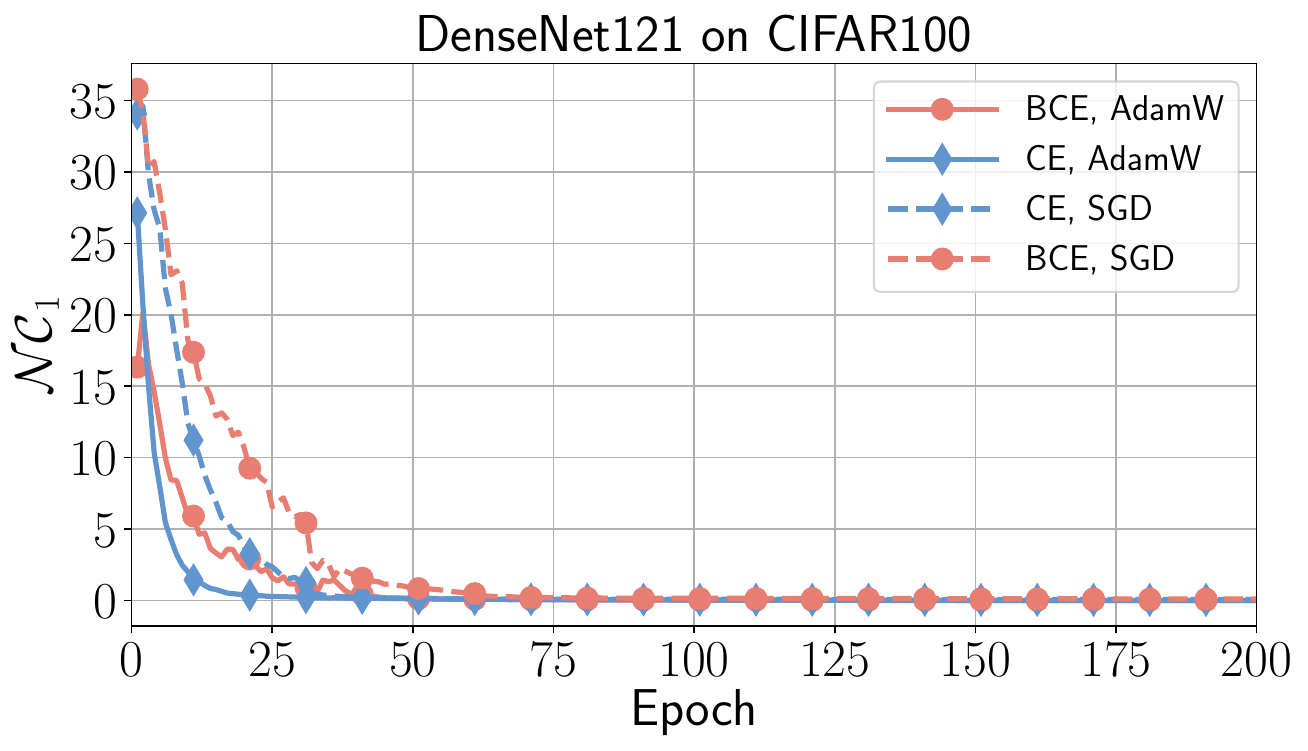}
            \includegraphics[width=0.3\textwidth,max width=\linewidth,max height=0.74\textheight,keepaspectratio]{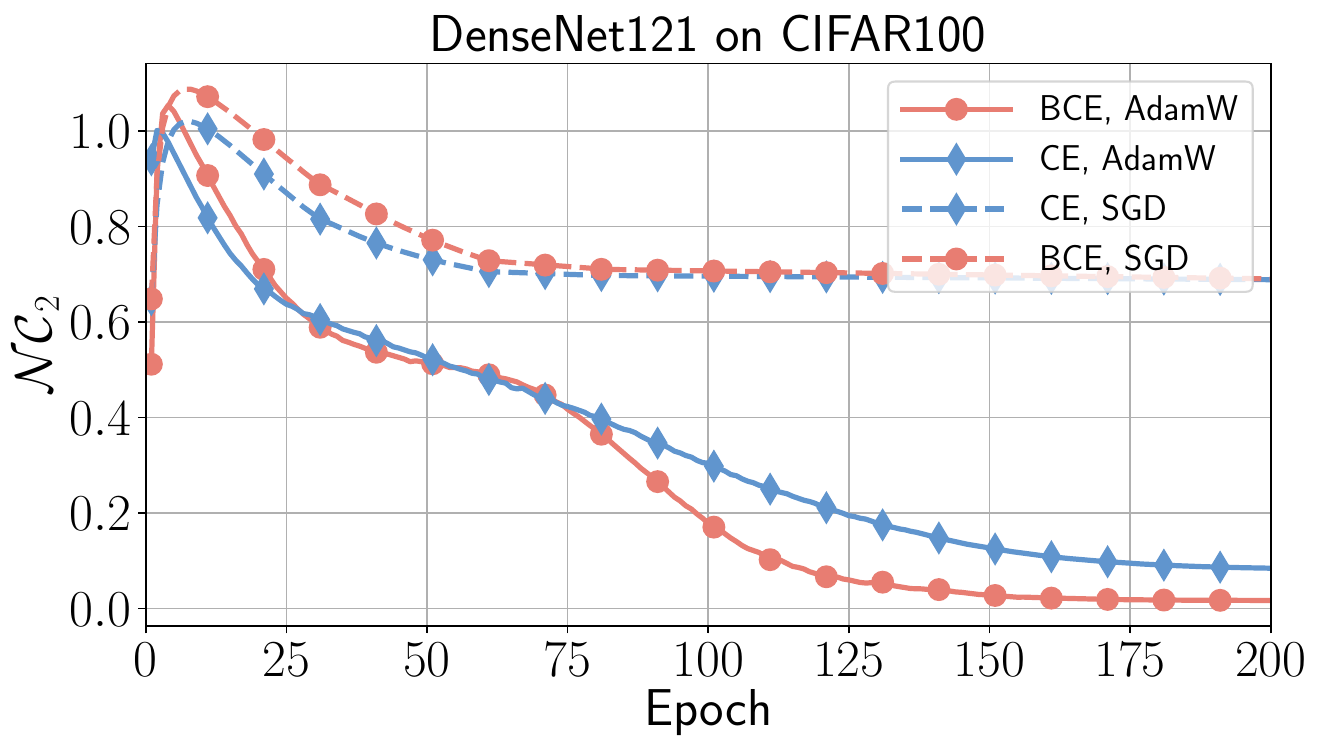}
            \includegraphics[width=0.3\textwidth,max width=\linewidth,max height=0.74\textheight,keepaspectratio]{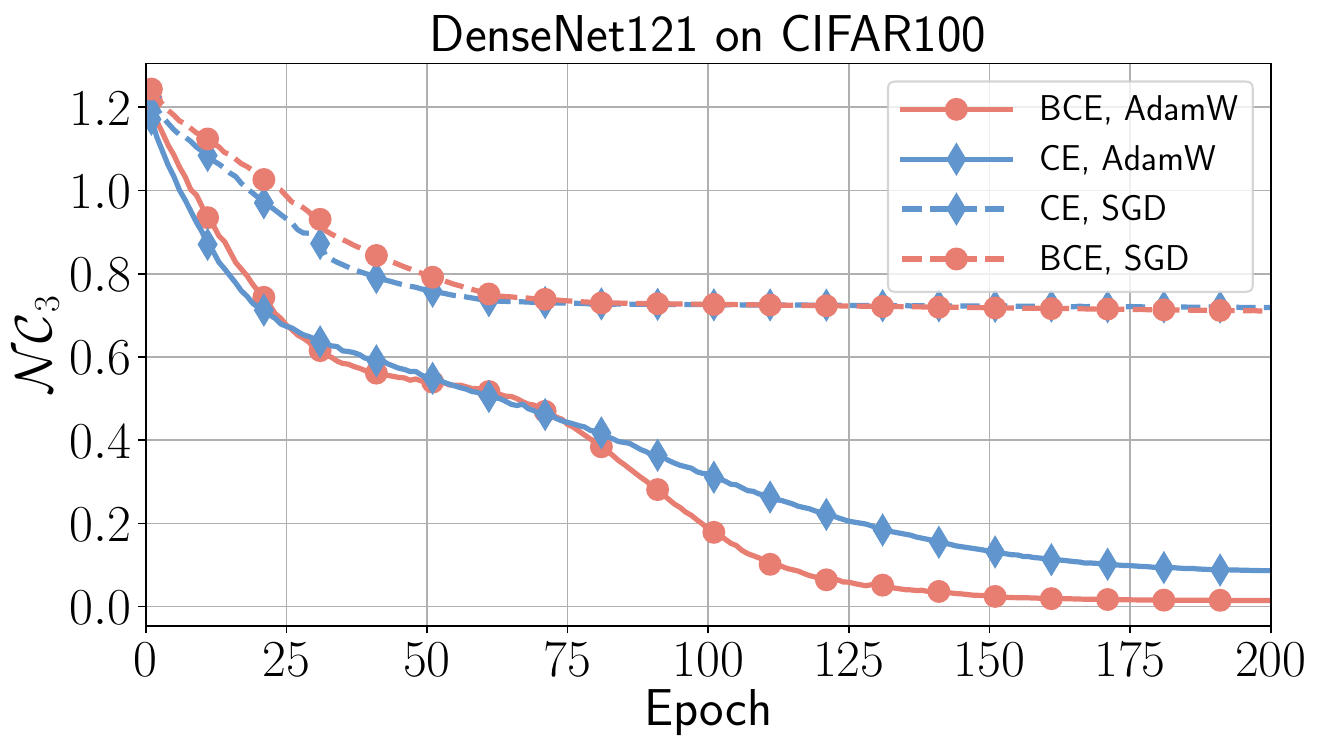}
            \label{fig_densenet121_cifar100_NC-supp}}
    \caption{The evolution of the three NC metrics in the training of ResNet18 (top), ResNet50 (middle) and DenseNet121 (bottom) on CIFAR100 with CE and BCE losses using SGD and AdamW, respectively}
	\label{fig_NC_result_of_resnet_and_densenet_cifar100-supp}
\end{figure}

\begin{figure}[!t]
	\centering
        \subfloat[$\mathcal{NC}_1$, $\mathcal{NC}_2$, and $\mathcal{NC}_3$ of ViT on CIFAR10]{
            \includegraphics[width=0.3\textwidth,max width=\linewidth,max height=0.74\textheight,keepaspectratio]{ViT_cifar10_NC1.pdf}
            \includegraphics[width=0.3\textwidth,max width=\linewidth,max height=0.74\textheight,keepaspectratio]{ViT_cifar10_NC2.pdf}
            \includegraphics[width=0.3\textwidth,max width=\linewidth,max height=0.74\textheight,keepaspectratio]{ViT_cifar10_NC3.pdf}
            \label{fig_ViT_cifar10_NC-supp}}

        \subfloat[$\mathcal{NC}_1$, $\mathcal{NC}_2$, and $\mathcal{NC}_3$ of ViT on CIFAR100]{
            \includegraphics[width=0.3\textwidth,max width=\linewidth,max height=0.74\textheight,keepaspectratio]{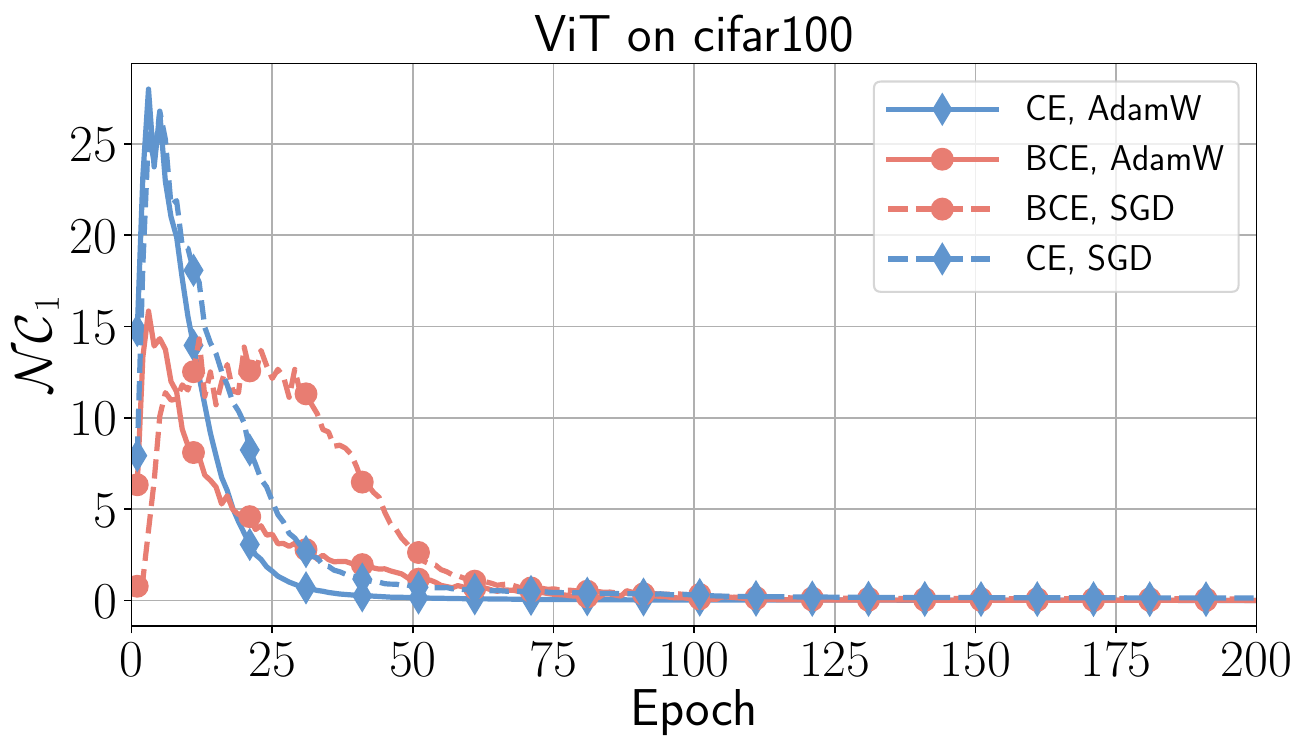}
            \includegraphics[width=0.3\textwidth,max width=\linewidth,max height=0.74\textheight,keepaspectratio]{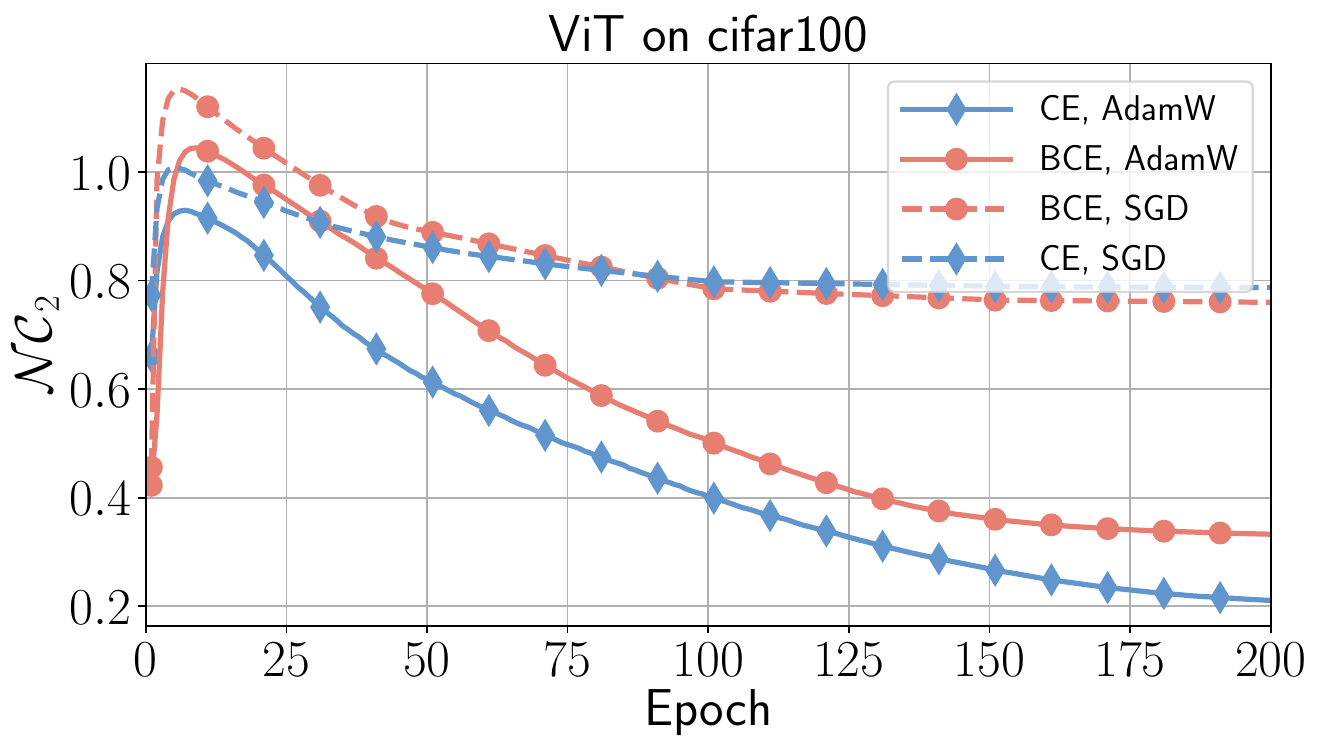}
            \includegraphics[width=0.3\textwidth,max width=\linewidth,max height=0.74\textheight,keepaspectratio]{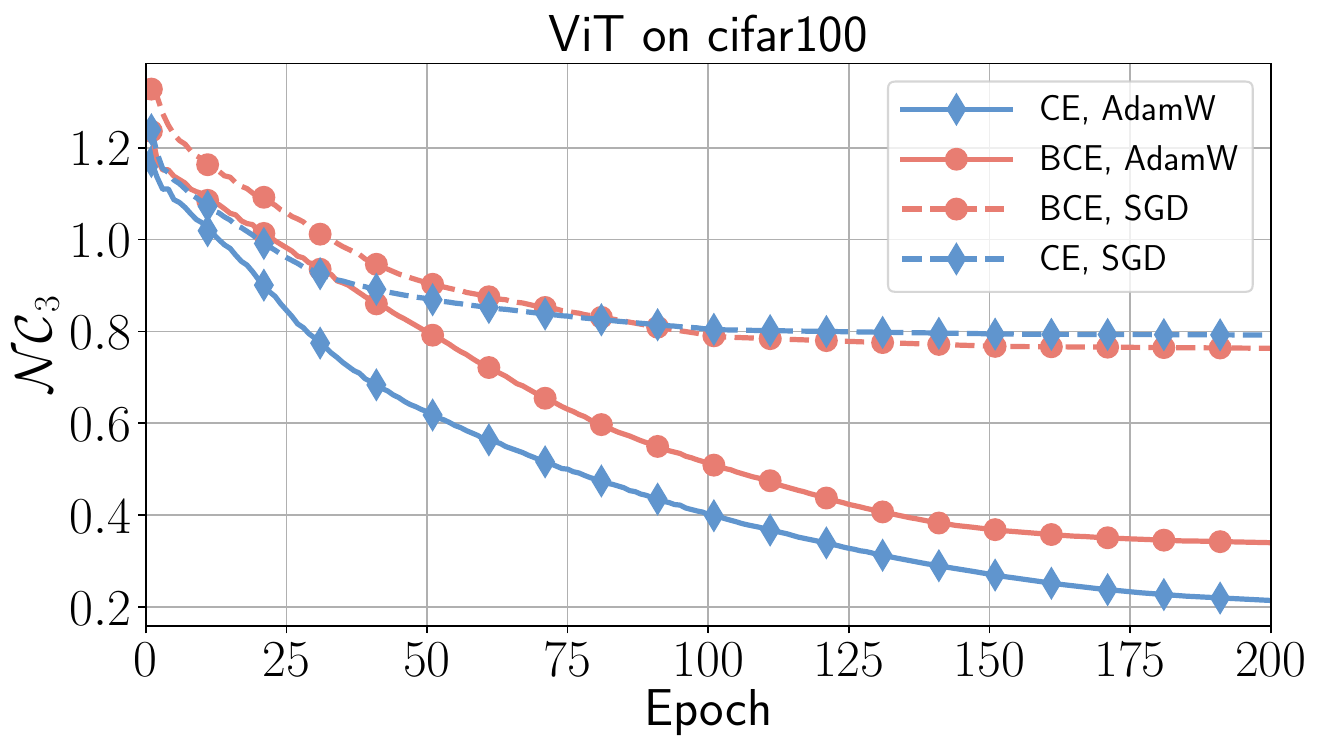}
            \label{fig_ViT_cifar100_NC-supp}}
    \caption{The evolution of the three NC metrics in the training of ViT on CIFAR10 (top) and CIFAR100 (bottom) with CE and BCE losses using SGD and AdamW, respectively}
	\label{fig_NC_result_of_ViT_cifar-supp}
\end{figure}

When $\gamma = 8$, Fig. \ref{fig_NC_result_of_resnet_and_densenet_mnist-supp} - \ref{fig_NC_result_of_ViT_cifar-supp} show the evolution in the training of ResNet18, ResNet50, DenseNet121, and ViT on the training datasets of MNIST, CIFAR10, and CIFAR100 with CE and BCE losses.
In the training on MNIST and CIFAR10 dataset, the three NC metrics of both CE and BCE losses approach zero at the terminal phase of training, indicating that both have approached the neural collapse. Additionally, it is obvious that BCE decreases faster and converges to a higher degree than CE in the first 50 epochs.
However, during the training on CIFAR100, which is a more challenging dataset than MNIST and CIFAR10, the NC metrics of models trained with the SGD optimizer, especially $\mathcal{NC}_2$ and $\mathcal{NC}_3$, do not approach zero, whereas those of models trained with AdamW approach zero. One can still observe that in most cases, the NC metrics of the BCE loss decreases faster and converges better than the CE loss.

\subsection{Impact of Classifier Biases and Scale Factor}

\textbf{The mean of initial classifier biases $\boldsymbol{b}$.}\quad
In Sec. \ref{section_nc}, we point out that classifier biases differs substantially between the BCE and CE losses; specifically, at the global minimizer, the final bias of the BCE loss is uniquely determined, whereas in the CE loss there are infinitely many solutions. To validate this finding, we present the distribution of the final classifier biases and positive/negative unbiased decision scores for ResNet18 trained on MNIST, and ViT trained from scratch on CIFAR10
using different optimizer, with varying mean of initial classifier biases.

\begin{figure}[!bt]
    \centering{\includegraphics[width=0.95\textwidth,max width=\linewidth,max height=0.74\textheight,keepaspectratio]{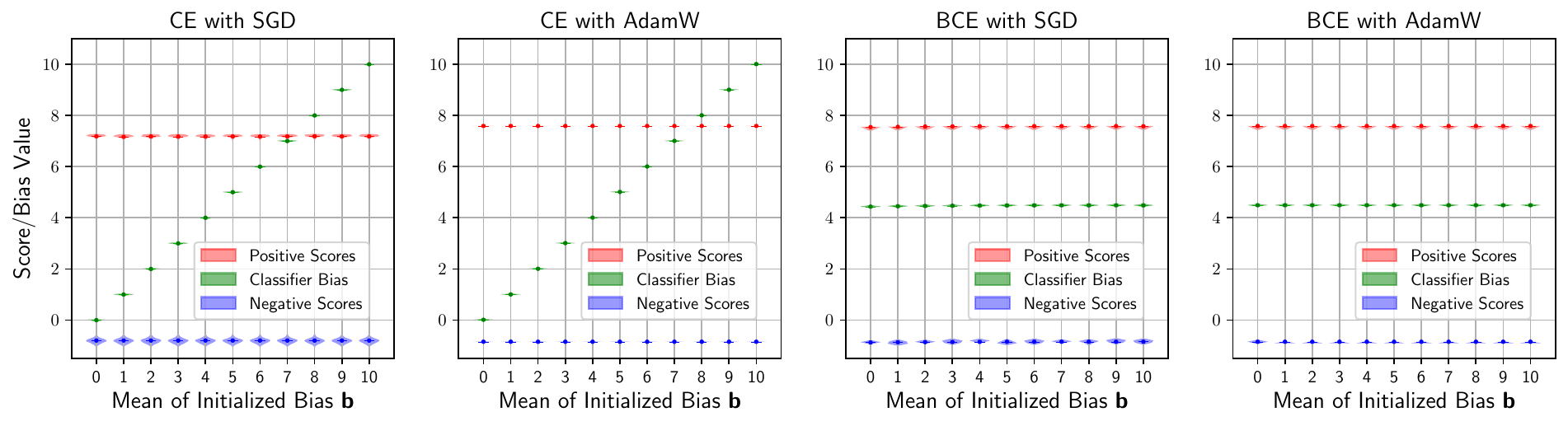}}
    \caption{The distribution of the final classifier biases and positive/negative unbiased decision scores for ResNet18 trained on MNIST with varying mean of initial classifier biases, while the variance of initial classifier biases is $0$ and $\gamma =8$. The final classifier biases of CE-trained models are determined by their initial values, while that of BCE-trained models always converge to the same value}
    \label{resnet18_mnist_bias_mean_gamma_8}
\end{figure}

\begin{figure}[!bt]
    \centering{\includegraphics[width=0.95\textwidth,max width=\linewidth,max height=0.74\textheight,keepaspectratio]{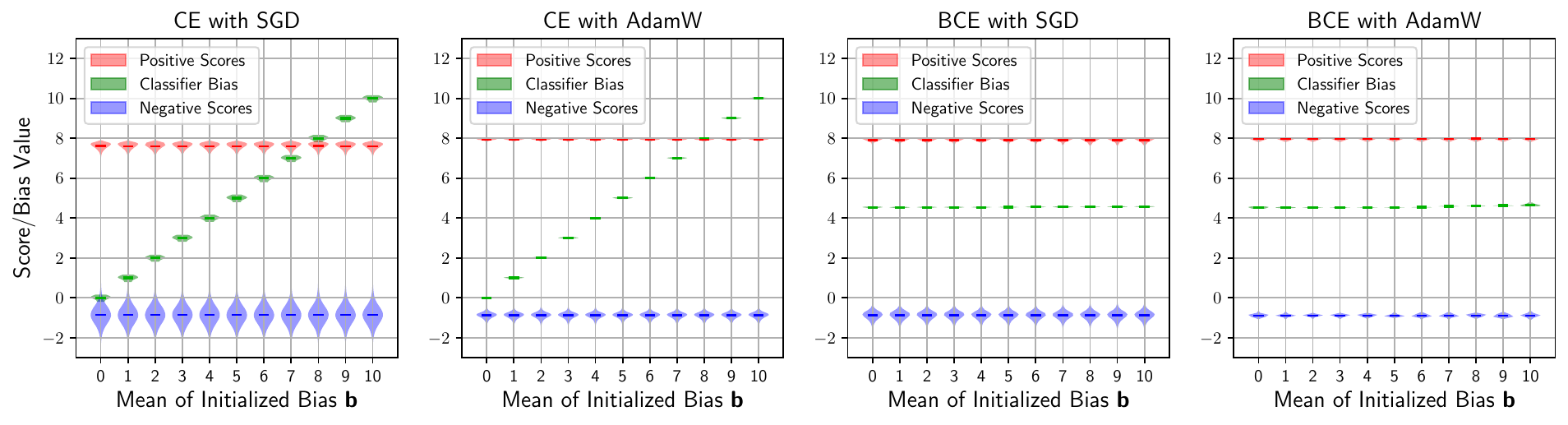}}
    \caption{The distribution of the final classifier biases and positive/negative unbiased decision scores for ViT trained on CIFAR10 with varying mean of initial classifier biases, while the variance of initial classifier biases is $0$ and $\gamma =8$. The final classifier biases of CE-trained models are determined by their initial values, while that of BCE-trained models always converge to the same value}
    \label{ViT_cifar10_bias_mean_gamma_8}
\end{figure}

Fig. \ref{resnet18_mnist_bias_mean_gamma_8} and \ref{ViT_cifar10_bias_mean_gamma_8} show that the final classifier biases of CE-trained models are determined by their initial values, while the biases of BCE-trained models always converge to the same value, consistent with our theoretical analysis in the paper.

It is notable to point out a difference between CNNs and Transformers. For the CNN models such as ResNet and DenseNet, the output sample features are processed by the ReLU activation function, which makes that each component of the generated features is a nonnegative number and results in the sum of all sample features not being equal to zero (unless all the features are zero vectors, which is meaningless), conflicting with Eqs. (\ref{eq_feature_zero}) and (\ref{equ-fce_hmean_w=h_condition}) required for proving Theorems. Therefore, the perfect results for NC cannot be achieved based on the ResNet and DenseNet, as demonstrated by the obvious difference between the converged positive unbiased decision scores and the theoretical value $\gamma = 8$, regardless of whether the model is trained using CE or BCE. This phenomenon has been found and discussed in~\citet{Zhu2021_geometric}.
In contrast, the features output by ViT are not processed by ReLU or any other activation functions, which aligns more closely with the unconstrained feature model~\citep{Zhu2021_geometric,Yaras2022_NeurIPS}. When training ViT with the CE or BCE, the converged positive and negative unbiased decision scores can converge to $\gamma = 8$ and $-\frac{\gamma}{K-1} = -\frac{8}{9}$, respectively.

\textbf{The variance of initial classifier biases $\boldsymbol{b}$.}\quad
It is well known that the variance of the parameters at model initialization determines whether gradient signals propagate stably in deep neural networks, thereby affecting convergence speed, training stability, and final generalization performance. Based on this consensus, to more comprehensively validate our theoretical findings, we conduct experiments with a fixed mean of initial classifier biases of 0, when the variances of initial classifier biases are set to $0, 0.1, 0.5, 1, 2, 3, 4, 5, 6$, respectively, to further compare the BCE and CE losses.

\begin{figure}[!tbp]
	\centering
        \subfloat[ResNet18 on MNIST, $\gamma=8$]{
            \includegraphics[width=0.95\textwidth,max width=\linewidth,max height=0.74\textheight,keepaspectratio]{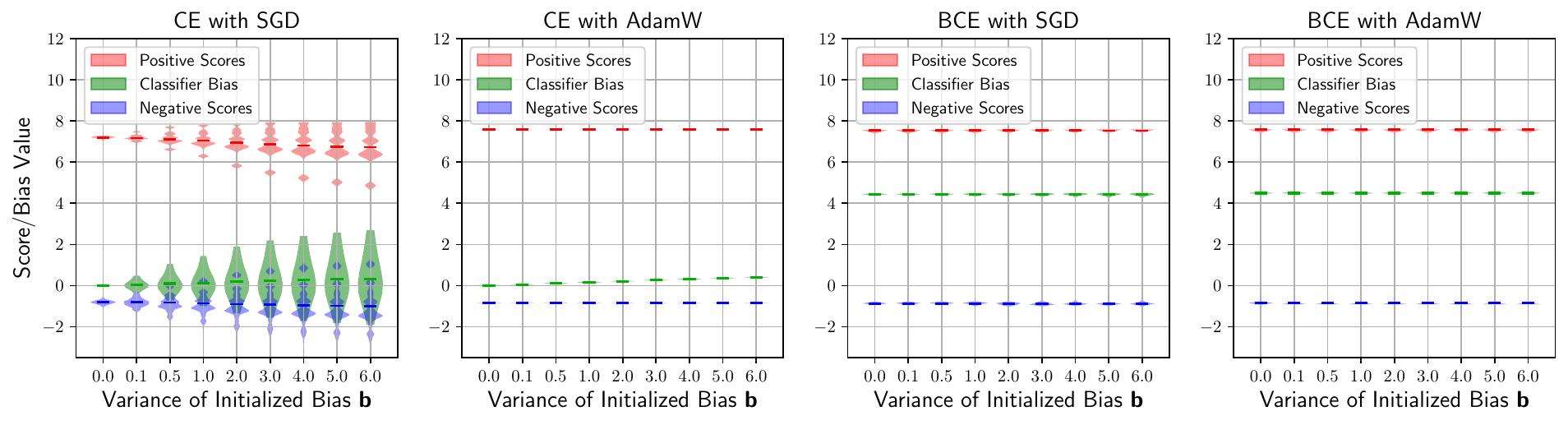}
            \label{fig_resnet18_mnist_var_gamma_8}}

        \subfloat[ResNet18 on MNIST, $\gamma=16$]{
            \includegraphics[width=0.95\textwidth,max width=\linewidth,max height=0.74\textheight,keepaspectratio]{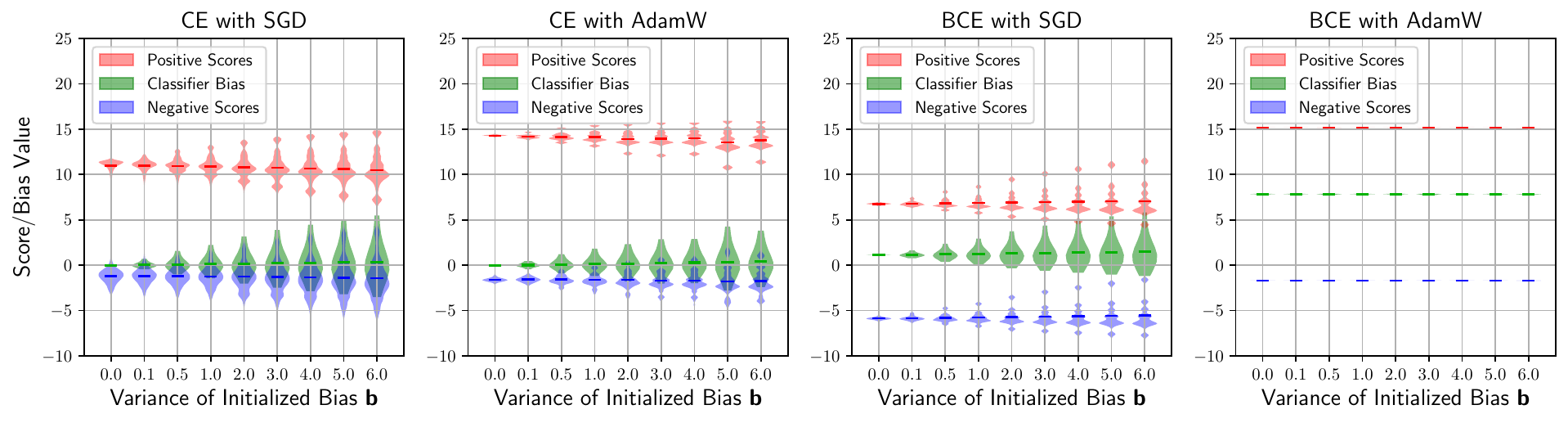}
            \label{fig_resnet18_mnist_var_gamma_16}}

        \subfloat[ResNet18 on MNIST, $\gamma=32$]{
            \includegraphics[width=0.95\textwidth,max width=\linewidth,max height=0.74\textheight,keepaspectratio]{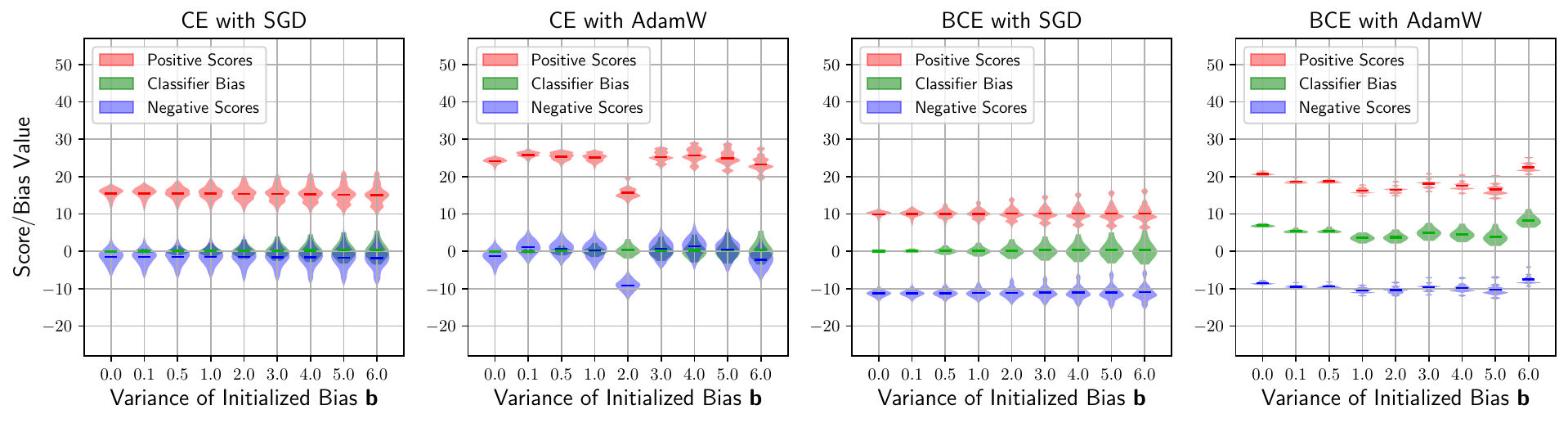}
            \label{fig_resnet18_mnist_var_gamma_32}}

        \subfloat[ResNet18 on MNIST, $\gamma=64$]{
            \includegraphics[width=0.95\textwidth,max width=\linewidth,max height=0.74\textheight,keepaspectratio]{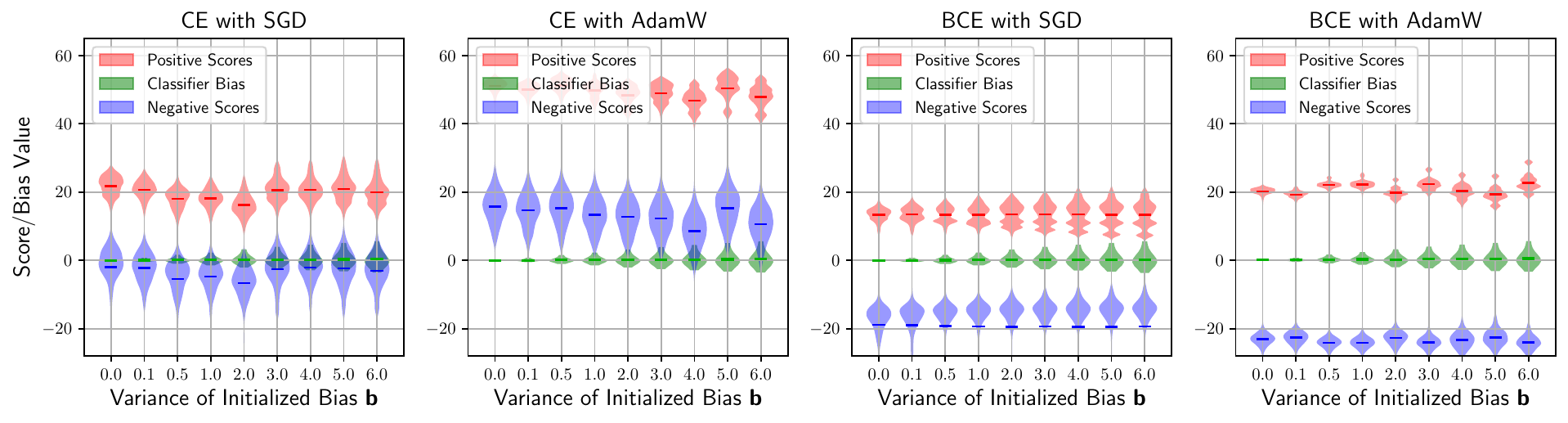}
            \label{fig_resnet18_mnist_var_gamma_64}}
	\caption{The distribution of the final classifier biases and positive/negative unbiased decision scores of ResNet18 trained on MNIST with varying initial variance of the classifier biases. The different scale factors, $\gamma=8$ (1-st row), $\gamma=16$ (2-nd row), $\gamma=32$ (3-rd row), and $\gamma=64$ (4-th row), also significantly impact the convergence of decision scores and the final classifier biases}
	\label{fig_result_of_resnet18_mnist_bias_var-supp}
\end{figure}

\begin{figure}[!tbp]
	\centering
        \subfloat[ViT on CIFAR10, $\gamma=8$]{
            \includegraphics[width=0.95\textwidth,max width=\linewidth,max height=0.74\textheight,keepaspectratio]{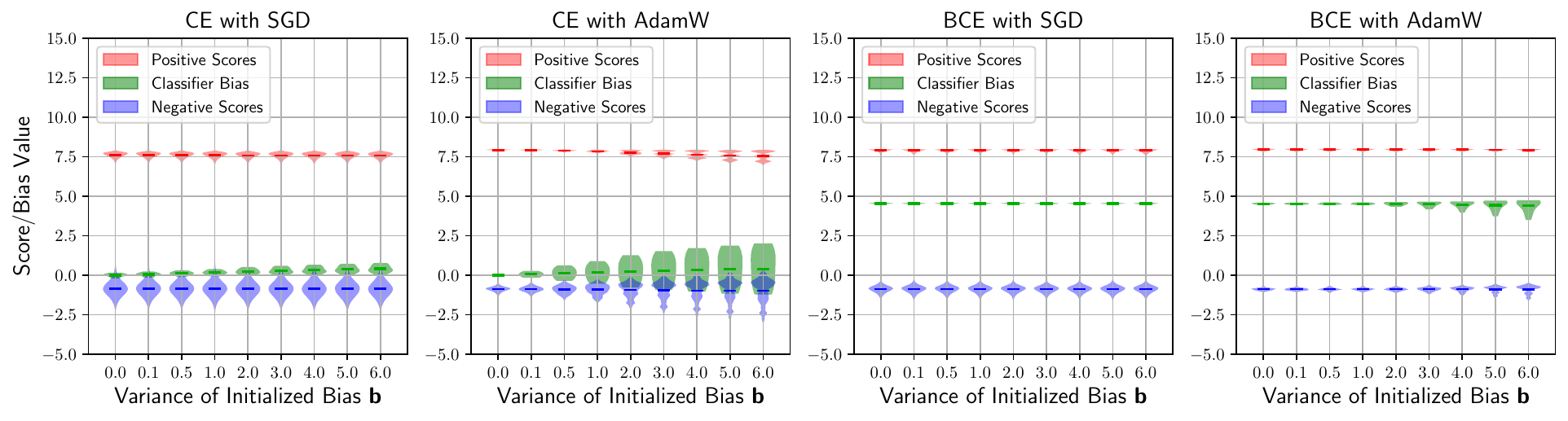}
            \label{fig_vit_cifar10_var_gamma_8}}

        \subfloat[ViT on CIFAR10, $\gamma=16$]{
            \includegraphics[width=0.95\textwidth,max width=\linewidth,max height=0.74\textheight,keepaspectratio]{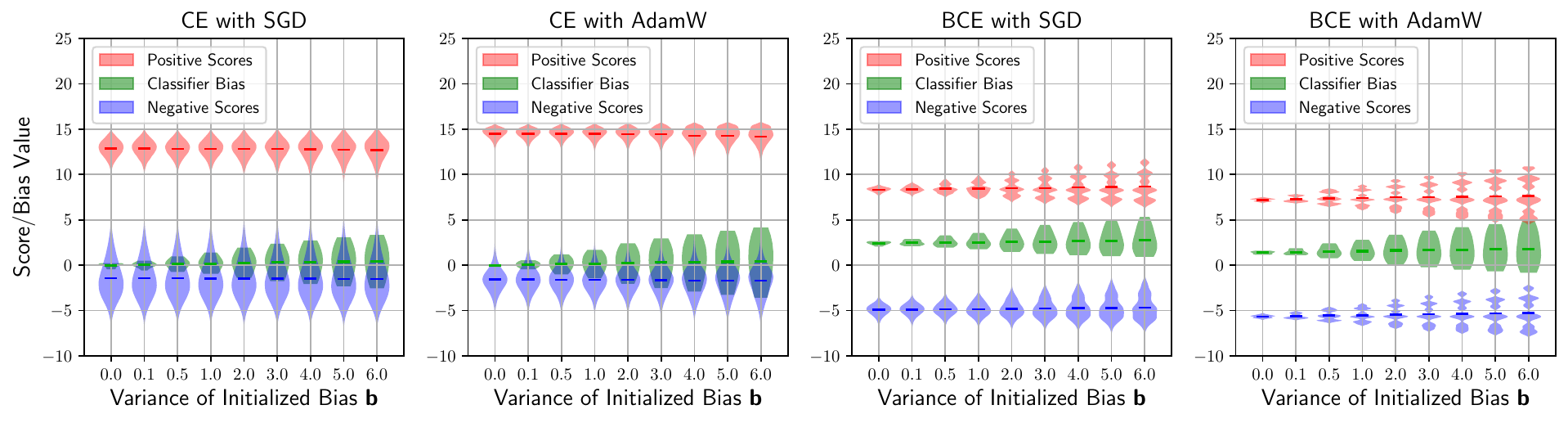}
            \label{fig_vit_cifar10_var_gamma_16}}

        \subfloat[ViT on CIFAR10, $\gamma=32$]{
            \includegraphics[width=0.95\textwidth,max width=\linewidth,max height=0.74\textheight,keepaspectratio]{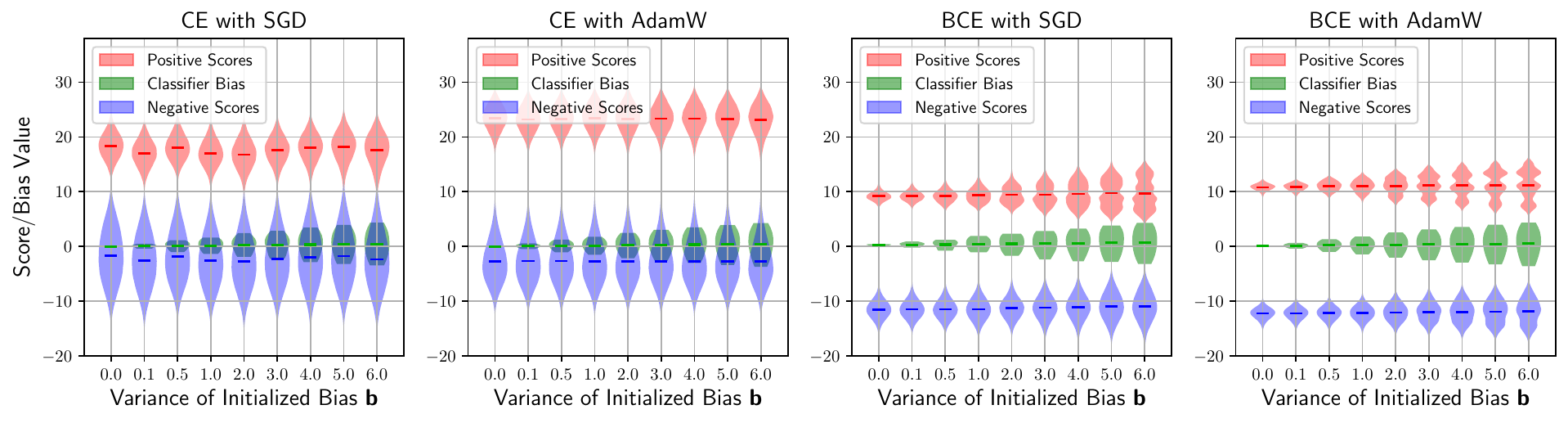}
            \label{fig_vit_cifar10_var_gamma_32}}
	\caption{The distribution of the final classifier bias and unbiased positive/negative decision scores of ViT trained on CIFAR10 with varying initial variance of the classifier bias. The different scale factors, $\gamma=8$ (1-st row), $\gamma=16$ (2-nd row), and $\gamma=32$ (3-rd row), also significantly impact the convergence of decision scores and the final classifier bias}
	\label{fig_result_of_vit_cifar10_bias_var-supp}
\end{figure}

Figures \ref{fig_result_of_resnet18_mnist_bias_var-supp} and \ref{fig_result_of_vit_cifar10_bias_var-supp} visually show the distributions of the final unbiased decision scores and classifier biases for ResNet18 trained on MNIST and that for ViT trained from scratch on CIFAR10.
For the different network architectures, optimizers, and scale factors, the BCE generally achieves better convergence compared to the CE, especially when $\gamma = 8$.
When $\gamma = 16, 32$ or $64$, the large normalized feature space makes it difficult for the models trained with either CE or BCE to converge, but the BCE-trained models usually yield better convergence.

\subsection{Impact of Scale Factor to Model Training}
\textbf{The scale factor $\gamma$.}\quad
To illustrate the effect of the scale factor for model training in the normalized feature space, we trained ResNet18 and ViT from scratch on the MNIST and CIFAR10 when $\gamma$ varies from $0.1$ to $64$.

\begin{figure}[!tbp]
	\centering
        \subfloat[ResNet18 on MNIST, $\text{Mean}(\bm b) = 0$, and $\text{Var}(\bm b) = 0$]{
            \includegraphics[width=0.95\textwidth,max width=\linewidth,max height=0.74\textheight,keepaspectratio]{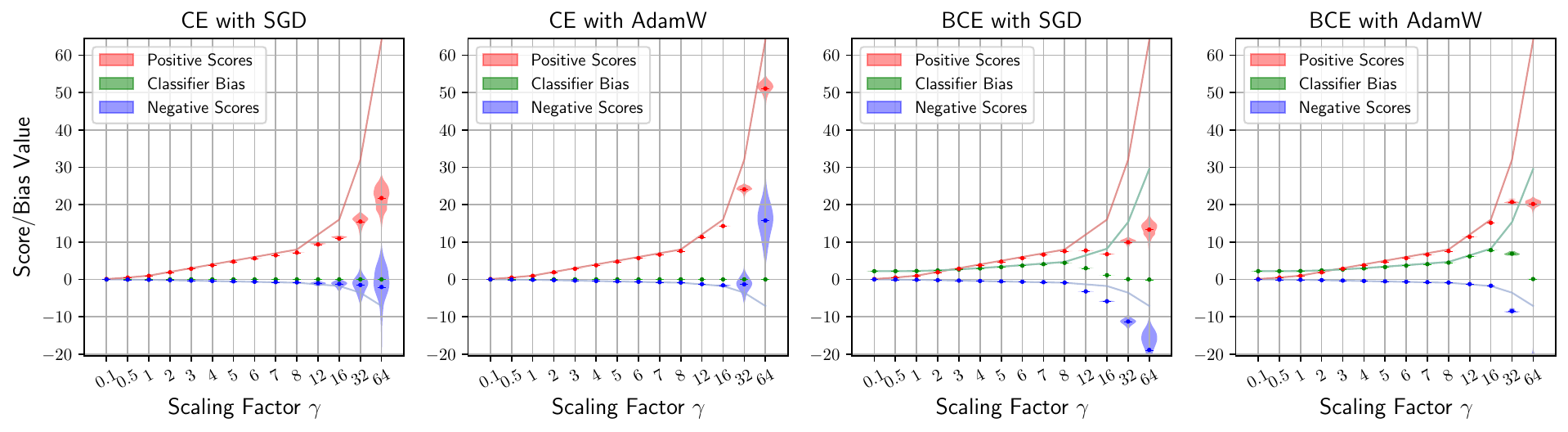}
            \label{fig_resnet18_mnist_gamma_bias_mean_0-supp}}
            
        \subfloat[ResNet18 on MNIST, $\text{Mean}(\bm b) = 10$, and $\text{Var}(\bm b) = 0$]{
            \includegraphics[width=0.95\textwidth,max width=\linewidth,max height=0.74\textheight,keepaspectratio]{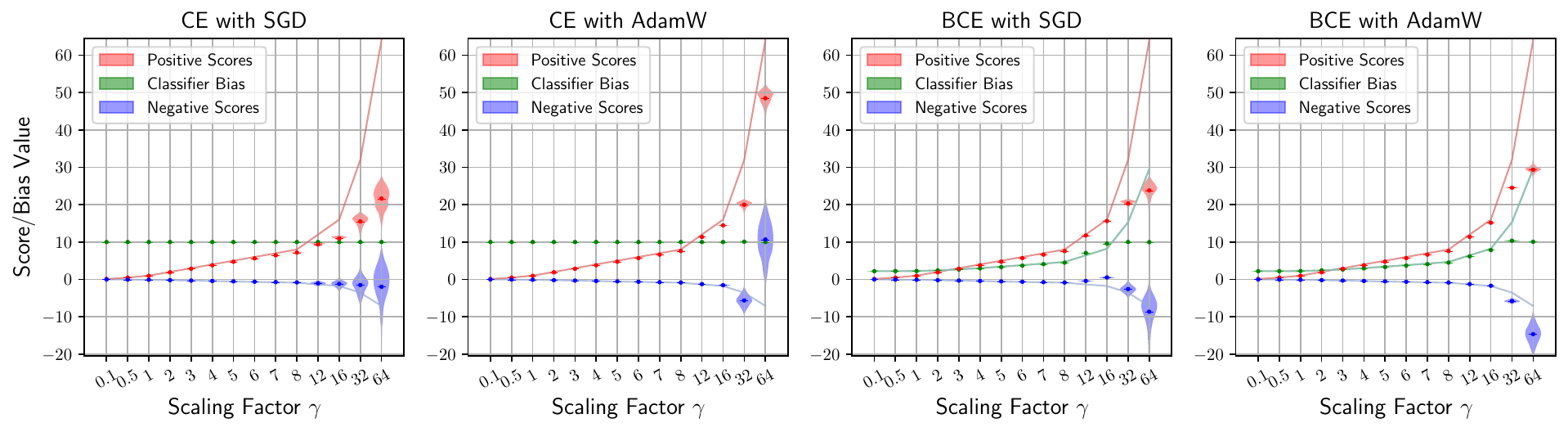}
            \label{fig_resnet18_mnist_gamma_bias_mean_10-supp}}
            
        \subfloat[ResNet18 on MNIST, $\text{Mean}(\bm b) = 20$, and $\text{Var}(\bm b) = 0$]{
            \includegraphics[width=0.95\textwidth,max width=\linewidth,max height=0.74\textheight,keepaspectratio]{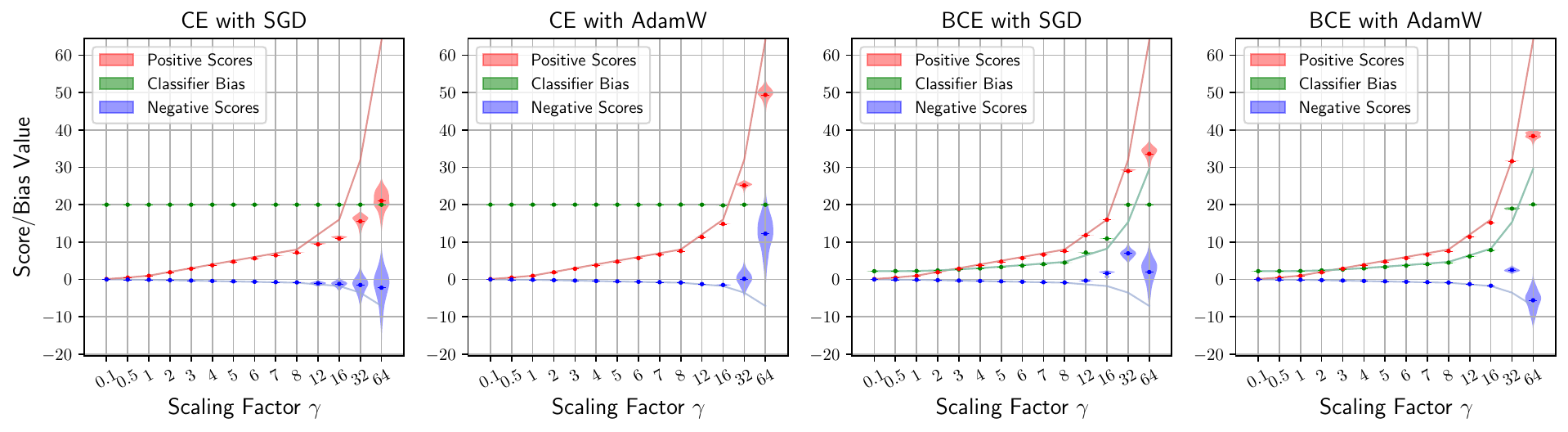}
            \label{fig_resnet18_mnist_gamma_bias_mean_20-supp}}
            
        \subfloat[ResNet18 on MNIST, $\text{Mean}(\bm b) = 30$, and $\text{Var}(\bm b) = 0$]{
            \includegraphics[width=0.95\textwidth,max width=\linewidth,max height=0.74\textheight,keepaspectratio]{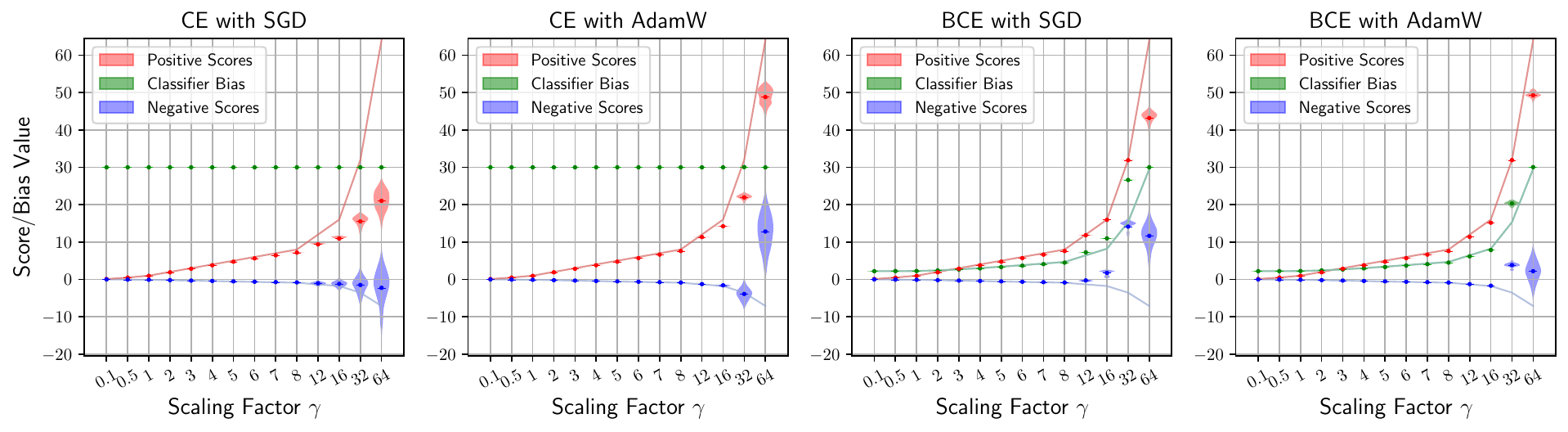}
            \label{fig_resnet18_mnist_gamma_bias_mean_30-supp}}
    \caption{The distribution of the final classifier biases and positive/negative unbiased decision scores of ResNet18 trained on MNIST by using various scale factors $\gamma$. The variance of initial classifier biases is set to 0, and their initial mean takes different values, $0, 10, 20$, and $30$, respectively}
	\label{fig_result_of_resnet18_mnist_gamma_1-supp}
\end{figure}

\begin{figure*}[!t]
	\centering
        \subfloat[ResNet18 on MNIST, $\text{Mean}(\bm b) = 5$, and $\text{Var}(\bm b) = 0$]{
            \includegraphics[width=0.95\textwidth,max width=\linewidth,max height=0.74\textheight,keepaspectratio]{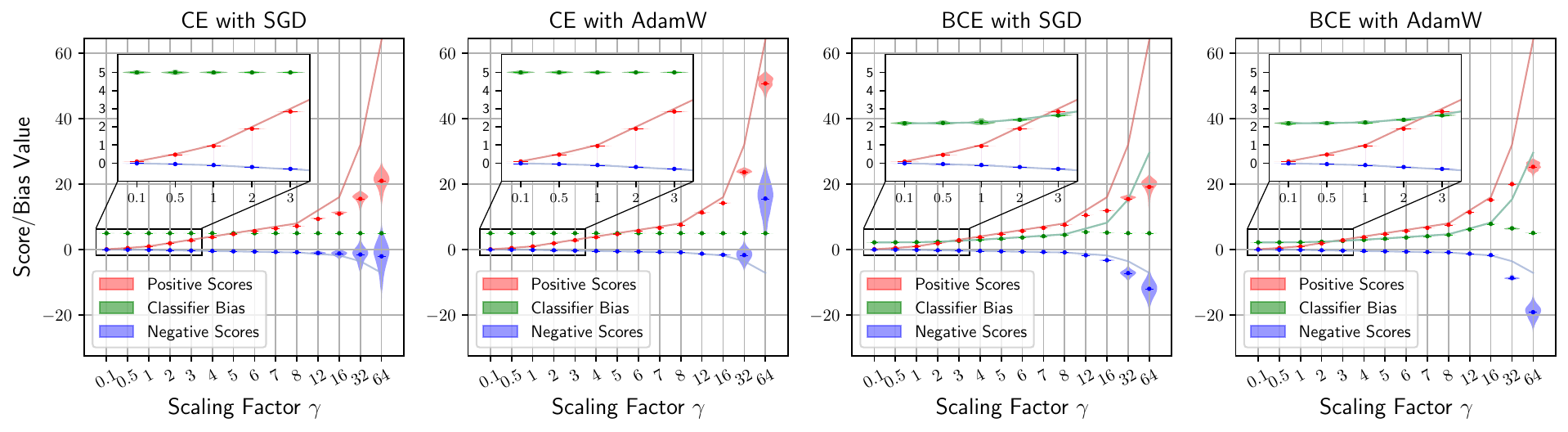}
            \label{fig_resnet18_mnist_gamma_bias_mean_5-supp}}
            
        \subfloat[ResNet18 on MNIST, $\text{Var}(\bm b) = 0$, while the mean ($\text{Mean}(\bm b)$) of initial classifier biases set to the theoretical value of the BCE loss (defined by \textit{Eq.} (\ref{eq_transcen_eq_b_bce}))]{
            \includegraphics[width=0.95\textwidth,max width=\linewidth,max height=0.74\textheight,keepaspectratio]{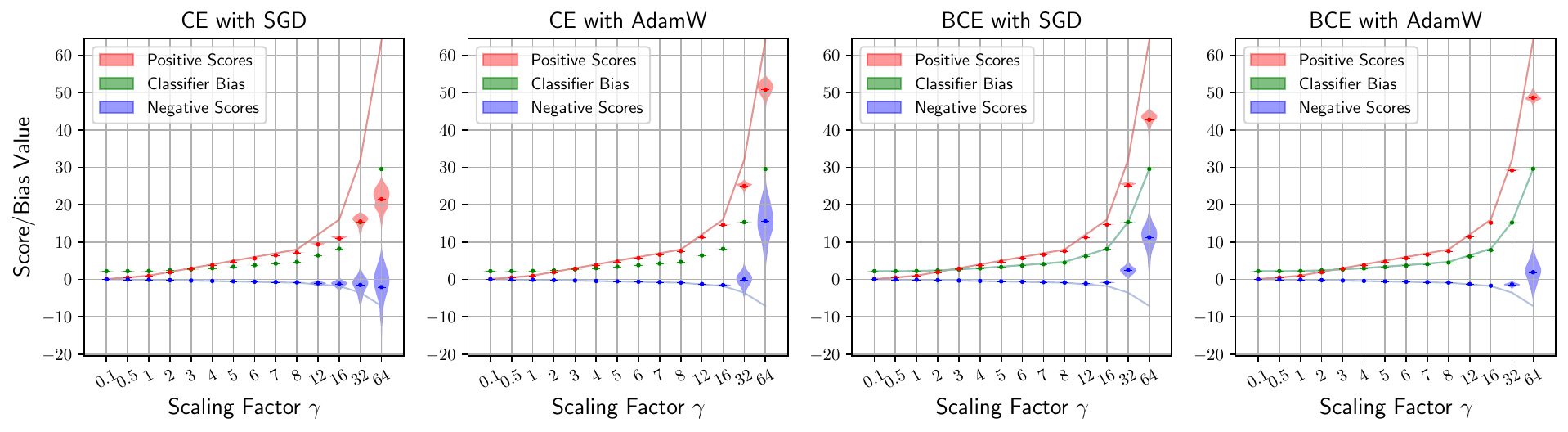}
            \label{fig_resnet18_mnist_gamma_bias_mean_theoretical-supp}}
    \caption{The distribution of the final classifier biases and positive/negative unbiased decision scores of ResNet18 trained on MNIST by using various scale factors $\gamma$. The variance of the initial classifier biases is set to 0}
	\label{fig_result_of_resnet18_mnist_gamma_2-supp}
\end{figure*}

\begin{figure}[!t]
    \centering{\includegraphics[width=0.95\textwidth,max width=\linewidth,max height=0.74\textheight,keepaspectratio]{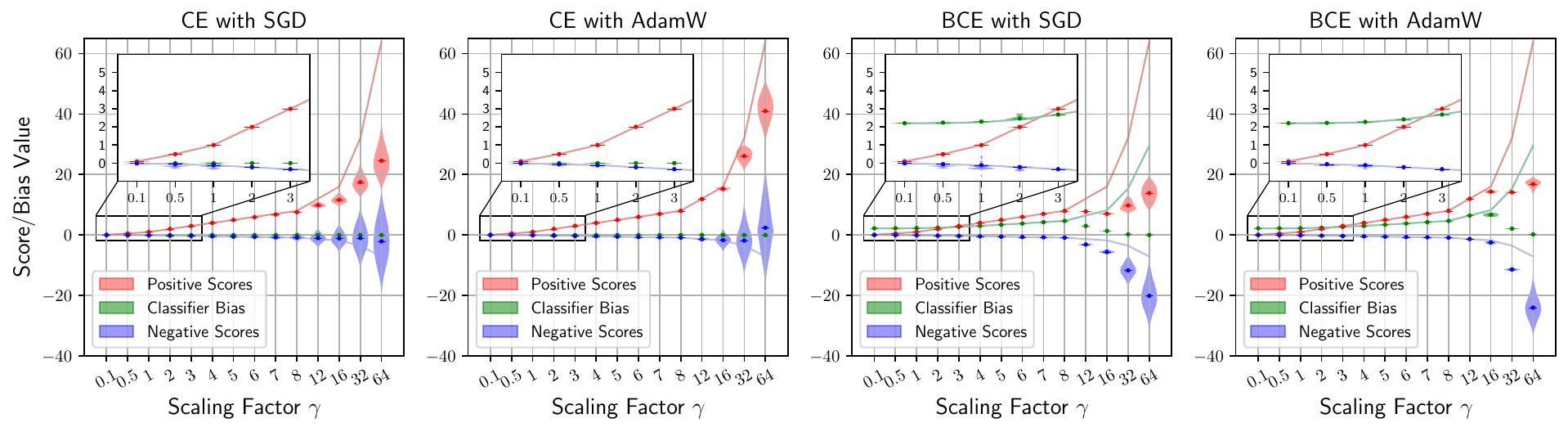}}
    \caption{The distribution of the final classifier biases and positive/negative unbiased decision scores of ResNet18 (without the final ReLU) trained on CIFAR10 by using various scale factors $\gamma$. The mean and variance of the initial classifier biases is both set to $0$}
    \label{resnet18_cifar_diff_gamma}
\end{figure}

\begin{figure*}[!t]
	\centering
        \subfloat[ViT on CIFAR10, $\text{Mean}(\bm b) = 0$, and $\text{Var}(\bm b) = 0$]{
            \includegraphics[width=0.95\textwidth,max width=\linewidth,max height=0.74\textheight,keepaspectratio]{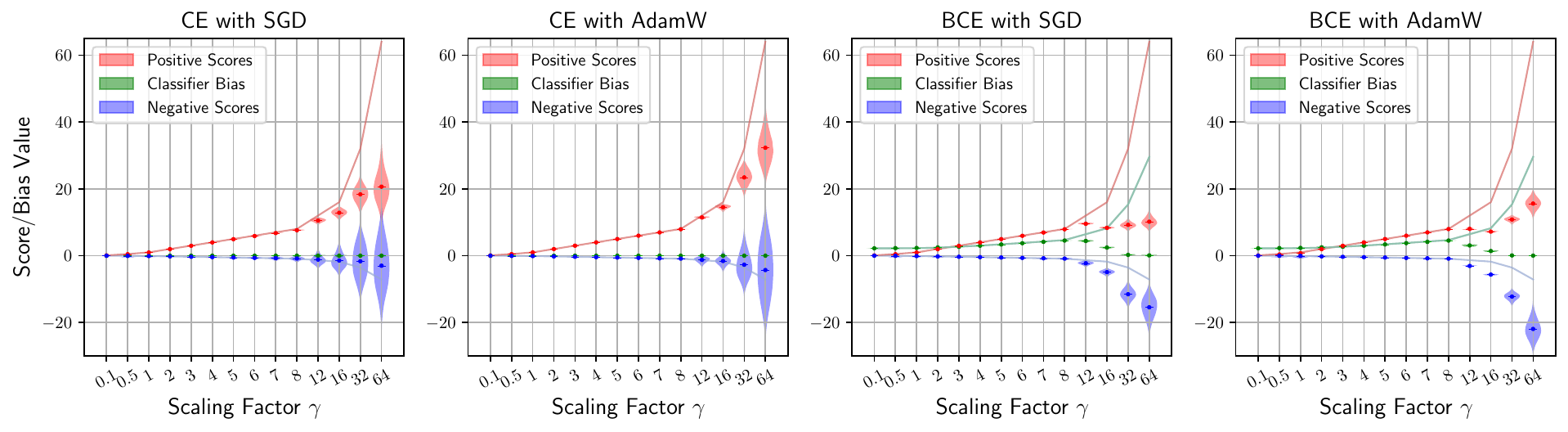}
            \label{fig_ViT_cifar10_gamma_bias_mean_0-supp}}
            
        \subfloat[ViT on CIFAR10, $\text{Mean}(\bm b) = 5$, and $\text{Var}(\bm b) = 0$]{
            \includegraphics[width=0.95\textwidth,max width=\linewidth,max height=0.74\textheight,keepaspectratio]{ViT_cifar10_gamma_bias_mean_5_zoom.pdf}
            \label{fig_ViT_cifar10_gamma_bias_mean_5-supp}}
            
        \subfloat[ViT on CIFAR10, $\text{Mean}(\bm b) = 10$, and $\text{Var}(\bm b) = 0$]{
            \includegraphics[width=0.95\textwidth,max width=\linewidth,max height=0.74\textheight,keepaspectratio]{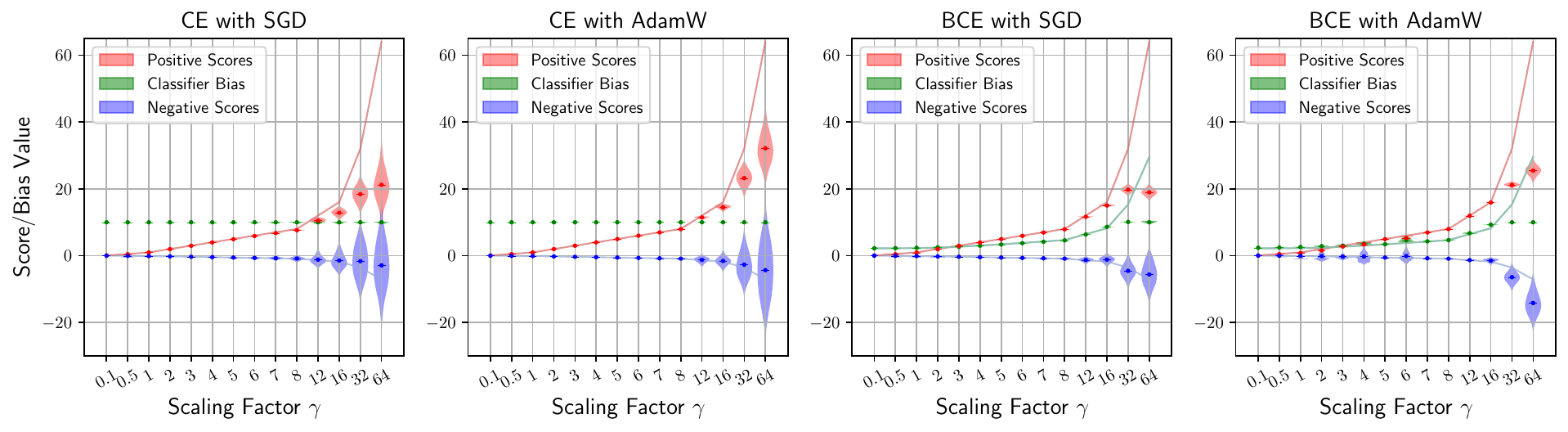}
            \label{fig_ViT_cifar10_gamma_bias_mean_10-supp}}
    \caption{The distribution of the final classifier biases and positive/negative unbiased decision scores of ViT trained on CIFAR10 by using various scale factors $\gamma$. The variance of the initial classifier biases is set to 0}
	\label{fig_result_of_ViT_cifar10_gamma-supp}
\end{figure*}

Fig. \ref{fig_result_of_resnet18_mnist_gamma_1-supp} - \ref{fig_result_of_ViT_cifar10_gamma-supp} visually show the distributions of the final decision scores and classifier biases for ResNet18 and ViT trained from scratch on the MNIST and CIFAR10 datasets. One can observe that, when $\gamma\leq8$ is small, the positive and negative unbiased decision scores, as well as the final classifier biases, of the models trained with BCE and CE losses closely align with the theoretical curves. However, as $\gamma$ increases, they increasingly deviate from their theoretical values.
Specifically, when $\gamma$ is reaching 32 or even 64, the distributions of the model's positive/negative unbiased decision scores span large ranges and do not converge.

Furthermore, one can find that, in the CE-trained models, the converged classifier biases largely depend on their initial values, providing no correlation with the final positive/negative unbiased decision scores. In contrast, with appropriately small $\gamma$, regardless of the initial value of the biases, the final classifier biases of the BCE-trained models converge to the theoretical values.

\subsection{Numerical Results of ResNet18 and ViT trained on CIFAR10}\label{sec_resnet_vit_on_cifar10-app}
In this section, Tables \ref{tab_resnet18_mnist_classification&properties} and \ref{tab_ViT_cifar10_classification&properties} provide the classification results and feature properties of ResNet18 and ViT trained from scratch on the MNIST and CIFAR10 datasets, when the scale factor $\gamma$ varies from $0.1$ to $64$.

It is evident that when $\gamma$ takes a very small value, it can significantly affect the model's classification performance, as the distance between the positive and negative decision scores becomes small. In this situation, even slight disturbance in the model training or small variance in the classifier biases can substantially impact the classification accuracy. However, based on the results of $\mathcal E_{\text{com}}$ and $\mathcal E_{\text{sep}}$, we conclude that poor classification accuracy $\mathcal A$ does not imply that the model has learned unsatisfactory sample features.

Moreover, as analyzed in Sec. \ref{section_nc}, the classifier biases of the BCE loss play a crucial role in the model training, explicitly constraining the model's positive and negative decision scores, which helps the model to learn features with better intra-class compactness and inter-class separability.
One can find that, when $\gamma > 5$, the BCE-trained models usually exhibit higher $\mathcal E_{\text{com}}$ and $\mathcal E_{\text{sep}}$ than the CE-trained models, which provides strong support for our theoretical conclusions. However, when $\gamma \leq 1$, due to the small normalized feature space, the  minor perturbations during the training could significantly affect the feature learning, overshadowing the role of classifier biases in the BCE-trained models, which prevents them from obtaining good feature properties.

\begin{tableorg}[!t]
	\tablebodyfont
	\centering
	\caption{The final positive/negative unbiased decision scores and classifier biases of ResNet18 (without the final ReLU) with various $\gamma$ on training data of CIFAR10 (mean $\pm$ standard deviation). The symbols $\boldsymbol{s}^{(kk)}$ and $\boldsymbol{s}^{(jk)}$ represent the positive and negative unbiased decision scores, respectively. The smaller standard deviation implies that the scores and biases tend to converge to similar values after training. The $\text{Mean}(\bm b) = 0$, and $\text{Var}(\bm b) = 0$}
	\label{tab_resnet18_mnist_parameter_converge}
		\begin{minipage}{\linewidth}\centering\textit{(a) ResNet18 on CIFAR10 with SGD optimizer}\par\smallskip 
			\centering
			\begin{tabular}{c|ccc|ccc}
				\toprule
				\multirow{2}{*}{$\gamma$} & \multicolumn{3}{c|}{CE with SGD} & \multicolumn{3}{c}{BCE with SGD} \\
				\cmidrule(lr){2-4}  \cmidrule(lr){5-7}
				& \multicolumn{1}{c}{$\boldsymbol{s}^{(kk)}$} & \multicolumn{1}{c}{$\boldsymbol{s}^{(jk)}$} & \multicolumn{1}{c|}{$\boldsymbol{b}$} & \multicolumn{1}{c}{$\boldsymbol{s}^{(kk)}$} & \multicolumn{1}{c}{$\boldsymbol{s}^{(jk)}$} & \multicolumn{1}{c}{$\boldsymbol{b}$} \\
				\midrule
				$0.1$ & $0.10 \pm 0.02$ & $-0.01 \pm 0.10$ & $0.00 \pm 0.02$ & $0.10 \pm 0.02$ & $-0.01 \pm 0.10$ & $2.20 \pm 0.02$ \\
				$0.5$ & $0.50 \pm 0.03$ & $-0.05 \pm 0.26$ & $0.00 \pm 0.02$ & $0.50 \pm 0.03$ & $-0.05 \pm 0.26$ & $2.24 \pm 0.02$ \\
				$1$ & $1.00 \pm 0.04$ & $-0.11 \pm 0.31$ & $0.00 \pm 0.04$ & $1.00 \pm 0.05$ & $-0.11 \pm 0.34$ & $2.29 \pm 0.03$ \\
				$2$ & $2.00 \pm 0.04$ & $-0.22 \pm 0.02$ & $0.00 \pm 0.00$ & $2.00 \pm 0.05$ & $-0.22 \pm 0.34$ & $2.46 \pm 0.14$ \\
				$3$ & $3.00 \pm 0.02$ & $-0.33 \pm 0.02$ & $0.00 \pm 0.00$ & $3.00 \pm 0.06$ & $-0.33 \pm 0.03$ & $2.68 \pm 0.00$ \\
				$4$ & $3.99 \pm 0.01$ & $-0.44 \pm 0.05$ & $0.00 \pm 0.00$ & $4.00 \pm 0.05$ & $-0.44 \pm 0.03$ & $3.02 \pm 0.00$ \\
				$5$ & $4.97 \pm 0.01$ & $-0.55 \pm 0.11$ & $0.00 \pm 0.00$ & $4.99 \pm 0.00$ & $-0.55 \pm 0.03$ & $3.40 \pm 0.00$ \\
				$6$ & $5.91 \pm 0.03$ & $-0.66 \pm 0.23$ & $0.00 \pm 0.02$ & $5.99 \pm 0.00$ & $-0.67 \pm 0.05$ & $3.81 \pm 0.00$ \\
				$7$ & $6.78 \pm 0.07$ & $-0.75 \pm 0.41$ & $0.00 \pm 0.04$ & $6.98 \pm 0.01$ & $-0.78 \pm 0.07$ & $4.23 \pm 0.00$ \\
				$8$ & $7.55 \pm 0.13$ & $-0.84 \pm 0.60$ & $0.00 \pm 0.03$ & $7.96 \pm 0.01$ & $-0.89 \pm 0.10$ & $4.65 \pm 0.01$ \\
				$12$ & $9.86 \pm 0.53$ & $-1.05 \pm 1.30$ & $0.00 \pm 0.02$ & $7.76 \pm 0.10$ & $-3.15 \pm 0.23$ & $3.00 \pm 0.01$ \\
				$16$ & $11.65 \pm 0.98$ & $-1.14 \pm 1.97$ & $0.00 \pm 0.01$ & $7.01 \pm 0.22$ & $-5.59 \pm 0.43$ & $1.35 \pm 0.03$ \\
				$32$ & $17.44 \pm 2.32$ & $-1.02 \pm 4.08$ & $0.00 \pm 0.01$ & $9.79 \pm 1.39$ & $-11.70 \pm 2.09$ & $0.20 \pm 0.04$ \\
				$64$ & $24.56 \pm 4.37$ & $-2.10 \pm 6.86$ & $0.00 \pm 0.02$ & $13.87 \pm 2.55$ & $-20.13 \pm 4.55$ & $0.02 \pm 0.16$ \\
				\bottomrule
			\end{tabular}
			\label{tab_resnet18_mnist_parameter_converge_a}
		\end{minipage}\par\medskip
		
			\begin{minipage}{\linewidth}\centering\textit{(b) ResNet18 on CIFAR10 with AdamW optimizer}\par\smallskip 
				\centering
				\begin{tabular}{c|ccc|ccc}
					\toprule
					\multirow{2}{*}{$\gamma$} & \multicolumn{3}{c|}{CE with AdamW} & \multicolumn{3}{c}{BCE with AdamW} \\
					\cmidrule(lr){2-4}  \cmidrule(lr){5-7}
					& \multicolumn{1}{c}{$\boldsymbol{s}^{(kk)}$} & \multicolumn{1}{c}{$\boldsymbol{s}^{(jk)}$} & \multicolumn{1}{c|}{$\boldsymbol{b}$} & \multicolumn{1}{c}{$\boldsymbol{s}^{(kk)}$} & \multicolumn{1}{c}{$\boldsymbol{s}^{(jk)}$} & \multicolumn{1}{c}{$\boldsymbol{b}$} \\
					\midrule
					$0.1$ & $0.10 \pm 0.02$ & $-0.01 \pm 0.10$ & $0.00 \pm 0.02$ & $0.10 \pm 0.02$ & $-0.01 \pm 0.10$ & $2.21 \pm 0.02$ \\
					$0.5$ & $0.50 \pm 0.00$ & $-0.06 \pm 0.15$ & $0.00 \pm 0.03$ & $0.50 \pm 0.00$ & $-0.06 \pm 0.18$ & $2.22 \pm 0.01$ \\
					$1$ & $1.00 \pm 0.00$ & $-0.11 \pm 0.18$ & $0.00 \pm 0.03$ & $1.00 \pm 0.00$ & $-0.11 \pm 0.25$ & $2.27 \pm 0.04$ \\
					$2$ & $2.00 \pm 0.00$ & $-0.22 \pm 0.00$ & $0.00 \pm 0.00$ & $2.00 \pm 0.00$ & $-0.22 \pm 0.00$ & $2.41 \pm 0.00$ \\
					$3$ & $3.00 \pm 0.00$ & $-0.33 \pm 0.00$ & $0.00 \pm 0.00$ & $3.00 \pm 0.00$ & $-0.33 \pm 0.00$ & $2.68 \pm 0.00$ \\
					$4$ & $4.00 \pm 0.00$ & $-0.44 \pm 0.01$ & $0.00 \pm 0.00$ & $4.00 \pm 0.00$ & $-0.44 \pm 0.01$ & $3.02 \pm 0.00$ \\
					$5$ & $5.00 \pm 0.00$ & $-0.56 \pm 0.02$ & $0.00 \pm 0.00$ & $5.00 \pm 0.00$ & $-0.56 \pm 0.01$ & $3.40 \pm 0.00$ \\
					$6$ & $6.00 \pm 0.00$ & $-0.67 \pm 0.03$ & $0.00 \pm 0.00$ & $6.00 \pm 0.00$ & $-0.67 \pm 0.02$ & $3.81 \pm 0.00$ \\
					$7$ & $7.00 \pm 0.00$ & $-0.78 \pm 0.03$ & $0.00 \pm 0.00$ & $7.00 \pm 0.00$ & $-0.78 \pm 0.02$ & $4.24 \pm 0.00$ \\
					$8$ & $7.99 \pm 0.00$ & $-0.89 \pm 0.05$ & $0.00 \pm 0.00$ & $8.00 \pm 0.00$ & $-0.89 \pm 0.03$ & $4.67 \pm 0.00$ \\
					$12$ & $11.90 \pm 0.08$ & $-1.32 \pm 0.45$ & $0.01 \pm 0.01$ & $11.99 \pm 0.00$ & $-1.33 \pm 0.05$ & $6.44 \pm 0.00$ \\
					$16$ & $15.26 \pm 0.37$ & $-1.68 \pm 1.52$ & $0.01 \pm 0.02$ & $14.35 \pm 0.32$ & $-2.48 \pm 0.37$ & $6.66 \pm 0.40$ \\
					$32$ & $26.09 \pm 1.84$ & $-1.85 \pm 4.64$ & $0.00 \pm 0.02$ & $14.14 \pm 0.21$ & $-11.40 \pm 0.33$ & $2.09 \pm 0.14$ \\
					$64$ & $40.98 \pm 5.59$ & $2.40 \pm 7.98$ & $0.00 \pm 0.06$ & $16.85 \pm 1.24$ & $-24.01 \pm 3.88$ & $0.17 \pm 0.04$ \\
					\bottomrule
				\end{tabular}
				\label{tab_resnet18_mnist_parameter_converge_b}
			\end{minipage}\par\medskip
\end{tableorg}

\begin{tableorg}[!t]
	\tablebodyfont
	\centering
	\caption{The final positive/negative unbiased decision scores and classifier biases of ViT with varying $\gamma$ on training data of CIFAR10 (mean $\pm$ standard deviation). The symbols $\boldsymbol{s}^{(kk)}$ and $\boldsymbol{s}^{(jk)}$ represent the positive and negative unbiased decision scores, respectively. The smaller standard deviation implies that the scores and biases tend to converge to similar values. The $\text{Mean}(\bm b) = 5$, and $\text{Var}(\bm b) = 0$}
	\label{tab_ViT_cifar10_parameter_converge}
		\begin{minipage}{\linewidth}\centering\textit{(a) ViT on CIFAR10 with SGD optimizer}\par\smallskip 
			\begin{tabular}{c|ccc|ccc}
				\toprule
				\multirow{2}{*}{$\gamma$} & \multicolumn{3}{c|}{CE with SGD} & \multicolumn{3}{c}{BCE with SGD} \\
				\cmidrule(lr){2-4}  \cmidrule(lr){5-7}
				& \multicolumn{1}{c}{$\boldsymbol{s}^{(kk)}$} & \multicolumn{1}{c}{$\boldsymbol{s}^{(jk)}$} & \multicolumn{1}{c|}{$\boldsymbol{b}$} & \multicolumn{1}{c}{$\boldsymbol{s}^{(kk)}$} & \multicolumn{1}{c}{$\boldsymbol{s}^{(jk)}$} & \multicolumn{1}{c}{$\boldsymbol{b}$} \\
				\midrule
				$0.1$ & $0.08 \pm 0.06$ & $-0.01 \pm 0.10$ & $5.00 \pm 0.01$ & $0.08 \pm 0.06$ & $-0.01 \pm 0.10$ & $2.20 \pm 0.02$ \\
				$0.5$ & $0.44 \pm 0.15$ & $-0.05 \pm 0.34$ & $5.00 \pm 0.02$ & $0.44 \pm 0.16$ & $-0.05 \pm 0.36$ & $2.25 \pm 0.01$ \\
				$1$ & $0.96 \pm 0.19$ & $-0.11 \pm 0.40$ & $5.00 \pm 0.02$ & $0.95 \pm 0.22$ & $-0.10 \pm 0.52$ & $2.35 \pm 0.05$ \\
				$2$ & $1.96 \pm 0.24$ & $-0.21 \pm 0.35$ & $5.00 \pm 0.08$ & $1.95 \pm 0.26$ & $-0.21 \pm 0.48$ & $2.48 \pm 0.12$ \\
				$3$ & $2.97 \pm 0.22$ & $-0.33 \pm 0.10$ & $5.00 \pm 0.00$ & $2.95 \pm 0.33$ & $-0.33 \pm 0.13$ & $2.66 \pm 0.00$ \\
				$4$ & $3.97 \pm 0.15$ & $-0.44 \pm 0.11$ & $5.00 \pm 0.01$ & $3.96 \pm 0.32$ & $-0.44 \pm 0.13$ & $2.99 \pm 0.00$ \\
				$5$ & $4.94 \pm 0.08$ & $-0.55 \pm 0.16$ & $5.00 \pm 0.02$ & $4.96 \pm 0.27$ & $-0.55 \pm 0.13$ & $3.35 \pm 0.01$ \\
				$6$ & $5.88 \pm 0.07$ & $-0.65 \pm 0.26$ & $5.00 \pm 0.04$ & $5.96 \pm 0.20$ & $-0.66 \pm 0.15$ & $3.74 \pm 0.01$ \\
				$7$ & $6.77 \pm 0.11$ & $-0.75 \pm 0.39$ & $5.00 \pm 0.07$ & $6.94 \pm 0.15$ & $-0.77 \pm 0.17$ & $4.14 \pm 0.01$ \\
				$8$ & $7.60 \pm 0.17$ & $-0.84 \pm 0.54$ & $5.00 \pm 0.11$ & $7.92 \pm 0.12$ & $-0.88 \pm 0.23$ & $4.55 \pm 0.02$ \\
				$12$ & $10.51 \pm 0.53$ & $-1.17 \pm 1.29$ & $5.00 \pm 0.22$ & $11.18 \pm 0.17$ & $-1.50 \pm 0.63$ & $5.70 \pm 0.14$ \\
				$16$ & $12.85 \pm 0.91$ & $-1.45 \pm 2.18$ & $5.00 \pm 0.22$ & $12.31 \pm 0.32$ & $-2.96 \pm 0.97$ & $5.61 \pm 0.17$ \\
				$32$ & $18.38 \pm 2.38$ & $-1.64 \pm 5.03$ & $5.00 \pm 0.11$ & $14.51 \pm 1.03$ & $-8.08 \pm 2.18$ & $5.11 \pm 0.10$ \\
				$64$ & $21.56 \pm 4.99$ & $-3.33 \pm 7.85$ & $5.00 \pm 0.13$ & $14.01 \pm 1.32$ & $-10.09 \pm 3.29$ & $5.14 \pm 0.58$ \\
				\bottomrule
			\end{tabular}
			\label{tab_ViT_cifar10_parameter_converge_a}
		\end{minipage}\par\medskip
		
			\begin{minipage}{\linewidth}\centering\textit{(b) ViT on CIFAR10 with AdamW optimizer}\par\smallskip 
				\centering
				\begin{tabular}{c|ccc|ccc}
					\toprule
					\multirow{2}{*}{$\gamma$} & \multicolumn{3}{c|}{CE with AdamW} & \multicolumn{3}{c}{BCE with AdamW} \\
					\cmidrule(lr){2-4}  \cmidrule(lr){5-7}
					& \multicolumn{1}{c}{$\boldsymbol{s}^{(kk)}$} & \multicolumn{1}{c}{$\boldsymbol{s}^{(jk)}$} & \multicolumn{1}{c|}{$\boldsymbol{b}$} & \multicolumn{1}{c}{$\boldsymbol{s}^{(kk)}$} & \multicolumn{1}{c}{$\boldsymbol{s}^{(jk)}$} & \multicolumn{1}{c}{$\boldsymbol{b}$} \\
					\midrule
					$0.1$ & $0.08 \pm 0.06$ & $-0.01 \pm 0.10$ & $5.00 \pm 0.01$ & $0.07 \pm 0.07$ & $-0.00 \pm 0.10$ & $2.20 \pm 0.01$ \\
					$0.5$ & $0.48 \pm 0.10$ & $-0.05 \pm 0.22$ & $5.00 \pm 0.01$ & $0.36 \pm 0.27$ & $-0.03 \pm 0.43$ & $2.28 \pm 0.06$ \\
					$1$ & $0.98 \pm 0.14$ & $-0.11 \pm 0.31$ & $5.00 \pm 0.04$ & $0.72 \pm 0.41$ & $-0.05 \pm 0.62$ & $2.38 \pm 0.11$ \\
					$2$ & $1.98 \pm 0.19$ & $-0.22 \pm 0.07$ & $5.00 \pm 0.00$ & $1.80 \pm 0.55$ & $-0.19 \pm 0.59$ & $2.52 \pm 0.14$ \\
					$3$ & $2.99 \pm 0.15$ & $-0.33 \pm 0.06$ & $5.00 \pm 0.01$ & $2.96 \pm 0.34$ & $-0.33 \pm 0.12$ & $2.67 \pm 0.00$ \\
					$4$ & $3.99 \pm 0.07$ & $-0.44 \pm 0.06$ & $5.00 \pm 0.01$ & $3.97 \pm 0.31$ & $-0.44 \pm 0.11$ & $3.00 \pm 0.00$ \\
					$5$ & $4.98 \pm 0.04$ & $-0.55 \pm 0.05$ & $5.00 \pm 0.01$ & $4.98 \pm 0.24$ & $-0.55 \pm 0.09$ & $3.37 \pm 0.00$ \\
					$6$ & $5.98 \pm 0.04$ & $-0.66 \pm 0.06$ & $5.00 \pm 0.01$ & $5.98 \pm 0.18$ & $-0.66 \pm 0.08$ & $3.76 \pm 0.00$ \\
					$7$ & $6.96 \pm 0.02$ & $-0.77 \pm 0.10$ & $5.00 \pm 0.02$ & $6.98 \pm 0.11$ & $-0.78 \pm 0.08$ & $4.14 \pm 0.02$ \\
					$8$ & $7.92 \pm 0.04$ & $-0.88 \pm 0.20$ & $5.00 \pm 0.03$ & $7.98 \pm 0.05$ & $-0.89 \pm 0.06$ & $4.53 \pm 0.01$ \\
					$12$ & $11.44 \pm 0.26$ & $-1.26 \pm 0.84$ & $5.00 \pm 0.03$ & $11.85 \pm 0.04$ & $-1.36 \pm 0.14$ & $5.82 \pm 0.09$ \\
					$16$ & $14.47 \pm 0.58$ & $-1.57 \pm 1.37$ & $5.00 \pm 0.02$ & $12.68 \pm 0.13$ & $-3.16 \pm 0.29$ & $5.32 \pm 0.06$ \\
					$32$ & $23.39 \pm 2.27$ & $-2.67 \pm 3.99$ & $5.00 \pm 0.01$ & $16.29 \pm 0.60$ & $-9.34 \pm 1.51$ & $5.02 \pm 0.01$ \\
					$64$ & $32.25 \pm 4.95$ & $-4.43 \pm 8.43$ & $5.00 \pm 0.01$ & $20.45 \pm 1.60$ & $-18.18 \pm 3.38$ & $5.01 \pm 0.01$ \\
					\bottomrule
				\end{tabular}
				\label{tab_ViT_cifar10_parameter_converge_b}
			\end{minipage}\par\medskip
\end{tableorg}
Furthermore, Tables \ref{tab_resnet18_mnist_parameter_converge} and \ref{tab_ViT_cifar10_parameter_converge} present the numerical results of the final positive/negative unbiased decision scores and classifier biases, for the CE- and BCE-trained ResNet18 and ViT, with $\gamma$ varying from $0.1$ to $64$.
When $\gamma$ is very small, the gap between the converged unbiased positive and negative decision scores is also very small.
When $\gamma \leq 0.5$, despite the ResNet18 trained on CIFAR10 being very close to neural collapse, as indicated by the relatively small variance of its positive and negative unbiased decision scores in Table \ref{tab_resnet18_mnist_parameter_converge}, which are essentially converged, even a slight variance in the classifier bias can still significantly affect its classification accuracy, as indicated by the significantly gap between the classification accuracy $\mathcal A$ and the unbiased one $\mathcal A^*$ in the Table \ref{tab_resnet18_mnist_classification&properties}.
When $\gamma \leq 2$, due to the disturbances during the model training, the ViT trained on CIFAR10 has not fully converged, which is reflected in the relatively large variance of its positive and negative unbiased decision scores. In fact, as shown in Figure \ref{fig_ViT_cifar10_gamma_bias_mean_5-supp}, when $\gamma \leq 2$, the distribution areas of the positive and negative unbiased decision scores of the ViT show significant overlapping, resulting in both low classification accuracy  $\mathcal A$ and the unbiased one $\mathcal A^*$ for the ViT in Table \ref{tab_ViT_cifar10_classification&properties}.

Conversely, a larger $\gamma$ expands the normalized feature space, leading to a large gap between the theoretical values of the converged positive and negative unbiased decision scores, which ensures robust classification; thus, the small perturbations such as the variance of the classifier biases cannot significantly affect the classification results. However, when
$\gamma$ is too large, the positive and negative decision scores struggle to converge, as evidenced by the large variance of the final unbiased decision scores, indicating that the sample features have poor intra-class compactness and inter-class separability.

\begin{tableorg}[tbp]
	\tablebodyfont
	\centering
	\caption{The classification accuracy and feature properties for ResNet18 on the test set of CIFAR10}
	\label{tab_resnet_cifar10_gamma_test_set}
	
	\noindent\textit{(a) CE with SGD --- smaller scales}\par\smallskip
	\begin{minipage}{\linewidth}\centering
		\begin{tabular}{c|ccccccc}
			\hline
			$\gamma $ & 0.1 & 0.5 & 1 & 2 & 3 & 4 & 5 \\
			$\mathcal A$ & 25.81 & 44.18 & 61.54 & 78.95 & 80.24 & 80.69 & 80.16 \\
			$\mathcal A^*$ & 25.62 & 46.94 & 61.60 & 78.94 & 80.24 & 80.68 & 80.16 \\
			$\mathcal E_{\text{com}}$ & 0.8887 & 0.8290 & 0.8240 & 0.8209 & 0.8293 & 0.8318 & 0.8232 \\
			$\mathcal E_{\text{sep}}$ & 0.5153 & 0.5347 & 0.5348 & 0.5348 & 0.5358 & 0.5363 & 0.5350 \\
			\hline
		\end{tabular}\par
	\end{minipage}\par\medskip
	\noindent\textit{(b) CE with SGD --- larger scales}\par\smallskip
	\begin{minipage}{\linewidth}\centering
		\begin{tabular}{c|ccccccc}
			\hline
			$\gamma $ & 6 & 7 & 8 & 12 & 16 & 32 & 64 \\
			$\mathcal A$ & 80.07 & 79.49 & 79.76 & 78.72 & 78.91 & 77.84 & 77.46 \\
			$\mathcal A^*$ & 80.08 & 79.49 & 79.76 & 78.71 & 78.91 & 77.79 & 77.44 \\
			$\mathcal E_{\text{com}}$ & 0.8136 & 0.7972 & 0.7823 & 0.7428 & 0.7251 & 0.6908 & 0.6849 \\
			$\mathcal E_{\text{sep}}$ & 0.5336 & 0.5273 & 0.5114 & 0.4159 & 0.3606 & 0.2658 & 0.2044 \\
			\hline
		\end{tabular}\par
	\end{minipage}\par\medskip
	\noindent\textit{(c) BCE with SGD --- smaller scales}\par\smallskip
	\begin{minipage}{\linewidth}\centering
		\begin{tabular}{c|ccccccc}
			\hline
			$\gamma $ & 0.1 & 0.5 & 1 & 2 & 3 & 4 & 5 \\
			$\mathcal A$ & 24.98 & 44.12 & 58.41 & 76.76 & 81.13 & 82.34 & 82.70 \\
			$\mathcal A^*$ & 26.55 & 45.67 & 59.71 & 76.75 & 81.13 & 82.34 & 82.70 \\
			$\mathcal E_{\text{com}}$ & 0.8891 & 0.8244 & 0.8254 & 0.8370 & 0.8324 & 0.8430 & 0.8464 \\
			$\mathcal E_{\text{sep}}$ & 0.5142 & 0.5349 & 0.5355 & 0.5343 & 0.5364 & 0.5375 & 0.5380 \\
			\hline
		\end{tabular}\par
	\end{minipage}\par\medskip
	\noindent\textit{(d) BCE with SGD --- larger scales}\par\smallskip
	\begin{minipage}{\linewidth}\centering
		\begin{tabular}{c|ccccccc}
			\hline
			$\gamma $ & 6 & 7 & 8 & 12 & 16 & 32 & 64 \\
			$\mathcal A$ & 82.54 & 82.62 & 81.83 & 81.63 & 82.14 & 81.69 & 82.54 \\
			$\mathcal A^*$ & 82.55 & 82.62 & 81.82 & 81.63 & 82.12 & 81.72 & 82.56 \\
			$\mathcal E_{\text{com}}$ & 0.8455 & 0.8457 & 0.8355 & 0.8226 & 0.8073 & 0.7491 & 0.7261 \\
			$\mathcal E_{\text{sep}}$ & 0.5379 & 0.5378 & 0.5362 & 0.4239 & 0.3680 & 0.3177 & 0.2934 \\
			\hline
		\end{tabular}\par
	\end{minipage}\par\medskip
\end{tableorg}


\subsection{Numerical Results of ResNet18 on Test Set of CIFAR10}
This paper focuses on the training set, providing an in-depth analysis of the differences between classification and feature learning. 
We believe that their difference can generalize to the test set, although there is still a lack of solid theoretical proof. 
Table \ref{tab_resnet_cifar10_gamma_test_set} offers the classification accuracy and feature properties of trained ResNet18 on the test set of CIFAR10, while the results on training set have been shown in Table \ref{tab_resnet18_mnist_classification&properties}.

From the table, one can observe that when $\gamma$ is small, such as 0.1 or 0.5, the models’ classification is poor, but the intra-class compactness and inter-class separability of the features are higher than 0.8 and 0.5, respectively. 
Conversely, when $\gamma$ is large, such as 32 or 64, the models achieve higher classification accuracy, but the feature properties significantly degrade. 
These results demonstrate that the differences between the tasks of classification and deep feature learning, as analyzed on the training set, also generalize to the test set.

\begin{tableorg}[tbp]
	\tablebodyfont
	\centering
	\caption{The classification accuracy (\%) and feature properties of ResNet18 (without the final ReLU) with various $\gamma$ on training data of CIFAR10. The accuracies of $\mathcal A$ and $\mathcal A^{*}$ are calculated using $\{\gamma\boldsymbol w_j^\top\boldsymbol h^{(k)}-b_j\}$ and $\{\gamma\boldsymbol w_j^\top\boldsymbol h^{(k)}\}$, respectively. The higher $\mathcal E_{\text{com}}$ and $\mathcal E_{\text{sep}}$ respectively indicate the smaller intra-class variability and the bigger inter-class separability. The $\text{Mean}(\bm b) = 0$, and $\text{Var}(\bm b) = 0$}
	\label{tab_resnet18_mnist_classification&properties}
	\noindent\textit{(a) Results with SGD}\par\smallskip
	\begin{minipage}{\linewidth}\centering
		\begin{tabular}{c|cccc|cccc}
			\toprule
			\multirow{2}{*}{$\gamma$} & \multicolumn{4}{c|}{CE with SGD} & \multicolumn{4}{c}{BCE with SGD} \\
			\cmidrule(lr){2-5}\cmidrule(lr){6-9}
			& $\mathcal{A}$ & $\mathcal{A}^{*}$ & $\mathcal E_{\text{com}}$ & $\mathcal E_{\text{sep}}$ & $\mathcal{A}$ & $\mathcal{A}^{*}$ & $\mathcal E_{\text{com}}$ & $\mathcal E_{\text{sep}}$ \\
			\midrule
			$0.1$ & $26.24$ & $27.27$ & $0.965$ & $0.526$ & $25.47$ & $27.77$ & $0.965$ & $0.526$ \\
			$0.5$ & $54.28$ & $57.00$ & $0.995$ & $0.554$ & $52.07$ & $54.75$ & $0.995$ & $0.554$ \\
			$1$ & $79.50$ & $80.01$ & $0.998$ & $0.555$ & $76.73$ & $77.76$ & $0.997$ & $0.555$ \\
			$2$ & $99.97$ & $99.97$ & $0.999$ & $0.555$ & $93.79$ & $93.78$ & $0.999$ & $0.553$ \\
			$3$ & $100.0$ & $100.0$ & $0.999$ & $0.555$ & $99.97$ & $99.97$ & $0.999$ & $0.555$ \\
			$4$ & $100.0$ & $100.0$ & $0.998$ & $0.555$ & $99.99$ & $99.99$ & $0.999$ & $0.555$ \\
			$5$ & $100.0$ & $100.0$ & $0.995$ & $0.555$ & $100.0$ & $100.0$ & $0.999$ & $0.555$ \\
			$6$ & $100.0$ & $100.0$ & $0.987$ & $0.553$ & $100.0$ & $100.0$ & $0.999$ & $0.555$ \\
			$7$ & $100.0$ & $100.0$ & $0.975$ & $0.547$ & $100.0$ & $100.0$ & $0.998$ & $0.555$ \\
			$8$ & $100.0$ & $100.0$ & $0.960$ & $0.535$ & $100.0$ & $100.0$ & $0.997$ & $0.554$ \\
			$12$ & $100.0$ & $100.0$ & $0.905$ & $0.454$ & $100.0$ & $100.0$ & $0.988$ & $0.459$ \\
			$16$ & $100.0$ & $100.0$ & $0.862$ & $0.394$ & $100.0$ & $100.0$ & $0.973$ & $0.410$ \\
			$32$ & $100.0$ & $100.0$ & $0.789$ & $0.288$ & $100.0$ & $100.0$ & $0.899$ & $0.354$ \\
			$64$ & $100.0$ & $100.0$ & $0.766$ & $0.221$ & $100.0$ & $100.0$ & $0.842$ & $0.313$ \\
			\bottomrule
		\end{tabular}\par
	\end{minipage}\par\medskip
	\noindent\textit{(b) Results with AdamW}\par\smallskip
	\begin{minipage}{\linewidth}\centering
		\begin{tabular}{c|cccc|cccc}
			\toprule
			\multirow{2}{*}{$\gamma$} & \multicolumn{4}{c|}{CE with AdamW} & \multicolumn{4}{c}{BCE with AdamW} \\
			\cmidrule(lr){2-5}\cmidrule(lr){6-9}
			& $\mathcal{A}$ & $\mathcal{A}^{*}$ & $\mathcal E_{\text{com}}$ & $\mathcal E_{\text{sep}}$ & $\mathcal{A}$ & $\mathcal{A}^{*}$ & $\mathcal E_{\text{com}}$ & $\mathcal E_{\text{sep}}$ \\
			\midrule
			$0.1$ & $19.96$ & $43.64$ & $0.986$ & $0.532$ & $19.93$ & $42.66$ & $0.980$ & $0.531$ \\
			$0.5$ & $70.00$ & $84.59$ & $1.0$ & $0.555$ & $60.00$ & $74.56$ & $1.0$ & $0.555$ \\
			$1$ & $90.00$ & $94.17$ & $1.0$ & $0.555$ & $80.00$ & $82.80$ & $1.0$ & $0.555$ \\
			$2$ & $100.0$ & $100.0$ & $1.0$ & $0.555$ & $100.0$ & $100.0$ & $1.0$ & $0.555$ \\
			$3$ & $100.0$ & $100.0$ & $1.0$ & $0.555$ & $100.0$ & $100.0$ & $1.0$ & $0.555$ \\
			$4$ & $100.0$ & $100.0$ & $1.0$ & $0.555$ & $100.0$ & $100.0$ & $1.0$ & $0.555$ \\
			$5$ & $100.0$ & $100.0$ & $1.0$ & $0.555$ & $100.0$ & $100.0$ & $1.0$ & $0.555$ \\
			$6$ & $100.0$ & $100.0$ & $1.0$ & $0.555$ & $100.0$ & $100.0$ & $1.0$ & $0.555$ \\
			$7$ & $100.0$ & $100.0$ & $1.0$ & $0.555$ & $100.0$ & $100.0$ & $1.0$ & $0.555$ \\
			$8$ & $100.0$ & $100.0$ & $0.999$ & $0.555$ & $100.0$ & $100.0$ & $1.0$ & $0.555$ \\
			$12$ & $100.0$ & $100.0$ & $0.992$ & $0.555$ & $100.0$ & $100.0$ & $0.999$ & $0.555$ \\
			$16$ & $100.0$ & $100.0$ & $0.957$ & $0.549$ & $100.0$ & $100.0$ & $0.998$ & $0.527$ \\
			$32$ & $100.0$ & $100.0$ & $0.883$ & $0.478$ & $100.0$ & $100.0$ & $0.993$ & $0.403$ \\
			$64$ & $100.0$ & $100.0$ & $0.809$ & $0.339$ & $100.0$ & $100.0$ & $0.928$ & $0.371$ \\
			\bottomrule
		\end{tabular}\par
	\end{minipage}\par\medskip
\end{tableorg}

\begin{tableorg}[tbp]
	\tablebodyfont
	\centering
	\caption{The classification accuracy (\%) and feature properties of ViT with various $\gamma$ on training data of CIFAR10. The accuracies of $\mathcal A$ and $\mathcal A^{*}$ are calculated using the decision score $\{\gamma\boldsymbol w_j^\top\boldsymbol h^{(k)}-b_j\}$ and $\{\gamma\boldsymbol w_j^\top\boldsymbol h^{(k)}\}$, respectively. The higher $\mathcal E_{\text{com}}$ and $\mathcal E_{\text{sep}}$ respectively indicate the smaller intra-class variability and the bigger inter-class separability. The $\text{Mean}(\bm b) = 5$, and $\text{Var}(\bm b) = 0$}
	\label{tab_ViT_cifar10_classification&properties}
	\noindent\textit{(a) Results with SGD}\par\smallskip
	\begin{minipage}{\linewidth}\centering
		\begin{tabular}{c|cccc|cccc}
			\toprule
			\multirow{2}{*}{$\gamma$} & \multicolumn{4}{c|}{CE with SGD} & \multicolumn{4}{c}{BCE with SGD} \\
			\cmidrule(lr){2-5}\cmidrule(lr){6-9}
			& $\mathcal{A}$ & $\mathcal{A}^{*}$ & $\mathcal E_{\text{com}}$ & $\mathcal E_{\text{sep}}$ & $\mathcal{A}$ & $\mathcal{A}^{*}$ & $\mathcal E_{\text{com}}$ & $\mathcal E_{\text{sep}}$ \\
			\midrule
			$0.1$ & $27.19$ & $27.20$ & $0.826$ & $0.523$ & $26.42$ & $26.45$ & $0.823$ & $0.524$ \\
			$0.5$ & $45.63$ & $45.65$ & $0.892$ & $0.542$ & $41.51$ & $41.54$ & $0.886$ & $0.542$ \\
			$1$ & $69.31$ & $69.32$ & $0.962$ & $0.551$ & $57.02$ & $57.00$ & $0.954$ & $0.549$ \\
			$2$ & $92.14$ & $92.14$ & $0.982$ & $0.552$ & $90.57$ & $90.57$ & $0.977$ & $0.551$ \\
			$3$ & $99.52$ & $99.52$ & $0.990$ & $0.554$ & $98.94$ & $98.94$ & $0.985$ & $0.554$ \\
			$4$ & $99.87$ & $99.87$ & $0.992$ & $0.555$ & $99.43$ & $99.43$ & $0.990$ & $0.554$ \\
			$5$ & $99.98$ & $99.98$ & $0.988$ & $0.554$ & $99.75$ & $99.75$ & $0.992$ & $0.555$ \\
			$6$ & $100.0$ & $100.0$ & $0.981$ & $0.553$ & $99.90$ & $99.90$ & $0.993$ & $0.555$ \\
			$7$ & $100.0$ & $100.0$ & $0.969$ & $0.551$ & $99.96$ & $99.96$ & $0.992$ & $0.555$ \\
			$8$ & $100.0$ & $100.0$ & $0.954$ & $0.549$ & $99.98$ & $99.98$ & $0.990$ & $0.554$ \\
			$12$ & $100.0$ & $100.0$ & $0.890$ & $0.542$ & $100.0$ & $100.0$ & $0.972$ & $0.541$ \\
			$16$ & $100.0$ & $100.0$ & $0.837$ & $0.536$ & $100.0$ & $100.0$ & $0.950$ & $0.501$ \\
			$32$ & $100.0$ & $100.0$ & $0.722$ & $0.513$ & $100.0$ & $100.0$ & $0.882$ & $0.470$ \\
			$64$ & $100.0$ & $100.0$ & $0.716$ & $0.474$ & $100.0$ & $100.0$ & $0.896$ & $0.447$ \\
			\bottomrule
		\end{tabular}\par
	\end{minipage}\par\medskip
	\noindent\textit{(b) Results with AdamW}\par\smallskip
	\begin{minipage}{\linewidth}\centering
		\begin{tabular}{c|cccc|cccc}
			\toprule
			\multirow{2}{*}{$\gamma$} & \multicolumn{4}{c|}{CE with AdamW} & \multicolumn{4}{c}{BCE with AdamW} \\
			\cmidrule(lr){2-5}\cmidrule(lr){6-9}
			& $\mathcal{A}$ & $\mathcal{A}^{*}$ & $\mathcal E_{\text{com}}$ & $\mathcal E_{\text{sep}}$ & $\mathcal{A}$ & $\mathcal{A}^{*}$ & $\mathcal E_{\text{com}}$ & $\mathcal E_{\text{sep}}$ \\
			\midrule
			$0.1$ & $23.45$ & $23.47$ & $0.791$ & $0.519$ & $20.57$ & $20.58$ & $0.760$ & $0.513$ \\
			$0.5$ & $65.21$ & $65.24$ & $0.961$ & $0.551$ & $24.47$ & $24.35$ & $0.769$ & $0.519$ \\
			$1$ & $79.29$ & $79.28$ & $0.983$ & $0.553$ & $35.28$ & $35.25$ & $0.762$ & $0.513$ \\
			$2$ & $99.23$ & $99.23$ & $0.991$ & $0.555$ & $76.80$ & $76.82$ & $0.905$ & $0.542$ \\
			$3$ & $99.79$ & $99.79$ & $0.997$ & $0.555$ & $98.93$ & $98.93$ & $0.988$ & $0.554$ \\
			$4$ & $99.98$ & $99.98$ & $0.997$ & $0.555$ & $99.48$ & $99.48$ & $0.993$ & $0.555$ \\
			$5$ & $100.0$ & $100.0$ & $0.997$ & $0.555$ & $99.80$ & $99.80$ & $0.997$ & $0.555$ \\
			$6$ & $100.0$ & $100.0$ & $0.997$ & $0.555$ & $99.92$ & $99.92$ & $0.997$ & $0.555$ \\
			$7$ & $100.0$ & $100.0$ & $0.995$ & $0.555$ & $99.98$ & $99.98$ & $0.997$ & $0.555$ \\
			$8$ & $100.0$ & $100.0$ & $0.991$ & $0.555$ & $100.0$ & $100.0$ & $0.997$ & $0.555$ \\
			$12$ & $100.0$ & $100.0$ & $0.959$ & $0.550$ & $100.0$ & $100.0$ & $0.995$ & $0.553$ \\
			$16$ & $100.0$ & $100.0$ & $0.920$ & $0.544$ & $100.0$ & $100.0$ & $0.989$ & $0.500$ \\
			$32$ & $100.0$ & $100.0$ & $0.795$ & $0.520$ & $100.0$ & $100.0$ & $0.940$ & $0.422$ \\
			$64$ & $100.0$ & $100.0$ & $0.728$ & $0.369$ & $100.0$ & $100.0$ & $0.876$ & $0.311$ \\
			\bottomrule
		\end{tabular}\par
	\end{minipage}\par\medskip
\end{tableorg}

\clearpage
\subsection{Experimental verification for Theorems \ref{them:CE_decay_of_decision_score} and \ref{them:BCE_decay_of_decision_score}} \label{gamma_decay_experiment}
In Sec. \ref{section_normalization_practical}, we theoretically analyze how the convergence rate of unbiased positive and negative decision scores changes as $\gamma$ linearly increases or decreases, which reflect the convergence behavior of the CE and BCE losses.
Specifically, (1) when $\gamma$ decreases linearly toward $0$, the convergence rate of the losses decays linearly, meaning that the number of training epochs required to reach their minima increases linearly;
(2) when CC I or II is satisfied, as $\gamma$ increases linearly, the convergence rate of the losses decays exponentially, meaning that the number of training epochs required to reach their minima increases exponentially.

\begin{figure}[!t]
\color{black}
	\centering
        \subfloat[fitted functions for the CE loss]{
            \includegraphics*[width=0.44\linewidth, viewport=33 5 455 320,max height=0.74\textheight,keepaspectratio]{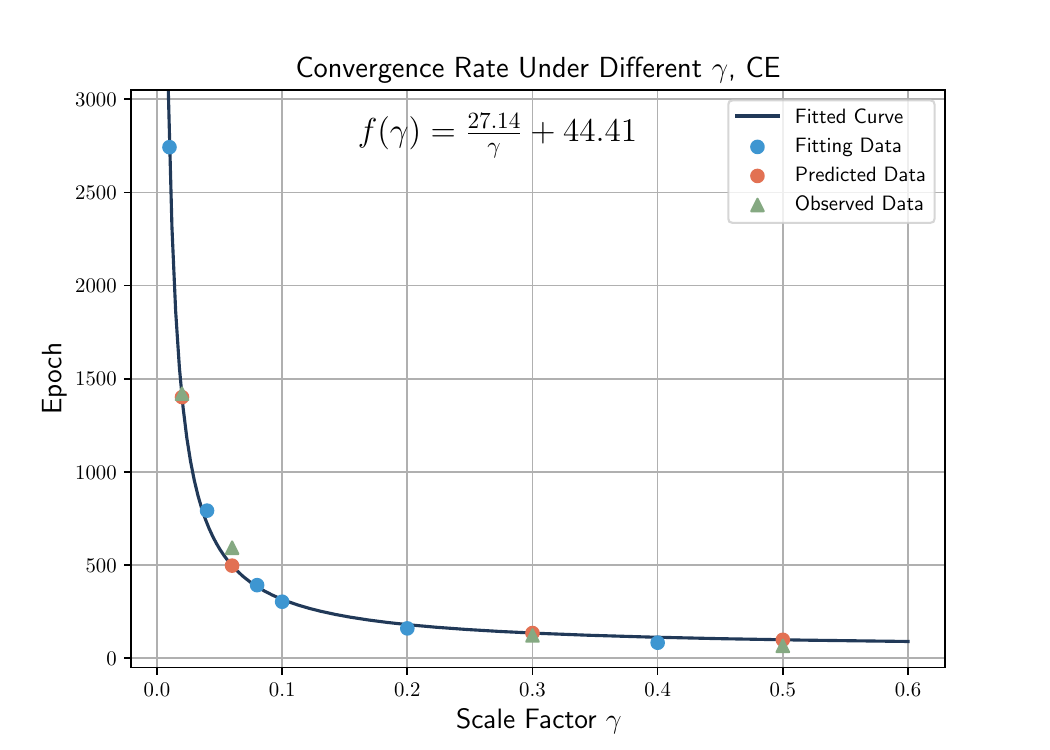}
            \includegraphics*[width=0.44\linewidth, viewport=33 5 455 320,max height=0.74\textheight,keepaspectratio]{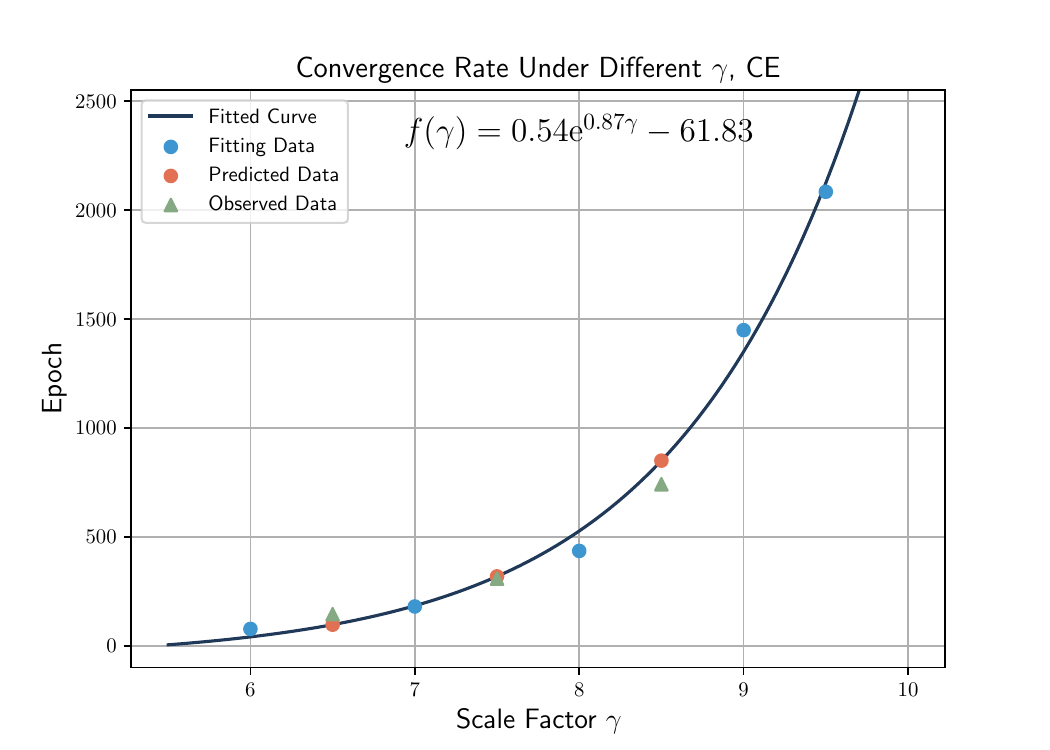}
            \label{gamma_theorem_figure_a}}
            
        \subfloat[fitted functions for the BCE loss]{
            \includegraphics*[width=0.44\linewidth, viewport=33 5 455 320,max height=0.74\textheight,keepaspectratio]{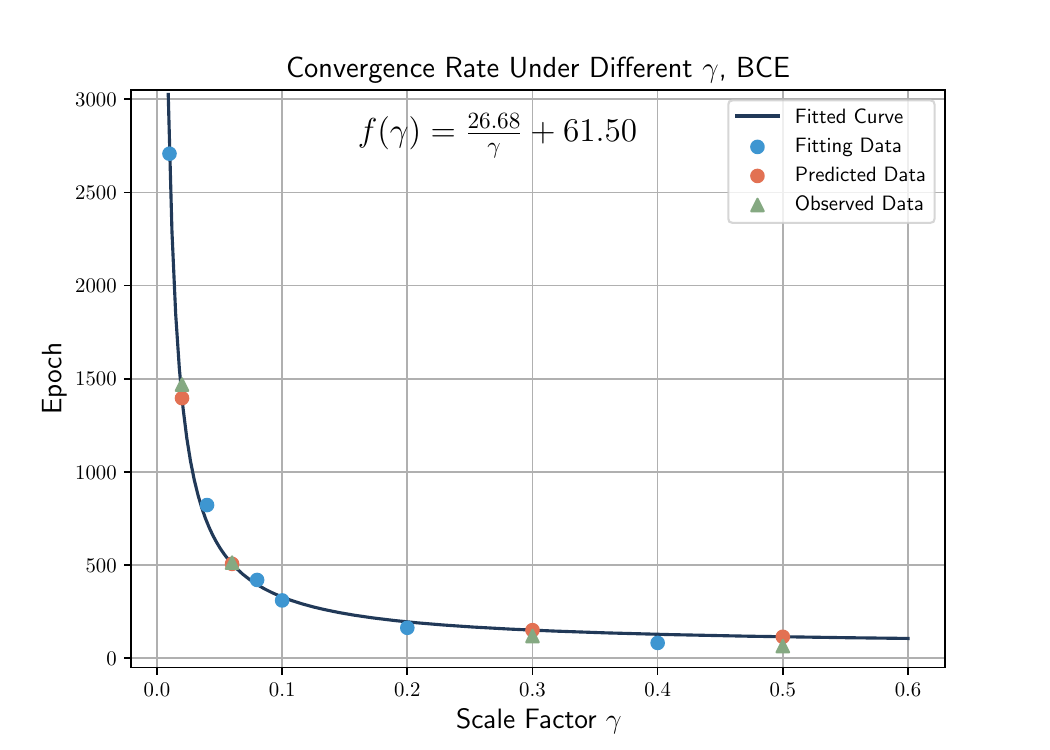}
            \includegraphics*[width=0.44\linewidth, viewport=33 5 455 320,max height=0.74\textheight,keepaspectratio]{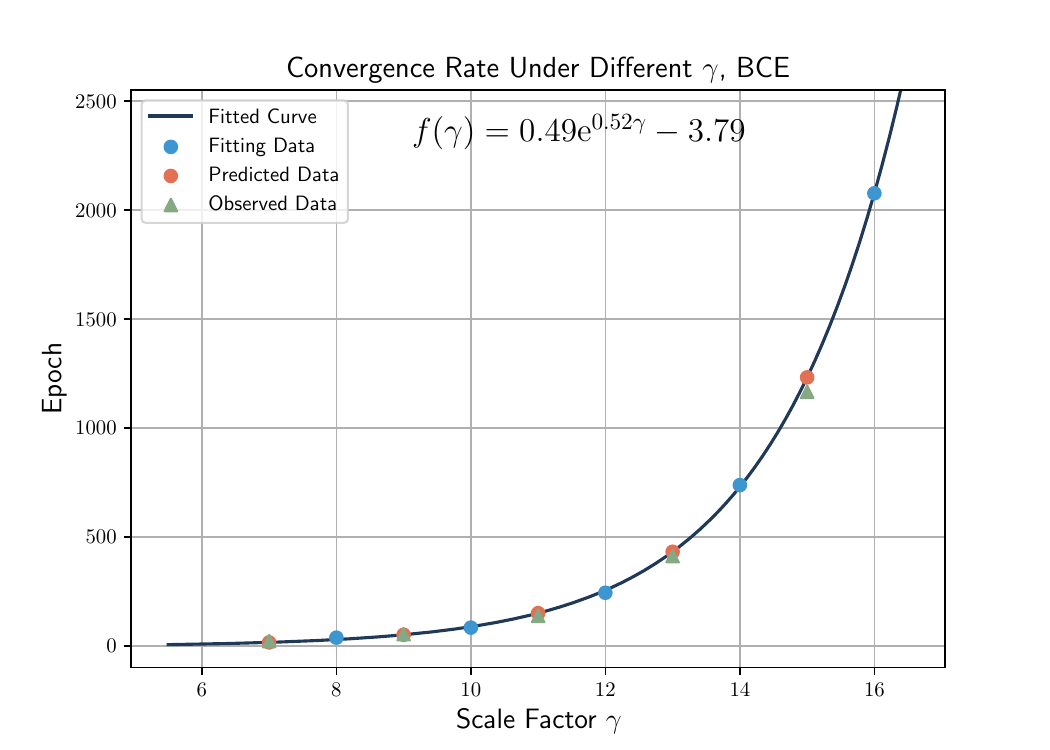}
            \label{gamma_theorem_figure_b}}
    \caption{The fitted functions describing the variation of epoch number with respect to $\gamma$ for CE (top) and BCE (bottom) losses.
    The epoch numbers in the figure indicate the training duration required to achieve minimum values (i.e., the criterion in Eq. (\ref{gamma_stop_condition})) of the two losses for training ResNet18 on MNIST, with different scale factors $\gamma$.
    The blue dots represent that the data was used for function fitting, the red dots represent the predicted values, and the green triangles represent the observed data in experiments}
	\label{gamma_theorem_figure}
\end{figure}

To directly verify the above conclusions, we set the NC structure at the minimum point as the stopping criterion, and train ResNet18 on MNIST with varying gamma. In these experiments, the stopping criterion is set as
\begin{equation}
\begin{aligned}
    &\Big|\boldsymbol{w}_k^{\top}\boldsymbol{h}_i^{(k)}-1\Big| \leq t,\quad \forall k\in[K], i\in[n], \\
    &\Big|\boldsymbol{w}_j^{\top}\boldsymbol{h}_i^{(k)}+\frac{1}{K-1}\Big| \leq t,\quad \forall j\neq k\in[K],i\in[n]. \label{gamma_stop_condition}
\end{aligned}
\end{equation}
When $\gamma < 5$, we set $t = 0.001$; when $\gamma\geq 5$, we set $t=0.0035$.
Moreover, we use SGD optimizer and fix the learning rate at $0.01$.
Table \ref{gamma_theorem_table_icml} and Figure \ref{gamma_theorem_figure} show the epoch numbers required by CE and BCE losses when they satisfying the criterion.

For cases with $\gamma < 0.5$ and $\gamma > 5$, we took six and five data points to respectively fit the dependence of epoch number on the scale factor $\gamma$ using linear, quadratic, and cubic polynomial functions, as well as exponential and logarithmic functions in terms of $\frac{1}{\gamma}$ (for $\gamma < 0.5$) and $\gamma$ (for $\gamma > 1$). 
The remaining data points were used to evaluate the predication quality of the fitted functions, and the evaluate metrics are coefficient of determination $R^2$ and root mean squared error (RMSE)
\begin{align}
    R^2 &=1-\frac{\sum_{i=1}^n(y_i-\hat{y}_i)^2}{\sum_{i=1}^n (y_i-\bar{y})^2}, \\
    \text{RMSE} &= \sqrt{\frac{1}{n}\sum_{i=1}^n (y_i-\hat{y}_i)^2},
\end{align}
where $\{y_i\}_{i=1}^n$ are the observed true data points, $\{\hat{y}_i\}_{i=1}^n$ are the predicted values by the fitted function, and $\bar{y}$ denotes the mean of the true values.
Tables \ref{small_gamma_fit_results-supp} and \ref{big_gamma_fit_results-supp} show the fitting results, together with the metrics of $R^2$ and RMSE on the fitting and prediction data points.

It can be seen that when $\gamma<0.5$, a linear function in $1/\gamma$ (i.e., an inverse proportional function) already provides a very good fitting to the number of epochs required by CE and BCE loss to reach their minima. In particular, their predictions achieve very high $R^2$ and low RMSE. Although for BCE a quadratic polynomial in $1/\gamma$ yields a better overall fitting, by examining its coefficients together with the $R^2$ and RMSE on both the fitted and predicted data, one can clearly observe that the inverse-proportional term is the main factor determining how the required epoch number varies with $\gamma$. 

For $\gamma>5$, it can be observed that linear and logarithmic functions do not conform to the trend of epoch number varying with $\gamma$; consequently, their fitting results are poor. Compared to quadratic and cubic polynomials, the exponential function achieved superior fitting results, particularly on the prediction data. This is consistent with the conclusion in the paper that the convergence rate of the losses exhibits exponential decay as $\gamma$ approaches infinity. Furthermore, it is noteworthy that as the polynomial order increases, the corresponding fitting results also improve progressively. Given that exponential functions can be expanded through Taylor series, the phenomenon is reasonable.

\begin{table}[!htp]
    \centering
    \caption{Fitted functions and their evaluation results for the epoch number used by the CE and BCE with $\gamma < 0.5$}
    \label{small_gamma_fit_results-supp}
    \begin{tabular}{c|c|cccc}
        \toprule
        \multirow{2}{*}{Loss} & \multirow{2}{*}{Fitted function} & \multicolumn{2}{c}{Fitting data} & \multicolumn{2}{c}{Predicted data} \\
        \cmidrule(lr){3-4} \cmidrule(lr){5-6}
        &  & $R^2 \uparrow$ & RMSE $\downarrow$& $R^2 \uparrow$ & RMSE $\downarrow$ \\
        \hline
        \multirow{5}{*}{CE} & $27.14/\gamma+44.41$ & $0.9987$ & $32.93$ & $\boldsymbol{0.9908}$ & $\boldsymbol{52.08}$ \\
        & $-0.06/\gamma^2+33.05/\gamma-7.52$ & $0.9999$ & $7.92$ & $0.9897$ & $55.16$ \\
        & $-0.002/\gamma^3+0.16/\gamma^2+2.83/\gamma+1.20$ & $\boldsymbol{1.0000}$ & $\boldsymbol{3.39}$ & $0.9638$ & $103.11$ \\
        & $4002.21\exp(-45.08\gamma)+185.11$ & $0.9943$ & $69.52$ & $0.8388$ & $217.72$ \\
        & $-710.69\log(5.63\gamma)+175.75$ & $0.8213$ & $389.56$ & $0.3936$ & $422.27$ \\
        \hline
        \multirow{5}{*}{BCE} & $26.68/\gamma+61.50$ & $0.9973$ & $47.11$ & $0.9930$ & $47.16$ \\
        & $-0.08/\gamma^2+35.22/\gamma-13.43$ & $\boldsymbol{0.9999}$ & $9.62$ & $\boldsymbol{0.9937}$ & $\boldsymbol{44.55}$ \\
        & $-0.001/\gamma^3+0.07/\gamma^2+3.17/\gamma+1.07$ & $0.9999$ & $\boldsymbol{8.02}$ & $0.9796$ & $80.37$ \\
        & $3849.46\exp(-42.68\gamma)+185.50$ & $0.9936$ & $72.78$ & $0.8842$ & $191.46$ \\
        & $-704.07\log(5.33\gamma)+147.95$ & $0.8326$ & $370.97$ & $0.4346$ & $423.15$ \\
        \bottomrule
    \end{tabular}
\end{table}

\begin{table}[!htp]
    \centering
    \caption{Fitted functions and their evaluation results for the epoch number used by the CE and BCE with $\gamma > 5$}
    \label{big_gamma_fit_results-supp}
    \begin{tabular}{c|c|cccc}
        \toprule
        \multirow{2}{*}{Loss} & \multirow{2}{*}{Fitted function} & \multicolumn{2}{c}{Fitting data} & \multicolumn{2}{c}{Predicted data} \\
        \cmidrule(lr){3-4} \cmidrule(lr){5-6}
        &  & $R^2 \uparrow$ & RMSE $\downarrow$& $R^2 \uparrow$ & RMSE $\downarrow$ \\
        \hline
        \multirow{5}{*}{CE} & $568.39\gamma-3645.68$ & $0.8556$ & $299.08$ & $-0.5945$ & $317.58$ \\
        & $236.01\gamma^2-3098.39\gamma+10205.49$ & $0.9939$ & $61.59$ & $0.7463$ & $126.69$ \\
        & $40.65\gamma^3-706.28\gamma^2+4076.80\gamma-7729.46$ & $\boldsymbol{0.9969}$ & $\boldsymbol{43.85}$ & $0.9038$ & $77.99$ \\
        & $0.54\exp(0.87\gamma)-61.83$ & $0.9932$ & $64.98$ & $\boldsymbol{0.9253}$ & $\boldsymbol{68.72}$ \\
        & $4.22\log(0.32\gamma)-3.01$ & $0.8094$ & $343.40$ & $-0.9326$ & $349.63$ \\
        \hline
        \multirow{5}{*}{BCE} & $236.60\gamma-2204.00$ & $0.7714$ & $364.33$ & $0.5566$ & $849.72$ \\
        & $52.14\gamma^2-1014.83\gamma+4887.43$ & $0.9812$ & $104.54$ & $0.8140$ & $550.31$ \\
        & $7.61\gamma^3-221.98\gamma^2+2171.11\gamma-7027.87$ & $0.9996$ & $15.55$ & $0.8837$ & $435.17$ \\
        & $0.49\exp(0.52\gamma)-3.79$ & $\boldsymbol{0.9999}$ & $\boldsymbol{7.56}$ & $\boldsymbol{0.9128}$ & $\boldsymbol{376.77}$ \\
        & $2.58\log(0.25\gamma)-2.11$ & $0.6871$ & $426.22$ & $0.4633$ & $934.77$ \\
        \bottomrule
    \end{tabular}
\end{table}

\subsection{Balanced Classification}
To further show the potential of BCE loss, we provide the classification performance on balanced dataset here.
Table \ref{balanced_classification} shows the classification results of the models trained on the balanced datasets, CIFAR10, CIFAR100, and Tiny-ImageNet dataset~\citep{wu2017tiny}. It can be observed that, the classification performance of models trained with BCE loss is comparable to or even surpassing that of models trained with CE loss.

When training ViT on the CIFAR10, CIFAR100, and Tiny-ImageNet datasets, its classification is significantly worse than that of the ResNets, which is intuitive and reasonable.
As noted by~\citet{zhu2023understanding} and~\citet{zhang2022bootstrapping}, due to the influence of factors such as model architecture and inductive biases, the ViT trained on small datasets is hugely different from ViT trained on large datasets, which may be the reason why the performance drops a lot on these small datasets. Setting aside the inherent differences between these models, we still find that training with the BCE loss can achieve performance on par with, or even surpassing, that of the CE loss. The BCE loss helps the model learn deep features with better compactness and separability, which is usually and intuitively beneficial for classification.

\begin{table}[t]
\centering
\caption{The classification accuracy (\%) on the test datasets of CIFAR10, CIFAR100 and Tiny-ImageNet. We conducted learning rate optimization for each experimental combination while keeping other hyperparameters fixed, and the $\gamma=32$}
    \begin{tabular}{c c cc cc cc}
        \toprule
        \multirow{2}{*}{$\mathcal{M}$} & \multirow{2}{*}{$\mathrm{Loss}$} & \multicolumn{2}{c}{$\mathrm{CIFAR10}$} & \multicolumn{2}{c}{$\mathrm{CIFAR100}$} & \multicolumn{2}{c}{Tiny-ImageNet} \\
        \cmidrule(lr){3-4} \cmidrule(lr){5-6} \cmidrule(lr){7-8}
        & & $\mathrm{SGD}$ & $\mathrm{AdamW}$ & $\mathrm{SGD}$ & $\mathrm{AdamW}$ & $\mathrm{SGD}$ & $\mathrm{AdamW}$ \\
        \midrule
        \multirow{2}{*}{ResNet18} & $\mathrm{CE}$ & $95.90$ & $\boldsymbol{96.24}$ & $78.39$ & $78.05$ & $68.10$ & $67.13$ \\
        & $\mathrm{BCE}$ & $\boldsymbol{95.97}$ & $95.96$ & $\boldsymbol{78.46}$ & $\boldsymbol{78.38}$ & $\boldsymbol{69.10}$ & $\boldsymbol{69.12}$ \\
        \midrule
        \multirow{2}{*}{ResNet50} & $\mathrm{CE}$ & $96.30$ & $\boldsymbol{96.51}$ & $\boldsymbol{80.89}$ & $78.06$ & $71.65$ & $69.49$ \\
        & $\mathrm{BCE}$ & $\boldsymbol{96.40}$ & $96.24$ & $80.87$ & $\boldsymbol{79.89}$ & $\boldsymbol{72.95}$ & $\boldsymbol{70.82}$ \\
        \midrule
        \multirow{2}{*}{ViT} & $\mathrm{CE}$ & $79.51$ & $87.00$ & $\boldsymbol{64.82}$ & $63.19$ & $50.42$ & $\boldsymbol{54.83}$ \\
        & $\mathrm{BCE}$ & $\boldsymbol{82.00}$ & $\boldsymbol{87.57}$ & $64.79$ & $\boldsymbol{64.91}$ & $\boldsymbol{51.15}$ & $53.91$ \\
        \bottomrule
    \end{tabular}
    \label{balanced_classification}
\end{table}

\subsection{Open-set Recognition Performance}
In Sec. \ref{Experiments}, we present the open-set recognition (OSR) performance of models trained by BCE and CE losses. To further analyze their differences and the impact of classifier biases in practice, we conducted additional OSR experiments with and without the classifier biases, respectively setting the mean of initial classifier biases as $0$, $5$, and $10$ when the classifier biases were included in the losses. Table \ref{tab_osr_reslts_more} presents the OSR results.

It can be observed that when the losses taking the classifier biases, regardless of their initial mean ($0,5$, or $10$), the BCE-trained models consistently outperforms the CE-trained ones in the OSR experiments. In contrast, without the biases, the OSR of BCE-trained models experiences a dramatic decline (over $8.92\%$ degradation), compared with the CE-trained models, especially on MNIST and SVHN, which demonstrate again the crucial role of classifier biases in the BCE loss.

\begin{table}[!htp]
	\caption{The results of open-set recognition. Results are averaged among five randomized trials and the scale factor $\gamma=32$}
	\label{tab_osr_reslts_more}
	\centering
		\begin{tabular}{cccccccccc}
			\toprule
			\multirow{2}{*}{Bias Mean} & \multirow{2}{*}{Metrics} & \multicolumn{2}{c}{MNIST} & \multicolumn{2}{c}{SVHN} & \multicolumn{2}{c}{CIFAR10} & \multicolumn{2}{c}{CIFAR+50} \\
			\cmidrule(lr){3-4} \cmidrule(lr){5-6} \cmidrule(lr){7-8} \cmidrule(lr){9-10}
			& & CE & BCE & CE & BCE & CE & BCE & CE & BCE \\
			\midrule
			\multirow{2}{*}{--} & AUROC & $\boldsymbol{96.74}$ & $90.22$ & $\boldsymbol{94.44}$ & $66.60$ & $\boldsymbol{74.34}$ & $68.14$ & $\boldsymbol{78.34}$ &$62.17$ \\
			& OSCR & $\boldsymbol{96.59}$ & $90.04$ & $\boldsymbol{92.56}$ & $71.00$& $60.10$ & $\boldsymbol{67.00}$ & $61.22$ & $\boldsymbol{63.87}$ \\
			\cdashline{1-10}[2pt/2pt]
			\multirow{2}{*}{0} & AUROC & $98.63$ & $\boldsymbol{99.29}$ & $94.42$ & $\boldsymbol{95.21}$ & $83.86$ & $\boldsymbol{85.70}$ & $87.82$ & $\boldsymbol{90.06}$ \\
			& OSCR & $98.50$ & $\boldsymbol{99.09}$ & $92.53$ & $\boldsymbol{93.49}$& $81.54$ & $\boldsymbol{83.50}$ & $85.81$ & $\boldsymbol{88.43}$ \\
			\cdashline{1-10}[2pt/2pt]
			\multirow{2}{*}{5} & AUROC & $98.58$ & $\boldsymbol{99.15}$ & $94.30$ & $\boldsymbol{95.07}$ & $84.17$ & $\boldsymbol{85.78}$ & $87.76$ & $\boldsymbol{90.15}$ \\
			& OSCR & $98.46$ & $\boldsymbol{98.96}$ & $92.42$ & $\boldsymbol{93.37}$& $81.87$ & $\boldsymbol{83.66}$ & $85.71$ & $\boldsymbol{88.51}$ \\
			\cdashline{1-10}[2pt/2pt]
			\multirow{2}{*}{10} & AUROC & $98.66$ & $\boldsymbol{99.21}$ & $94.49$ & $\boldsymbol{95.07}$ & $84.02$ & $\boldsymbol{85.51}$ & $87.55$ & $\boldsymbol{89.72}$ \\
			& OSCR & $98.53$ & $\boldsymbol{99.04}$ & $92.64$ & $\boldsymbol{93.37}$& $81.72$ & $\boldsymbol{83.35}$ & $85.62$ & $\boldsymbol{88.08}$ \\
			\bottomrule
		\end{tabular}
\end{table}


\section{Proof of Theorems}\label{proof}
In the normalized feature space,
Theorems \ref{them:CE collapse} and \ref{them:BCE collapse} have concluded the neural collapse of the CE and BCE losses on balanced datasets.
In this section, we present their detail proofs.

\subsection{Basics}

Before providing the proofs, we first formally present the definition of canonical $K$-simplex ETF in $d$-dimension space, and two lemmas required by the proofs.

\begin{definition}  \label{simplex_ETF_definition}
    (K-simplex ETF) A K-simplex ETF is a collection of vectors in $\mathbb{R}^d$ specified by the columns of matrix
    \begin{equation}
        \boldsymbol{M}=\sqrt{\frac{K}{K-1}}\boldsymbol{P}\left(\boldsymbol{I}_{K}-\frac{1}{K}\boldsymbol{1}_{K}\boldsymbol{1}_{K}^{\top}\right),
    \end{equation}
    where $\boldsymbol{I}_{K} \in \mathbb{R}^{K\times K}$ is the identity matrix and $\boldsymbol{1}_{K} \in \mathbb{R}^K$ is the ones vector, and $\boldsymbol{P} \in \mathbb{R}^{d\times K} (d\geq K)$ is a partial-orthogonal matrix such that $\boldsymbol{P}^{\top}\boldsymbol{P}=\boldsymbol{I}_K$.

    Note that the matrix $\boldsymbol{M}$ satisfies:
    \begin{equation}
        \boldsymbol{M}^{\top}\boldsymbol{M}=\frac{K}{K-1}\left(\boldsymbol{I}_{K}-\frac{1}{K}\mathbf{1}_{K}\mathbf{1}_{K}^{\top}\right).
    \end{equation}
\end{definition}

\begin{lemma}   \label{young_inequality_definition}
    (Young's Inequality) Let $p, q$ be positive numbers satisfying $\frac{1}{p}+\frac{1}{q}=1$. Then for any $a, b \in \mathbb{R}$, we have
    \begin{equation}
        |ab|\leq\frac{|a|^p}{p}+\frac{|b|^q}{q},
    \end{equation}
    where the equality holds if and only if $|a|^p=|b|^q$. The case for $p=q=2$ is just the AM-GM inequality which is $|ab|\leq\frac{1}{2}(a^2+b^2)$, where the equality holds if and only if $|a|=|b|$.
\end{lemma}

\begin{lemma}   \label{jensen_inequality_definition}
    (Jensen's Inequality) Let $f$ be a convex function on an interval $I$. Then
    \begin{equation}
        f(\lambda_1x_1+\cdots+\lambda_nx_n)\leq\lambda_1f(x_1)+\cdots+\lambda_nf(x_n),
    \end{equation}
    where
    \begin{equation}
        x_1,x_2,\cdots,x_n\in I,\quad \lambda_1,\lambda_2,\cdots,\lambda_n\in[0,1].
    \end{equation}
    When $f$ is strictly convex, let
    \begin{equation}
        I_1=\{k:\lambda_k\in(0,1]\}, ~\text{and}~ I_2=\{k:\lambda_k=0\}.
    \end{equation}
    The equality holds if and only if all $x_{k}, k \in I_1$ are equal. In practice, if the function $f$ is concave like $\exp(x)$, $\log(x)$, or $\log(1+\e^x)$, Jensen's Inequality can be written as
    \begin{align}
        \exp\left({\frac{\sum_{i=1}^{n}x_{i}}{n}}\right)&\leq{\frac{\sum_{i=1}^{n}\exp(x_{i})}{n}},\\
        \log\left({\frac{\sum_{i=1}^{n}x_{i}}{n}}\right)&\geq{\frac{\sum_{i=1}^{n}\log(x_{i})}{n}}, \\ \log\left(1+\exp\left(\frac{\sum_{i=1}^{n}x_{i}}{n}\right)\right)&\geq\frac{1}{n}\sum_{i=1}^{n}\log\big(1+\exp(x_{i})\big).
    \end{align}
\end{lemma}

\subsection{Proof of Theorem \ref{them:BCE collapse}}\label{proof:BCE collapse}
\begin{theorem}\label{them:BCE collapse_supp}
    Assume that the feature dimension $d$ is larger than the number of classes $K$, i.e., $d\geq K-1$, and $\gamma>0$. In the normalized feature space, any global minimizer $(\boldsymbol{W}^{\star}, \boldsymbol{H}^{\star}, \boldsymbol{b}^{\star})$ of
    \begin{equation} \label{equ-minf_object}
        f_{\mathrm{bce}}(\boldsymbol W,\boldsymbol H,\boldsymbol{b})\triangleq\frac{1}{nK}\sum_{k=1}^K\sum_{i=1}^n\mathcal{L}_{\mathrm{bce}}(\gamma \boldsymbol W\boldsymbol{h}_i^{(k)}-\boldsymbol{b})
    \end{equation}
    with $\|\boldsymbol w_j\|_2=\|\boldsymbol h_i^{(k)}\|_2=1$ for $\forall j,k\in[K], i\in[n]$,
    obeys 
    \begin{align}
        &\boldsymbol{w}_k^{\star} =\boldsymbol{h}_i^{(k)\star}, \quad \forall k\in[K], ~i\in[n], \label{equ-equality_w_h}\\
        &\tilde{\boldsymbol{h}}_{i}^{\star} :=\frac{1}{K}\sum_{j=1}^K\boldsymbol{h}_{i}^{(k)\star}=\boldsymbol{0},\quad\forall i\in[n],  \label{equ-equality_h_0}\\
        &\boldsymbol{W}^{\star}\boldsymbol{W}^{\star\top} =\frac K{K-1}\left(\boldsymbol{I}_K-\frac1K\boldsymbol{1}_K\boldsymbol{1}_K^\top\right), \label{equ-equality_w_w}\\
        &\boldsymbol{b}^{\star} =b^{\star}\mathbf{1},  \label{equ-equality_b}
    \end{align}
    where
    \begin{align}
            b^{\star} = \log\frac{(K-2)\e^{-\frac{\gamma}{K-1}} + \sqrt{(K-2)^{2}\e^{{-\frac{2\gamma}{K-1}}}+4(K-1)\e^{-\frac{\gamma}{K-1} + \gamma}}}{2}.\label{equ-b_equation_1}
    \end{align}
\end{theorem}

\begin{proof} For the function $f_\mathrm{bce}(\boldsymbol{W},\boldsymbol{H},\boldsymbol{b})$, according to Lemma \ref{lem:lemmaA.2},
    \begin{equation} \label{equ-fbce_lowerBound_1}
        f_{\mathrm{bce}}(\boldsymbol{W},\boldsymbol{H},\boldsymbol{b})
        \geq -2\gamma - (K-2)b^{\star} + \log\big(1+\e^{\gamma-b^{\star}}\big) + (K-1)\log\Big(1+\e^{b^{\star}+\frac{\gamma}{K-1}}\Big).
    \end{equation}

    According to Lemma \ref{lem:lemmaA.3}, the inequality (\ref{equ-fbce_lowerBound_1}) achieves its equality when Eqs. (\ref{equ-equality_w_h}), (\ref{equ-equality_h_0}), (\ref{equ-equality_w_w}), (\ref{equ-equality_b}), and (\ref{equ-b_equation_1}) hold.
    For any fixed $\gamma>0$ and $K$, the right side of the inequality (\ref{equ-fbce_lowerBound_1}) is a constant. Therefore, when $f_\mathrm{bce}(\boldsymbol{W},\boldsymbol{H},\boldsymbol{b})$ reaching this constant, it gets its minimum.
\end{proof}

\begin{lemma} \label{lem:lemmaA.1}
    For any $\bm h_i^{(k)}$ with $c_1, c_2 > 0$ and $\gamma > 0$, the normalized BCE loss is lower bounded by
    \begin{equation} \label{equ-Lbce_lowerBound}
        \mathcal{L}_{\mathrm{bce}}(\gamma \boldsymbol{W}\boldsymbol{h}_{i}^{(k)}-\boldsymbol{b}) \geq \frac{1}{1+c_{1}}\Big(-\gamma \boldsymbol{w}_{k}^{\top}\boldsymbol{h}_{i}^{(k)}+b_{k}\Big)+\frac{1}{1+c_{2}}\sum_{j=1 \atop j\neq k}^{K}\Big(\gamma \boldsymbol{w}_{j}^{\top}\boldsymbol{h}_{i}^{(k)}-b_{j}\Big)+C,
    \end{equation}
    where
    \begin{equation} \label{equ-C_formula_2}
        C=\log(1+c_1)-\frac{c_1}{1+c_1}\log(c_1)+(K-1)\bigg[\log(1+c_2)-\frac{c_2}{1+c_2}\log(c_2)\bigg].
    \end{equation}
    The inequality becomes an equality if and only if
    \begin{align}
        &c_1=\exp\Big(\gamma \boldsymbol{w}_k^\top\boldsymbol{h}_i^{(k)}-b_k\Big), \quad \forall k\in [K], \label{equ-c1_1}\\
        & c_2=\exp\Big(b_j-\gamma \boldsymbol{w}_j^\top\boldsymbol{h}_i^{(k)}\Big),\quad \forall j\neq k \in [K]. \label{equ-c2_1}
    \end{align}
\end{lemma}

\begin{proof}
    By the concavities of $\log(x)$, and according to Lemma \ref{jensen_inequality_definition}, the normalized BCE loss can be lower bounded by
    \begin{align}
        ~&\mathcal{L}_{\mathrm{bce}}(\gamma \boldsymbol{W}\boldsymbol{h}_{i}^{(k)}-\boldsymbol{b}) \nonumber \\
        =~ &\log\Big(1+\exp\big(-\gamma \boldsymbol{w}_k^\top\boldsymbol{h}_i^{(k)}+b_k\big)\Big)+\sum_{j=1 \atop j\neq k}^K\log\Big(1+\exp\big(\gamma \boldsymbol{w}_j^\top\boldsymbol{h}_i^{(k)}-b_j\big)\Big) \\
        =~ &\log\bigg(\frac{c_1}{1+c_1}\frac{1+c_1}{c_1}+\frac{1+c_1}{1+c_1}\exp\big(-\gamma \boldsymbol{w}_k^\top\boldsymbol{h}_i^{(k)}+b_k\big)\bigg)\nonumber\\
          &~+\sum_{j=1 \atop j\neq k}^K\log\bigg(\frac{c_2}{1+c_2}\frac{1+c_2}{c_2}+\frac{1+c_2}{1+c_2}\exp\big(\gamma \boldsymbol{w}_j^\top\boldsymbol{h}_i^{(k)}-b_j\big)\bigg) \\
        \geq~ & \frac{c_1}{1+c_1}\log\bigg(\frac{1+c_1}{c_1}\bigg) + \frac{1}{1+c_1}\log\Big[(1+c_1)\exp\big(-\gamma \boldsymbol{w}_k^\top\boldsymbol{h}_i^{(k)}+b_k\big)\Big]\nonumber\\
            &~+\frac{(K-1)c_2}{1+c_2}\log\bigg(\frac{1+c_2}{c_2}\bigg) + \frac{1}{1+c_2}\sum_{j=1 \atop j\neq k}^K\log\Big[(1+c_2)\exp\big(\gamma \boldsymbol{w}_j^\top\boldsymbol{h}_i^{(k)}-b_j\big)\Big]\\
        =~ &\frac{1}{1+c_1}\bigg(-\gamma \boldsymbol{w}_k^\top\boldsymbol{h}_i^{(k)}+b_k\bigg)+\frac1{1+c_2}\sum_{j=1 \atop j \neq k}^K\bigg(\gamma \boldsymbol{w}_j^\top\boldsymbol{h}_i^{(k)}-b_j\bigg) \nonumber \\
        &~+\underbrace{\log(1+c_1)-\frac{c_1}{1+c_1}\log(c_1)+(K-1)\bigg[\log(1+c_2)-\frac{c_2}{1+c_2}\log(c_2)\bigg]}_{C}. \label{use_C}
    \end{align}
    The inequality is derived from the concavity of $\log(x)$, which achieves the equality if and only if
    \begin{align}
        \frac{1+c_1}{c_1} &=\left(1+c_1\right)\exp\left(-\gamma \boldsymbol{w}_k^\top\boldsymbol{h}_i^{(k)}+b_k\right) ~\text{or}~ c_1=0 ~\text{or}~ c_1=+\infty, ~\text{and} \\
        \frac{1+c_2}{c_2} &=\left(1+c_2\right)\exp\Big(\gamma \boldsymbol{w}_j^\top\boldsymbol{h}_i^{(k)}-b_j\Big) ~\text{or}\, c_2=0 ~\text{or}~ c_2=+\infty,\quad \forall j\neq k \in[K].
    \end{align}
    However, these equalities are trivial when \(c_1 = 0\) or \(c_1 = +\infty\) and \(c_2 = 0\) or \(c_2 = +\infty\). 
    Therefore, if and only if
    \begin{align}
        c_1 &=\exp\left(\gamma \boldsymbol{w}_k^\top\boldsymbol{h}_i^{(k)}-b_k\right), \\
        c_2 &=\exp\left(b_j-\gamma \boldsymbol{w}_j^\top\boldsymbol{h}_i^{(k)}\right), \quad\forall j\neq k\in [K],
    \end{align}
    Eq. (\ref{equ-Lbce_lowerBound}) becomes an equality.
\end{proof}

\begin{lemma} \label{lem:lemmaA.2}
    For the function $f_{\mathrm{bce}}(\boldsymbol W,\boldsymbol H,\boldsymbol{b})$ defined in \textit{Eq.} (\ref{equ-minf_object}),
    \begin{equation} \label{equ-fbce_lowerBound_2}
        f_{\mathrm{bce}}(\boldsymbol{W},\boldsymbol{H},\boldsymbol{b})\geq -2\gamma - (K-2)b^{\star} + \log\big(1+\e^{\gamma-b^{\star}}\big) + (K-1)\log\Big(1+\e^{b^{\star}+\frac{\gamma}{K-1}}\Big).
    \end{equation}
\end{lemma}

\begin{proof}
    According to Lemma \ref{lem:lemmaA.1}, \textit{Eq.} (\ref{equ-Lbce_lowerBound}) holds for any $c_1, c_2 > 0$ and any $\boldsymbol{h}_i^{(k)}$ with $k\in[K], ~i\in[n]$. We take the same $c_1$ and $c_2$ for all $\boldsymbol{h}_i^{(k)}$ with $\forall i\in[n], \forall k\in[K]$, then
    \allowdisplaybreaks
    \begin{align}
        ~&f_{\mathrm{bce}}(\boldsymbol{W},\boldsymbol{H}, \bm b) = \frac{1}{nK}\sum_{k=1}^K\sum_{i=1}^n\mathcal{L}_\mathrm{bce}\big(\gamma \boldsymbol{W}\boldsymbol{h}_i^{(k)}-\boldsymbol{b}\big) \\
        \geq~&\frac{1}{nK}\sum_{k=1}^K\sum_{i=1}^n\bigg[\frac{1}{1+c_1}\Big(-\gamma \boldsymbol{w}_k^\top\boldsymbol{h}_i^{(k)}+b_k\Big)+\frac{1}{1+c_2}\sum_{j=1\atop j\neq k}^K\Big(\gamma \boldsymbol{w}_j^\top\boldsymbol{h}_i^{(k)}-b_j\Big)\bigg] + C\\
        =~&\frac{1}{nK}\frac{1}{1+c_2}\sum_{k=1}^K\sum_{i=1}^n\sum_{j=1 \atop j\neq k}^K\Big(\gamma \boldsymbol{w}_j^\top\boldsymbol{h}_i^{(k)}-b_j\Big)-\frac1{nK}\frac1{1+c_1}\sum_{k=1}^K\sum_{i=1}^n\Big(\gamma \boldsymbol{w}_k^\top\boldsymbol{h}_i^{(k)}-b_k\Big)  + C \\
        =~&\frac1{nK}\frac1{1+c_2}\sum_{k=1}^K\sum_{i=1}^n\sum_{j=1}^K\left(\gamma \boldsymbol{w}_j^\top\boldsymbol{h}_i^{(k)}-b_j-\gamma \boldsymbol{w}_k^\top\boldsymbol{h}_i^{(k)}+b_k\right) + C \nonumber \\
        &~+\frac1{nK}\frac1{1+c_2}\sum_{k=1}^K\sum_{i=1}^n\sum_{j=1 \atop j\neq k}^K\left(\gamma \boldsymbol{w}_k^\top\boldsymbol{h}_i^{(k)}-b_k\right)-\frac1{nK}\frac1{1+c_1}\sum_{k=1}^K\sum_{i=1}^n\left(\gamma \boldsymbol{w}_k^\top\boldsymbol{h}_i^{(k)}-b_k\right)\\
        =~&\frac1{nK}\frac1{1+c_2}\sum_{i=1}^n\bigg[\sum_{k=1}^K\Big(\sum_{j=1}^K\gamma \boldsymbol{w}_k^\top\boldsymbol{h}_i^{(j)}-{K}\gamma \boldsymbol{w}_k^\top\boldsymbol{h}_i^{(k)}\Big)\underbrace{-\sum_{k=1}^K\sum_{j=1}^{K}b_j+\sum_{k=1}^K\sum_{j=1}^{K}b_k}_{0}\bigg] \nonumber \\
        &~+\frac1{nK}\bigg(\frac{K-1}{1+c_2}-\frac1{1+c_1}\bigg)\sum_{k=1}^K\sum_{i=1}^n\gamma \boldsymbol{w}_k^\top\boldsymbol{h}_i^{(k)}-\frac1{K}\bigg(\frac{K-1}{1+c_2}-\frac1{1+c_1}\bigg)\sum_{k=1}^{K}b_k  + C\\
        =~&\frac1n\frac1{1+c_2}\sum_{i=1}^n\sum_{k=1}^K\gamma \boldsymbol{w}_k^\top\left(\tilde{\boldsymbol{h}}_i-\boldsymbol{h}_i^{(k)}\right)-\frac1{K}\bigg(\frac{K-1}{1+c_2}-\frac1{1+c_1}\bigg)\sum_{k=1}^{K}b_k \nonumber \\
        &~+\frac1{nK}\bigg(\frac{K-1}{1+c_2}-\frac1{1+c_1}\bigg)\sum_{k=1}^K\sum_{i=1}^n\gamma \boldsymbol{w}_k^\top\boldsymbol{h}_i^{(k)}  + C {\color{black}~\triangleq~\hat{f}_{\text{bce}}(\boldsymbol{W},\boldsymbol{H}, \bm b),
        }
    \end{align}
    {\color{black}
    where in the last equality, $\tilde{\boldsymbol{h}}_{i}=\frac{1}{K}\sum_{j=1}^{K}\boldsymbol{h}_{i}^{(j)}$.

    For the function $\hat{f}_{\text{bce}}(\boldsymbol{W},\boldsymbol{H}, \bm b)$, its global minimum must occur at its stationary points that satisfies the first-order necessary condition,
    \begin{equation}
        0 = \frac{\partial \hat{f}_{\mathrm{bce}}}{\partial b_k} = -\frac1{K}\bigg(\frac{K-1}{1+c_2}-\frac1{1+c_1}\bigg),\quad k\in[K]. \label{b_condition}
    \end{equation}
    Therefore,
    \begin{align}
        \hat{f}_{\mathrm{bce}}(\boldsymbol{W},\boldsymbol{H}, \bm b) \geq \frac{\gamma}{n}\frac1{1+c_2}\sum_{i=1}^n\sum_{k=1}^K \boldsymbol{w}_k^\top\Big(\tilde{\boldsymbol{h}}_i-\boldsymbol{h}_i^{(k)}\Big) + C.
    \end{align}
    Furthermore, according to the AM-GM inequality (see Lemma \ref{young_inequality_definition}), we have
    \begin{equation}
        \boldsymbol{u}^\top\boldsymbol{v}\geq - \Big(\frac{c}{2}\|\boldsymbol{u}\|_2^2+\frac{1}{2c}\|\boldsymbol{v}\|_2^2\Big), \quad\forall\boldsymbol{u}, \boldsymbol{v}\in\mathbb{R}^d, ~\forall c > 0,
    \end{equation}
    which becomes an equality when $\boldsymbol{v} = -c\boldsymbol{u}$. Then,
    \begin{align}
        &~f_{\mathrm{bce}}(\boldsymbol{W},\boldsymbol{H}, \bm b) \geq \hat{f}_{\mathrm{bce}}(\boldsymbol{W},\boldsymbol{H}, \bm b) \geq \frac{\gamma}{n}\frac1{1+c_2}\sum_{i=1}^n\sum_{k=1}^K \boldsymbol{w}_k^\top\Big(\tilde{\boldsymbol{h}}_i-\boldsymbol{h}_i^{(k)}\Big) + C \nonumber \\
        \geq& -\frac{\gamma}n\frac1{1+c_2}\bigg(\frac{c_3}2\sum_{i=1}^n\sum_{k=1}^K\|\boldsymbol{w}_k\|_2^2+\frac1{2c_3}\sum_{i=1}^n\sum_{k=1}^K\|\tilde{\boldsymbol{h}}_i-\boldsymbol{h}_i^{(k)}\|_2^2\bigg) + C \label{c3_ineq} \\
        =& -\frac{\gamma}n\frac1{1+c_2}\bigg[\frac{c_3}2\sum_{i=1}^n\sum_{k=1}^K\|\boldsymbol{w}_k\|_2^2+\frac1{2c_3}\sum_{i=1}^n\bigg(\sum_{k=1}^K\|\boldsymbol{h}_i^{(k)}\|_2^2-K\|\tilde{\boldsymbol{h}}_i\|_2^2\bigg)\bigg] + C \\
        =& -\frac{\gamma}{n}\frac1{1+c_2}\bigg(\frac{c_3}{2} nK +\frac{1}{2c_3}nK\bigg) +\frac{\gamma}{n}\frac1{1+c_2}\frac{1}{2c_3}\sum_{i=1}^nK\|\tilde{\boldsymbol{h}}_i\|_2^2 + C\\
        \geq& -\frac{\gamma}{n}\frac1{1+c_2}\bigg(\frac{c_3}{2} nK +\frac{1}{2c_3}nK\bigg) + C. \label{h_mean=0}
    \end{align}
    Inequality (\ref{c3_ineq}) becomes an equality if and only if
    \begin{equation}
        c_{3} \boldsymbol{w}_{k} =\boldsymbol{h}_{i}^{(k)}-\tilde{\boldsymbol{h}}_{i}, ~\forall k\in[K], \quad i\in[n], ~\gamma>0,~\forall c_3 > 0, \label{equ-c3_hold}
    \end{equation}
    and, the inequality (\ref{h_mean=0}) becomes an equality if and only if
    \begin{equation}
        \tilde{\boldsymbol{h}}_{i} = \bm 0, \quad \forall i\in[n]. \label{equ-hmean=0}
    \end{equation}
    Eqs. (\ref{equ-c3_hold}) and (\ref{equ-hmean=0}) imply
    \begin{equation}
        c_3 \boldsymbol{w}_k=\boldsymbol{h}_i^{(k)}\Rightarrow\|c_3 \boldsymbol{w}_k\|_2^2=\|\boldsymbol{h}_i^{(k)}\|_2^2\Rightarrow c_3^2=1\Rightarrow c_3=1. \label{equ-c3_solution}
    \end{equation}
    Combining Eqs. (\ref{h_mean=0}) and (\ref{equ-c3_solution}), we get
    \begin{align}
        f_{\mathrm{bce}}(\boldsymbol{W}, \boldsymbol{H}, \bm b) \geq& -\frac{\gamma}{n}\frac1{1+c_2}\bigg(\frac{c_3}{2} nK +\frac{1}{2c_3}nK\bigg) + C = -\frac{\gamma K}{1+c_2} + C, \label{f_bce_low}
    \end{align}
    which becomes an equality if and only if \textit{Eqs.} (\ref{equ-c1_1}), (\ref{equ-c2_1}), (\ref{equ-c3_hold}) and (\ref{equ-hmean=0}) hold.

    Furthermore, according to Lemma \ref{lem:lemmaA.3}, when $f_{\mathrm{bce}}(\boldsymbol{W}, \boldsymbol{H}, \bm b)$ achieves its lower bound, one can get $b_1=b_2=\cdots=b_k=b^{\star}$, and
    \begin{equation}
        c_1=\exp\left(\gamma-b^\star\right) \quad \text{and} \quad c_2=\exp\left(b^\star+\frac{\gamma}{K-1}\right). \label{c1_c2_solution}
    \end{equation}
    Then, combining Eqs. (\ref{use_C}), (\ref{b_condition}), (\ref{f_bce_low}) and (\ref{c1_c2_solution}), we have
    \begin{align}
        & f_{\mathrm{bce}}(\boldsymbol W,\boldsymbol H,\boldsymbol{b})
        \geq -\frac{\gamma K}{1+c_2} + C \nonumber \\
        =~& -\frac{\gamma K}{1+c_2} + \log\left(1+c_1\right)-\frac{c_1}{1+c_1}\log(c_1)+(K-1)\bigg[\log(1+c_2)-\frac{c_2}{1+c_2}\log(c_2)\bigg]\\
        =~& -2\gamma - (K-2)b^{\star} + \log\big(1+\e^{\gamma-b^{\star}}\big) + (K-1)\log\Big(1+\e^{b^{\star}+\frac{\gamma}{K-1}}\Big),
    \end{align}
    which is desired.}
\end{proof}

\begin{lemma} \label{lem:lemmaA.3}
    Under the same assumption of Lemma \ref{lem:lemmaA.2}, the lower bound in \textit{Eq.} (\ref{equ-fbce_lowerBound_2}) is achieved if and only if
    \begin{align}
        &\boldsymbol{w}_k=\boldsymbol{h}_i^{(k)}, \quad\forall k\in[K], ~i\in[n], \\
        &\tilde{\boldsymbol{h}}_{i}:=\frac{1}{K}\sum\limits_{k=1}^K\boldsymbol{h}_{i}^{(k)}=\boldsymbol{0}, \quad\forall i\in[n],  \\
        &\boldsymbol{W}\boldsymbol{W}^{\top}=\frac K{K-1}\left(\boldsymbol{I}_K-\frac{1}{K}\boldsymbol{1}_K\boldsymbol{1}_K^\top\right), \\
        &\boldsymbol{b}^{\star}=b^{\star}\mathbf{1},  \\
        &c_1=\exp\left(\gamma-b^\star\right)\quad\text{and} \quad c_2=\exp\left(b^\star+\frac{\gamma}{K-1}\right),
    \end{align}
    where $b^{\star}$ satisfies
    \begin{equation}
        b^{\star} = \log\frac{(K-2)\e^{-\frac{\gamma}{K-1}} + \sqrt{(K-2)^{2}\e^{{-\frac{2\gamma}{K-1}}}+4(K-1)\e^{-\frac{\gamma}{K-1} + \gamma}}}{2}.
    \end{equation}
\end{lemma}

\begin{proof}
    According to the proof of Lemma \ref{lem:lemmaA.2}, when the loss function $f_\text{bce}$ achieves its lower bound if and only if \textit{Eqs.} (\ref{equ-c1_1}), (\ref{equ-c2_1}), (\ref{equ-c3_hold}) and (\ref{equ-hmean=0}) hold.
    From Eqs. (\ref{equ-c3_hold}) and (\ref{equ-hmean=0}),
    \begin{align}
         &\boldsymbol{w}_{k}=\boldsymbol{h}_{i}^{(k)}, \quad\forall k\in[K],~i\in[n],\label{eq_feature_zero}\\
       \text{and}\qquad &\sum_{k=1}^{K} \boldsymbol{w}_{k}=\sum_{k=1}^{K}\boldsymbol{h}_{i}^{(k)}=\boldsymbol{0}.\label{equ-sumw=sumh}
    \end{align}
    Then, we have
    \begin{align}
        c_1 &=\exp\left(\gamma \boldsymbol{w}_k^\top\boldsymbol{h}_i^{(k)}-b_k\right)=\exp\left(\gamma\| \boldsymbol{w}_k\|_2^2-b_k\right), \quad\forall k \in [K], \label{equ-c1_formula_simple} \\
        c_2 &=\exp\left(b_j-\gamma \boldsymbol{w}_j^\top\boldsymbol{h}_i^{(k)}\right)=\exp\left(b_j-\gamma \boldsymbol{w}_k^\top\boldsymbol{w}_j\right), \quad \forall j\neq k \in [K]. \label{equ-c2_formula_simple}
    \end{align}
    Since the $c_1, c_2 > 0$ are the same for different $j \neq k \in [K]$, we have
    \begin{align}
        &\gamma\|\boldsymbol{w}_k\|_2^2-b_k=\gamma\|\boldsymbol{w}_j\|_2^2-b_j, \quad \forall k,j\in[K], \label{equ-c1_same_for_all} \\
        &\gamma \boldsymbol{w}_k^\top\boldsymbol{w}_j-b_j=\gamma \boldsymbol{w}_k^\top\boldsymbol{w}_\ell-b_\ell, \quad \forall j\neq\ell\in[K], ~\forall k\in[K]. \label{equ-c2_same_for_all}
    \end{align}
    In the normalized feature space, \textit{Eq.} (\ref{equ-c1_same_for_all}) implies
    \begin{align}
        \gamma\|\boldsymbol{w}_k\|_2^2-b_k &=\gamma\|\boldsymbol{w}_j\|_2^2-b_j \Rightarrow \gamma-b_k=\gamma-b_j \Leftrightarrow b_k=b_j, \quad \forall k, j \in [K], \label{equ-bk=bj}
    \end{align}
    Therefore, one can write $\boldsymbol{b}=b^\star\mathbf{1}$ for some $b \in \mathbb R$. 
    Then, from Eq. (\ref{equ-c2_same_for_all}), 
    \begin{align}
         \gamma \boldsymbol{w}_k^\top\boldsymbol{w}_j=\gamma \boldsymbol{w}_k^\top\boldsymbol{w}_\ell \Rightarrow \boldsymbol{w}_k^\top\boldsymbol{w}_j=\boldsymbol{w}_k^\top\boldsymbol{w}_\ell, \quad\forall j\neq\ell\in[K], \quad \forall k\in[K],
    \end{align}
    and combining with \textit{Eq.} (\ref{equ-sumw=sumh}), 
    \begin{align}
        1 = \|\boldsymbol{w}_{k}\|_{2}^{2}&=-\sum_{j=1 \atop j \neq k}^{K}\boldsymbol{w}_{k}^\top\boldsymbol{w}_{j}=-\left(K-1\right)\boldsymbol{w}_{k}^\top\boldsymbol{w}_{j} \notag \\
        &\Rightarrow \boldsymbol{w}_{k}^\top\boldsymbol{w}_{j}=-\frac{1}{K-1}, \quad \forall j\neq k\in[K]. \label{equ-wkwj}
    \end{align}
    Therefore, 
    \begin{equation}
        \boldsymbol{W}\boldsymbol{W}^\top=\frac K{K-1}\left(\boldsymbol{I}_K-\frac1K\boldsymbol{1}_K\boldsymbol{1}_K^\top\right).
    \end{equation}
    Plugging \textit{Eqs.} (\ref{equ-bk=bj}), (\ref{equ-wkwj}) into \textit{Eqs.} (\ref{equ-c1_formula_simple}), (\ref{equ-c2_formula_simple}),
    \begin{align}
        c_1 &=\exp(\gamma-b^\star)\quad\text{and}\quad c_2 =\exp\bigg(b^\star+\frac{\gamma}{K-1}\bigg).
    \end{align}

    When $f_\mathrm{bce}$ defined in \textit{Eq.} (\ref{equ-minf_object}) achieves its minimum value, it theoretically satisfies
    \begin{equation}
        0 = \frac{\partial f_{\mathrm{bce}}}{\partial b_{k}}=\frac{1}{nK}\bigg(n-\sum_{j=1}^{K}\sum_{i=1}^{n}\left(1+\exp\left(-\gamma \boldsymbol{w}_{k}^\top\boldsymbol{h}_{i}^{(j)}+b_{k}\right)\right)^{-1}\bigg), \quad\forall k\in[K],
    \end{equation}
    thus, combining \textit{Eqs.} (\ref{equ-bk=bj}) and (\ref{equ-wkwj}), 
    \begin{align}
        &~n-\sum_{j=1}^{K}\sum_{i=1}^{n}\left(1+\exp\left(-\gamma \boldsymbol{w}_{k}^\top\boldsymbol{h}_{i}^{(j)}+b_{k}\right)\right)^{-1} \nonumber \\
        =&~n-\sum_{i=1}^{n}\Bigg(\frac{1}{1+\exp(-\gamma+b)} + \sum_{j\neq k}^{K}\frac{1}{1+\exp(\frac{\gamma}{K-1}+b)}\Bigg) \\
        =&~n-n\Bigg(\frac{1}{1+\exp(-\gamma+b)} + \frac{K-1}{1+\exp(\frac{\gamma}{K-1}+b)}\Bigg)=0,
    \end{align}
    then, 
    \begin{equation}
        b^{\star} = \log\frac{(K-2)\e^{-\frac{\gamma}{K-1}} + \sqrt{(K-2)^{2}\e^{{-\frac{2\gamma}{K-1}}}+4(K-1)\e^{-\frac{\gamma}{K-1} + \gamma}}}{2},
    \end{equation}
    which satisfies Eq. (\ref{b_condition}).
\end{proof}

\begin{lemma} \label{lem:lemmaA.4}
    When the class number $K > 2$, the scale factor $\gamma > 0$, and
    \begin{equation}
        \log\left(2K-3\right)(1-\frac{1}{K})<\gamma,
    \end{equation}
    the final critical bias $\boldsymbol{b}^{\star}$ could separate the all positive unbiased decision scores
    \begin{equation}
        \left\{\gamma \boldsymbol{w}_k^{\star \top}\boldsymbol{h}_i^{(k)\star}: ~k\in[K], ~i\in[n]\right\},
    \end{equation}
    and the all negative ones
    \begin{equation}
        \left\{\gamma \boldsymbol{w}_j^{\star \top}\boldsymbol{h}_i^{(k)\star}: ~k,j\in[K], ~i\in[n], ~k\neq j\right\}.
    \end{equation}
\end{lemma}

\begin{proof}
    According to Lemma \ref{lem:lemmaA.3}, when the function $f_{\mathrm{bce}}$ achieves the lower bound, we have
    \begin{align}
        \gamma \boldsymbol{w}_{k}^{\star \top}\boldsymbol{h}_{i}^{(k) \star} &=\gamma, \quad \forall k\in[K], ~i \in [n], \\
        \gamma \boldsymbol{w}_{j}^{\star \top}\boldsymbol{h}_{i}^{(k) \star} &=-\frac{\gamma}{K-1}, \quad \forall j\neq k\in[K], ~i \in[n],
    \end{align}
    Let $b_\mathrm{pos}=\gamma, b_\mathrm{neg}=-\frac{\gamma}{K-1}$. Then the critical $b^{\star}$ separating the all positive and negative decision scores means
    \begin{equation}
        b_{\mathrm{neg}}=-\frac{\gamma}{K-1}<b^{\star}<\gamma=b_{\mathrm{pos}}.
    \end{equation}
    Due to {
    \allowdisplaybreaks
    \begin{align}
        -\frac{\gamma}{K-1}<b^{\star} &\Leftrightarrow 2\mathrm{~e}^{-\frac{\gamma}{K-1}}<\left(K-2\right)\mathrm{e}^{-\frac\gamma{K-1}}+\sqrt{\left(K-2\right)^2\mathrm{e}^{-\frac{2\gamma}{K-1}}+4\left(K-1\right)\mathrm{e}^{-\frac\gamma{K-1}+\gamma}} \nonumber\\
        &\Leftarrow 2\mathrm{~e}^{-\frac{\gamma}{K-1}}<\left(K-2\right)\mathrm{e}^{-\frac\gamma{K-1}}+\sqrt{4\left(K-1\right)\mathrm{e}^{(-\frac\gamma{K-1}+\gamma)}} \nonumber\\
        &\Leftrightarrow 2<\left(K-2\right)+2\sqrt{K-1}\mathrm{e}^{(\frac\gamma{2(K-1)}+\frac{\gamma}{2})} \quad \Leftarrow\quad 2<\left(K-2\right)+2\sqrt{K-1} \nonumber\\
        &\Leftarrow 2<K, \\
        b^{\star}<\gamma &\Leftrightarrow \left(K-2\right)\mathrm{e}^{-\frac\gamma{K-1}}+\sqrt{\left(K-2\right)^2\mathrm{e}^{-\frac{2\gamma}{K-1}}+4\left(K-1\right)\mathrm{e}^{-\frac\gamma{K-1}+\gamma}}<2\mathrm{e}^\gamma \nonumber\\
        &\Leftrightarrow \left(K-1\right)\mathrm{e}^{-\frac\gamma{K-1}}<\mathrm{e}^{\gamma}-\left(K-2\right)\mathrm{e}^{-\frac{\gamma}{K-1}} \nonumber\\
        &\Leftrightarrow \log\left(2K-3\right)(1-\frac{1}{K})<\gamma \quad \Leftarrow \quad \log\left(2K-3\right)<\gamma \nonumber\\
        &\Leftrightarrow K<\frac{\mathrm{e}^{\gamma}+3}{2},
    \end{align}}
    which completes the proof.
\end{proof}

\subsection{Proof of Theorem \ref{them:CE collapse}}\label{proof:CE collapse}
\begin{theorem} \label{them:lemmaB.1}
    Assume that the feature dimension $d$ is larger than the number of classes $K$, i.e., $d\geq K-1$, and the scale factor $\gamma>0$. In the normalized feature space, any global minimizer $(\boldsymbol{W}^{\star},\boldsymbol{H}^{\star},\boldsymbol{b}^{\star})$ of
    \begin{equation} \label{equ-min_fce_formula}
        \min_{\boldsymbol{W},\boldsymbol{H},\boldsymbol{b}}f_{\mathrm{ce}}(\boldsymbol W,\boldsymbol H,\boldsymbol{b}):=\frac{1}{nK}\sum_{k=1}^K\sum_{i=1}^n\mathcal{L}_{\mathrm{ce}}(\gamma \boldsymbol W\boldsymbol{h}_i^{(k)}-\boldsymbol{b})
    \end{equation}
    with $\|\boldsymbol{w}_j\|=\|\boldsymbol{h}_i^{(k)}\|=1$,
    obeys the following
    \begin{align}
        &\boldsymbol{w}_k^{\star}=\boldsymbol{h}_i^{(k)\star}, \quad \forall k\in[K], i\in[n], \label{eq_ce_w_h}\\ &\tilde{\boldsymbol{h}}_{i}^{\star}:=\frac{1}{K}\sum\limits_{j=1}^K\boldsymbol{h}_{i}^{(k)\star}=\boldsymbol{0}, \quad \forall i\in[n],\label{eq_ce_h_center}\\
        &\boldsymbol{W}^{\star}\boldsymbol{W}^{\star\top}=\frac K{K-1}\left(\boldsymbol{I}_K-\frac1K\boldsymbol{1}_K\boldsymbol{1}_K^\top\right),\label{eq_ce_w_etf}\\
        &\boldsymbol{b}^{\star}=b\mathbf{1}, \quad \forall b\in\mathbb R. \label{eq_ce_b_identity}
    \end{align}
\end{theorem}

\begin{proof}
    According to Lemma \ref{lem:lemmaB.2}, for any fixed $\gamma > 0$, we have
    \begin{equation} \label{equ-fce_lowerBound_1}
        f_{\mathrm{ce}}(\boldsymbol{W}, \boldsymbol{H}, \boldsymbol{b})\geq\log\Big(1+(K-1)\e^{-\frac{\gamma K}{K-1}}\Big).
    \end{equation}
    According to Lemma \ref{lem:lemmaB.3}, the inequality (\ref{equ-fce_lowerBound_1}) achieves its equality when \textit{Eqs.} (\ref{eq_ce_w_h}, \ref{eq_ce_h_center}, \ref{eq_ce_w_etf}, \ref{eq_ce_b_identity}) hold, which finishes the proof.
\end{proof}

\begin{lemma} \label{lem:lemmaB.1}
    For any $\bm h_i^{(k)}$ with $c'_1 > 0$ and $\gamma > 0$, the normalized CE loss is lower bounded by
    \begin{equation} \label{equ-Lce_lowerBound}
        \mathcal{L}_{\mathrm{ce}}(\gamma \boldsymbol{W}\boldsymbol{h}_{i}^{(k)}-\boldsymbol{b}) \geq \frac{\sum_{j=1}^K(\gamma\boldsymbol{w}_j^{\top}\boldsymbol{h}_i^{(k)}-b_j)-K(\gamma\boldsymbol{w}_k^{\top}\boldsymbol{h}_i^{(k)}-b_k)}{\left(1+c'_1\right)\left(K-1\right)}+C',
    \end{equation}
    where
    \begin{equation} \label{equ-fce_c2_formula_2}
        C' =\frac{c'_1}{1+c'_1}\log\left(\frac{1+c'_1}{c'_1}\right)+\frac{1}{1+c'_1}\log\left[\left(1+c'_1\right)\left(K-1\right)\right].
    \end{equation}
    The inequality becomes an equality when
    \begin{equation} \label{equ-jensen_hold_condition_a}
        c'_1=\left[\left(K-1\right)\exp\left(\frac{\sum_{j=1}^K(\gamma\boldsymbol{w}_j^{\top}\boldsymbol{h}_i^{(k)}-b_j)-K(\gamma\boldsymbol{w}_k^{\top}\boldsymbol{h}_i^{(k)}-b_k)}{K-1}\right)\right]^{-1},
    \end{equation}
    and
    \begin{equation} \label{equ-jensen_hold_condition_b}
        \gamma\boldsymbol{w}_j^{\top}\boldsymbol{h}_i^{(k)}-b_j=\gamma\boldsymbol{w}_\ell^{\top}\boldsymbol{h}_i^{(k)}-b_\ell,~ \forall j,\ell \neq k \in [K].
    \end{equation}
\end{lemma}

\begin{proof}
    According to Jensen's inequality in Lemma \ref{jensen_inequality_definition}, the normalized CE loss can be lower bounded,
    \begin{align}
        \mathcal{L}_{\mathrm{ce}}(\gamma \boldsymbol{W}\boldsymbol{h}_{i}^{(k)}-\boldsymbol{b}) 
        =&~ \log\left(1+\sum_{j=1 \atop j\neq k}^K\frac{\exp\left(\gamma\boldsymbol{w}_j^{\top}\boldsymbol{h}_i^{(k)}-b_j\right)}{\exp\left(\gamma\boldsymbol{w}_k^{\top}\boldsymbol{h}_i^{(k)}-b_k\right)}\right) \\
        =&~ \log\left(1+\left(K-1\right)\sum_{j=1\atop j\neq k}^K\frac{\exp\left(\gamma\boldsymbol{w}_j^{\top}\boldsymbol{h}_i^{(k)}-b_j\right)}{(K-1)\exp\left(\gamma\boldsymbol{w}_k^{\top}\boldsymbol{h}_i^{(k)}-b_k\right)}\right) \\
        \geq&~ \log\left(1+(K-1)\exp\left(\sum\limits_{j=1\atop j\neq k}^K\frac{\gamma\boldsymbol{w}_j^{\top}\boldsymbol{h}_i^{(k)}-b_j-\gamma\boldsymbol{w}_k^{\top}\boldsymbol{h}_i^{(k)}+b_k}{K-1}\right)\right) \\
        =&~ \log\left(1+\left(K-1\right)\exp\left(\frac{\sum_{j=1}^K(\gamma\boldsymbol{w}_j^{\top}\boldsymbol{h}_i^{(k)}-b_j)-K(\gamma\boldsymbol{w}_k^{\top}\boldsymbol{h}_i^{(k)}-b_k)}{K-1}\right)\right) \label{equ-Lce_Jensen},
    \end{align}
    which achieves the equality if and only if $\gamma\boldsymbol{w}_j^{\top}\boldsymbol{h}_i^{(k)}-b_j=\gamma\boldsymbol{w}_\ell^{\top}\boldsymbol{h}_i^{(k)}-b_\ell$ for all $j,\ell \neq k \in [K]$.

    Second, by the concavity of the function $\log(1+\mathrm{e}^x)$, for any $c'_1>0$, we get
    \begin{align}
        &\mathcal{L}_{\mathrm{ce}}(\gamma \boldsymbol{W}\boldsymbol{h}_{i}^{(k)}-\boldsymbol{b}) \nonumber \\
        \geq &\log\left(1+\left(K-1\right)\exp\left(\frac{\sum_{j=1}^K(\gamma\boldsymbol{w}_j^{\top}\boldsymbol{h}_i^{(k)}-b_j)-K(\gamma\boldsymbol{w}_k^{\top}\boldsymbol{h}_i^{(k)}-b_k)}{K-1}\right)\right) \\
        = &\log\left(\frac{c'_1}{1+c'_1}\frac{1+c'_1}{c'_1}+\frac{1+c'_1}{1+c'_1}\left(K-1\right)\exp\left(\frac{\sum_{j=1}^K(\gamma\boldsymbol{w}_j^{\top}\boldsymbol{h}_i^{(k)}-b_j)-K(\gamma\boldsymbol{w}_k^{\top}\boldsymbol{h}_i^{(k)}-b_k)}{K-1}\right)\right) \\
        \geq &\frac{c'_1}{1+c'_1}\log\left(\frac{1+c'_1}{c'_1}\right) \nonumber \\
        &+\frac{1}{1+c'_1}\log\left(\left(1+c'_1\right)\left(K-1\right)\exp\left(\frac{\sum_{j=1}^K(\gamma\boldsymbol{w}_j^{\top}\boldsymbol{h}_i^{(k)}-b_j)-K(\gamma\boldsymbol{w}_k^{\top}\boldsymbol{h}_i^{(k)}-b_k)}{K-1}\right)\right) \\
        = &\frac1{1+c'_1}\frac{\sum_{j=1}^K(\gamma\boldsymbol{w}_j^{\top}\boldsymbol{h}_i^{(k)}-b_j)-K(\gamma\boldsymbol{w}_k^{\top}\boldsymbol{h}_i^{(k)}-b_k)}{K-1} \nonumber \\
        &+\underbrace{\frac{c'_1}{1+c'_1}\log\left(\frac{1+c'_1}{c'_1}\right)+\frac{1}{1+c'_1}\log\left[\left(1+c_1\right)\left(K-1\right)\right]}_{C'}.
    \end{align}
    The last inequality achieves its equality if and only if
    \begin{align}
        \frac{1+c'_1}{c'_1} &=\left(1+c'_1\right)\left(K-1\right)\exp\left(\frac{\sum_{j=1}^K(\gamma\boldsymbol{w}_j^{\top}\boldsymbol{h}_i^{(k)}-b_j)-K(\gamma\boldsymbol{w}_k^{\top}\boldsymbol{h}_i^{(k)}-b_k)}{K-1}\right), \\
        \text{or}~~ c'_1 &=0 ~~\text{or}~~ c'_1=+\infty.
    \end{align}
    Actually, when $c'_1 = 0$ or $c'_1 = +\infty$, the equality is trivial. Therefore, we have
    \begin{equation}
        c'_1=\left[\left(K-1\right)\exp\left(\frac{\sum_{j=1}^K(\gamma\boldsymbol{w}_j^{\top}\boldsymbol{h}_i^{(k)}-b_j)-K(\gamma\boldsymbol{w}_k^{\top}\boldsymbol{h}_i^{(k)}-b_k)}{K-1}\right)\right]^{-1},
    \end{equation}
    as desired.
\end{proof}

\begin{lemma} \label{lem:lemmaB.2} For the function $f_{\mathrm{ce}}(\boldsymbol W,\boldsymbol H,\boldsymbol{b})$ defined in \textit{Eq.} (\ref{equ-min_fce_formula}),
    \begin{equation} \label{equ-fce_lowerBound_2}
        f_{\mathrm{ce}}(\boldsymbol{W},\boldsymbol{H},\boldsymbol{b})\geq\log\Big(1+(K-1)\e^{-\frac{\gamma K}{K-1}}\Big).
    \end{equation}
\end{lemma}

\begin{proof}
    According to Lemma \ref{lem:lemmaB.1}, \textit{Eq.} (\ref{equ-Lce_lowerBound}) holds for any $c'_1 > 0$ and any $\boldsymbol{h}_i^{(k)}$ with $k\in[K], ~i\in[n]$. We take the same $c'_1$ for all $\boldsymbol{h}_i^{(k)}$, then we have the following lower bound for the function $f_{\mathrm{ce}}$ as {
    \allowdisplaybreaks
    \begin{align}
        &\left(1+c'_1\right)\left(K-1\right)\left[f_{\mathrm{ce}}( \boldsymbol{W},\boldsymbol{H},\boldsymbol{b})-C\right] \\
        = &\left(1+c'_1\right)\left(K-1\right)\left[\frac{1}{nK}\sum_{k=1}^K\sum_{i=1}^n\mathcal{L}_\mathrm{ce}(\gamma \boldsymbol{W}\boldsymbol{h}_i^{(k)}-\boldsymbol{b})-C\right] \\
        \geq &\frac1{nK}\sum_{k=1}^K\sum_{i=1}^n\left[\sum_{j=1}^K\left(\gamma \boldsymbol{w}_j^\top\boldsymbol{h}_i^{(k)}-b_j\right)-K\left(\gamma \boldsymbol{w}_k^\top\boldsymbol{h}_i^{(k)}-b_k\right)\right] \\
        = &\frac{1}{nK}\sum_{i=1}^{n}\Bigg[\left(\sum_{k=1}^{K}\sum_{j=1}^{K}\gamma \boldsymbol{w}_{j}^{\top}\boldsymbol{h}_{i}^{(k)}-K\sum_{k=1}^{K}\gamma \boldsymbol{w}_{k}^{\top}\boldsymbol{h}_{i}^{(k)}\right)+\underbrace{\sum_{k=1}^{K}\sum_{j=1}^{K}\left(b_{k}-b_{j}\right)}_{0}\Bigg] \\
        = &\frac{1}{nK}\sum_{i=1}^{n}\left(\sum_{k=1}^{K}\sum_{j=1}^{K}\gamma \boldsymbol{w}_{k}^{\top}\boldsymbol{h}_{i}^{(j)}-K\sum_{k=1}^{K}\gamma \boldsymbol{w}_{k}^{\top}\boldsymbol{h}_{i}^{(k)}\right) \\
        = &\frac{1}{n}\sum_{i=1}^{n}\sum_{k=1}^{K}\left[\gamma \boldsymbol{w}_{k}^{\top}\left(\frac{1}{K}\sum_{j=1}^{K}\boldsymbol{h}_{i}^{(j)}-\boldsymbol{h}_{i}^{(k)}\right)\right] \\
        = &\frac1n\sum_{i=1}^n\sum_{k=1}^K\gamma \boldsymbol{w}_k^\top\left(\tilde{\boldsymbol{h}}_i-\boldsymbol{h}_i^{(k)}\right), \label{equ-fce_medal_lower_bound}
    \end{align}}
    where $\tilde{\boldsymbol{h}}_{i}=\frac{1}{K}\sum_{j=1}^{K}\boldsymbol{h}_{i}^{(j)}$.

    Then, according to the AM-GM inequality (see Lemma \ref{young_inequality_definition}), we have
    \begin{equation}
        \boldsymbol{u}^\top\boldsymbol{v}\leq\frac{c}{2}\|\boldsymbol{u}\|_2^2+\frac{1}{2c}\|\boldsymbol{v}\|_2^2, ~\forall\boldsymbol{u},\boldsymbol{v}\in\mathbb{R}^d, ~\forall c > 0,
    \end{equation}
    which becomes an equality when $\boldsymbol{v}=c\boldsymbol{u}$. Then, based on \textit{Eq.} (\ref{equ-fce_medal_lower_bound}), we get {
    \allowdisplaybreaks
    \begin{align}
        &~\left(1+c'_1\right)\left(K-1\right)\left[f_{\mathrm{ce}}( \boldsymbol{W},\boldsymbol{H},\boldsymbol{b})-C'\right] \nonumber \\
        \geq&~ \frac1n\sum_{i=1}^n\sum_{k=1}^K\gamma \boldsymbol{w}_k^\top\left(\tilde{\boldsymbol{h}}_i-\boldsymbol{h}_i^{(k)}\right) \\
        \geq&~ -\frac{c'_2}2\sum_{k=1}^K\|\gamma \boldsymbol{w}_k\|_2^2 - \frac1{2{c'_2}n}\sum_{i=1}^n\sum_{k=1}^K\|\tilde{\boldsymbol{h}}_i-\boldsymbol{h}_i^{(k)}\|_2^2 \\
        =&~ -\frac{c'_2}2\sum_{k=1}^K\gamma^2\|\boldsymbol{w}_k\|_2^2 \notag - \frac1{2{c'_2}n}\sum_{i=1}^n\left[\sum_{k=1}^K\|\boldsymbol{h}_i^{(k)}\|_2^2-K\|\tilde{\boldsymbol{h}}_i\|_2^2\right] \\
        = &-\frac{c'_2}{2}\gamma^2 K - \frac1{2{c'_2}n}\left(nK -K\sum_{i=1}^n\|\tilde{\boldsymbol{h}}_i\|_2^2\right),
    \end{align}}
    where the second inequality becomes an equality if and only if
    \begin{equation}
        c'_2\gamma \boldsymbol{w}_{k}=\boldsymbol{h}_{i}^{(k)}-\tilde{\boldsymbol{h}}_{i}, ~\forall k\in[K], ~i\in[n], ~\gamma>0. \label{equ-fce_c3_hold}
    \end{equation}
    Therefore,
    \begin{align}
        f_{\mathrm{ce}}(\boldsymbol{W},\boldsymbol{H},\boldsymbol{b}) 
        \geq &-\frac{c'_2\gamma^2K}{2\left(1+c'_1\right)\left(K-1\right)}-\frac K{2c'_2n\left(1+c'_1\right)\left(K-1\right)}\left(n-\sum_{i=1}^n\|\tilde{\boldsymbol{h}}_i\|_2^2\right) + C' \\
        \geq &-\frac{c'_2\gamma^2K}{2\left(1+c'_1\right)\left(K-1\right)}-\frac K{2c'_2\left(1+c'_1\right)\left(K-1\right)} + C' \\
        = &-\frac K{2\left(1+c'_1\right)\left(K-1\right)}\left(c'_2\gamma^2+\frac1{c'_2}\right) + C', \label{equ-gce_final_equation}
    \end{align}
    where the second inequality becomes an equality if and only if
    \begin{equation}
        \tilde{\boldsymbol{h}}_i=\frac{1}{K}\sum_{j=1}^{K}\boldsymbol{h}_{i}^{(j)}=\boldsymbol{0}, ~\forall i \in [n]. \label{equ-fce_h_mean=0}
    \end{equation}
    Based on \textit{Eqs.} (\ref{equ-fce_c3_hold}) and (\ref{equ-fce_h_mean=0}), we get
    \begin{equation} \label{equ-fce_c3_solution}
        c'_2\gamma \boldsymbol{w}_k=\boldsymbol{h}_i^{(k)}\Rightarrow\|c'_2\gamma \boldsymbol{w}_k\|_2^2=\|\boldsymbol{h}_i^{(k)}\|_2^2\Rightarrow (c'_2\gamma)^2=1\Rightarrow c'_2=\frac1\gamma.
    \end{equation}
    Plugging \textit{Eq.} (\ref{equ-fce_c3_solution}) into (\ref{equ-gce_final_equation}), we have
    \begin{align}
        f_{\mathrm{ce}}(\boldsymbol W,\boldsymbol H,\boldsymbol{b}) \geq -\frac {K(c'_2\gamma^2+\frac1{c'_2})}{2\left(1+c'_1\right)\left(K-1\right)} + C' =-\frac{\gamma K}{\left(1+c'_1\right)\left(K-1\right)}+C'.
    \end{align}

    Furthermore, according to Lemma \ref{lem:lemmaB.3}, when the objective function $f_\text{ce}$ achieves its lower bound, we have $\frac{\partial f_\text{ce}}{\partial b_k}\equiv 0$, and $c'_1$ satisfies
    \begin{equation}
        c'_1=\left[\left(K-1\right)\exp\left(-\frac{\gamma K}{K-1}\right)\right]^{-1}.
    \end{equation}
    Then, we have
    \begin{equation}
        f_{\mathrm{ce}}(\boldsymbol W,\boldsymbol H,\boldsymbol{b}) \geq-\frac{\gamma K}{\left(1+c'_1\right)\left(K-1\right)}+C' =\log\Big(1+(K-1)\mathrm{e}^{-\frac{\gamma K}{K-1}}\Big),
    \end{equation}
    as suggested in \textit{Eq.} (\ref{equ-fce_lowerBound_2}).
\end{proof}

\begin{lemma} \label{lem:lemmaB.3}
    Under the same assumption of Lemma \ref{lem:lemmaB.2}, the lower bound in \textit{Eq.} (\ref{equ-fce_lowerBound_2}) is achieved for any critical point $(\boldsymbol{W},\boldsymbol{H},\boldsymbol{b})$ of \textit{Eq.} (\ref{equ-min_fce_formula}) if and only if the following hold
    \begin{align}
        &\boldsymbol{w}_k=\boldsymbol{h}_i^{(k)}, \quad \forall k\in[K], ~i\in[n],\\
        & \tilde{\boldsymbol{h}}_{i}:=\frac{1}{K}\sum\limits_{k=1}^K\boldsymbol{h}_{i}^{(k)}=\boldsymbol{0}, \quad \forall i\in[n],  \\
        &\boldsymbol{W}\boldsymbol{W}^{\top}=\frac K{K-1}\left(\boldsymbol{I}_K-\frac1K\boldsymbol{1}_K\boldsymbol{1}_K^\top\right), \\
        &\boldsymbol{b}^{\star}=b^{\star}\mathbf{1}, \quad \forall b^\star\in\mathbb R,\\
        &c'_1=\left[\left(K-1\right)\exp\left(-\frac{\gamma K}{K-1}\right)\right]^{-1}.
    \end{align}
\end{lemma}

\begin{proof}
    With the proof of Lemma \ref{lem:lemmaB.2}, to achieve the lower bound, we need at least \textit{Eqs.} (\ref{equ-fce_c3_hold}) and (\ref{equ-fce_h_mean=0}) to hold, \textit{i.e},
    \begin{equation} \label{equ-fce_hmean_w=h_condition}
        \tilde{\boldsymbol{h}}_{i}=\frac{1}{K}\sum_{k=1}^{K}\boldsymbol{h}_{i}^{(k)}=\mathbf{0}, ~\forall i\in[n] \quad \text{and} \quad \boldsymbol{w}_{k}=\boldsymbol{h}_{i}^{(k)}, ~\forall k\in[K], ~i\in[n],
    \end{equation}
    which further implies that
    \begin{equation} \label{equ-fce_sumw=sumh=0}
        \sum_{k=1}^{K}\boldsymbol{w}_{k}=\sum_{k=1}^{K}\boldsymbol{h}_{i}^{(k)}=\boldsymbol{0}.
    \end{equation}
    Then,
    \begin{align}
        \sum_{j=1}^{K}(\gamma \boldsymbol{w}_j^\top\boldsymbol{h}_i^{(k)}-b_j) &=\gamma\left(\sum_{j=1}^{K}\boldsymbol{w}_j^\top\right)\boldsymbol{h}_i^{(k)}-\sum_{j=1}^{K}b_j =0-\sum_{j=1}^{K}b_j=-K\bar{b}, \label{equ-sum_zji}
    \end{align}
    where $\bar{b}=\frac{1}{K}\sum_{j=1}^{K}b_j$, and
    \begin{equation}\label{equ-k_zki}
        K(\gamma \boldsymbol{w}_k^\top\boldsymbol{h}_i^{(k)}-b_k)=K\left(\gamma\|\boldsymbol{w}_k\|_2^2-b_k\right)=K\gamma-Kb_k.
    \end{equation}
    Combining \textit{Eqs.} (\ref{equ-jensen_hold_condition_a}), (\ref{equ-jensen_hold_condition_b}), (\ref{equ-sum_zji}), and (\ref{equ-k_zki}), we have
    \begin{align}
        c'_1 &=\left[\left(K-1\right)\exp\left(\frac{\sum_{j=1}^K(\gamma \boldsymbol{w}_j^\top\boldsymbol{h}_i^{(k)}-b_j)-K(\gamma \boldsymbol{w}_k^\top\boldsymbol{h}_i^{(k)}-b_k)}{K-1}\right)\right]^{-1} \nonumber \\
        &=\left[(K-1)\exp\left(\frac{-K\bar{b}-\left(K\gamma-Kb_k\right)}{K-1}\right)\right]^{-1} \nonumber \\
        &=\left[(K-1)\exp\left(\frac{K}{K-1}\left(b_k-\overline{b}-\gamma\right)\right)\right]^{-1}. \label{equ-fce_c1_medal_solution}
    \end{align}
    Since the $c'_1 > 0$ can be arbitrary number, we choose the same $c'_1$ for all $k \in [K]$. Then, we have
    \begin{align}
        b_k-\bar{b}-\gamma &=b_j-\bar{b}-\gamma \Leftrightarrow b_k=b_j, ~\forall j \neq k \in [K]. \label{equ-fce_bj=bk}
    \end{align}
    Therefore, we can write $\boldsymbol{b}=b\mathbf{1}_K$. Moreover, plugging \textit{Eqs.} (\ref{equ-fce_hmean_w=h_condition}) and (\ref{equ-fce_bj=bk}) into \textit{Eq.} (\ref{equ-k_zki}), we get
    \begin{align}
        \gamma\boldsymbol{w}_j^{\top}\boldsymbol{h}_i^{(k)}-b_j=\gamma\boldsymbol{w}_\ell^{\top}\boldsymbol{h}_i^{(k)}-b_\ell &\Leftrightarrow\boldsymbol{w}_j^\top\boldsymbol{h}_i^{(k)}=\boldsymbol{w}_\ell^\top\boldsymbol{h}_i^{(k)} \Leftrightarrow\boldsymbol{w}_j^\top\boldsymbol{w}_k=\boldsymbol{w}_\ell^\top\boldsymbol{w}_k.
    \end{align}

    Then, combining with \textit{Eq.} (\ref{equ-fce_sumw=sumh=0}), we have
    \begin{align}
        1 = \|\boldsymbol{w}_{k}\|_{2}^{2}&=-\sum_{j=1 \atop j \neq k}^{K}\boldsymbol{w}_{k}^{\top}\boldsymbol{w}_{j}=-\left(K-1\right)\boldsymbol{w}_{k}^{\top}\boldsymbol{w}_{j} ~\Rightarrow~ \boldsymbol{w}_{k}^{\top}\boldsymbol{w}_{j}=-\frac{1}{K-1}, ~\forall j\neq k\in[K]. \label{equ-fce_wkwj_solution}
    \end{align}
    Therefore,
    \begin{equation}
        \boldsymbol{W}\boldsymbol{W}^{\top}=\frac K{K-1}\left(\boldsymbol{I}_K-\frac1K\boldsymbol{1}_K\boldsymbol{1}_K^\top\right).
    \end{equation}
    Finally, plugging the results in \textit{Eq.} (\ref{equ-fce_bj=bk}) into (\ref{equ-fce_c1_medal_solution}), we have
    \begin{align}
       c'_1 &=\left[(K-1)\exp\left(\frac{K}{K-1}\left(b_k-\overline{b}-\gamma\right)\right)\right]^{-1} \nonumber =\left[(K-1)\exp\left(-\frac{\gamma K}{K-1}\right)\right]^{-1}.
    \end{align}

    When $f_\mathrm{ce}$ defined in \textit{Eq.} (\ref{equ-min_fce_formula}) achieves its lower bound, it theoretically satisfies
    \begin{equation}
        \frac{\partial f_{\mathrm{ce}}}{\partial b_{k}}=\frac1{nK}\left(n-\sum_{j=1}^K\sum_{i=1}^n\frac{\exp(\gamma \boldsymbol{w}_k^\top\boldsymbol{h}_i^{(j)}-b_k)}{\sum_{\ell=1}^K\exp(\gamma \boldsymbol{w}_\ell^\top\boldsymbol{h}_i^{(j)}-b_\ell)}\right)=0, ~\forall k,\ell\in[K].
    \end{equation}
    However, combining with \textit{Eqs.} (\ref{equ-fce_bj=bk}) and (\ref{equ-fce_wkwj_solution}), we have
    \begin{align}
        &n-\sum_{j=1}^K\sum_{i=1}^n\frac{\exp(\gamma \boldsymbol{w}_k^\top\boldsymbol{h}_i^{(j)}-b_k)}{\sum_{\ell=1}^K\exp(\gamma \boldsymbol{w}_\ell^\top\boldsymbol{h}_i^{(j)}-b_\ell)} \notag \\
        ={}&n-\sum_{j=1}^K\sum_{i=1}^n\frac{\exp(\gamma \boldsymbol{w}_k^\top\boldsymbol{h}_i^{(j)}-b_k)}{\sum_{\ell\neq j}^K\exp(-\frac\gamma{K-1}-b_\ell)+\exp(\gamma-b_j)} \\
        =~&n-\sum_{i=1}^n\left(\frac{\sum_{j\neq k}^K\exp(-\frac{\gamma}{K-1}-b)+\exp(\gamma-b)}{(K-1)\exp(-\frac{\gamma}{K-1}-b)+\exp(\gamma-b)}\right) 
        =n-n 
        \equiv0,
    \end{align}
    which means when the function $f_\mathrm{ce}$ achieves its lower bound, the classifier biases $\boldsymbol{b}^{\star}$ only needs to satisfies $\boldsymbol{b}^{\star}=b^{\star}\mathbf{1}$. In other words, in terms of the classifier biases, the function $f_{\text{ce}}$ has infinitely many minima.
\end{proof}

\clearpage
\subsection{Proof of Theorems \ref{them:CE_decay_of_decision_score} and \ref{them:BCE_decay_of_decision_score}}\label{proof:coro}
\begin{lemma}\label{lemma_decay_exponential_bce}
    (1) For a polynomial $\bm P(\gamma)$ and Sigmoid function $\sigma(s\gamma-b) = \frac{1}{1+\exp(-s\gamma+b)}$ with $s, b\in\mathbb R$, as $\gamma$ approaches positive infinity, their product exhibits exponential decay, converging to zero,
    \begin{align}
        &\lim_{\gamma\rightarrow+\infty} \bm P(\gamma)\sigma(s\gamma-b) = \lim_{\gamma\rightarrow+\infty} \bm P(\gamma)\frac{1}{1+\exp(-s\gamma+b)} = 0,\quad~\forall s<0,~b\in\mathbb R,     \label{eq_limit_bce_infinity_neg}\\
        &\lim_{\gamma\rightarrow+\infty} \bm P(\gamma)\big(1-\sigma(s\gamma-b)\big) = \lim_{\gamma\rightarrow+\infty} \bm P(\gamma)\frac{\exp(-s\gamma+b)}{1+\exp(-s\gamma+b)} = 0,\quad~\forall s>0,~b\in\mathbb R.  \label{eq_limit_bce_infinity_pos}
    \end{align}
    (2) For a polynomial $\bm P(\gamma)$ without constant term and Sigmoid function $\sigma(s\gamma-b) = \frac{1}{1+\exp(-s\gamma+b)}$ with $s, b\in\mathbb R$, their product approaches $0$ as $\gamma$ approaches zero, i.e.,
    \begin{align}
        &\lim_{\gamma\rightarrow0} \bm P(\gamma)\sigma(s\gamma-b) = \lim_{\gamma\rightarrow0} \bm P(\gamma)\frac{1}{1+\exp(-s\gamma+b)} = 0,\quad~\forall s, b\in\mathbb R,\\
        &\lim_{\gamma\rightarrow0} \bm P(\gamma)\big(1-\sigma(s\gamma-b)\big) = \lim_{\gamma\rightarrow0} \bm P(\gamma)\frac{\exp(-s\gamma+b)}{1+\exp(-s\gamma+b)} = 0,\quad~\forall s, b\in\mathbb R.
    \end{align}
\end{lemma}
\begin{proof}
(1) Suppose that $\bm P(\gamma)$ is an $p$-degree polynomial in terms of $\gamma$, with the leading coefficient being $a_p$. Then by applying L'hospital's rule $p$ times, we have
\begin{align}
    &\lim_{\gamma\rightarrow+\infty} \bm P(\gamma)\big(1-\sigma(s\gamma-b)\big) =
      \lim_{\gamma\rightarrow+\infty} \bm P(\gamma)\frac{\exp(-s\gamma+b)}{1+\exp(-s\gamma+b)}\nonumber\\
    = &\lim_{\gamma\rightarrow+\infty}\frac{\bm P(\gamma)}{1+\exp(s\gamma-b)}
    = \lim_{\gamma\rightarrow+\infty}\frac{\mathtt{d}\bm P(\gamma)/\mathtt{d}\gamma}{s\exp(s\gamma-b)} = \cdots\nonumber\\
    = &\lim_{\gamma\rightarrow+\infty}\frac{\mathtt{d}^p\bm P(\gamma)/\mathtt{d}\gamma^p}{s^p\exp(s\gamma-b)} 
    = \lim_{\gamma\rightarrow+\infty}\frac{p!\,a_p}{s^p\exp(s\gamma-b)} = 0,\quad \forall s > 0.
\end{align}
When $s<0$, one can similarly prove Eq. (\ref{eq_limit_bce_infinity_neg}).

(2) As $\bm P(\gamma)$ is without constant term, $\lim_{\gamma\rightarrow0} \bm P(\gamma) = 0$. Therefore,
\begin{align}
    &\lim_{\gamma\rightarrow0} \bm P(\gamma)\sigma(s\gamma-b) =\lim_{\gamma\rightarrow0} \bm P(\gamma)\lim_{\gamma\rightarrow0} \sigma(s\gamma-b) = 0\cdot \lim_{\gamma\rightarrow0}\frac{1}{1+\exp(-s\gamma+b)} = 0\cdot \frac{1}{1+\e^b} = 0,\nonumber\\
    &\lim_{\gamma\rightarrow0} \bm P(\gamma)\big(1-\sigma(s\gamma-b)\big)
     =\lim_{\gamma\rightarrow0} \bm P(\gamma)\lim_{\gamma\rightarrow0} \big(1-\sigma(s\gamma-b)\big) \nonumber\\
     &\qquad\qquad\qquad\qquad\qquad\quad= 0\cdot \lim_{\gamma\rightarrow0}\frac{\exp(-s\gamma+b)}{1+\exp(-s\gamma+b)} = 0\cdot \frac{\e^b}{1+\e^b} = 0,\nonumber
\end{align}
which are desired.
\end{proof}

\begin{lemma}\label{lemma_decay_exponential_ce}
Let $\{s_j\}_{j=1}^K \subset \mathbb R$ and $\{b_j\}_{j=1}^K \subset \mathbb R$ be two sets of real numbers and
\begin{align}
    s_k = \max \{s_j\}_{j=1}^K > \max \{s_j\}_{j=1\atop j\neq k}^K.
\end{align}
    (1) For a polynomial $\bm P(\gamma)$ and Softmax function $\mathtt{Softmax}_\ell(\gamma) = \frac{\exp(s_\ell\gamma - b_\ell)}{\sum_{j=1}^K\exp(s_j\gamma - b_j)}$, as $\gamma$ approaches positive infinity, their product exhibits exponential decay, converging to zero,
    \begin{align}
        &\lim_{\gamma\rightarrow+\infty} \bm P(\gamma)\mathtt{Softmax}_\ell(\gamma) = \lim_{\gamma\rightarrow+\infty} \bm P(\gamma)\frac{\exp(s_\ell\gamma - b_\ell)}{\sum_{j=1}^K\exp(s_j\gamma - b_j)} = 0,\qquad \forall \ell\neq k,\\
        &\lim_{\gamma\rightarrow+\infty} \bm P(\gamma)\big(1-\mathtt{Softmax}_k(\gamma)\big) = \lim_{\gamma\rightarrow+\infty} \bm P(\gamma)\frac{\sum_{j=1\atop j\neq k}^K\exp(s_j\gamma - b_j)}{\sum_{j=1}^K\exp(s_j\gamma - b_j)} = 0.
    \end{align}
    (2) For a polynomial $\bm P(\gamma)$ without constant term and $\mathtt{Softmax}_\ell(\gamma) = \frac{\exp(s_\ell\gamma - b_\ell)}{\sum_{j=1}^K\exp(s_j\gamma - b_j)}$, their product approaches 0 as $\gamma$ approaches $0$, and
    \begin{align}
        &\lim_{\gamma\rightarrow0} \bm P(\gamma)\mathtt{Softmax}_\ell(\gamma) = \lim_{\gamma\rightarrow0} \bm P(\gamma)\frac{\exp(s_\ell\gamma - b_\ell)}{\sum_{j=1}^K\exp(s_j\gamma - b_j)} = 0,\qquad \forall j\neq k,\\
        &\lim_{\gamma\rightarrow0} \bm P(\gamma)\big(1-\mathtt{Softmax}_k(\gamma)\big) = \lim_{\gamma\rightarrow0} \bm P(\gamma)\frac{\sum_{j=1\atop j\neq k}^K\exp(s_j\gamma - b_j)}{\sum_{j=1}^K\exp(s_j\gamma - b_j)} = 0.
    \end{align}
\end{lemma}
\begin{proof}
    (1) Suppose $\bm P(\gamma) = a_p \gamma^p + a_{p-1}\gamma^{p-1}+\cdots+a_1\gamma+a_0$. Then,
    \begin{align}
        0&\leq\Big|\bm P(\gamma)\mathtt{Softmax}_\ell(\gamma)\Big| \leq \sum_{i=1}^p|a_i|\gamma^i\frac{\exp(s_\ell\gamma - b_\ell)}{\sum_{j=1}^K\exp(s_j\gamma - b_j)}\nonumber\\
        &\leq \sum_{i=1}^p|a_i|\gamma^i \frac{\exp(s_\ell\gamma - b_\ell)}{\exp(s_k\gamma - b_k)} = \sum_{i=1}^p|a_i|\gamma^i \frac{1}{\e^{(s_k-s_\ell)\gamma - (b_k-b_\ell)}}.
    \end{align}
    As $s_k - s_\ell > 0$ for $\ell \neq k$, the right side approaches zero when $\gamma$ approaches positive infinity, according to Lemma \ref{lemma_decay_exponential_bce}(1). Therefore, $\lim_{\gamma\rightarrow0} \bm P(\gamma)\mathtt{Softmax}_\ell(\gamma) = 0$.

    By applying the arithmetic operations of limits,
    \begin{align}
        \lim_{\gamma\rightarrow+\infty} \bm P(\gamma)\big(1-\mathtt{Softmax}_k(\gamma)\big) = \sum_{\ell =1\atop \ell\neq k}^K\lim_{\gamma\rightarrow+\infty} \bm P(\gamma)\frac{\exp(s_\ell\gamma - b_\ell)}{\sum_{j=1}^K\exp(s_j\gamma - b_j)} = \sum_{\ell =1\atop \ell\neq k}^K 0 = 0.
    \end{align}

    (2) As $\bm P(\gamma)$ is without constant term, $\lim_{\gamma\rightarrow0} \bm P(\gamma) = 0$. Therefore,
    \begin{align}
        &0 \leq \lim_{\gamma\rightarrow0} \Big|\bm P(\gamma)\mathtt{Softmax}_\ell(\gamma)\Big| \leq \lim_{\gamma\rightarrow0} \Big|\bm P(\gamma)\Big|\lim_{\gamma\rightarrow0}\Big|\mathtt{Softmax}_\ell(\gamma)\Big| \leq 0\cdot 1 = 0\quad\text{and}\nonumber\\
        &0 \leq \lim_{\gamma\rightarrow0} \Big|\bm P(\gamma)\big(1-\mathtt{Softmax}_\ell(\gamma)\big)\Big| \leq \lim_{\gamma\rightarrow0} \Big|\bm P(\gamma)\Big|\lim_{\gamma\rightarrow0}\Big|1-\mathtt{Softmax}_\ell(\gamma)\Big| \leq 0\cdot 1 = 0,\nonumber
    \end{align}
    which lead to the conclusions.
\end{proof}

\begin{theorem}\label{them:BCE_decay_of_decision_score_supp}
When training the model using the BCE loss $f_{\text{bce}}$,
as $\gamma$ approaches zero, the linear decrease in $\gamma$ leads to a linear decay in the convergence rate of unbiased decision scores.
In contrary, once the \textbf{critical condition II} is satisfied, as $\gamma$ linearly approaches positive infinity, the convergence rate of unbiased decision scores decay exponentially.
\end{theorem}
\begin{proof}
In the model training, for any unbiased decision score $s_i^{(kj)} = \gamma \bm w_k^\top \bm h_i^{(j)}$,
its update step is determined by that of the classifier vector $\bm w_k$ and the feature $\bm h_i^{(j)}$.
For an unconstrained feature model (UFM), $\{\bm w_k\}_k$ and $\{\bm h_i^{(j)}\}_{i,j}$ are independent variables.
Therefore, with an iteration, they are updated as
\begin{align}
    \hat{\bm w}_k = \bm w_k - \eta \frac{\partial f_\mu}{\partial \bm w_k}\quad \text{and}\quad
    \hat{\bm h}_i^{(j)} = \bm h_i^{(j)} - \eta \frac{\partial f_\mu}{\partial \bm h_i^{(j)}},
\end{align}
where $\eta$ is the learning rate and $\mu\in\{\text{ce}, \text{bce}\}$.
Then, the unbiased decision score is updated as
\begin{align}
    \hat{s}_i^{(kj)} & = \gamma \hat{\bm w}_k^\top \hat{\bm h}_i^{(j)}
                     = \gamma \Big(\bm w_k - \eta \frac{\partial f_\mu}{\partial \bm w_k}\Big)^\top \Big(\bm h_i^{(j)} - \eta \frac{\partial f_\mu}{\partial \bm h_i^{(j)}}\Big)\\
                     & = \gamma \bm w_k^\top \bm h_i^{(j)} - \gamma\eta \bm w_k^\top \frac{\partial f_\mu}{\partial \bm h_i^{(j)}}
                        - \gamma\eta\Big(\frac{\partial f_\mu}{\partial \bm w_k}\Big)^\top\bm h_i^{(j)}
                        + \gamma\eta^2\Big(\frac{\partial f_\mu}{\partial \bm w_k}\Big)^\top\frac{\partial f_\mu}{\partial \bm h_i^{(j)}},
\end{align}
and its update step is
\begin{align}
    \triangle(s_i^{(kj)}) = s_i^{(kj)} - \hat{s}_i^{(kj)} = - \gamma\eta \bm w_k^\top \frac{\partial f_\mu}{\partial \bm h_i^{(j)}}
                            - \gamma\eta\Big(\frac{\partial f_\mu}{\partial \bm w_k}\Big)^\top\bm h_i^{(j)}
                            + \gamma\eta^2\Big(\frac{\partial f_\mu}{\partial \bm w_k}\Big)^\top\frac{\partial f_\mu}{\partial \bm h_i^{(j)}}.
\end{align}

For the BCE loss,
\begin{align}
    \frac{\partial f_{\text{bce}}}{\partial\bm h_i^{(j)}} & = \frac{\gamma}{nK}\bigg(
             -\frac{\e^{-\gamma\bm w_j^\top\bm h_i^{(j)} + b_j}}{1+\e^{-\gamma\bm w_j^\top\bm h_i^{(j)} + b_j}}\bm w_j
            +\sum_{\ell=1\atop \ell\neq j}^K\frac{\e^{\gamma\bm w_\ell^\top\bm h_i^{(j)} - b_\ell}}{1+\e^{\gamma\bm w_\ell^\top\bm h_i^{(j)} - b_\ell}}\bm w_\ell\bigg),\\
    \frac{\partial f_{\text{bce}}}{\partial\bm w_k} & = \frac{\gamma}{nK}\sum_{i'=1}^n\bigg(
             -\frac{\e^{-\gamma\bm w_k^\top\bm h_{i'}^{(k)} + b_k}}{1+\e^{-\gamma\bm w_k^\top\bm h_{i'}^{(k)} + b_k}}\bm h_{i'}^{(k)}
            +\sum_{\ell'=1\atop \ell'\neq k}^K\frac{\e^{\gamma\bm w_k^\top\bm h_{i'}^{(\ell')} - b_k}}{1+\e^{\gamma\bm w_k^\top\bm h_{i'}^{(\ell')} - b_k}}\bm h_{i'}^{(\ell')}\bigg),
\end{align}
and
\begin{align}
        \triangle(s_i^{(kj)}) = ~& \frac{\gamma^2\eta}{nK}\bm w_k^\top\bigg(
             \frac{\e^{-\gamma\bm w_j^\top\bm h_i^{(j)} + b_j}}{1+\e^{-\gamma\bm w_j^\top\bm h_i^{(j)} + b_j}}\bm w_j
            -\sum_{\ell=1\atop \ell\neq j}^K\frac{\e^{\gamma\bm w_{\ell}^\top\bm h_i^{(j)} - b_\ell}}{1+\e^{\gamma\bm w_\ell^\top\bm h_i^{(j)} - b_\ell}}\bm w_\ell\bigg)\nonumber\\
           & + \frac{\gamma^2\eta}{nK}\sum_{i'=1}^n\bigg(
             \frac{\e^{-\gamma\bm w_k^\top\bm h_{i'}^{(k)} + b_k}}{1+\e^{-\gamma\bm w_k^\top\bm h_{i'}^{(k)} + b_k}}\bm h_{i'}^{(k)}
            -\sum_{\ell'=1\atop \ell'\neq k}^K\frac{\e^{\gamma\bm w_k^\top\bm h_{i'}^{(\ell')} - b_k}}{1+\e^{\gamma\bm w_k^\top\bm h_{i'}^{(\ell')} - b_k}}\bm h_{i'}^{(\ell')}\bigg)^\top\bm h_i^{(j)}\nonumber\\
           & +\frac{\gamma^3\eta^2}{n^2K^2}\sum_{i'=1}^n\bigg(
             \frac{\e^{-\gamma\bm w_k^\top\bm h_{i'}^{(k)} + b_k}}{1+\e^{-\gamma\bm w_k^\top\bm h_{i'}^{(k)} + b_k}}\bm h_{i'}^{(k)}
            -\sum_{\ell'=1\atop \ell'\neq k}^K\frac{\e^{\gamma\bm w_k^\top\bm h_{i'}^{(\ell')} - b_k}}{1+\e^{\gamma\bm w_k^\top\bm h_{i'}^{(\ell')} - b_k}}\bm h_{i'}^{(\ell')}\bigg)^\top \nonumber\\
            & \qquad\qquad \bigg(\frac{\e^{-\gamma\bm w_j^\top\bm h_i^{(j)} + b_j}}{1+\e^{-\gamma\bm w_j^\top\bm h_i^{(j)} + b_j}}\bm w_j
            -\sum_{\ell=1\atop \ell\neq j}^K\frac{\e^{\gamma\bm w_\ell^\top\bm h_i^{(j)} - b_\ell}}{1+\e^{\gamma\bm w_\ell^\top\bm h_i^{(j)} - b_\ell}}\bm w_\ell\bigg)\\
            = ~& \frac{\gamma^2\eta}{nK} \bigg(
                    \frac{\e^{-\gamma\bm w_j^\top\bm h_i^{(j)} + b_j}}{1+\e^{-\gamma\bm w_j^\top\bm h_i^{(j)} + b_j}}\bm w_k^\top \bm w_j
              -\sum_{\ell=1\atop \ell\neq j}^K
                    \frac{\e^{\gamma\bm w_\ell^\top\bm h_i^{(j)} - b_\ell}}{1+\e^{\gamma\bm w_\ell^\top\bm h_i^{(j)} - b_\ell}}\bm w_k^\top\bm w_\ell\bigg)\nonumber\\
           & + \frac{\gamma^2\eta}{nK}\sum_{i'=1}^n\bigg(
                    \frac{\e^{-\gamma\bm w_k^\top\bm h_{i'}^{(k)} + b_k}}{1+\e^{-\gamma\bm w_k^\top\bm h_{i'}^{(k)} + b_k}}\bm h_{i'}^{(k)\top}\bm h_i^{(j)}
            -\sum_{\ell'=1\atop \ell'\neq k}^K
                    \frac{\e^{\gamma\bm w_k^\top\bm h_{i'}^{(\ell')} - b_k}}{1+\e^{\gamma\bm w_k^\top\bm h_{i'}^{(\ell')} - b_k}}\bm h_{i'}^{(\ell')\top}\bm h_i^{(j)}\bigg)\nonumber\\
           & + \frac{\gamma^3\eta^2}{n^2K^2}\bigg(\sum_{i'=1}^n
                \frac{\e^{-\gamma\bm w_k^\top\bm h_{i'}^{(k)} + b_k}}{1+\e^{-\gamma\bm w_k^\top\bm h_{i'}^{(k)} + b_k}}
                \frac{\e^{-\gamma\bm w_j^\top\bm h_i^{(j)} + b_j}}{1+\e^{-\gamma\bm w_j^\top\bm h_i^{(j)} + b_j}}\bm w_j^\top\bm h_{i'}^{(k)}\nonumber\\
           & \qquad\qquad  - \sum_{i'=1}^n \sum_{\ell=1\atop \ell\neq j}^K
                \frac{\e^{-\gamma\bm w_k^\top\bm h_{i'}^{(k)} + b_k}}{1+\e^{-\gamma\bm w_k^\top\bm h_{i'}^{(k)} + b_k}}
                \frac{\e^{\gamma\bm w_{\ell}^\top\bm h_i^{(j)} - b_{\ell}}}{1+\e^{\gamma\bm w_{\ell}^\top\bm h_i^{(j)} - b_{\ell}}}\bm w_{\ell}^\top\bm h_{i'}^{(k)} \nonumber\\
           & \qquad\qquad  - \sum_{i'=1}^n\sum_{\ell'=1\atop \ell'\neq k}^K
                \frac{\e^{\gamma\bm w_k^\top\bm h_{i'}^{(\ell')} - b_k}}{1+\e^{\gamma\bm w_k^\top\bm h_{i'}^{(\ell')} - b_k}}
                \frac{\e^{-\gamma\bm w_j^\top\bm h_i^{(j)} + b_j}}{1+\e^{-\gamma\bm w_j^\top\bm h_i^{(j)} + b_j}}\bm w_j^\top\bm h_{i'}^{(\ell')}\nonumber\\
           & \qquad\qquad  + \sum_{i'=1}^n\sum_{\ell'=1\atop \ell'\neq k}^K\sum_{\ell=1\atop \ell\neq k}^K
                \frac{\e^{\gamma\bm w_k^\top\bm h_{i'}^{(\ell')} - b_k}}{1+\e^{\gamma\bm w_k^\top\bm h_{i'}^{(\ell')} - b_k}}
                \frac{\e^{\gamma\bm w_{\ell}^\top\bm h_i^{(j)} - b_{\ell}}}{1+\e^{\gamma\bm w_{\ell}^\top\bm h_i^{(j)} - b_{\ell}}}\bm w_{\ell}^\top\bm h_{i'}^{(\ell')}\bigg).
\end{align}
Then,
\begin{align}
           0\leq\Big|\triangle(s_i^{(kj)})\Big|\leq ~& \frac{\gamma^2\eta}{nK} \bigg(
                    \frac{\e^{-\gamma\bm w_j^\top\bm h_i^{(j)} + b_j}}{1+\e^{-\gamma\bm w_j^\top\bm h_i^{(j)} + b_j}}
              +\sum_{\ell=1\atop \ell\neq j}^K
                    \frac{\e^{\gamma\bm w_\ell^\top\bm h_i^{(j)} - b_\ell}}{1+\e^{\gamma\bm w_\ell^\top\bm h_i^{(j)} - b_\ell}}\bigg)\nonumber\\
           & + \frac{\gamma^2\eta}{nK}\sum_{i'=1}^n\bigg(
                    \frac{\e^{-\gamma\bm w_k^\top\bm h_{i'}^{(k)} + b_k}}{1+\e^{-\gamma\bm w_k^\top\bm h_{i'}^{(k)} + b_k}}
            +\sum_{\ell'=1\atop \ell'\neq k}^K
                    \frac{\e^{\gamma\bm w_k^\top\bm h_{i'}^{(\ell')} - b_k}}{1+\e^{\gamma\bm w_k^\top\bm h_{i'}^{(\ell')} - b_k}}\bigg)\nonumber\\
           & + \frac{\gamma^3\eta^2}{n^2K^2}\bigg(\sum_{i'=1}^n
                \frac{\e^{-\gamma\bm w_k^\top\bm h_{i'}^{(k)} + b_k}}{1+\e^{-\gamma\bm w_k^\top\bm h_{i'}^{(k)} + b_k}}
                \frac{\e^{-\gamma\bm w_j^\top\bm h_i^{(j)} + b_j}}{1+\e^{-\gamma\bm w_j^\top\bm h_i^{(j)} + b_j}}\nonumber\\
           & \qquad\qquad  + \sum_{i'=1}^n \sum_{\ell=1\atop \ell\neq j}^K
                \frac{\e^{-\gamma\bm w_k^\top\bm h_{i'}^{(k)} + b_k}}{1+\e^{-\gamma\bm w_k^\top\bm h_{i'}^{(k)} + b_k}}
                \frac{\e^{\gamma\bm w_{\ell}^\top\bm h_i^{(j)} - b_{\ell}}}{1+\e^{\gamma\bm w_{\ell}^\top\bm h_i^{(j)} - b_{\ell}}} \nonumber\\
           & \qquad\qquad  + \sum_{i'=1}^n\sum_{\ell'=1\atop \ell'\neq k}^K
                \frac{\e^{\gamma\bm w_k^\top\bm h_{i'}^{(\ell')} - b_k}}{1+\e^{\gamma\bm w_k^\top\bm h_{i'}^{(\ell')} - b_k}}
                \frac{\e^{-\gamma\bm w_j^\top\bm h_i^{(j)} + b_j}}{1+\e^{-\gamma\bm w_j^\top\bm h_i^{(j)} + b_j}}\nonumber\\
           & \qquad\qquad  + \sum_{i'=1}^n\sum_{\ell'=1\atop \ell'\neq k}^K\sum_{\ell=1\atop \ell\neq k}^K
                \frac{\e^{\gamma\bm w_k^\top\bm h_{i'}^{(\ell')} - b_k}}{1+\e^{\gamma\bm w_k^\top\bm h_{i'}^{(\ell')} - b_k}}
                \frac{\e^{\gamma\bm w_{\ell}^\top\bm h_i^{(j)} - b_{\ell}}}{1+\e^{\gamma\bm w_{\ell}^\top\bm h_i^{(j)} - b_{\ell}}}\bigg). \label{eq_step_bce_decision_score}
\end{align}
Each term in the right side of Eq. (\ref{eq_step_bce_decision_score}) is product of a polynomial without constant term and Sigmoid functions in terms of $\gamma$.
When the \textbf{critical condition II} holds, i.e.,
\begin{align}
    \min\bigcup_{k=1}^K\big\{\bm w_k^\top\bm h_i^{(k)}\big\}_{i=1}^n > 0 > \max\bigcup_{k=1}^K\bigcup_{j=1\atop j\neq k}^K\big\{\bm w_k^\top\bm h_i^{(j)}\big\}_{i=1}^n,
\end{align}
every term in Eq. (\ref{eq_step_bce_decision_score}) satisfies the requirements of Lemma \ref{lemma_decay_exponential_bce}(1);
then, as $\gamma$ increases, $\triangle(s_i^{(kj)})$ decays exponentially and converges to $0$.

When $\gamma$ approaches zero, each term in the right side of Eq. (\ref{eq_step_bce_decision_score}) satisfies the requirements of Lemma \ref{lemma_decay_exponential_bce}(2).
Then, as $\gamma$ decreases linearly toward 0, every term in the right side of Eq. (\ref{eq_step_bce_decision_score}) decays to $0$ at a quadratic or cubic rate; consequently, $\triangle(s_i^{(kj)})$ decays quadratically to $0$.
Consider that the optimal values of unbiased decision scores are $\gamma$ or $-\frac{\gamma}{K-1}$, which are linearly decay as $\gamma$ decreases.
Therefore, the convergence rate of the unbiased decision scores decay linearly when $\gamma$ linearly decreases to zero.
\end{proof}

\begin{theorem}\label{them:CE_decay_of_decision_score_supp}
When training the model using the CE loss $f_{\text{ce}}$, as $\gamma$ approaches zero,
the linear decrease in $\gamma$ leads to a linear decay in the convergence rate of unbiased decision scores.
Once the \textbf{critical condition I} is satisfied, as $\gamma$ linearly approaches positive infinity, the convergence rate of unbiased decision scores decay exponentially.
\end{theorem}
\begin{proof}
For the CE loss,
\begin{align}
    \frac{\partial f_{\text{ce}}}{\partial\bm h_i^{(j)}} & = \frac{\gamma}{nK}\bigg[
        \bigg(\frac{\e^{\gamma\bm w_j^\top\bm h_i^{(j)}-b_j}}{\sum_{m=1}^K\e^{\gamma\bm w_m^\top\bm h_i^{(j)}-b_m}}-1\bigg)\bm w_j
        +\sum_{\ell=1\atop\ell\neq j}^K\frac{\e^{\gamma\bm w_\ell^\top\bm h_i^{(j)}-b_\ell}}{\sum_{m=1}^K\e^{\gamma\bm w_m^\top\bm h_i^{(j)}-b_m}}\bm w_\ell
        \bigg],\\
    \frac{\partial f_{\text{ce}}}{\partial\bm w_k} & = \frac{\gamma}{nK}\sum_{i'=1}^n\bigg[
        \bigg(\frac{\e^{\gamma\bm w_k^\top\bm h_{i'}^{(k)}-b_k}}{\sum_{m=1}^K\e^{\gamma\bm w_m^\top\bm h_{i'}^{(k)}-b_m}}-1\bigg)\bm h_{i'}^{(k)}
        +\sum_{\ell'=1\atop\ell'\neq k}^K\frac{\e^{\gamma\bm w_k^\top\bm h_{i'}^{(\ell')}-b_k}}{\sum_{m=1}^K\e^{\gamma\bm w_m^\top\bm h_{i'}^{(\ell')}-b_m}}\bm h_{i'}^{(\ell')}
        \bigg].
\end{align}
The critical condition I is
\begin{align}
     \bm w_k \bm h_i^{(k)} > \max \{\bm w_j \bm h_i^{(k)}: j=1,2,\cdots,K, \text{and}~j\neq k\},\quad\forall k,i.
\end{align}
Similar to the proof of Theorem \ref{them:BCE_decay_of_decision_score_supp}, one can easily get the conclusions with help of Lemma \ref{lemma_decay_exponential_ce}.
\end{proof}

%% file: MLJ_main.bbl
\begin{thebibliography}{}
\renewcommand{\doi}[1]{\url{https://doi.org/#1}}
\bibcommenthead

\bibitem [\protect \citeauthoryear {%
Alshammari%
, Wang%
, Ramanan%
\BCBL {}\ \BBA {} Kong%
}{%
Alshammari%
\ \protect \BOthers {.}}{%
{\protect \APACyear {2022}}%
}]{%
alshammari2022long}
\APACinsertmetastar {%
alshammari2022long}%
\begin{APACrefauthors}%
Alshammari, S.%
, Wang, Y\BHBI X.%
, Ramanan, D.%
\BCBL {} Kong, S.%
\end{APACrefauthors}%
\unskip\
\newblock
\APACrefYearMonthDay{2022}{}{}.
\newblock
{\BBOQ}\APACrefatitle {Long-tailed recognition via weight balancing}
  {Long-tailed recognition via weight balancing}.{\BBCQ}
\newblock
 \APACrefbtitle {Proceedings of the {IEEE/CVF} conference on computer vision
  and pattern recognition} {Proceedings of the {IEEE/CVF} conference on
  computer vision and pattern recognition}\ (\BPGS\ 6897--6907).
\PrintBackRefs{\CurrentBib}

\bibitem [\protect \citeauthoryear {%
Azad%
\ \protect \BOthers {.}}{%
Azad%
\ \protect \BOthers {.}}{%
{\protect \APACyear {2024}}%
}]{%
Azad2024medical}
\APACinsertmetastar {%
Azad2024medical}%
\begin{APACrefauthors}%
Azad, R.%
, Aghdam, E.K.%
, Rauland, A.%
, Jia, Y.%
, Avval, A.H.%
, Bozorgpour, A.%
\BDBL {}Merhof, D.%
\end{APACrefauthors}%
\unskip\
\newblock
\APACrefYearMonthDay{2024}{}{}.
\newblock
{\BBOQ}\APACrefatitle {Medical Image Segmentation Review: The Success of
  {U-Net}} {Medical image segmentation review: The success of {U-Net}}.{\BBCQ}
\newblock
\APACjournalVolNumPages{IEEE Transactions on Pattern Analysis and Machine
  Intelligence}{46}{12}{10076--10095,}
\newblock
\begin{APACrefDOI} \doi{10.1109/TPAMI.2024.3435571} \end{APACrefDOI}
\newblock

\newblock

\PrintBackRefs{\CurrentBib}

\bibitem [\protect \citeauthoryear {%
Cao%
, Wei%
, Gaidon%
, Arechiga%
\BCBL {}\ \BBA {} Ma%
}{%
Cao%
\ \protect \BOthers {.}}{%
{\protect \APACyear {2019}}%
}]{%
cao2019learning}
\APACinsertmetastar {%
cao2019learning}%
\begin{APACrefauthors}%
Cao, K.%
, Wei, C.%
, Gaidon, A.%
, Arechiga, N.%
\BCBL {} Ma, T.%
\end{APACrefauthors}%
\unskip\
\newblock
\APACrefYearMonthDay{2019}{}{}.
\newblock
{\BBOQ}\APACrefatitle {Learning Imbalanced Datasets with
  Label-Distribution-Aware Margin Loss} {Learning imbalanced datasets with
  label-distribution-aware margin loss}.{\BBCQ}
\newblock
 \APACrefbtitle {Advances in Neural Information Processing Systems} {Advances
  in neural information processing systems}\ (\BVOL~32).
\newblock
\APACrefnote{Available at
  \url{https://proceedings.neurips.cc/paper_files/paper/2019/file/621461af90cadfdaf0e8d4cc25129f91-Paper.pdf}}
\PrintBackRefs{\CurrentBib}

\bibitem [\protect \citeauthoryear {%
G.~Chen%
, Peng%
, Wang%
\BCBL {}\ \BBA {} Tian%
}{%
G.~Chen%
\ \protect \BOthers {.}}{%
{\protect \APACyear {2022}}%
}]{%
chen2021_arpl}
\APACinsertmetastar {%
chen2021_arpl}%
\begin{APACrefauthors}%
Chen, G.%
, Peng, P.%
, Wang, X.%
\BCBL {} Tian, Y.%
\end{APACrefauthors}%
\unskip\
\newblock
\APACrefYearMonthDay{2022}{}{}.
\newblock
{\BBOQ}\APACrefatitle {Adversarial Reciprocal Points Learning for Open Set
  Recognition} {Adversarial reciprocal points learning for open set
  recognition}.{\BBCQ}
\newblock
\APACjournalVolNumPages{IEEE Transactions on Pattern Analysis and Machine
  Intelligence}{44}{11}{8065--8081,}
\newblock
\begin{APACrefDOI} \doi{10.1109/TPAMI.2021.3106743} \end{APACrefDOI}
\newblock

\newblock

\PrintBackRefs{\CurrentBib}

\bibitem [\protect \citeauthoryear {%
Z.~Chen%
\ \protect \BOthers {.}}{%
Z.~Chen%
\ \protect \BOthers {.}}{%
{\protect \APACyear {2026}}%
}]{%
chen2025decompose}
\APACinsertmetastar {%
chen2025decompose}%
\begin{APACrefauthors}%
Chen, Z.%
, Xu, C.%
, Hu, T.%
, Wang, L.%
, Chen, J.%
\BCBL {} Li, J.%
\end{APACrefauthors}%
\unskip\
\newblock
\APACrefYearMonthDay{2026}{}{}.
\newblock
{\BBOQ}\APACrefatitle {Decompose-Compose Feature Augmentation for Imbalanced
  Crack Recognition in Industrial Scenarios} {Decompose-compose feature
  augmentation for imbalanced crack recognition in industrial
  scenarios}.{\BBCQ}
\newblock
\APACjournalVolNumPages{IEEE Transactions on Automation Science and
  Engineering}{23}{}{3329--3339,}
\newblock
\begin{APACrefDOI} \doi{10.1109/TASE.2025.3535764} \end{APACrefDOI}
\newblock

\newblock

\PrintBackRefs{\CurrentBib}

\bibitem [\protect \citeauthoryear {%
Z.~Chen%
\ \protect \BOthers {.}}{%
Z.~Chen%
\ \protect \BOthers {.}}{%
{\protect \APACyear {2024}}%
}]{%
chen2024neural}
\APACinsertmetastar {%
chen2024neural}%
\begin{APACrefauthors}%
Chen, Z.%
, Zhang, M.%
, Cui, S.%
, Li, H.%
, Niu, G.%
, Gong, M.%
\BDBL {}Zhang, K.%
\end{APACrefauthors}%
\unskip\
\newblock
\APACrefYearMonthDay{2024}{}{}.
\newblock
{\BBOQ}\APACrefatitle {Neural Collapse Inspired Feature Alignment for
  Out-of-Distribution Generalization} {Neural collapse inspired feature
  alignment for out-of-distribution generalization}.{\BBCQ}
\newblock
 \APACrefbtitle {The Thirty-eighth Annual Conference on Neural Information
  Processing Systems.} {The thirty-eighth annual conference on neural
  information processing systems.}
\newblock
\begin{APACrefURL} {https://openreview.net/forum?id=wQpNG9JnPK}
  \end{APACrefURL}
\PrintBackRefs{\CurrentBib}

\bibitem [\protect \citeauthoryear {%
Chhabra%
}{%
Chhabra%
}{%
{\protect \APACyear {2024}}%
}]{%
s-chh_ViT_python}
\APACinsertmetastar {%
s-chh_ViT_python}%
\begin{APACrefauthors}%
Chhabra, S.%
\end{APACrefauthors}%
\unskip\
\newblock
\APACrefYearMonthDay{2024}{}{}.
\newblock
\APACrefbtitle {PyTorch-Scratch-Vision-Transformer-ViT.}
  {Pytorch-scratch-vision-transformer-vit.}
\newblock
\APAChowpublished
  {\url{https://github.com/s-chh/PyTorch-Scratch-Vision-Transformer-ViT}}.
\newblock
\APACaddressPublisher{}{GitHub}.
\newblock
\APACrefnote{Accessed: 2025-09-01}
\PrintBackRefs{\CurrentBib}

\bibitem [\protect \citeauthoryear {%
Dan%
\ \protect \BOthers {.}}{%
Dan%
\ \protect \BOthers {.}}{%
{\protect \APACyear {2024}}%
}]{%
dan2024topofr}
\APACinsertmetastar {%
dan2024topofr}%
\begin{APACrefauthors}%
Dan, J.%
, Liu, Y.%
, Deng, J.%
, Xie, H.%
, Li, S.%
, Sun, B.%
\BCBL {} Luo, S.%
\end{APACrefauthors}%
\unskip\
\newblock
\APACrefYearMonthDay{2024}{}{}.
\newblock
{\BBOQ}\APACrefatitle {Topo{FR}: A Closer Look at Topology Alignment on Face
  Recognition} {Topo{FR}: A closer look at topology alignment on face
  recognition}.{\BBCQ}
\newblock
 \APACrefbtitle {The Thirty-eighth Annual Conference on Neural Information
  Processing Systems.} {The thirty-eighth annual conference on neural
  information processing systems.}
\newblock
\begin{APACrefURL} {https://openreview.net/forum?id=KVAx5tys2p}
  \end{APACrefURL}
\PrintBackRefs{\CurrentBib}

\bibitem [\protect \citeauthoryear {%
Dang%
, Huu%
, Nguyen%
\BCBL {}\ \BBA {} Ho%
}{%
Dang%
\ \protect \BOthers {.}}{%
{\protect \APACyear {2024}}%
}]{%
dang2024neural}
\APACinsertmetastar {%
dang2024neural}%
\begin{APACrefauthors}%
Dang, H.%
, Huu, T.T.%
, Nguyen, T.M.%
\BCBL {} Ho, N.%
\end{APACrefauthors}%
\unskip\
\newblock
\APACrefYearMonthDay{2024}{}{}.
\newblock
{\BBOQ}\APACrefatitle {Neural Collapse for Cross-entropy Class-Imbalanced
  Learning with Unconstrained Re{LU} Features Model} {Neural collapse for
  cross-entropy class-imbalanced learning with unconstrained re{LU} features
  model}.{\BBCQ}
\newblock
 \APACrefbtitle {Forty-first International Conference on Machine Learning.}
  {Forty-first international conference on machine learning.}
\newblock
\begin{APACrefURL} {https://openreview.net/forum?id=YBetKvUlF7}
  \end{APACrefURL}
\PrintBackRefs{\CurrentBib}

\bibitem [\protect \citeauthoryear {%
Deng%
, Guo%
, Xue%
\BCBL {}\ \BBA {} Zafeiriou%
}{%
Deng%
\ \protect \BOthers {.}}{%
{\protect \APACyear {2019}}%
}]{%
deng2019arcface}
\APACinsertmetastar {%
deng2019arcface}%
\begin{APACrefauthors}%
Deng, J.%
, Guo, J.%
, Xue, N.%
\BCBL {} Zafeiriou, S.%
\end{APACrefauthors}%
\unskip\
\newblock
\APACrefYearMonthDay{2019}{}{}.
\newblock
{\BBOQ}\APACrefatitle {Arc{F}ace: Additive angular margin loss for deep face
  recognition} {Arc{F}ace: Additive angular margin loss for deep face
  recognition}.{\BBCQ}
\newblock
 \APACrefbtitle {Proceedings of the {IEEE/CVF} conference on computer vision
  and pattern recognition} {Proceedings of the {IEEE/CVF} conference on
  computer vision and pattern recognition}\ (\BPGS\ 4690--4699).
\PrintBackRefs{\CurrentBib}

\bibitem [\protect \citeauthoryear {%
Dosovitskiy%
\ \protect \BOthers {.}}{%
Dosovitskiy%
\ \protect \BOthers {.}}{%
{\protect \APACyear {2021}}%
{\protect \APACexlab {{\protect \BCnt {1}}}}}]{%
dosovitskiy2021an}
\APACinsertmetastar {%
dosovitskiy2021an}%
\begin{APACrefauthors}%
Dosovitskiy, A.%
, Beyer, L.%
, Kolesnikov, A.%
, Weissenborn, D.%
, Zhai, X.%
, Unterthiner, T.%
\BDBL {}Houlsby, N.%
\end{APACrefauthors}%
\unskip\
\newblock
\APACrefYearMonthDay{2021{\protect \BCnt {1}}}{}{}.
\newblock
{\BBOQ}\APACrefatitle {An Image is Worth 16x16 Words: Transformers for Image
  Recognition at Scale} {An image is worth 16x16 words: Transformers for image
  recognition at scale}.{\BBCQ}
\newblock
 \APACrefbtitle {International Conference on Learning Representations.}
  {International conference on learning representations.}
\newblock
\begin{APACrefURL} {https://openreview.net/forum?id=YicbFdNTTy}
  \end{APACrefURL}
\PrintBackRefs{\CurrentBib}

\bibitem [\protect \citeauthoryear {%
Dosovitskiy%
\ \protect \BOthers {.}}{%
Dosovitskiy%
\ \protect \BOthers {.}}{%
{\protect \APACyear {2021}}%
{\protect \APACexlab {{\protect \BCnt {2}}}}}]{%
dosovitskiy2020_vit}
\APACinsertmetastar {%
dosovitskiy2020_vit}%
\begin{APACrefauthors}%
Dosovitskiy, A.%
, Beyer, L.%
, Kolesnikov, A.%
, Weissenborn, D.%
, Zhai, X.%
, Unterthiner, T.%
\BDBL {}Houlsby, N.%
\end{APACrefauthors}%
\unskip\
\newblock
\APACrefYearMonthDay{2021{\protect \BCnt {2}}}{}{}.
\newblock
{\BBOQ}\APACrefatitle {An Image is Worth 16x16 Words: Transformers for Image
  Recognition at Scale} {An image is worth 16x16 words: Transformers for image
  recognition at scale}.{\BBCQ}
\newblock
 \APACrefbtitle {International Conference on Learning Representations.}
  {International conference on learning representations.}
\newblock
\begin{APACrefURL} {https://openreview.net/forum?id=YicbFdNTTy}
  \end{APACrefURL}
\PrintBackRefs{\CurrentBib}

\bibitem [\protect \citeauthoryear {%
Fang%
, He%
, Long%
\BCBL {}\ \BBA {} Su%
}{%
Fang%
\ \protect \BOthers {.}}{%
{\protect \APACyear {2021}}%
}]{%
fang2021exploring}
\APACinsertmetastar {%
fang2021exploring}%
\begin{APACrefauthors}%
Fang, C.%
, He, H.%
, Long, Q.%
\BCBL {} Su, W.J.%
\end{APACrefauthors}%
\unskip\
\newblock
\APACrefYearMonthDay{2021}{}{}.
\newblock
{\BBOQ}\APACrefatitle {Exploring deep neural networks via layer-peeled model:
  Minority collapse in imbalanced training} {Exploring deep neural networks via
  layer-peeled model: Minority collapse in imbalanced training}.{\BBCQ}
\newblock
\APACjournalVolNumPages{Proceedings of the National Academy of
  Sciences}{118}{43}{e2103091118,}
\newblock
\begin{APACrefDOI} \doi{10.1073/pnas.2103091118} \end{APACrefDOI}
\newblock
\begin{APACrefURL} {https://www.pnas.org/doi/abs/10.1073/pnas.2103091118}
  \end{APACrefURL}
\newblock

\newblock

\PrintBackRefs{\CurrentBib}

\bibitem [\protect \citeauthoryear {%
FISHER%
}{%
FISHER%
}{%
{\protect \APACyear {1936}}%
}]{%
fisher1936use}
\APACinsertmetastar {%
fisher1936use}%
\begin{APACrefauthors}%
FISHER, R.A.%
\end{APACrefauthors}%
\unskip\
\newblock
\APACrefYearMonthDay{1936}{}{}.
\newblock
{\BBOQ}\APACrefatitle {THE USE OF MULTIPLE MEASUREMENTS IN TAXONOMIC PROBLEMS}
  {The use of multiple measurements in taxonomic problems}.{\BBCQ}
\newblock
\APACjournalVolNumPages{Annals of Eugenics}{7}{2}{179--188,}
\newblock
\begin{APACrefDOI} \doi{https://doi.org/10.1111/j.1469-1809.1936.tb02137.x}
  \end{APACrefDOI}
\newblock
\begin{APACrefURL}
  {https://onlinelibrary.wiley.com/doi/abs/10.1111/j.1469-1809.1936.tb02137.x}
  \end{APACrefURL}
\newblock

\newblock

\PrintBackRefs{\CurrentBib}

\bibitem [\protect \citeauthoryear {%
Gao%
, Zhao%
, Guo%
\BCBL {}\ \BBA {} Zha%
}{%
Gao%
\ \protect \BOthers {.}}{%
{\protect \APACyear {2024}}%
}]{%
gao2024distribution}
\APACinsertmetastar {%
gao2024distribution}%
\begin{APACrefauthors}%
Gao, J.%
, Zhao, H.%
, Guo, D.D.%
\BCBL {} Zha, H.%
\end{APACrefauthors}%
\unskip\
\newblock
\APACrefYearMonthDay{2024}{21--27 Jul}{}.
\newblock
{\BBOQ}\APACrefatitle {Distribution Alignment Optimization through Neural
  Collapse for Long-tailed Classification} {Distribution alignment optimization
  through neural collapse for long-tailed classification}.{\BBCQ}
\newblock
 \APACrefbtitle {Proceedings of the 41st International Conference on Machine
  Learning} {Proceedings of the 41st international conference on machine
  learning}\ (\BVOL~235, \BPGS\ 14969--14987).
\newblock
\APACaddressPublisher{}{PMLR}.
\newblock
\begin{APACrefURL} {https://proceedings.mlr.press/v235/gao24s.html}
  \end{APACrefURL}
\PrintBackRefs{\CurrentBib}

\bibitem [\protect \citeauthoryear {%
Graf%
, Hofer%
, Niethammer%
\BCBL {}\ \BBA {} Kwitt%
}{%
Graf%
\ \protect \BOthers {.}}{%
{\protect \APACyear {2021}}%
}]{%
graf2021dissecting}
\APACinsertmetastar {%
graf2021dissecting}%
\begin{APACrefauthors}%
Graf, F.%
, Hofer, C.%
, Niethammer, M.%
\BCBL {} Kwitt, R.%
\end{APACrefauthors}%
\unskip\
\newblock
\APACrefYearMonthDay{2021}{}{}.
\newblock
{\BBOQ}\APACrefatitle {Dissecting supervised contrastive learning} {Dissecting
  supervised contrastive learning}.{\BBCQ}
\newblock
 \APACrefbtitle {International Conference on Machine Learning} {International
  conference on machine learning}\ (\BPGS\ 3821--3830).
\PrintBackRefs{\CurrentBib}

\bibitem [\protect \citeauthoryear {%
Han%
}{%
Han%
}{%
{\protect \APACyear {2023}}%
}]{%
han2023wrapped}
\APACinsertmetastar {%
han2023wrapped}%
\begin{APACrefauthors}%
Han, B.%
\end{APACrefauthors}%
\unskip\
\newblock
\APACrefYearMonthDay{2023}{}{}.
\newblock
{\BBOQ}\APACrefatitle {Wrapped cauchy distributed angular softmax for
  long-tailed visual recognition} {Wrapped cauchy distributed angular softmax
  for long-tailed visual recognition}.{\BBCQ}
\newblock
 \APACrefbtitle {International Conference on Machine Learning} {International
  conference on machine learning}\ (\BPGS\ 12368--12388).
\PrintBackRefs{\CurrentBib}

\bibitem [\protect \citeauthoryear {%
He%
, Fan%
, Wu%
, Xie%
\BCBL {}\ \BBA {} Girshick%
}{%
He%
\ \protect \BOthers {.}}{%
{\protect \APACyear {2020}}%
}]{%
he2020momentum}
\APACinsertmetastar {%
he2020momentum}%
\begin{APACrefauthors}%
He, K.%
, Fan, H.%
, Wu, Y.%
, Xie, S.%
\BCBL {} Girshick, R.%
\end{APACrefauthors}%
\unskip\
\newblock
\APACrefYearMonthDay{2020}{}{}.
\newblock
{\BBOQ}\APACrefatitle {Momentum contrast for unsupervised visual representation
  learning} {Momentum contrast for unsupervised visual representation
  learning}.{\BBCQ}
\newblock
 \APACrefbtitle {Proceedings of the {IEEE/CVF} conference on computer vision
  and pattern recognition} {Proceedings of the {IEEE/CVF} conference on
  computer vision and pattern recognition}\ (\BPGS\ 9729--9738).
\PrintBackRefs{\CurrentBib}

\bibitem [\protect \citeauthoryear {%
He%
, Zhang%
, Ren%
\BCBL {}\ \BBA {} Sun%
}{%
He%
\ \protect \BOthers {.}}{%
{\protect \APACyear {2016}}%
}]{%
He2016_deep}
\APACinsertmetastar {%
He2016_deep}%
\begin{APACrefauthors}%
He, K.%
, Zhang, X.%
, Ren, S.%
\BCBL {} Sun, J.%
\end{APACrefauthors}%
\unskip\
\newblock
\APACrefYearMonthDay{2016}{}{}.
\newblock
{\BBOQ}\APACrefatitle {Deep Residual Learning for Image Recognition} {Deep
  residual learning for image recognition}.{\BBCQ}
\newblock
 \APACrefbtitle {Proceedings of the IEEE Conference on Computer Vision and
  Pattern Recognition (CVPR)} {Proceedings of the ieee conference on computer
  vision and pattern recognition (cvpr)}\ (\BPGS\ 770--778).
\newblock
\APACaddressPublisher{Las Vegas, USA}{IEEE}.
\PrintBackRefs{\CurrentBib}

\bibitem [\protect \citeauthoryear {%
Howard%
\ \protect \BOthers {.}}{%
Howard%
\ \protect \BOthers {.}}{%
{\protect \APACyear {2017}}%
}]{%
howard2017mobilenets}
\APACinsertmetastar {%
howard2017mobilenets}%
\begin{APACrefauthors}%
Howard, A.%
, Zhu, M.%
, Chen, B.%
, Kalenichenko, D.%
, Wang, W.%
, Weyand, T.%
\BDBL {}Adam, H.%
\end{APACrefauthors}%
\unskip\
\newblock
\APACrefYearMonthDay{2017}{}{}.
\newblock
\APACrefbtitle {MobileNets: Efficient Convolutional Neural Networks for Mobile
  Vision Applications.} {Mobilenets: Efficient convolutional neural networks
  for mobile vision applications.}
\newblock
\APAChowpublished {arXiv preprint arXiv:1704.04861}.
\newblock
\begin{APACrefURL} {https://arxiv.org/abs/1704.04861} \end{APACrefURL}
\PrintBackRefs{\CurrentBib}

\bibitem [\protect \citeauthoryear {%
Huang%
, Liu%
, van~der Maaten%
\BCBL {}\ \BBA {} Weinberger%
}{%
Huang%
\ \protect \BOthers {.}}{%
{\protect \APACyear {2017}}%
}]{%
Huang2017_densely}
\APACinsertmetastar {%
Huang2017_densely}%
\begin{APACrefauthors}%
Huang, G.%
, Liu, Z.%
, van~der Maaten, L.%
\BCBL {} Weinberger, K.Q.%
\end{APACrefauthors}%
\unskip\
\newblock
\APACrefYearMonthDay{2017}{}{}.
\newblock
{\BBOQ}\APACrefatitle {Densely Connected Convolutional Networks} {Densely
  connected convolutional networks}.{\BBCQ}
\newblock
 \APACrefbtitle {Proceedings of the IEEE Conference on Computer Vision and
  Pattern Recognition (CVPR)} {Proceedings of the ieee conference on computer
  vision and pattern recognition (cvpr)}\ (\BPGS\ 4700--4708).
\newblock
\APACaddressPublisher{Honolulu, HI, USA}{IEEE}.
\PrintBackRefs{\CurrentBib}

\bibitem [\protect \citeauthoryear {%
Jain%
\ \protect \BOthers {.}}{%
Jain%
\ \protect \BOthers {.}}{%
{\protect \APACyear {2024}}%
}]{%
jain2024fusion}
\APACinsertmetastar {%
jain2024fusion}%
\begin{APACrefauthors}%
Jain, D.K.%
, Zhao, X.%
, Gan, C.%
, Shukla, P.K.%
, Jain, A.%
\BCBL {} Sharma, S.%
\end{APACrefauthors}%
\unskip\
\newblock
\APACrefYearMonthDay{2024}{}{}.
\newblock
{\BBOQ}\APACrefatitle {Fusion-driven deep feature network for enhanced object
  detection and tracking in video surveillance systems} {Fusion-driven deep
  feature network for enhanced object detection and tracking in video
  surveillance systems}.{\BBCQ}
\newblock
\APACjournalVolNumPages{Information Fusion}{109}{}{102429,}
\newblock
\begin{APACrefDOI} \doi{https://doi.org/10.1016/j.inffus.2024.102429}
  \end{APACrefDOI}
\newblock
\begin{APACrefURL}
  {https://www.sciencedirect.com/science/article/pii/S1566253524002070}
  \end{APACrefURL}
\newblock

\newblock

\PrintBackRefs{\CurrentBib}

\bibitem [\protect \citeauthoryear {%
Kim%
\ \BBA {} Kim%
}{%
Kim%
\ \BBA {} Kim%
}{%
{\protect \APACyear {2024}}%
}]{%
kim2024fixed}
\APACinsertmetastar {%
kim2024fixed}%
\begin{APACrefauthors}%
Kim, H.%
\BCBT {}\ \BBA {} Kim, K.%
\end{APACrefauthors}%
\unskip\
\newblock
\APACrefYearMonthDay{2024}{}{}.
\newblock
{\BBOQ}\APACrefatitle {Fixed Non-negative Orthogonal Classifier: Inducing
  Zero-mean Neural Collapse with Feature Dimension Separation} {Fixed
  non-negative orthogonal classifier: Inducing zero-mean neural collapse with
  feature dimension separation}.{\BBCQ}
\newblock
 \APACrefbtitle {The Twelfth International Conference on Learning
  Representations.} {The twelfth international conference on learning
  representations.}
\newblock
\begin{APACrefURL} {https://openreview.net/forum?id=F4bmOrmUwc}
  \end{APACrefURL}
\PrintBackRefs{\CurrentBib}

\bibitem [\protect \citeauthoryear {%
Kobayashi%
}{%
Kobayashi%
}{%
{\protect \APACyear {2023}}%
}]{%
kobayashi2023two}
\APACinsertmetastar {%
kobayashi2023two}%
\begin{APACrefauthors}%
Kobayashi, T.%
\end{APACrefauthors}%
\unskip\
\newblock
\APACrefYearMonthDay{2023}{}{}.
\newblock
{\BBOQ}\APACrefatitle {Two-way multi-label loss} {Two-way multi-label
  loss}.{\BBCQ}
\newblock
 \APACrefbtitle {Proceedings of the {IEEE/CVF} Conference on Computer Vision
  and Pattern Recognition} {Proceedings of the {IEEE/CVF} conference on
  computer vision and pattern recognition}\ (\BPGS\ 7476--7485).
\PrintBackRefs{\CurrentBib}

\bibitem [\protect \citeauthoryear {%
Koromilas%
, Bouritsas%
, Giannakopoulos%
, Nicolaou%
\BCBL {}\ \BBA {} Panagakis%
}{%
Koromilas%
\ \protect \BOthers {.}}{%
{\protect \APACyear {2024}}%
}]{%
koromilas2024bridging}
\APACinsertmetastar {%
koromilas2024bridging}%
\begin{APACrefauthors}%
Koromilas, P.%
, Bouritsas, G.%
, Giannakopoulos, T.%
, Nicolaou, M.%
\BCBL {} Panagakis, Y.%
\end{APACrefauthors}%
\unskip\
\newblock
\APACrefYearMonthDay{2024}{}{}.
\newblock
{\BBOQ}\APACrefatitle {Bridging Mini-Batch and Asymptotic Analysis in
  Contrastive Learning: From InfoNCE to Kernel-Based Losses} {Bridging
  mini-batch and asymptotic analysis in contrastive learning: From infonce to
  kernel-based losses}.{\BBCQ}
\newblock
 \APACrefbtitle {International Conference on Machine Learning} {International
  conference on machine learning}\ (\BPGS\ 25276--25301).
\PrintBackRefs{\CurrentBib}

\bibitem [\protect \citeauthoryear {%
Krizhevsky%
}{%
Krizhevsky%
}{%
{\protect \APACyear {2009}}%
}]{%
krizhevsky2009_learning}
\APACinsertmetastar {%
krizhevsky2009_learning}%
\begin{APACrefauthors}%
Krizhevsky, A.%
\end{APACrefauthors}%
\unskip\
\newblock
\APACrefYearMonthDay{2009}{}{}.
\newblock
\APACrefbtitle {Learning multiple layers of features from tiny images}
  {Learning multiple layers of features from tiny images}\
  \APACbVolEdTR{}{\BTR{}}.
\newblock
\APACaddressInstitution{}{University of Toronto}.
\newblock
\begin{APACrefURL}
  {https://www.cs.toronto.edu/~kriz/learning-features-2009-TR.pdf}
  \end{APACrefURL}
\PrintBackRefs{\CurrentBib}

\bibitem [\protect \citeauthoryear {%
Lecun%
, Bottou%
, Bengio%
\BCBL {}\ \BBA {} Haffner%
}{%
Lecun%
\ \protect \BOthers {.}}{%
{\protect \APACyear {1998}}%
}]{%
Lecun1998_gradient}
\APACinsertmetastar {%
Lecun1998_gradient}%
\begin{APACrefauthors}%
Lecun, Y.%
, Bottou, L.%
, Bengio, Y.%
\BCBL {} Haffner, P.%
\end{APACrefauthors}%
\unskip\
\newblock
\APACrefYearMonthDay{1998}{}{}.
\newblock
{\BBOQ}\APACrefatitle {Gradient-based learning applied to document recognition}
  {Gradient-based learning applied to document recognition}.{\BBCQ}
\newblock
\APACjournalVolNumPages{Proceedings of the IEEE}{86}{11}{2278--2324,}
\newblock
\begin{APACrefDOI} \doi{10.1109/5.726791} \end{APACrefDOI}
\newblock

\newblock

\PrintBackRefs{\CurrentBib}

\bibitem [\protect \citeauthoryear {%
Lee%
, Lee%
\BCBL {}\ \BBA {} Song%
}{%
Lee%
\ \protect \BOthers {.}}{%
{\protect \APACyear {2021}}%
}]{%
lee2021vision}
\APACinsertmetastar {%
lee2021vision}%
\begin{APACrefauthors}%
Lee, S.H.%
, Lee, S.%
\BCBL {} Song, B.C.%
\end{APACrefauthors}%
\unskip\
\newblock
\APACrefYearMonthDay{2021}{}{}.
\newblock
\APACrefbtitle {Vision transformer for small-size datasets.} {Vision
  transformer for small-size datasets.}
\newblock
\APAChowpublished {arXiv preprint arXiv:2112.13492}.
\newblock
\begin{APACrefURL} {https://arxiv.org/abs/2112.13492} \end{APACrefURL}
\PrintBackRefs{\CurrentBib}

\bibitem [\protect \citeauthoryear {%
J.~Li%
\ \protect \BOthers {.}}{%
J.~Li%
\ \protect \BOthers {.}}{%
{\protect \APACyear {2023}}%
}]{%
li2023fcc}
\APACinsertmetastar {%
li2023fcc}%
\begin{APACrefauthors}%
Li, J.%
, Meng, Z.%
, Shi, D.%
, Song, R.%
, Diao, X.%
, Wang, J.%
\BCBL {} Xu, H.%
\end{APACrefauthors}%
\unskip\
\newblock
\APACrefYearMonthDay{2023}{}{}.
\newblock
{\BBOQ}\APACrefatitle {Fcc: Feature clusters compression for long-tailed visual
  recognition} {Fcc: Feature clusters compression for long-tailed visual
  recognition}.{\BBCQ}
\newblock
 \APACrefbtitle {Proceedings of the {IEEE/CVF} conference on computer vision
  and pattern recognition} {Proceedings of the {IEEE/CVF} conference on
  computer vision and pattern recognition}\ (\BPGS\ 24080--24089).
\PrintBackRefs{\CurrentBib}

\bibitem [\protect \citeauthoryear {%
Q.~Li%
, Xiao%
\BCBL {}\ \BBA {} Shen%
}{%
Q.~Li%
\ \protect \BOthers {.}}{%
{\protect \APACyear {2025}}%
}]{%
li2025_bce}
\APACinsertmetastar {%
li2025_bce}%
\begin{APACrefauthors}%
Li, Q.%
, Xiao, H.%
\BCBL {} Shen, L.%
\end{APACrefauthors}%
\unskip\
\newblock
\APACrefYearMonthDay{2025}{}{}.
\newblock
{\BBOQ}\APACrefatitle {{BCE} vs. {CE} in Deep Feature Learning} {{BCE} vs. {CE}
  in deep feature learning}.{\BBCQ}
\newblock
 \APACrefbtitle {Forty-second International Conference on Machine Learning.}
  {Forty-second international conference on machine learning.}
\newblock
\begin{APACrefURL} {https://openreview.net/forum?id=aoLFIUlyPE}
  \end{APACrefURL}
\PrintBackRefs{\CurrentBib}

\bibitem [\protect \citeauthoryear {%
Liang%
, Zhang%
, Zhang%
, Zhang%
\BCBL {}\ \BBA {} Tang%
}{%
Liang%
\ \protect \BOthers {.}}{%
{\protect \APACyear {2024}}%
}]{%
liang2024nonlocal}
\APACinsertmetastar {%
liang2024nonlocal}%
\begin{APACrefauthors}%
Liang, R.%
, Zhang, S.%
, Zhang, W.%
, Zhang, G.%
\BCBL {} Tang, J.%
\end{APACrefauthors}%
\unskip\
\newblock
\APACrefYearMonthDay{2024}{}{}.
\newblock
{\BBOQ}\APACrefatitle {Nonlocal hybrid network for long-tailed image
  classification} {Nonlocal hybrid network for long-tailed image
  classification}.{\BBCQ}
\newblock
\APACjournalVolNumPages{ACM Transactions on Multimedia Computing,
  Communications and Applications}{20}{4}{1--22,}
\newblock
\begin{APACrefDOI} \doi{10.1145/3630256} \end{APACrefDOI}
\newblock

\newblock

\PrintBackRefs{\CurrentBib}

\bibitem [\protect \citeauthoryear {%
J.~Liu%
, Sun%
, Han%
, Dou%
\BCBL {}\ \BBA {} Li%
}{%
J.~Liu%
\ \protect \BOthers {.}}{%
{\protect \APACyear {2020}}%
}]{%
liu2020deep}
\APACinsertmetastar {%
liu2020deep}%
\begin{APACrefauthors}%
Liu, J.%
, Sun, Y.%
, Han, C.%
, Dou, Z.%
\BCBL {} Li, W.%
\end{APACrefauthors}%
\unskip\
\newblock
\APACrefYearMonthDay{2020}{}{}.
\newblock
{\BBOQ}\APACrefatitle {Deep representation learning on long-tailed data: A
  learnable embedding augmentation perspective} {Deep representation learning
  on long-tailed data: A learnable embedding augmentation perspective}.{\BBCQ}
\newblock
 \APACrefbtitle {Proceedings of the {IEEE/CVF} conference on computer vision
  and pattern recognition} {Proceedings of the {IEEE/CVF} conference on
  computer vision and pattern recognition}\ (\BPGS\ 2970--2979).
\PrintBackRefs{\CurrentBib}

\bibitem [\protect \citeauthoryear {%
W.~Liu%
, Wen%
, Raj%
, Singh%
\BCBL {}\ \BBA {} Weller%
}{%
W.~Liu%
, Wen%
\BCBL {}\ \protect \BOthers {.}}{%
{\protect \APACyear {2023}}%
}]{%
liu2022sphereface}
\APACinsertmetastar {%
liu2022sphereface}%
\begin{APACrefauthors}%
Liu, W.%
, Wen, Y.%
, Raj, B.%
, Singh, R.%
\BCBL {} Weller, A.%
\end{APACrefauthors}%
\unskip\
\newblock
\APACrefYearMonthDay{2023}{}{}.
\newblock
{\BBOQ}\APACrefatitle {SphereFace Revived: Unifying Hyperspherical Face
  Recognition} {Sphereface revived: Unifying hyperspherical face
  recognition}.{\BBCQ}
\newblock
\APACjournalVolNumPages{IEEE Transactions on Pattern Analysis and Machine
  Intelligence}{45}{2}{2458--2474,}
\newblock
\begin{APACrefDOI} \doi{10.1109/TPAMI.2022.3159732} \end{APACrefDOI}
\newblock

\newblock

\PrintBackRefs{\CurrentBib}

\bibitem [\protect \citeauthoryear {%
W.~Liu%
\ \protect \BOthers {.}}{%
W.~Liu%
\ \protect \BOthers {.}}{%
{\protect \APACyear {2017}}%
}]{%
liu2017sphereface}
\APACinsertmetastar {%
liu2017sphereface}%
\begin{APACrefauthors}%
Liu, W.%
, Wen, Y.%
, Yu, Z.%
, Li, M.%
, Raj, B.%
\BCBL {} Song, L.%
\end{APACrefauthors}%
\unskip\
\newblock
\APACrefYearMonthDay{2017}{}{}.
\newblock
{\BBOQ}\APACrefatitle {{S}phere{F}ace: Deep hypersphere embedding for face
  recognition} {{S}phere{F}ace: Deep hypersphere embedding for face
  recognition}.{\BBCQ}
\newblock
 \APACrefbtitle {Proceedings of the IEEE conference on computer vision and
  pattern recognition} {Proceedings of the ieee conference on computer vision
  and pattern recognition}\ (\BPGS\ 212--220).
\PrintBackRefs{\CurrentBib}

\bibitem [\protect \citeauthoryear {%
W.~Liu%
, Yu%
, Weller%
\BCBL {}\ \BBA {} Sch{\"o}lkopf%
}{%
W.~Liu%
, Yu%
\BCBL {}\ \protect \BOthers {.}}{%
{\protect \APACyear {2023}}%
}]{%
liu2023generalizing}
\APACinsertmetastar {%
liu2023generalizing}%
\begin{APACrefauthors}%
Liu, W.%
, Yu, L.%
, Weller, A.%
\BCBL {} Sch{\"o}lkopf, B.%
\end{APACrefauthors}%
\unskip\
\newblock
\APACrefYearMonthDay{2023}{}{}.
\newblock
{\BBOQ}\APACrefatitle {Generalizing and Decoupling Neural Collapse via
  Hyperspherical Uniformity Gap} {Generalizing and decoupling neural collapse
  via hyperspherical uniformity gap}.{\BBCQ}
\newblock
 \APACrefbtitle {The Eleventh International Conference on Learning
  Representations.} {The eleventh international conference on learning
  representations.}
\newblock
\begin{APACrefURL} {https://openreview.net/forum?id=inU2quhGdNU}
  \end{APACrefURL}
\PrintBackRefs{\CurrentBib}

\bibitem [\protect \citeauthoryear {%
Z.~Liu%
\ \protect \BOthers {.}}{%
Z.~Liu%
\ \protect \BOthers {.}}{%
{\protect \APACyear {2021}}%
}]{%
liu2021swin}
\APACinsertmetastar {%
liu2021swin}%
\begin{APACrefauthors}%
Liu, Z.%
, Lin, Y.%
, Cao, Y.%
, Hu, H.%
, Wei, Y.%
, Zhang, Z.%
\BDBL {}Guo, B.%
\end{APACrefauthors}%
\unskip\
\newblock
\APACrefYearMonthDay{2021}{}{}.
\newblock
{\BBOQ}\APACrefatitle {Swin transformer: Hierarchical vision transformer using
  shifted windows} {Swin transformer: Hierarchical vision transformer using
  shifted windows}.{\BBCQ}
\newblock
 \APACrefbtitle {Proceedings of the {IEEE/CVF} international conference on
  computer vision} {Proceedings of the {IEEE/CVF} international conference on
  computer vision}\ (\BPGS\ 10012--10022).
\PrintBackRefs{\CurrentBib}

\bibitem [\protect \citeauthoryear {%
Z.~Liu%
\ \protect \BOthers {.}}{%
Z.~Liu%
\ \protect \BOthers {.}}{%
{\protect \APACyear {2022}}%
}]{%
liu2022convnet}
\APACinsertmetastar {%
liu2022convnet}%
\begin{APACrefauthors}%
Liu, Z.%
, Mao, H.%
, Wu, C\BHBI Y.%
, Feichtenhofer, C.%
, Darrell, T.%
\BCBL {} Xie, S.%
\end{APACrefauthors}%
\unskip\
\newblock
\APACrefYearMonthDay{2022}{}{}.
\newblock
{\BBOQ}\APACrefatitle {A convnet for the 2020s} {A convnet for the
  2020s}.{\BBCQ}
\newblock
 \APACrefbtitle {Proceedings of the {IEEE/CVF} conference on computer vision
  and pattern recognition} {Proceedings of the {IEEE/CVF} conference on
  computer vision and pattern recognition}\ (\BPGS\ 11976--11986).
\PrintBackRefs{\CurrentBib}

\bibitem [\protect \citeauthoryear {%
Lu%
\ \BBA {} Steinerberger%
}{%
Lu%
\ \BBA {} Steinerberger%
}{%
{\protect \APACyear {2022}}%
}]{%
LU2022}
\APACinsertmetastar {%
LU2022}%
\begin{APACrefauthors}%
Lu, J.%
\BCBT {}\ \BBA {} Steinerberger, S.%
\end{APACrefauthors}%
\unskip\
\newblock
\APACrefYearMonthDay{2022}{}{}.
\newblock
{\BBOQ}\APACrefatitle {Neural collapse under cross-entropy loss} {Neural
  collapse under cross-entropy loss}.{\BBCQ}
\newblock
\APACjournalVolNumPages{Applied and Computational Harmonic
  Analysis}{59}{}{224--241,}
\newblock

\newblock

\PrintBackRefs{\CurrentBib}

\bibitem [\protect \citeauthoryear {%
Neal%
, Olson%
, Fern%
, Wong%
\BCBL {}\ \BBA {} Li%
}{%
Neal%
\ \protect \BOthers {.}}{%
{\protect \APACyear {2018}}%
}]{%
neal2018open}
\APACinsertmetastar {%
neal2018open}%
\begin{APACrefauthors}%
Neal, L.%
, Olson, M.%
, Fern, X.%
, Wong, W\BHBI K.%
\BCBL {} Li, F.%
\end{APACrefauthors}%
\unskip\
\newblock
\APACrefYearMonthDay{2018}{}{}.
\newblock
{\BBOQ}\APACrefatitle {Open set learning with counterfactual images} {Open set
  learning with counterfactual images}.{\BBCQ}
\newblock
 \APACrefbtitle {Proceedings of the European conference on computer vision
  (ECCV)} {Proceedings of the european conference on computer vision (eccv)}\
  (\BPGS\ 613--628).
\PrintBackRefs{\CurrentBib}

\bibitem [\protect \citeauthoryear {%
Netzer%
\ \protect \BOthers {.}}{%
Netzer%
\ \protect \BOthers {.}}{%
{\protect \APACyear {2011}}%
}]{%
netzer2011reading}
\APACinsertmetastar {%
netzer2011reading}%
\begin{APACrefauthors}%
Netzer, Y.%
, Wang, T.%
, Coates, A.%
, Bissacco, A.%
, Wu, B.%
\BCBL {} Ng, A.Y.%
\end{APACrefauthors}%
\unskip\
\newblock
\APACrefYearMonthDay{2011}{}{}.
\newblock
{\BBOQ}\APACrefatitle {Reading digits in natural images with unsupervised
  feature learning} {Reading digits in natural images with unsupervised feature
  learning}.{\BBCQ}
\newblock
 \APACrefbtitle {NIPS workshop on deep learning and unsupervised feature
  learning.} {Nips workshop on deep learning and unsupervised feature
  learning.}
\PrintBackRefs{\CurrentBib}

\bibitem [\protect \citeauthoryear {%
Papyan%
, Han%
\BCBL {}\ \BBA {} Donoho%
}{%
Papyan%
\ \protect \BOthers {.}}{%
{\protect \APACyear {2020}}%
}]{%
Papyan2020_prevalence}
\APACinsertmetastar {%
Papyan2020_prevalence}%
\begin{APACrefauthors}%
Papyan, V.%
, Han, X.Y.%
\BCBL {} Donoho, D.L.%
\end{APACrefauthors}%
\unskip\
\newblock
\APACrefYearMonthDay{2020}{}{}.
\newblock
{\BBOQ}\APACrefatitle {Prevalence of neural collapse during the terminal phase
  of deep learning training} {Prevalence of neural collapse during the terminal
  phase of deep learning training}.{\BBCQ}
\newblock
\APACjournalVolNumPages{Proceedings of the National Academy of
  Sciences}{117}{40}{24652--24663,}
\newblock
\begin{APACrefDOI} \doi{10.1073/pnas.2015509117} \end{APACrefDOI}
\newblock
\begin{APACrefURL} {https://www.pnas.org/doi/abs/10.1073/pnas.2015509117}
  \end{APACrefURL}
\newblock

\newblock

\PrintBackRefs{\CurrentBib}

\bibitem [\protect \citeauthoryear {%
Vasu%
, Pouransari%
, Faghri%
, Vemulapalli%
\BCBL {}\ \BBA {} Tuzel%
}{%
Vasu%
\ \protect \BOthers {.}}{%
{\protect \APACyear {2024}}%
}]{%
Vasu_2024_CVPR}
\APACinsertmetastar {%
Vasu_2024_CVPR}%
\begin{APACrefauthors}%
Vasu, P.K.A.%
, Pouransari, H.%
, Faghri, F.%
, Vemulapalli, R.%
\BCBL {} Tuzel, O.%
\end{APACrefauthors}%
\unskip\
\newblock
\APACrefYearMonthDay{2024}{June}{}.
\newblock
{\BBOQ}\APACrefatitle {{MobileCLIP}: Fast Image-Text Models through Multi-Modal
  Reinforced Training} {{MobileCLIP}: Fast image-text models through
  multi-modal reinforced training}.{\BBCQ}
\newblock
 \APACrefbtitle {Proceedings of the {IEEE/CVF} Conference on Computer Vision
  and Pattern Recognition (CVPR)} {Proceedings of the {IEEE/CVF} conference on
  computer vision and pattern recognition (cvpr)}\ (\BPGS\ 15963--15974).
\PrintBackRefs{\CurrentBib}

\bibitem [\protect \citeauthoryear {%
F.~Wang%
, Xiang%
, Cheng%
\BCBL {}\ \BBA {} Yuille%
}{%
F.~Wang%
\ \protect \BOthers {.}}{%
{\protect \APACyear {2017}}%
}]{%
wang2017normface}
\APACinsertmetastar {%
wang2017normface}%
\begin{APACrefauthors}%
Wang, F.%
, Xiang, X.%
, Cheng, J.%
\BCBL {} Yuille, A.L.%
\end{APACrefauthors}%
\unskip\
\newblock
\APACrefYearMonthDay{2017}{}{}.
\newblock
{\BBOQ}\APACrefatitle {Norm{F}ace: ${L}_2$ hypersphere embedding for face
  verification} {Norm{F}ace: ${L}_2$ hypersphere embedding for face
  verification}.{\BBCQ}
\newblock
 \APACrefbtitle {Proceedings of the 25th ACM international conference on
  Multimedia} {Proceedings of the 25th acm international conference on
  multimedia}\ (\BPGS\ 1041--1049).
\PrintBackRefs{\CurrentBib}

\bibitem [\protect \citeauthoryear {%
H.~Wang%
\ \protect \BOthers {.}}{%
H.~Wang%
\ \protect \BOthers {.}}{%
{\protect \APACyear {2018}}%
}]{%
wang2018cosface}
\APACinsertmetastar {%
wang2018cosface}%
\begin{APACrefauthors}%
Wang, H.%
, Wang, Y.%
, Zhou, Z.%
, Ji, X.%
, Gong, D.%
, Zhou, J.%
\BDBL {}Liu, W.%
\end{APACrefauthors}%
\unskip\
\newblock
\APACrefYearMonthDay{2018}{}{}.
\newblock
{\BBOQ}\APACrefatitle {Cos{F}ace: Large margin cosine loss for deep face
  recognition} {Cos{F}ace: Large margin cosine loss for deep face
  recognition}.{\BBCQ}
\newblock
 \APACrefbtitle {Proceedings of the IEEE conference on computer vision and
  pattern recognition} {Proceedings of the ieee conference on computer vision
  and pattern recognition}\ (\BPGS\ 5265--5274).
\PrintBackRefs{\CurrentBib}

\bibitem [\protect \citeauthoryear {%
Wen%
, Liu%
, Weller%
, Raj%
\BCBL {}\ \BBA {} Singh%
}{%
Wen%
\ \protect \BOthers {.}}{%
{\protect \APACyear {2022}}%
}]{%
wensphereface2}
\APACinsertmetastar {%
wensphereface2}%
\begin{APACrefauthors}%
Wen, Y.%
, Liu, W.%
, Weller, A.%
, Raj, B.%
\BCBL {} Singh, R.%
\end{APACrefauthors}%
\unskip\
\newblock
\APACrefYearMonthDay{2022}{}{}.
\newblock
{\BBOQ}\APACrefatitle {Sphere{F}ace2: Binary Classification is All You Need for
  Deep Face Recognition} {Sphere{F}ace2: Binary classification is all you need
  for deep face recognition}.{\BBCQ}
\newblock
 \APACrefbtitle {International Conference on Learning Representations.}
  {International conference on learning representations.}
\PrintBackRefs{\CurrentBib}

\bibitem [\protect \citeauthoryear {%
Wu%
, Zhang%
\BCBL {}\ \BBA {} Xu%
}{%
Wu%
\ \protect \BOthers {.}}{%
{\protect \APACyear {2017}}%
}]{%
wu2017tiny}
\APACinsertmetastar {%
wu2017tiny}%
\begin{APACrefauthors}%
Wu, J.%
, Zhang, Q.%
\BCBL {} Xu, G.%
\end{APACrefauthors}%
\unskip\
\newblock
\APACrefYearMonthDay{2017}{}{}.
\newblock
\APACrefbtitle {{Tiny ImageNet} challenge} {{Tiny ImageNet} challenge}\
  \APACbVolEdTR{}{\BTR{}}.
\newblock
\APACaddressInstitution{}{Stanford University}.
\newblock
\begin{APACrefURL} {https://cs231n.stanford.edu/reports/2017/pdfs/930.pdf}
  \end{APACrefURL}
\PrintBackRefs{\CurrentBib}

\bibitem [\protect \citeauthoryear {%
Yaras%
, Wang%
, Zhu%
, Balzano%
\BCBL {}\ \BBA {} Qu%
}{%
Yaras%
\ \protect \BOthers {.}}{%
{\protect \APACyear {2022}}%
}]{%
Yaras2022_NeurIPS}
\APACinsertmetastar {%
Yaras2022_NeurIPS}%
\begin{APACrefauthors}%
Yaras, C.%
, Wang, P.%
, Zhu, Z.%
, Balzano, L.%
\BCBL {} Qu, Q.%
\end{APACrefauthors}%
\unskip\
\newblock
\APACrefYearMonthDay{2022}{}{}.
\newblock
{\BBOQ}\APACrefatitle {Neural Collapse with Normalized Features: A Geometric
  Analysis over the Riemannian Manifold} {Neural collapse with normalized
  features: A geometric analysis over the riemannian manifold}.{\BBCQ}
\newblock
 S.~Koyejo, S.~Mohamed, A.~Agarwal, D.~Belgrave, K.~Cho\BCBL {}\ \BBA {} A.~Oh\
  (\BEDS), \APACrefbtitle {Advances in Neural Information Processing Systems}
  {Advances in neural information processing systems}\ (\BPGS\ 11547--11560).
\newblock
\APACaddressPublisher{New Orleans}{Curran Associates, Inc.}
\PrintBackRefs{\CurrentBib}

\bibitem [\protect \citeauthoryear {%
Zhang%
\ \protect \BOthers {.}}{%
Zhang%
\ \protect \BOthers {.}}{%
{\protect \APACyear {2022}}%
}]{%
zhang2022bootstrapping}
\APACinsertmetastar {%
zhang2022bootstrapping}%
\begin{APACrefauthors}%
Zhang, H.%
, Duan, J.%
, Xue, M.%
, Song, J.%
, Sun, L.%
\BCBL {} Song, M.%
\end{APACrefauthors}%
\unskip\
\newblock
\APACrefYearMonthDay{2022}{}{}.
\newblock
{\BBOQ}\APACrefatitle {Bootstrapping ViTs: Towards liberating vision
  transformers from pre-training} {Bootstrapping vits: Towards liberating
  vision transformers from pre-training}.{\BBCQ}
\newblock
 \APACrefbtitle {Proceedings of the {IEEE/CVF} Conference on Computer Vision
  and Pattern Recognition} {Proceedings of the {IEEE/CVF} conference on
  computer vision and pattern recognition}\ (\BPGS\ 8944--8953).
\PrintBackRefs{\CurrentBib}

\bibitem [\protect \citeauthoryear {%
Zhao%
\ \protect \BOthers {.}}{%
Zhao%
\ \protect \BOthers {.}}{%
{\protect \APACyear {2025}}%
}]{%
zhao2025global}
\APACinsertmetastar {%
zhao2025global}%
\begin{APACrefauthors}%
Zhao, W.%
, Zhu, X.%
, Shi, H.%
, Zhang, X\BHBI Y.%
, Zhao, G.%
\BCBL {} Lei, Z.%
\end{APACrefauthors}%
\unskip\
\newblock
\APACrefYearMonthDay{2025}{}{}.
\newblock
{\BBOQ}\APACrefatitle {Global Cross-Entropy Loss for Deep Face Recognition}
  {Global cross-entropy loss for deep face recognition}.{\BBCQ}
\newblock
\APACjournalVolNumPages{IEEE Transactions on Image
  Processing}{34}{}{1672--1685,}
\newblock
\begin{APACrefDOI} \doi{10.1109/TIP.2025.3546481} \end{APACrefDOI}
\newblock

\newblock

\PrintBackRefs{\CurrentBib}

\bibitem [\protect \citeauthoryear {%
Zheng%
\ \BBA {} Yang%
}{%
Zheng%
\ \BBA {} Yang%
}{%
{\protect \APACyear {2024}}%
}]{%
zheng2024knowledge}
\APACinsertmetastar {%
zheng2024knowledge}%
\begin{APACrefauthors}%
Zheng, K.%
\BCBT {}\ \BBA {} Yang, E\BHBI H.%
\end{APACrefauthors}%
\unskip\
\newblock
\APACrefYearMonthDay{2024}{}{}.
\newblock
{\BBOQ}\APACrefatitle {Knowledge Distillation Based on Transformed Teacher
  Matching} {Knowledge distillation based on transformed teacher
  matching}.{\BBCQ}
\newblock
 \APACrefbtitle {The Twelfth International Conference on Learning
  Representations.} {The twelfth international conference on learning
  representations.}
\newblock
\begin{APACrefURL} {https://openreview.net/forum?id=MJ3K7uDGGl}
  \end{APACrefURL}
\PrintBackRefs{\CurrentBib}

\bibitem [\protect \citeauthoryear {%
Zhou%
, Jia%
, Li%
, Shen%
\BCBL {}\ \BBA {} Duan%
}{%
Zhou%
\ \protect \BOthers {.}}{%
{\protect \APACyear {2023}}%
}]{%
zhou2023uniface}
\APACinsertmetastar {%
zhou2023uniface}%
\begin{APACrefauthors}%
Zhou, J.%
, Jia, X.%
, Li, Q.%
, Shen, L.%
\BCBL {} Duan, J.%
\end{APACrefauthors}%
\unskip\
\newblock
\APACrefYearMonthDay{2023}{}{}.
\newblock
{\BBOQ}\APACrefatitle {Uni{F}ace: Unified cross-entropy loss for deep face
  recognition} {Uni{F}ace: Unified cross-entropy loss for deep face
  recognition}.{\BBCQ}
\newblock
 \APACrefbtitle {Proceedings of the {IEEE/CVF} International Conference on
  Computer Vision} {Proceedings of the {IEEE/CVF} international conference on
  computer vision}\ (\BPGS\ 20730--20739).
\PrintBackRefs{\CurrentBib}

\bibitem [\protect \citeauthoryear {%
Zhou%
\ \protect \BOthers {.}}{%
Zhou%
\ \protect \BOthers {.}}{%
{\protect \APACyear {2022}}%
}]{%
Zhou2022_areall}
\APACinsertmetastar {%
Zhou2022_areall}%
\begin{APACrefauthors}%
Zhou, J.%
, You, C.%
, Li, X.%
, Liu, K.%
, Liu, S.%
, Qu, Q.%
\BCBL {} Zhu, Z.%
\end{APACrefauthors}%
\unskip\
\newblock
\APACrefYearMonthDay{2022}{}{}.
\newblock
{\BBOQ}\APACrefatitle {Are All Losses Created Equal: A Neural Collapse
  Perspective} {Are all losses created equal: A neural collapse
  perspective}.{\BBCQ}
\newblock
 S.~Koyejo, S.~Mohamed, A.~Agarwal, D.~Belgrave, K.~Cho\BCBL {}\ \BBA {} A.~Oh\
  (\BEDS), \APACrefbtitle {Advances in Neural Information Processing Systems}
  {Advances in neural information processing systems}\ (\BPGS\ 31697--31710).
\newblock
\APACaddressPublisher{New Orleans}{Curran Associates, Inc.}
\PrintBackRefs{\CurrentBib}

\bibitem [\protect \citeauthoryear {%
H.~Zhu%
, Chen%
\BCBL {}\ \BBA {} Yang%
}{%
H.~Zhu%
\ \protect \BOthers {.}}{%
{\protect \APACyear {2023}}%
}]{%
zhu2023understanding}
\APACinsertmetastar {%
zhu2023understanding}%
\begin{APACrefauthors}%
Zhu, H.%
, Chen, B.%
\BCBL {} Yang, C.%
\end{APACrefauthors}%
\unskip\
\newblock
\APACrefYearMonthDay{2023}{}{}.
\newblock
\APACrefbtitle {Understanding why {ViT} trains badly on small datasets: An
  intuitive perspective.} {Understanding why {ViT} trains badly on small
  datasets: An intuitive perspective.}
\newblock
\APAChowpublished {arXiv preprint arXiv:2302.03751}.
\newblock
\begin{APACrefURL} {https://arxiv.org/abs/2302.03751} \end{APACrefURL}
\PrintBackRefs{\CurrentBib}

\bibitem [\protect \citeauthoryear {%
Z.~Zhu%
\ \protect \BOthers {.}}{%
Z.~Zhu%
\ \protect \BOthers {.}}{%
{\protect \APACyear {2021}}%
}]{%
Zhu2021_geometric}
\APACinsertmetastar {%
Zhu2021_geometric}%
\begin{APACrefauthors}%
Zhu, Z.%
, Ding, T.%
, Zhou, J.%
, Li, X.%
, You, C.%
, Sulam, J.%
\BCBL {} Qu, Q.%
\end{APACrefauthors}%
\unskip\
\newblock
\APACrefYearMonthDay{2021}{}{}.
\newblock
{\BBOQ}\APACrefatitle {A Geometric Analysis of Neural Collapse with
  Unconstrained Features} {A geometric analysis of neural collapse with
  unconstrained features}.{\BBCQ}
\newblock
 M.~Ranzato, A.~Beygelzimer, Y.~Dauphin, P.~Liang\BCBL {}\ \BBA {}
  J.W.~Vaughan\ (\BEDS), \APACrefbtitle {Advances in Neural Information
  Processing Systems} {Advances in neural information processing systems}\
  (\BPGS\ 29820--29834).
\newblock
\APACaddressPublisher{}{Curran Associates, Inc.}
\PrintBackRefs{\CurrentBib}

\end{thebibliography}
